\documentclass{article}

\usepackage[utf8]{inputenc} 
\usepackage[T1]{fontenc}    
\usepackage{hyperref}       
\usepackage{url}            
\usepackage{float}
\usepackage{booktabs}       
\usepackage{mathtools}
\usepackage{amsmath,amssymb,amsfonts}       
\usepackage{nicefrac}       
\usepackage{microtype}      
\usepackage[noabbrev,capitalize]{cleveref}       
\usepackage{lipsum}         
\usepackage{graphicx}
\usepackage{natbib}
\usepackage{doi}
\usepackage{xcolor,xspace}
\usepackage{derivative}
\usepackage{subcaption}
\usepackage{enumitem}
\usepackage{array}
\newcolumntype{H}{>{\setbox0=\hbox\bgroup}c<{\egroup}@{}}
\usepackage{multirow}
\usepackage{authblk}

\newlength{\defbaselineskip}
\date{}

\newcommand{\printfnsymbol}[1]{%
  \textsuperscript{\@fnsymbol{#1}}%
}

\newcommand{\xdomain}{{\Omega}}
\newcommand{\xboundary}{{\partial\xdomain}}

\newcommand{\tdomain}{{[0, T]}}
\newcommand{\tr}{{\text{train}}}
\newcommand{\te}{{\text{test}}}
\newcommand{\supp}{{\text{supp}}}
\newcommand{\operator}{{A}}
\newcommand{\trueoperator}{{\operator^\dagger}}
\newcommand{\cD}{{\mathcal{D}}}
\newcommand{\posterior}{{p(\rmW | \cD^\tr)}}
\newcommand{\diverse}{{\text{diverse}}}

\newcommand{\boundaryop}{{\mathcal{B}}}

\newcommand{\outputvarmethod}{VarianceNO\xspace}
\newcommand{\bayesiannomethod}{BayesianNO\xspace}
\newcommand{\ensemblenomethod}{EnsembleNO\xspace}
\newcommand{\mcdropoutnomethod}{MC-DropoutNO\xspace}
\newcommand{\method}{\textsc{DiverseNO}\xspace}
\newcommand{\probconservno}{\textsc{Operator-ProbConserv}\xspace}
\newcommand{\probconserv}{\textsc{ProbConserv}\xspace}
\newcommand{\probconservanp}{\textsc{ProbConserv-ANP}\xspace}

\usepackage{amsmath,amsfonts,bm}

\def\eqref#1{equation~\ref{#1}}

\def\1{\bm{1}}

\def\rmI{{\mathbf{I}}}

\def\rmW{{\mathbf{\theta}}}

\def\vzero{{\bm{0}}}
\def\vone{{\bm{1}}}

\DeclareMathAlphabet{\mathsfit}{\encodingdefault}{\sfdefault}{m}{sl}
\SetMathAlphabet{\mathsfit}{bold}{\encodingdefault}{\sfdefault}{bx}{n}

\def\sR{{\mathbb{R}}}

\DeclareMathOperator*{\argmax}{arg\,max}
\DeclareMathOperator*{\argmin}{arg\,min}

\begin{document}

\title{Using Uncertainty Quantification to Characterize and Improve \\ Out-of-Domain Learning for PDEs}

\author[a]{S. Chandra Mouli\thanks{Work completed during an internship at AWS AI Labs.}}
\author[b]{Danielle C. Maddix\footnote{Correspondence to: Danielle C. Maddix <dmmaddix@amazon.com>.}}
\author[b]{Shima Alizadeh}
\author[b]{Gaurav Gupta}
\author[c,d]{Andrew Stuart}
\author[e]{Michael W. Mahoney}
\author[b]{Yuyang Wang}
\affil[a]{ Dept. of Computer Science, Purdue Univ. (305 N University St., West Lafayette, IN 47907) }
\affil[b]{AWS AI Labs (2795 Augustine Dr., Santa Clara, CA 95054)}
\affil[c]{Dept. of Computing and Mathematical Sciences, Caltech (1200 E. California Blvd., Pasadena, CA 91125)}
\affil[d]{Amazon Search (271 South Chester Ave., Pasadena, CA 91106)}
\affil[e]{Amazon Supply Chain Optimization Technologies (7 West 34th St., NY, NY 10001)}
\affil[ ]
{{\texttt{chandr@purdue.edu}, \  \{\texttt{dmmaddix, alizshim, gauravaz, andrxstu, zmahmich, yuyawang}\}\texttt{@amazon.com}}}

\date{}

\maketitle

\begin{abstract}
Existing work in scientific machine learning (SciML) has shown that data-driven learning of solution operators can provide a fast approximate alternative to classical numerical partial differential equation (PDE) solvers.
Of these, Neural Operators (NOs) have emerged as particularly promising.
We observe that several uncertainty quantification (UQ) methods for NOs fail for test inputs that are even moderately out-of-domain (OOD), even when the model approximates the solution well for in-domain tasks. 
To address this limitation, we show that ensembling several NOs can identify high-error regions and provide good uncertainty estimates that are well-correlated with prediction errors.  
Based on this, we propose a cost-effective alternative, \method, that mimics the properties of the ensemble by encouraging diverse predictions from its multiple heads in the last feed-forward layer. 
We then introduce \probconservno, a method that uses these well-calibrated UQ estimates within the \probconserv framework to update the model.
Our empirical results show that \probconservno enhances OOD model performance for a variety of challenging PDE problems and satisfies physical constraints such as conservation laws.
\end{abstract}


\newcommand{\figsizeintro}{0.32}
\newcommand{\figscaleintro}{0.37}
\newcommand{\figsizeexp}{0.35}
\newcommand{\figscaleexp}{0.37}
\newcommand{\figsizeapp}{0.3}
\newcommand{\figscaleapp}{0.28}

\section{Introduction}

A promising approach to scientific machine learning (SciML) involves so-called Neural Operators (NOs)~\citep{lu2019deeponet,li2020fourier,kovachki2021neural,li2020graphkernel,gupta2021multiwavelet, nicholas2023} (as well as other operator learning methods such as DeepONet~\citep{lu2019deeponet}).
These data-driven methods use neural networks (NNs) to try to learn a mapping  from the input data, e.g., initial conditions, boundary conditions, and partial differential equation (PDE) coefficients, to the PDE solution.
Advantages of NOs are that, if properly designed, they are discretization-invariant and can learn a mapping usable on different underlying discrete meshes.
Furthermore, NOs can be orders of magnitude faster at inference than conventional numerical solvers, especially for moderate accuracy
levels. Once trained, they can solve a PDE for different values of the physical PDE parameters efficiently. 
For instance, consider the 1-d heat equation, $\partial u/\partial t = k\cdot \partial^2u/\partial x^2$, where $u(x, t)$ denotes the solution as a function of 
space $x$ and time $t$, and where $k$ denotes the diffusivity. 
A NO can be trained 
to learn the parameter mapping between the diffusivity parameter $k$ (input) and the solution $u(x,t)$ (output) for fixed initial and boundary conditions.

\begin{figure}[t]
    \hspace{-0.15in}
    \centering
    \begin{subfigure}[t]{\figsizeintro\textwidth}
    \centering
    \includegraphics[scale=\figscaleintro]{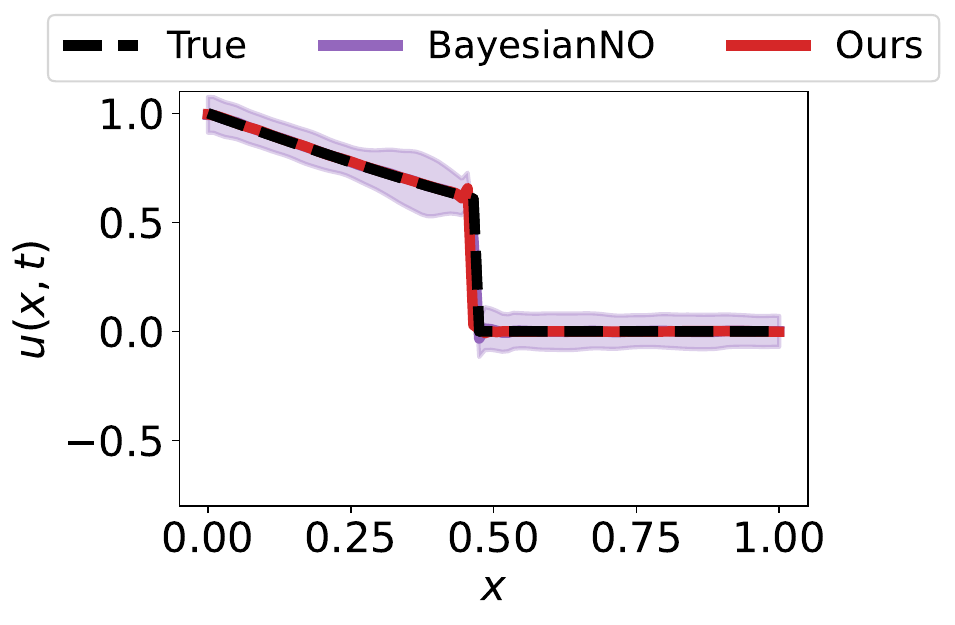}
    \caption{In-domain}
    \label{fig:solutions_stefan_id_bno}
    \end{subfigure}
    ~~~~
    \begin{subfigure}[t]{\figsizeintro\textwidth}
    \centering
    \includegraphics[scale=\figscaleintro]{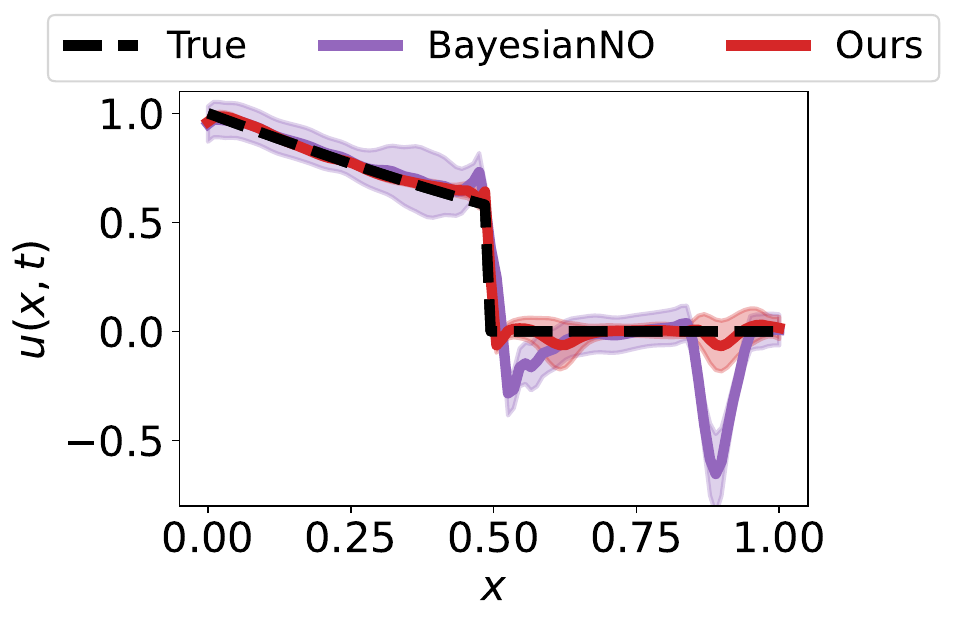}
    \caption{OOD}, 
 \label{fig:solutions_stefan_ood}
    \end{subfigure}
    ~~~~
    \caption{{\bf 1-d Stefan Equation.} In-domain (a) and OOD (b) predictions from FNO with post-hoc uncertainty estimates (3 standard deviations) obtained by \bayesiannomethod~\citep{magnani2022} and the proposed method at time $t = 0.5$. 
    \bayesiannomethod outputs an accurate solution with reasonable uncertainty bounds in-domain (a), where the training and test input parameters are drawn from the same distribution, i.e., $u{^*}^{\tr}, u{^*}^{\te} \in [0.6,0.7]$. 
    It fails to do so when the test inputs are OOD (b), 
    with $u{^*}^{\te} \in [0.5,0.55]$. The large oscillation occurs at different locations for different seeds.
    Proposed method (red) outputs accurate solution in-domain and improves the OOD predictions. 
    } 
\label{fig:solutions_stef_mot}
    \vspace{-.2cm}
\end{figure}

NOs have been shown to approximate the ground truth solution operator well for in-domain tasks ~\citep{lu2019deeponet,li2020fourier, saad2022guiding, alesiani2022}.
However, to be useful, they also need to be robust for out-of-domain (OOD) applications.
The robustness of OOD predictions is necessary in practical applications, as the test-time 
PDE parameters are typically not known during training.  
\cref{fig:solutions_stef_mot} shows the sharp moving (discontinuous) shock solution to the ``hard'' Stefan problem, where the PDE parameter $u^*$ denotes the solution value at the shock point. We see that while a NO model trained to map the input PDE parameter $u^*$
to the solution of the Stefan problem accurately captures the sharp dynamics for in-domain values of $u^*$, 
it fails for OOD values of $u^*$, i.e., those values of $u^*$ that are outside the range covered by the training data. 
This failure 
shows that NOs are not robust in OOD scenarios and thus may not effectively capture the ``true'' underlying operator map in practical applications.  
It is important when developing these models to use reliable mechanisms to detect and correct such inaccurate predictions, as this limits their widespread deployment in critical scientific applications.

To identify and correct OOD robustness issues, tools from 
regression diagnostics \citep{ChatterjeeHadi88}  and uncertainty quantification (UQ) \citep{Schwaiger2020IsUQ} may be used. Recent work \citep{magnani2022,psaros2023Uncertainty} has shown that several UQ methods (Bayesian approaches, ensembles, variational methods)~\citep{graves2011practical, gal2016dropout,lakshminarayanan2017Simple,teye2018bayesian, yang2022} originally proposed for standard NNs provide well-calibrated uncertainty estimates for NOs when the test inputs are in-domain. 
In parallel with this, \citet{hansen2023learning} proposed \probconserv, a method to use well-calibrated uncertainty estimates to help constrain PDE solutions to satisfy conservation laws.   
A reliable UQ framework can offer valuable insights more generally: into sensitivity analysis; to identify regions with high sensitivities and study how perturbations (epistemic or aleatoric) impact the PDE solution \citep{xiu2002, maitre2012, Rezaeiravesh2020}; and potentially to enhance the accuracy and reliability of the original PDE method. 

Problems with the existing UQ methods for NOs are that they either \textbf{(i)} fail to provide good uncertainty estimates for NOs for OOD predictions, or \textbf{(ii)} are computationally expensive, limiting the practical advantages of NOs over classical solvers.
For example, \cref{fig:solutions_stefan_id_bno} shows that the Bayesian Neural Operator (\bayesiannomethod)~\citep{magnani2022} captures the associated uncertainty accurately for in-domain predictions, while \cref{fig:solutions_stefan_ood} shows that it 
is inaccurate for OOD tasks. 
This example shows that the uncertainty estimates from existing UQ methods may not be correlated with the prediction errors OOD, highlighting the need for improved UQ metrics and methods to detect this issue. 

In this work, we address this challenge. 
We start by identifying failure modes of most existing UQ methods for NOs OOD, and we 
demonstrate that ensembling methods outperform other methods in this context, characterizing the
benefits of the output diversity. However, ensembling can be expensive.
Inspired by the success of ensembling, we propose \method, a method for scalable UQ for NOs that detects and reduces OOD prediction errors.  
Lastly, we use these uncertainty estimates within the recently-developed \probconserv framework \citep{hansen2023learning}  
in \probconservno, 
which further improves the OOD performance by satisfying known physical constraints of the problem. 
Our main contributions are as~follows: 
\begin{itemize}[noitemsep,topsep=0pt]
    \item 
    We identify an important challenge when using NOs to solve PDE problems in practical OOD settings. 
    In particular, we show that NOs fail to provide accurate solutions when test-time PDE parameters are outside the domain of the training data, even under mild domain shifts and when the model performs well in-domain. 
    We also show that existing UQ methods for NOs (e.g., \bayesiannomethod~\citep{magnani2022}) that provide ``good'' in-domain uncertainty estimates fail to do so for OOD and/or are computationally very expensive. 
    (See \cref{subsec:ood_failures_uq}.) 
    \item 
    We demonstrate empirically that ensembles of NOs provide improved uncertainty estimates OOD, compared with existing UQ methods; and we identify diversity in the predictions of the individual models as the key reason for better OOD UQ. 
    (See \cref{subsec:why_ensembles_good_uq}.) 
    \item
     We propose \method, a simple cost-efficient alternative to ensembling that encourages diverse predictions (via regularization) to mimic the OOD properties of an ensemble.
    (See \cref{subsec:proposed}.)
    \item  
    We use the error-correlated uncertainty estimates from \method as an input to the 
    \probconserv framework \citep{hansen2023learning}. 
    Our resulting \probconservno uses the variance information to correct the prediction to satisfy known physical constraints, e.g., conservation laws. 
    (See \cref{subsec:probconservno}.)
    \item 
    We provide an extensive empirical evaluation across a wide range of OOD PDE tasks that shows \method achieves $2\times$ to $70\times$ improvement in the meaningful UQ metric n-MeRCI compared to other computationally cheap UQ methods. Cost-performance tradeoff curves demonstrate its computational efficiency. We also show that using the uncertainty estimates from \method in \probconservno further improves the OOD accuracy by up to $34\%$. 
    (See \cref{sec:results}.)
\end{itemize}

\section{Background and Problem Setup} 
\label{sec:problem}

In this section, we introduce the basic problem, and we provide background on relevant UQ and sensitivity analysis. 

\allowdisplaybreaks
\subsection{Problem Definition}
PDEs are used to describe the evolution of a physical quantity \(u\) with respect to space and/or time.
The general (differential) form of a PDE is given as:
\begin{align}
\label{eq:pde}
\mathcal{F}_\phi u(x, t) &= 0, \qquad \forall x\in \xdomain, \forall t\in \tdomain, \nonumber \\
u(x, 0) &= u_0(x), \qquad \forall x\in \xdomain, \nonumber \\
\boundaryop u(x, t) &= 0, \qquad \forall x\in \xboundary, \forall t\in \tdomain, 
\end{align}
where 
$\xdomain \subseteq \sR^d$ denotes a bounded domain with boundary $\xboundary$, 
$\mathcal{F}_\phi$ denotes a (potentially nonlinear) differential operator parameterized by $\phi: \xdomain \to \sR \in \Phi$ 
acting on the solution $u: \xdomain \times \tdomain \to \sR^{d_o}, u\in \mathcal{U}$ 
for some final time $T$, $u_0: \xdomain \to \sR \in \mathcal{U}_0$ denotes the initial condition at time $t=0$, $\boundaryop$ 
denotes the boundary constraint operator; 
here $\Phi, \mathcal{U}_0, \mathcal{U}$ denote appropriate Banach spaces. 
Our basic goal is to learn an operator $\operator: \Phi \to \mathcal{U}$ that
accurately approximates
the mapping from PDE parameters $\phi$ to the solution $u$.   
We consider $\phi \in \tilde{\Phi} \subseteq \Phi$ so that the problem defined in \cref{eq:pde} is well-posed: it has a unique solution $u \in \mathcal{U}$ depending continuously on the input $\phi$ from $\tilde{\Phi}.$ 

\paragraph{Training distribution and data.} 
We assume that the operator $\operator$ is learned with an under-specified training distribution $\mathcal{H}^\tr(\phi)$, i.e., the support of $\mathcal{H}^\tr(\phi)$ does not cover the entire subspace $\tilde{\Phi} \subset \Phi$ of inputs for which the problem is well-posed.
Samples from this distribution form our training data $\cD^\tr = \{\phi^{(i)}, u^{(i)}\}_{i=1}^N$, where $\phi^{(i)} \sim \mathcal{H}^\tr(\phi)$, $u^{(i)} = \trueoperator(\phi^{(i)})$ and $\trueoperator$ denotes the ground-truth operator.
Practically, discrete approximations of the functions $\phi^{(i)}$ and $u^{(i)}$ are evaluated at a given time $t$ (for time-dependent problems) on a grid $\{x_l\}_{l=1}^L \subset \xdomain$, where $L$ denotes the number of gridpoints.

\paragraph{OOD test distribution.}
We define the OOD task, where the learned operator is tested on inputs $\phi^\te \sim \mathcal{H}^\te(\phi)$ from the test distribution $\mathcal{H}^\te(\phi)$, such that the $\supp(\mathcal{H}^\te(\phi)) \neq \supp(\mathcal{H}^\tr(\phi))$.  
We assume $\phi^\te$ is evaluated on the same grid $\{x_l\}_{l=1}^L$ as during training.

\subsection{Sensitivity Analysis for Operator Learning} 
\label{subsec:uq}

When used with inputs close to the support of the training data, NOs can provide cost-efficient and accurate surrogates to augment traditional PDE solvers. 
Away from these inputs, however, it is important to be able to estimate potential uncertainties in the predictions made by the NO.
Learning NN parameters has been viewed as a fully Bayesian problem \citep{teye2018bayesian, magnani2022, apostolos2022, dandekar2022}. 
To do this, one assumes a prior on the parameters, and then one adopts a statistical model to describe the generation of the training pairs and their relationship to the NN, defining the likelihood.
Pursuing a fully Bayesian approach to learning the parameters of a NN is arguably unwise for (at least) three reasons: 
\textbf{(i)} it is computationally expensive; 
\textbf{(ii)} uncertainty in the
parameters of the NO does not necessarily translate into uncertainty in outputs; and
\textbf{(iii)} the model likelihood is likely to be mis-specified.
(These are in addition to the fact that nontrivial issues arise when the NNs are overparameterized \citep{hodgkinson2022,hodgkinson_IIC_TR}.)
Viewing the problem through a Bayesian lens, even if not adopting a fully Bayesian approach, can be helpful. 
For example, using a Bayesian perspective, and the idea of collections of candidate solutions that match the data, facilitates the study of the sensitivities of learned NNs
to perturbations of various kinds. 
In particular, for PDE operator learning, this approach has the potential to uncover regions of the physical domain which are most sensitive to perturbations resulting from deploying a NO outside the support of the training data.

We now describe the Bayesian model that we adopt. 
Consider the training data $\cD^\tr = \{\phi^{(i)}, u^{(i)}\}_{i=1}^N$ with $u^{(i)} := \trueoperator(\phi^{(i)})$.  
On the basis of this data, we 
attempt to learn the parameter $\rmW$ of a NO $\operator(\cdot; \rmW)$ so that
$\operator(\cdot; \rmW) \approx \trueoperator(\cdot).$
To this end, we place the prior $p(\rmW) = \mathcal{N}(\rmW; \vzero, \frac{1}{\alpha^2} \rmI)$
on the parameters for some $\alpha \in \mathbb{R}$. 
We can assume that the training data is given by
$
u^{(i)} = \operator(\phi^{(i)}; \rmW)+\eta,
$
where $\eta$ denotes a mean zero Gaussian random variable with block-diagonal covariance,
where the identity blocks are scaled, for each $i$, by the size of the ground truth operator, 
generating the data $\|\trueoperator(\phi^{(i)})\|_2^2.$ 
This model accounts for the fact that the training data is not actually drawn from a realization of the NN, and it leads to the likelihood
$$
p(u^{(i)} | \rmW, \phi^{(i)}) = \mathcal{N}(u^{(i)}; \operator(\phi^{(i)}; \rmW), \frac{1}{\alpha^2}\|\trueoperator(\phi^{(i)})\|^2_2 \rmI).
$$
 From Bayes Theorem, we obtain the posterior
 
\begin{equation}
    \posterior \propto \prod_{i=1}^N p(u^{(i)} | \rmW, \phi^{(i)})p(\rmW).
    \label{eqn:posterior}
\end{equation}
Estimating the exact posterior is intractable for most practical tasks. 
A common practice is to use the maximum a posteriori (MAP) estimate, $\rmW^\text{MAP} = \argmax \posterior$, ignoring the uncertainty information and obtaining a single point prediction.
 
If an approximation of the $\posterior$ given in \cref{eqn:posterior} is known, then
the probability distribution on output predictions $u^\te$ from a test input $\phi^\te$ 
can be obtained by the Bayesian model average (BMA), 
\begin{align}\label{eq:bma}
p(u^\te | \phi^\te, \cD^\tr) = \int_\rmW p(u^\te | \rmW, \phi^\te) \posterior d\rmW.
\end{align}
To capture the uncertainty in predictions, several approximate inference techniques have been proposed~\citep{magnani2022, lakshminarayanan2017Simple, graves2011practical,teye2018bayesian, gal2016dropout}; these use a Monte Carlo approximation of \cref{eq:bma}, i.e., $p(u^\te | \phi^\te, \cD^\tr) \approx \frac{1}{J} \sum_{j=1}^J p(u^\te | \rmW_j, \phi^\te)$ with $\rmW_j \sim \posterior$, and they differ in the (approximate) procedure used to obtain samples $\rmW_j$ from the posterior. 
(See related work in \cref{sec:related_work} for details.) 

\paragraph{Effect of training data underspecification.}
A key difference between our OOD setting and the traditional setting is the under-specification of the training data, i.e., the support of $\mathcal{H}^\tr(\phi)$ does not cover the entire space $\tilde{\Phi}$.  
There are many NOs that can achieve close to zero training loss while disagreeing on OOD inputs. 
The posterior $\posterior$  has multiple modes, and choosing only one of these models can result in losing uncertainty information (regarding the predictions) that is important for the OOD test inputs.

There is a large body of work on numerical and SciML methods for solving PDEs, UQ for NOs, and diversity in ensembles; see \cref{sec:related_work} for a detailed discussion.

\section{Do Existing Methods Give Good OOD~UQ?}  
\label{sec:ensembles_best_uq}
In this section, we use 
the 1-d heat equation
as an illustrative example to evaluate existing UQ methods~\citep{magnani2022,lakshminarayanan2017Simple,gal2016dropout} with the aim of determining whether uncertainty estimates from these methods are robust to OOD shifts. 
We consider the Fourier Neural Operator (FNO)~\citep{li2022fourier} as our base model, and we evaluate the OOD performance of several UQ methods. We show that \ensemblenomethod, which is based on DeepEnsembles in \citet{lakshminarayanan2017Simple}, performs better than the other UQ methods OOD.
\begin{figure*}[t]
    \centering
    \begin{subfigure}{0.22\textwidth}
    \centering
    \includegraphics[scale=0.27]{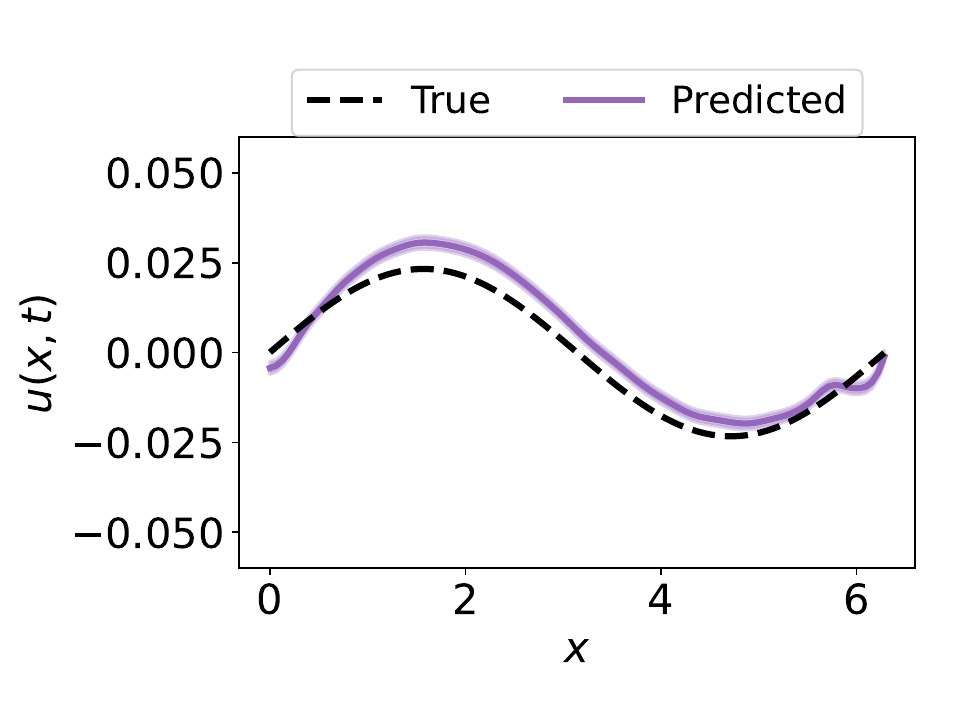}
    \caption{\bayesiannomethod}
    \end{subfigure}
    ~~~~
    \begin{subfigure}{0.22\textwidth}
    \centering
    \includegraphics[scale=0.27]{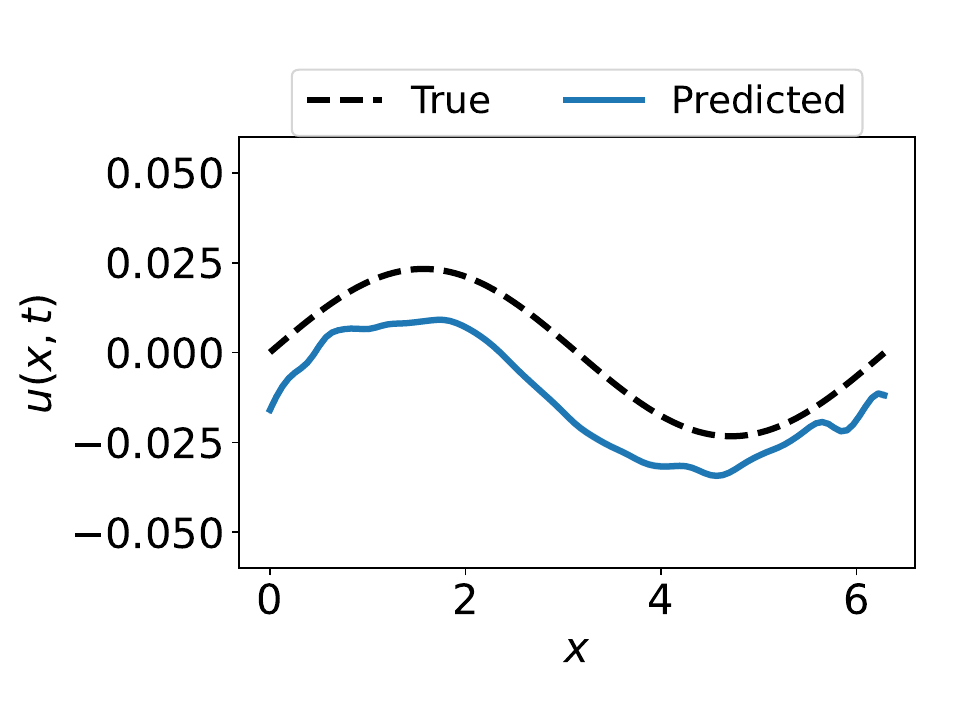}
    \caption{\outputvarmethod}
    \end{subfigure}
    ~~~~
    \begin{subfigure}{0.22\textwidth}
    \centering
    \includegraphics[scale=0.27]{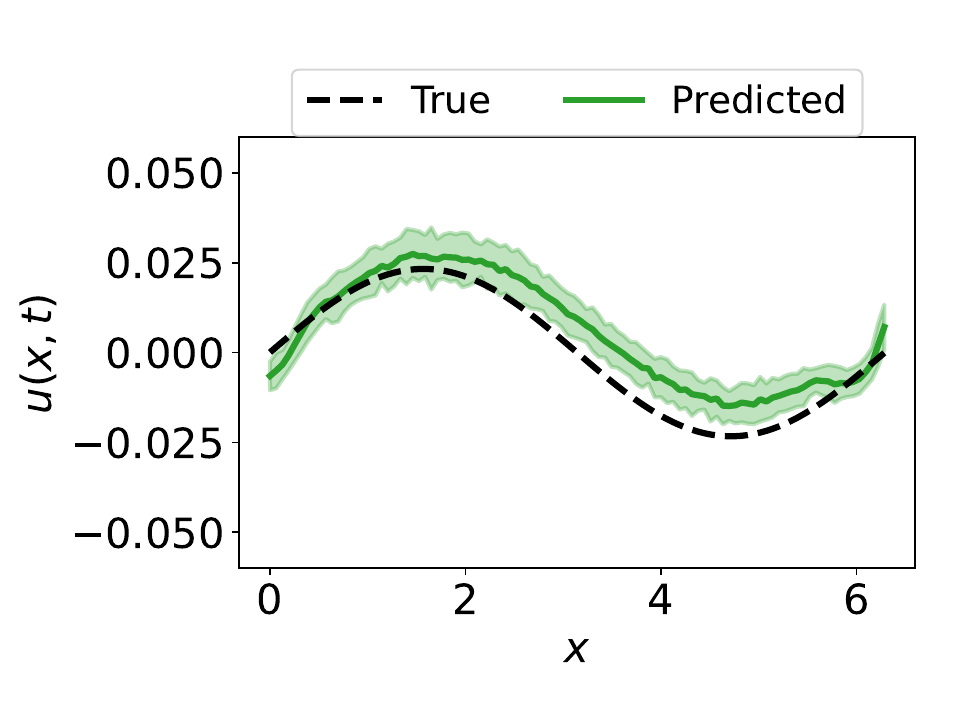}
    \caption{\mcdropoutnomethod}
    \end{subfigure}
    ~~~~
    \begin{subfigure}{0.22\textwidth}
    \centering
    \includegraphics[scale=0.27]{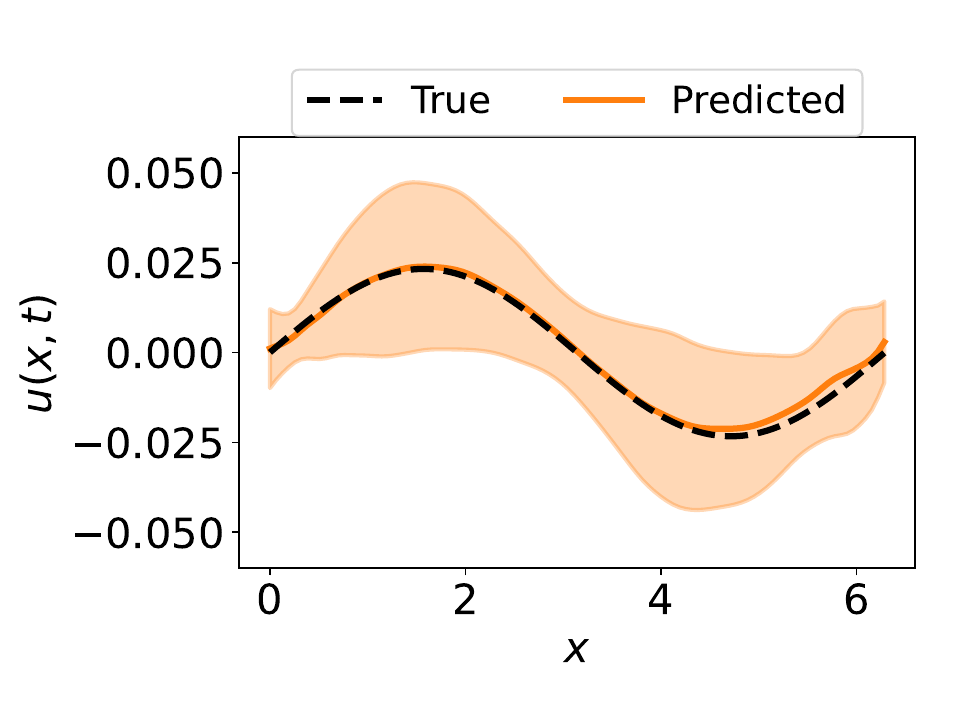}
    \caption{\ensemblenomethod}
    \end{subfigure}
    \caption{{\bf 1-d Heat equation, large OOD shift.} 
    Uncertainty estimates (3 standard deviations) from various UQ methods under large OOD shifts in the input diffusivity coefficient, where $k^\tr\in[1,5], k^\te\in[7,8]$. The  \ensemblenomethod prediction, which is based on DeepEnsembles \citep{lakshminarayanan2017Simple}, is contained within its uncertainty estimate, whereas the predictions from the other methods have narrow uncertainty estimates and are inaccurate on this ``easy'' task. 
    } 
    \label{fig:solutions_heat_ood3}
\end{figure*}

\subsection{OOD Failures of Existing UQ methods} \label{subsec:ood_failures_uq}
While the existing UQ methods perform well in-domain (e.g., see \bayesiannomethod in~\cref{fig:solutions_stefan_id_bno}), we identify several canonical cases where they fail OOD. 
\cref{fig:solutions_heat_ood3} shows the solution profiles and uncertainty estimates to the 1-d heat equation for a large OOD shift, where $k^{\tr} \in [1,5]$ and $k^{\te} \in [7,8]$.  We see that the predictions are inaccurate for most of the UQ methods, with the solution not contained within the uncertainty estimate, other than for \ensemblenomethod.   
Quantitatively, we use the Normalized Mean Rescaled Confidence Interval (n-MeRCI)~\citep{moukari2019n} (defined formally in \cref{eq:nmerci} in Section \ref{sec:results}), which measures how well the uncertainty estimates are correlated with the prediction errors, with lower values indicating better correlation. 
Surprisingly, 
\ensemblenomethod achieves significantly lower n-MeRCI value (e.g., 0.05 vs 0.8 in the heat equation) compared to the other UQ methods across various PDEs and OOD shifts. 
In contrast to the other UQ methods that output worse uncertainty estimates as the OOD shift increases, the ensemble model consistently outputs uncertainty estimates that are correlated with the prediction errors.
(See \cref{subsec:ensemble_perf_app} for similar results on a range of PDEs.)

\subsection{How do Ensembles Provide Better OOD UQ} 
\label{subsec:why_ensembles_good_uq}

\begin{figure*}[h]
    \centering
    \begin{subfigure}{0.22\textwidth}
    \centering
    \includegraphics[scale=0.5]{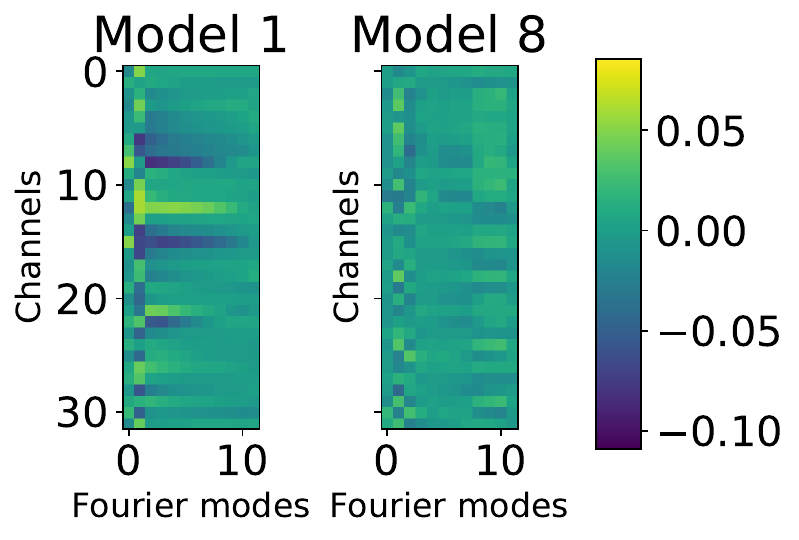}
    \caption{Heatmaps of first Fourier layer weights of two diverse FNO models.}
\label{fig:ensemble_diversity:heatmaps}
    \end{subfigure}
    ~~~~~~~~
    \begin{subfigure}{0.22\textwidth}
    \centering
    \includegraphics[scale=0.4]{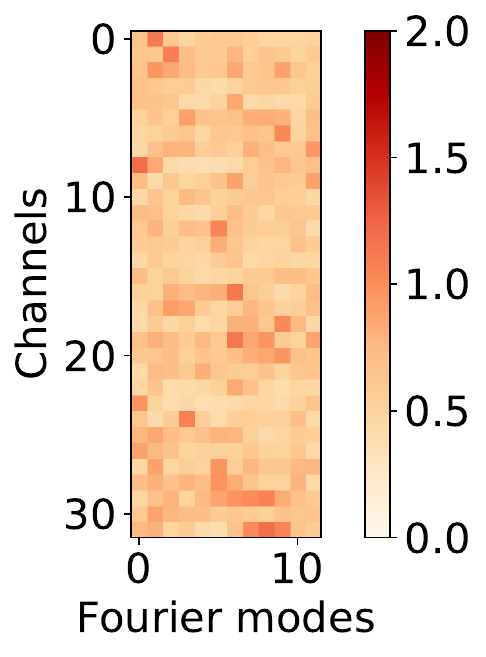}
    \caption{Coefficient of variation (std/mean) of the first Fourier layer weights}
    \label{fig:ensemble_diversity:coefvar}
    \end{subfigure}
    \hfill
    \begin{subfigure} {0.5\textwidth}
        \centering
        \begin{subfigure}[t]{0.45\textwidth}
         \centering
        \includegraphics[scale=0.22]{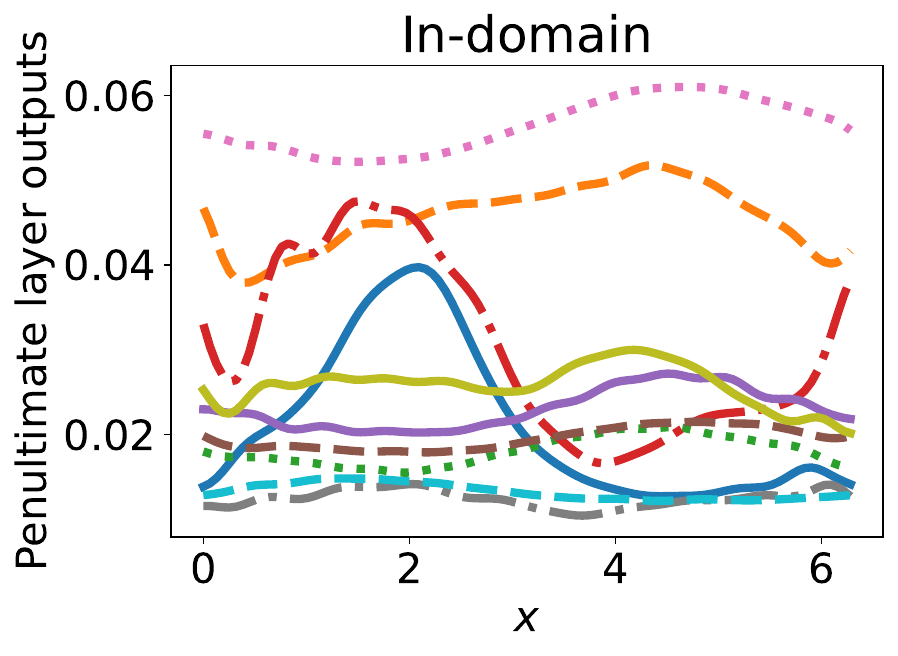}
        \caption{ID penultimate layer}
        \label{fig:ensemble_diversity_idpen}
        \end{subfigure}
        \begin{subfigure}[t]{0.45\textwidth}
         \centering
        \includegraphics[scale=0.22]{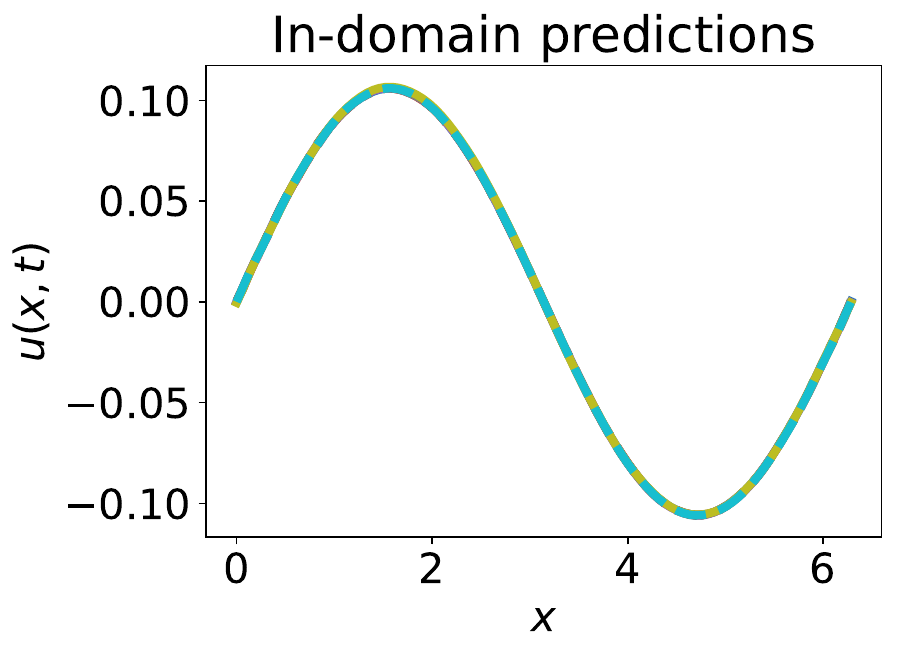}
         \caption{ID predictions}
        \label{fig:ensemble_diversity_idoutputs}
        \end{subfigure} 
        
        \begin{subfigure}[t]{0.45\textwidth}
         \centering
        \includegraphics[scale=0.22]{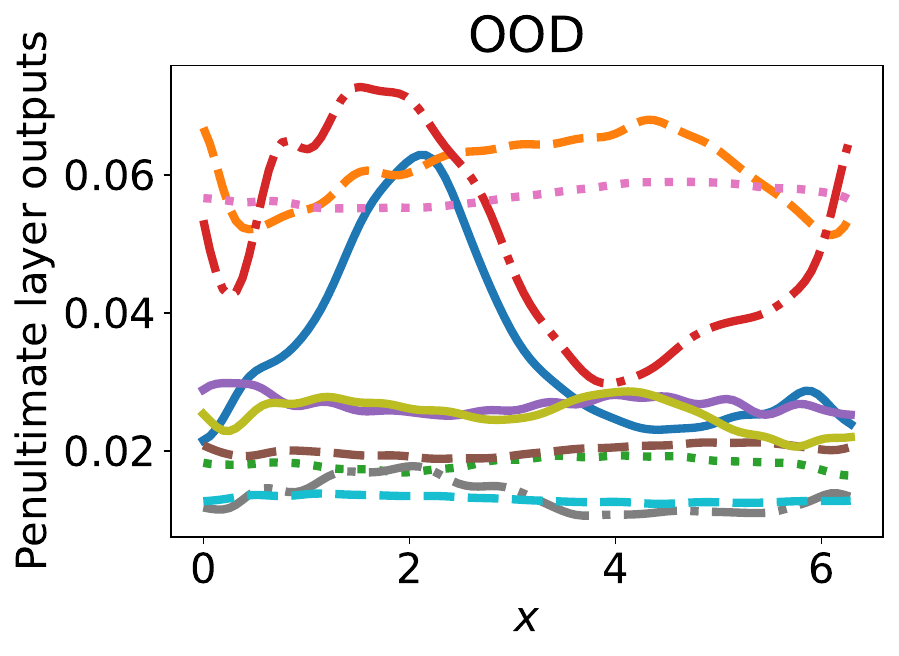}
        \caption{OOD penultimate layer}
        \label{fig:ensemble_diversity_oodpen}
        \end{subfigure}
        \begin{subfigure}[t]{0.45\textwidth}
         \centering
        \includegraphics[scale=0.22]{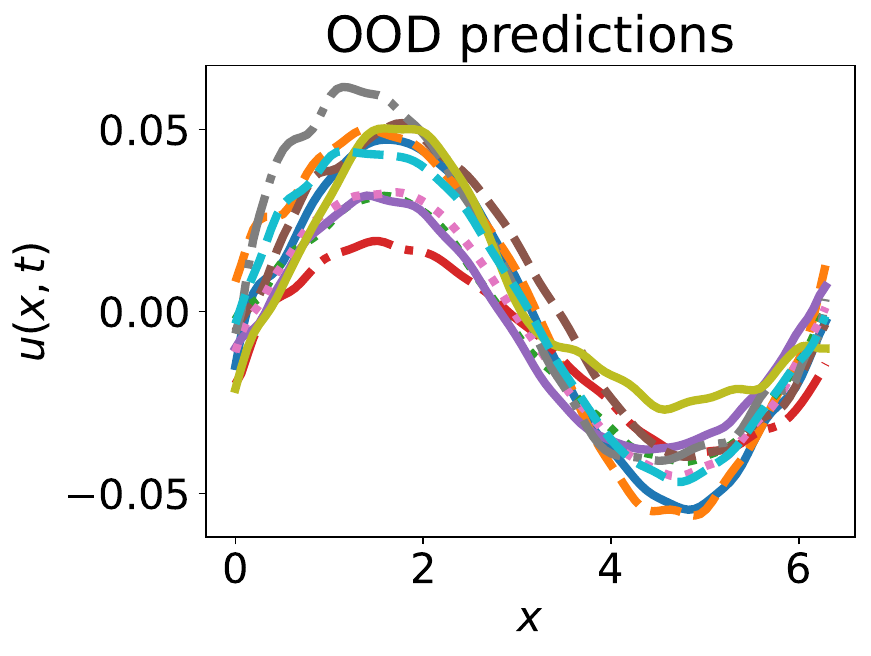}
         \caption{OOD predictions} 
    \label{fig:ensemble_diversity_oodoutputs}
        \end{subfigure} 
    \end{subfigure}
    \caption{{\bf{Diversity of Models in an Ensemble of NOs (\ensemblenomethod).}} 
    (a) Heatmaps of the weights,
    (b) coefficient of variation of the 10 models in the ensemble, and (c-f) penultimate and last layer outputs on a 1-d heat equation task. 
    (c-d) For in-domain (ID) input, the last layer removes the diversity from the intermediate outputs and maps the final output close to the ground truth. (e-f) For OOD input, this diversity remains in the output. 
    }
    \label{fig:ensemble_diversity}
\end{figure*}

In this subsection, we study the reason behind significantly better OOD uncertainty estimates from a seemingly simple ensemble of NOs i.e., \ensemblenomethod. 
\citet{wilson2020Bayesian} posit that ensembles are able to explore the different modes of the posterior better than other UQ methods that use, e.g., variational inference or a Laplace approximation, since these methods are only able to explore a single posterior mode due to the Gaussian assumption. 
They show that samples of weights $\rmW_j$ from a single posterior mode are not functionally very diverse, i.e., the corresponding predictions from these models $\operator(\cdot; \rmW_j)$ are not very different from each other. 
This leads to a suboptimal estimation of the BMA in \cref{eq:bma}, as the Monte Carlo sum contains redundant terms. In contrast, individual models of the ensembles reach distinct posterior modes due to the random initialization and the noise in SGD training. 
The resulting functional diversity allows the ensemble to better estimate \cref{eq:bma}. Next, we show that this diversity with the models in the ensemble holds empirically for operator learning as well. 

In \cref{fig:ensemble_diversity}, we illustrate the diversity in operator learning by visualizing the differences in the Fourier layer weights  and final layer outputs for the models in the ensemble on a 1-d heat equation task. (See \cref{app:heatmaps} for analogous results across a variety of PDEs and Fourier layers.) 
Different models seem to focus on different spectral characteristics of its input function. 
For example, the heatmaps of the weights in \cref{fig:ensemble_diversity:heatmaps} show that the first model (\textit{left}) uses the low-frequency components across most channels except a few, whereas another model (\textit{right}) uses the available Fourier components across all the channels.  \cref{fig:ensemble_diversity:coefvar} illustrates that this model diversity holds for all models in the ensemble since the coefficient of variation of the Fourier layer weights computed across the ten models is large. 

The intermediate outputs from the different models in the ensemble 
also show diversity in operator learning until the last layer with a key difference between the in-domain and OOD final outputs. 
Given the same in-domain input, the first feed-forward (lifting) layer of these models already produces diverse outputs which continues until the penultimate layer (\cref{fig:ensemble_diversity_idpen}). 
\cref{fig:ensemble_diversity_idoutputs}  shows that the last layer removes the diversity and maps the outputs to the same ground truth for an in-domain input.
\cref{fig:ensemble_diversity_oodpen,fig:ensemble_diversity_oodoutputs} illustrates that a different trend occurs for an OOD input. 
The OOD intermediate outputs are also diverse, 
the difference here is that this diversity does not vanish from the last layer outputs since the model is not able to match the ground truth data.
This representative example demonstrates how the ensemble model captures epistemic uncertainty in the OOD setting. 
We see that the individual models in the ensemble fit the training data with close to zero error, but are  diverse and disagree for OOD inputs. 
Ensembles can approximate the BMA in \cref{eq:bma} with functionally diverse models from the posterior, leading to better OOD uncertainty estimates.

\section{\method + \probconserv}
In this section, we first present our \method model, which encourages diverse OOD predictions, while being significantly computationally cheaper than \ensemblenomethod~\citep{lakshminarayanan2017Simple}. 
We then feed these black-box NO uncertainty estimates into the first step of the probabilistic \probconserv framework \citep{hansen2023learning} to enforce physical constraints and further improve the OOD model performance. As opposed to in \probconserv, here the constraint is applied on the OOD predictions.

\subsection{\method: Computationally Efficient OOD UQ} 
\label{subsec:proposed}
The primary challenge with ensembles is their high computational complexity during both training and inference. 
An ensemble-based model for surrogate modeling diminishes the computational benefits gained from using a data-driven approach over classical solvers.  
Here, we propose \method, a method which makes a simple modification
to the NO architecture, along with a diversity-enforcing regularization term, to emulate the favorable UQ properties of \ensemblenomethod, while being computationally cheaper. 

For the architecture change, we modify the last feed-forward layer to have $M$ output/prediction heads instead of one. (See \cref{subsec:ablation_num_heads} for a hyperparameter study on the choice of $M$.) 
The architecture may be viewed as an ensemble with the individual models sharing parameters up to the penultimate layer. 
While parameter-sharing significantly reduces the computational complexity, when compared to using a full ensemble, it simultaneously hinders the diversity within the ensemble, which we showed is a crucial element for generating good uncertainty estimates (see \cref{subsec:why_ensembles_good_uq}).

To encourage diverse predictions, we propose to maximize the following diversity measure among the last-layer weights corresponding to the $M$ prediction heads.
We also constrain the in-domain predictions from each of these heads to match the ground truth outputs. 
Formally, we solve:
\allowdisplaybreaks
\begin{align} \label{eq:optimization}
\hat{\rmW}  = \argmin_{\rmW}  
\underbrace{\frac{1}{N M} \sum_{i=1}^N \sum_{m=1}^M{\frac{||\hat{u}_m^{(i)}-  u^{(i)}||^2_{L_2}} {||u^{(i)}||^2_{L_2}}}}_{\text{unconstrained NO loss}} 
\underbrace{- \frac{2\lambda_\diverse}{M(M-1)}\sum_{m,k:m< k} ||\rmW_m - \rmW_k||^2_2}_{\text{diversity regularization}},
\end{align}
where $\rmW_m$ and $\rmW_k$ denote the last-layer weights corresponding to $m$-th and $k$-th prediction heads, respectively, $\hat{u}_m^{(i)}$ denotes the prediction from $m$-th output head for the $i$-th training example, and $u^{(i)}$ denotes the corresponding ground truth.
The first term in \cref{eq:optimization} is the relative $L_2$ loss standard in NO training. We add the regularizing second term to encourage diversity in the last-layer weights corresponding to the different head. 
The hyperparameter $\lambda_\diverse$ controls the strength of the diversity regularization relative to the prediction loss. 
A naive selection procedure for $\lambda_\diverse$ using only in-domain validation MSE selects an unconstrained model with no diversity penalty.
Instead, we select the maximum regularization strength $\lambda_\diverse$ that achieves an in-domain validation MSE within 10\% of the identified best in-domain validation MSE. 
10\% is an arbitrary tolerance that denotes the \% accuracy the practitioner is willing to forego to achieve better OOD UQ.
This procedure trades off in-domain prediction errors for higher diversity that is useful for OOD UQ. 
See \cref{subsec:diversity_reg_app} for a hyperparameter study on the diversity regularization strength $\lambda_{\text{diverse}}$; uncertainty metrics monotonically improve with higher $\lambda_{\text{diverse}}$. 
\cref{subsec:diversity_reg_app} also includes ablations with regularizations for ensemble NNs that directly diversify the predictions~\citep{bourel2020boosting,zhang2020diversified}.

\subsection{\probconservno}
\label{subsec:probconservno}
We detail how to use uncertainty estimates for NOs, e.g., \method, within the \probconserv framework~\citep{hansen2023learning} to improve OOD performance and incorporate physical constraints known to be satisfied by the PDEs we consider. 

The two-step procedure is given as:
\begin{enumerate}[noitemsep,topsep=0pt]
    \item Compute uncertainty estimates $\mu, \Sigma$ from the NO;
    \item Use the update rule from \probconserv, described in \cref{eqn:updated_mean_var} below, to improve to the model.
\end{enumerate}

Given the predictions $\mu$ and the covariance matrix $\Sigma$, \probconserv solves the constrained least squares problem:
\begin{equation}
\begin{aligned}
\tilde{\mu} = \argmin_y \frac{1}{2}||y - \mu||^2_{\Sigma^{-1}} \quad \text{s.t.} \quad Gy=b,
\label{eqn:probconserv_opt}
\end{aligned}
\end{equation}
where $G$ denotes the constraint matrix and $b$ denotes the values of the constraints. For example, conservation of mass in the 1-d heat equation with zero Dirichlet boundary conditions is given by the linear constraint $\int_x u(x, t)dx = 0$ and $G$ can take the form of a discretized integral. 
The optimization in \cref{eqn:probconserv_opt} can be solved in closed form with the following update: 
\begin{subequations}
\label{eqn:updated_mean_var}
\begin{align}
\label{eqn:updated_mean_var_MEAN_app}
   \tilde \mu &= \mu - \Sigma G^T (G \Sigma G^T)^{-1} (G\mu - b), \\
   \label{eqn:updated_mean_var_VAR_app}
   \tilde \Sigma &= \Sigma - \Sigma G^T (G \Sigma G^T)^{-1} G \Sigma.
\end{align}
\end{subequations}

For the UQ methods for NOs considered here, $\Sigma$ denotes a diagonal matrix with the variance estimates on the diagonal. 
The update can be viewed as an oblique projection of the unconstrained predictions $\mu$ onto the constrained subspace, while taking into account the variance (via $\Sigma$) in the predictions. 
When the uncertainty estimates are not uniform, \probconserv encourages higher corrections in regions of higher variance, and it is important to have uncertainty estimates that are correlated with the prediction errors.

In the original implementation of \probconserv \citep{hansen2023learning}, the Attentive Neural Process (ANP) \citep{kimAttentiveNeuralProcesses2019} is used as an instantiation of the framework in \probconservanp. 
There has yet to be an operator model, which is typically more suitable for SciML problems, used within \probconserv. We show its effectiveness with NO uncertainty estimates and its application to OOD tasks.

\section{Empirical Results} 
\label{sec:results}

In this section, we evaluate \method 
against several baselines for using UQ in PDE solving, with a specific aim to answer the following questions: 
\begin{enumerate}[noitemsep,topsep=0pt]
    \item 
    Is diversity in the weights of the last layer of \method enough to obtain good uncertainty estimates? 
    (See \cref{subsec:uq_est_divno}.) 
    \item 
    What are the computational savings of \method compared to ensembling? 
    (See \cref{subsec:comp_savings}.)
    \item Does using error-correlated UQ in \probconserv help improve OOD prediction errors in PDE tasks?
    (See \cref{subsec:probconserv_res}.)
\end{enumerate}

\paragraph{Uncertainty metrics.} 
Several metrics have been used to evaluate the goodness of uncertainty estimates, e.g., negative log-likelihood (NLL), root mean squared calibration error (RMSCE), sharpness and continuous ranked probability score (CRPS)~\citep{gneiting2007probabilistic,gneiting2007,kuleshov2018accurate,psaros2023Uncertainty}. 
We use the normalized Mean Rescaled Confidence Interval (n-MeRCI) metric~\citep{moukari2019n}, 
that evaluates how well the uncertainty estimates are correlated with the prediction errors. 
The n-MeRCI metric is given as:      \begin{align}\label{eq:nmerci}
    \text{n-MeRCI} = \frac{\frac{1}{N} \sum_{i=1}^N \tau \sigma_i - \text{MAE}}{\max(|\hat{u}_i - u_i|) - \text{MAE}} ,
    \end{align}
    where MAE denotes the mean absolute error and $\tau$ 
    denotes the 95$^{\text{th}}$ percentile of the ratios $|\hat{u}_i - u_i|/\sigma_i$. 
    This percentile of the ratios scales the uncertainty estimates so that the measure is scale-independent and robust to outliers. 
    Values closer to zero correspond to better-correlated uncertainty estimates, whereas values closer to one or greater correspond to uncorrelated/random estimates. 
    See \cref{subsec:metrics_app} for the MSE, NLL, RMSCE and CRPS metrics. 

\paragraph{Baselines.} 
We use the Fourier Neural Operator (FNO)~\citep{li2022fourier} as a base model and leave similar investigations of other operators to future work. 
To provide uncertainty estimates in PDE applications, we compare our proposed UQ method, \method \footnote{The code is available at \url{https://github.com/amazon-science/operator-probconserv}.} with the following four commonly-used baselines of approximate Bayesian inference~\citep{psaros2023Uncertainty}:
\textbf{(i)}  \bayesiannomethod~\citep{magnani2022}, that uses the (last-layer) Laplace approximation over the MAP estimate to approximate the posterior; 
\textbf{(ii)} 
\outputvarmethod~\citep{lakshminarayanan2017Simple}, which outputs the mean and variance and which is trained with the negative log-likelihood;
\textbf{(iii)} \mcdropoutnomethod~\citep{gal2016dropout}, where dropout in the feed-forward layers is used as approximate variational inference; and 
\textbf{(iv)} \ensemblenomethod~\citep{lakshminarayanan2017Simple}, where we train $K$ randomly initialized FNO models (in our experiments, $K=10$) and compute the empirical mean and variance of the predictions.

\paragraph{GPME Benchmarking Family of Equations.}
The Generalized Porous Medium Equation (GPME) is a family of PDEs parameterized by a (potentially nonlinear) coefficient $k(u)$ ~\citep{maddix2018_harmonic_avg, maddix2018temp_oscill}. 
The GPME models fluid flow through a porous medium, and it has additional applications in heat transfer, groundwater flow, and crystallization, to name a few~\citep{vazquez2007porous}. 
It can be written in the conservative form with flux $F(u) = -k(u)\nabla u$~as:
\begin{align}\label{eq:gpme}
u_t - \nabla \cdot (k(u) \nabla u) = 0, \qquad x \in \Omega, t\in \tdomain, 
\end{align}
where $k(u): \xdomain \to \sR$ denotes the diffusion coefficient. 
We consider three instances of the GPME on a 1-d domain $\xdomain=[0, S]$.
By varying $k(u)$, these correspond to increasing levels of difficulty~\citep{hansen2023learning}: 
\textbf{(i)} an ``easy'' case, with $k(u) = k$, the standard heat equation (linear, constant coefficient, parabolic); 
\textbf{(ii)} a ``medium'' case, with $k(u) = u^m, m\geq 1$, the Porous Medium Equation (PME) (nonlinear, degenerate parabolic); and 
\textbf{(iii)} a ``hard'' case, with $k(u) = \vone_{u \geq u^*}, u^* > 0$, the Stefan problem (nonlinear, discontinuous, degenerate parabolic). 

We aim to learn an operator that maps the constant parameter $c$ identifying the diffusion coefficient $k(u)$---constant $k$ in the  heat equation, degree $m$ in the PME, and $u^*$ in the Stefan equation---to the solution $u(\cdot, T)$ for some $T > 0$. 
Input to the NO is a constant scalar field taking the value $c$ for all spatiotemporal points.
Our training dataset consists of $N=400$ input/output pairs $\{\phi^{(i)}, u^{(i)}\}_{i=1}^N$, where $\phi^{(i)}(x, t):=c^{(i)}$ denotes a constant function 
identifying the diffusion coefficient with value $c^{(i)}$ over the domain and $u^{(i)}$ denotes the corresponding solutions at discrete times $t\in[0,T]$. 
We keep the initial and boundary conditions fixed for all the examples of a particular PDE.
During training, we sample the constant parameter $c^{(i)} \sim \mathcal{H}^\tr(c)$  
and we evaluate the trained neural operator on OOD shifts $c^{(i)} \sim \mathcal{H}^\te(c)$, where $\mathcal{H}^\te(c)$ has no support overlap with $\mathcal{H}^\tr(c)$.
We consider small, medium and large OOD shifts for each PDE task based on the distance between $\mathcal{H}^\tr(c)$ and $\mathcal{H}^\te(c)$.
See \cref{app:test_probs} for details about the PDE test problems and \cref{app:exp_details} for additional experimental settings.

\begin{figure}[t]
    \centering
    \includegraphics[scale=0.4]{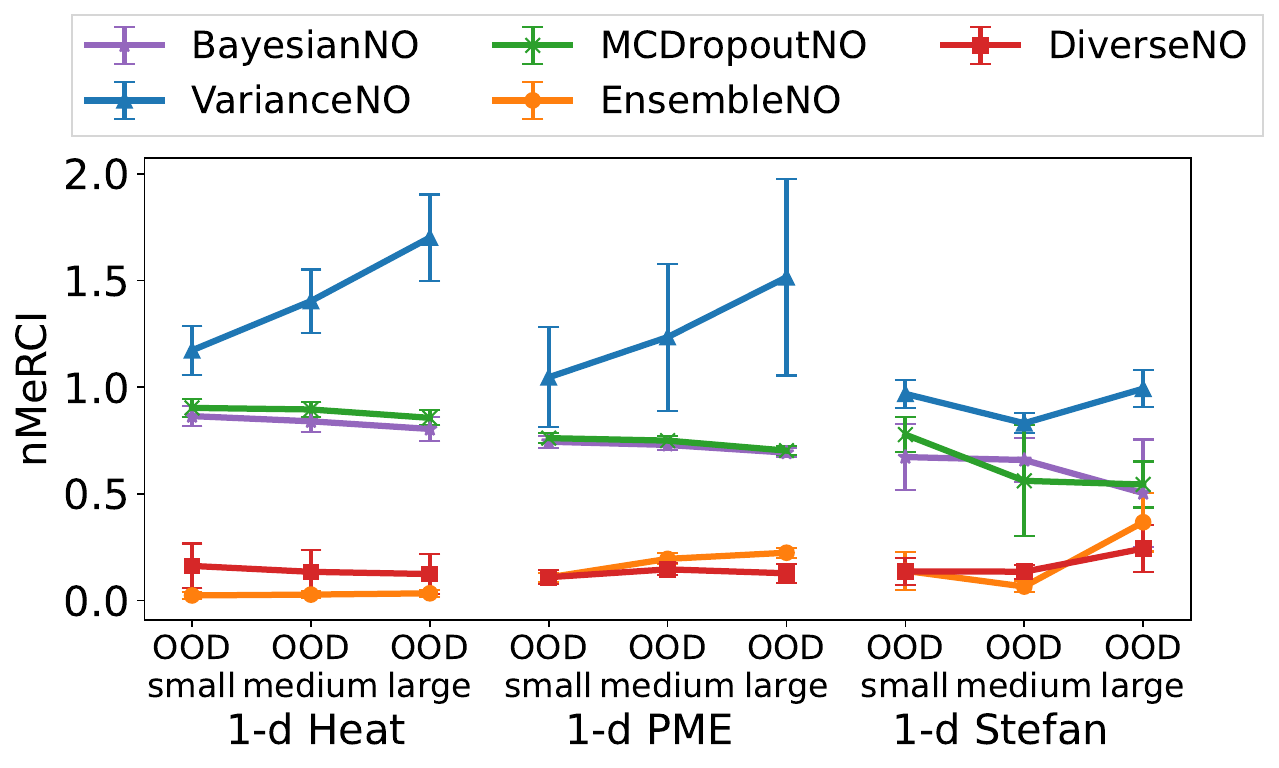}
    \label{fig:nmerci_ood}
    \caption{
    {\bf Performance vs. OOD shift.} n-MeRCI $\downarrow$ of the UQ methods with increasing OOD shifts in the ``easy,'' ``medium,'' and ``hard'' cases of the GPME.
     \ensemblenomethod and \method consistently achieve low n-MeRCI values, indicating higher correlation between their uncertainty estimates and prediction errors.
    } 
    \label{fig:perfvpdeood}
\end{figure}

\subsection{Uncertainty estimates from \method}
\label{subsec:uq_est_divno}
In \cref{sec:ensembles_best_uq}, we showed that \ensemblenomethod outputs better uncertainty estimates OOD due to the diversity in the predictions among each model in the ensemble.
Here, we evaluate whether \method can output good uncertainty estimates due to the diversity enforced over the last layer alone. 
\cref{fig:perfvpdeood} shows the n-MeRCI metric for all UQ methods on increasing OOD shifts on the ``easy'', ''medium'' and ``hard'' cases of GPME. 
\method consistently outperforms \bayesiannomethod, \outputvarmethod and \mcdropoutnomethod by $2\times$ to $70\times$ across all PDEs and OOD shifts, and is comparable (sometimes better) to the more expensive \ensemblenomethod that has diversity in weights of all layers (as seen in \cref{fig:ensemble_diversity}). 
The n-MeRCI metric is generally close to zero for both methods indicating that their uncertainty estimates are well-correlated with the prediction errors and can be used to detect OOD shifts. 
See \cref{subsec:metrics_app} for additional metrics, solution profiles, uncertainty~estimates and test problems, e.g., non-constant input and 2-d Darcy flow.

\subsection{Computational savings from \method}
\label{subsec:comp_savings} 
\begin{figure}[t]
    \hspace{-0.2in}
    \centering
    \begin{subfigure}{\figsizeintro\textwidth}
    \centering
    \includegraphics[scale=\figscaleintro]{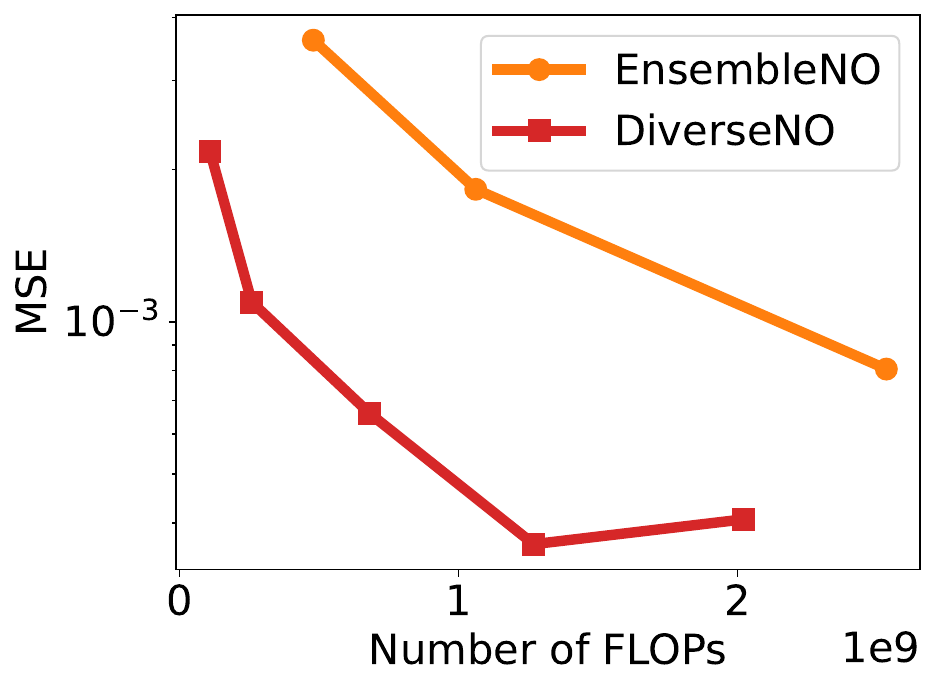}
    \caption{MSE $\downarrow$}
    \end{subfigure}
    ~~~~
    \begin{subfigure}{\figsizeintro\textwidth}
    \centering
    \includegraphics[scale=\figscaleintro]{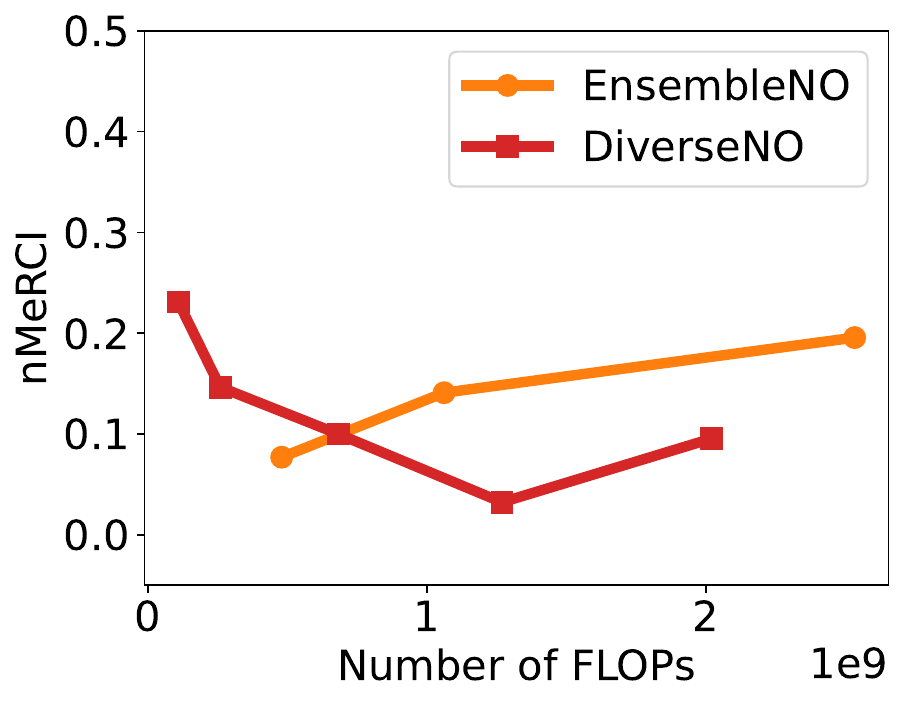}
    \caption{n-MeRCI $\downarrow$}
    \end{subfigure}
    \caption{{\bf Cost-performance tradeoff.} MSE $\downarrow$ and n-MeRCI $\downarrow$ vs. the number of floating point operations (FLOPs) for \ensemblenomethod and \method with varying number of parameters on the PME task with medium OOD shift, i.e., $m^\tr\in[2,3], m^\te\in[4,5]$.
    }
    \label{fig:costperf_heat_pme}
\end{figure}

\cref{fig:costperf_heat_pme} shows the computational savings of \method compared to \ensemblenomethod on the PME task with medium OOD shift, i.e., $m^\tr\in[2,3], m^\te\in[4,5]$. We plot the MSE and n-MeRCI metric as a function of the number of floating point operations (FLOPs) for different  
model sizes (using the derivation of FLOPs for FNO by~\citet{dehoop2022CostAccuracy}). We see that \method is significantly more efficient across the various PDEs: for similar number of FLOPs, \method is 49\% to 80\% better in MSE and comparable to \ensemblenomethod in the n-MeRCI metric. 
(See \cref{subsec:cost_perf_app} for the cost performance curves as a function of the number of parameters.) 

\subsection{OOD applications of UQ with \probconserv}
\label{subsec:probconserv_res}

\begin{figure}[t]
    \centering
    \includegraphics[scale=0.4]{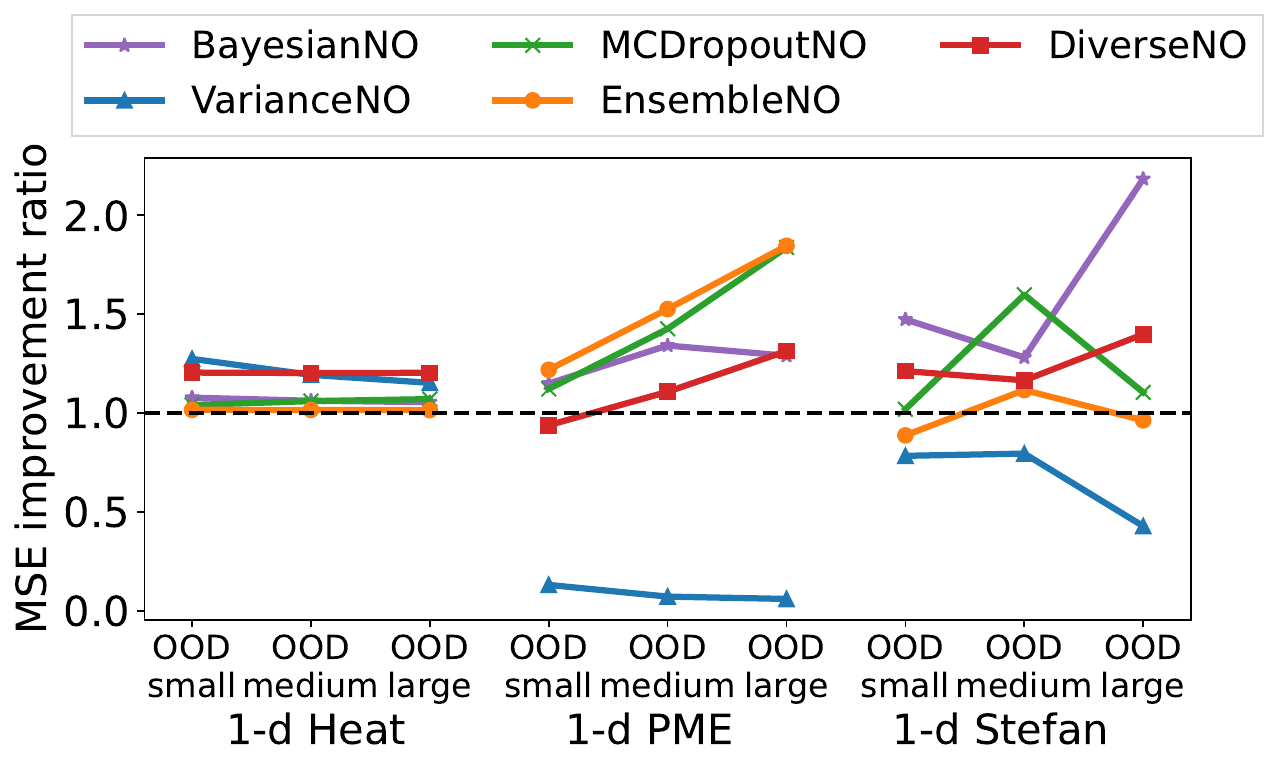}
    \caption{
    {\bf{MSE Improvement Ratio for \probconserv.}} The MSE ratio before and after applying \probconserv to the UQ methods ($\ge 1$ indicates improvement) for increasing OOD shifts in the ``easy'', ``medium'' and ``hard'' cases of the GPME.
    } 
    \label{fig:mse_improvement_probconserv}
\end{figure}

\begin{figure}[h]
    \centering
        \begin{subfigure}[t]{\figsizeexp\textwidth}
    \centering
    \includegraphics[scale=\figscaleexp] {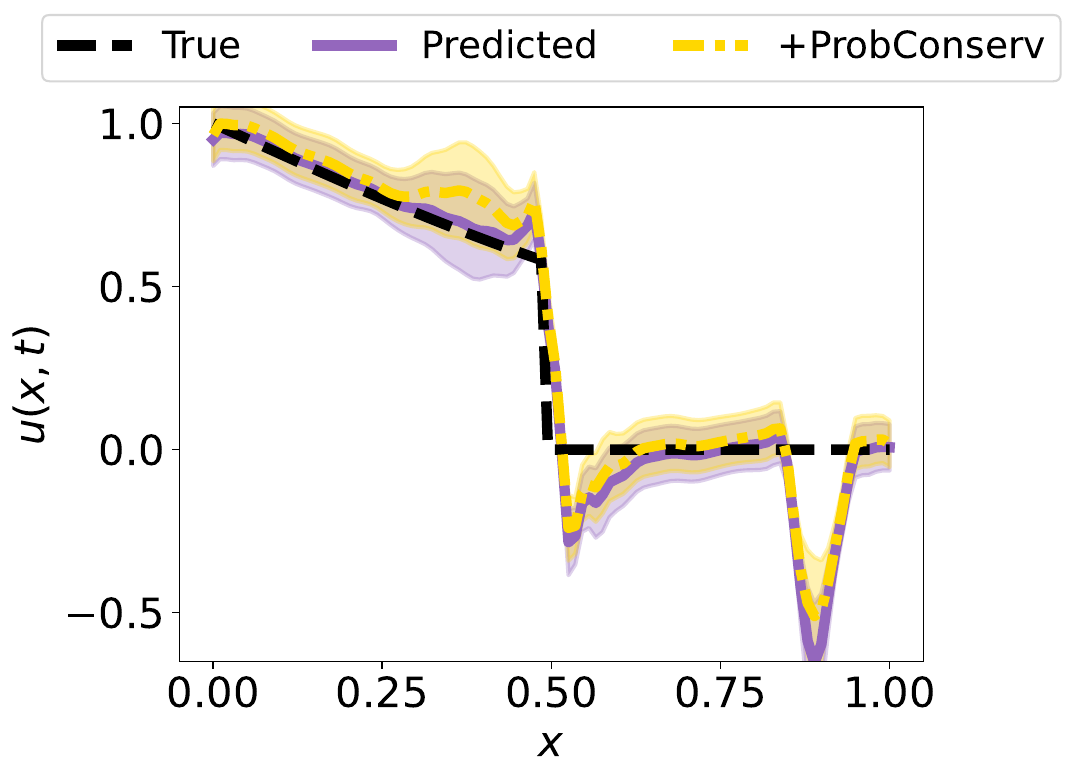}
    \caption{\bayesiannomethod+ \\ \probconserv}
    \label{fig:probconserv_bayesianno_stefan}
    \end{subfigure}
    ~~~~~~
    \begin{subfigure}[t]{\figsizeexp\textwidth}
    \centering
    \includegraphics[scale=\figscaleexp]{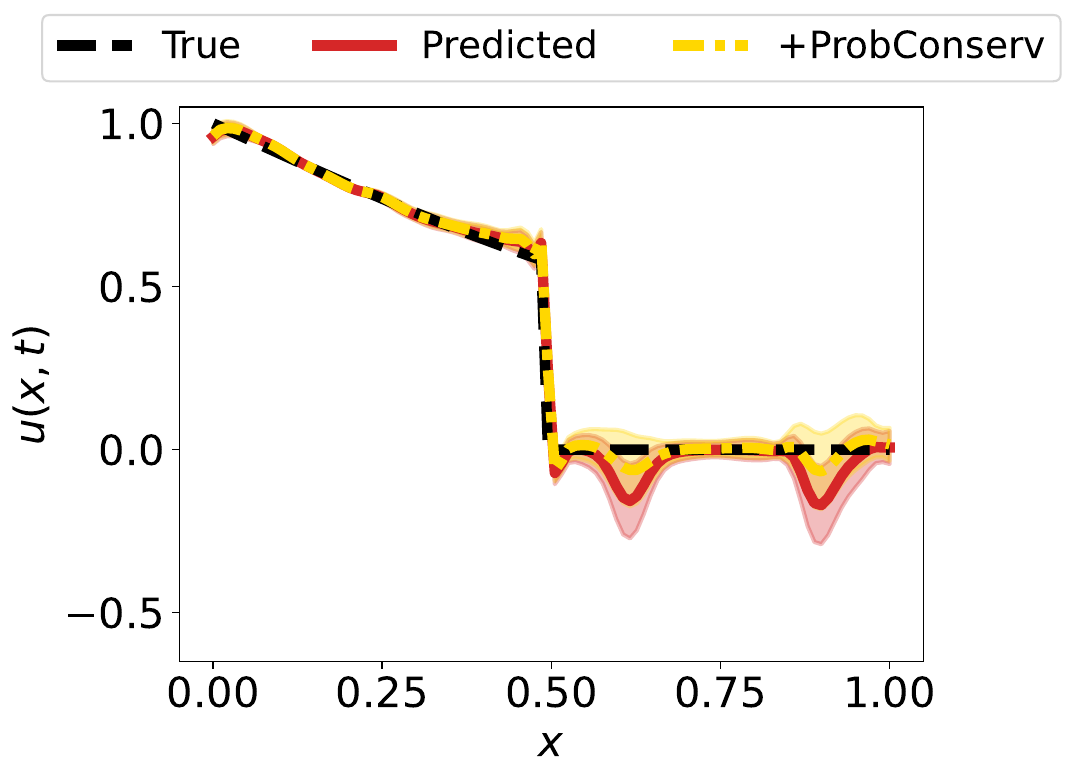}
    \caption{\method + \probconserv}
    \label{fig:probconserv_diverseno_stefan}
    \end{subfigure}
    \caption{
    {\bf Effect of \probconserv update.}
    Solution profiles before and after applying \probconserv over \method and \bayesiannomethod on 1-d Stefan with small OOD shift. {\bf (a)} \probconserv is able to correct the dips in \method's solution as the uncertainty estimates are high in those regions as well.  {\bf (b)} Uncertainty estimates from \bayesiannomethod are not correlated with the errors and \probconserv focuses on $x=0.4$ where the solution was already accurate instead of the dip at $x=0.9$. 
    }
    \label{fig:probconserv}
\end{figure}

We investigate whether uncertainty estimates obtained from the UQ methods can be used to improve OOD performance and to apply probabilistic physics-based constraints. 
We use \probconserv~\citep{hansen2023learning} to enforce a known linear conservation constraint of the form $\int_x u(x, t) dx = b(t)$ corresponding to each of the PDEs considered. (See \cref{subsec:probconservno}.) 
When applying \probconserv, we use the discretized form of the constraint $Gu = b + \sigma_G \epsilon$, where $\epsilon$ denotes a noise term that allows for a slack in the constraint for practical considerations. 
We use $\sigma_G=10^{-9}$ in our experiments. 
We report the conservation errors (CE) in \cref{subsec:probconserv} and show that conservation is typically violated OOD for the unconstrained NO methods (with CE $\approx\!10^{-2}$ for the heat equation, $\approx\!0.3$ for PME and $\approx\!0.4$ for Stefan) and \probconserv enforces the known physical constraint exactly for all UQ methods (i.e., with $\text{CE}=0$).

\cref{fig:mse_improvement_probconserv} shows the MSE ratio before and after applying \probconserv (larger than 1 indicates improvement) for all 3 PDEs and OOD shifts. (See 
\cref{subsec:probconserv} for the exact metrics.)
Similar to the original \probconserv \citep{hansen2023learning} work for in-domain problems, we also observe differing behaviors of applying \probconserv on ``easy'' to ``hard'' PDEs for OOD problems. 
In the ``easy'' heat equation, \probconserv improves OOD MSE for all the UQ methods, even when the uncertainty estimates are uncalibrated (e.g., for \outputvarmethod). \method improves the most with \probconserv ($1.4\times$) and subsequently achieves the lowest MSE of all methods in every OOD shift ($1.8\times$ to $5.5\times$ better than baselines).
For ``medium'' and ``hard'' cases of GPME, the effect of \probconserv depends on the goodness of uncertainty estimates. 
For instance, \probconserv degrades the performance of \outputvarmethod drastically (by 1300\% and 31\% for small OOD shifts in the 2 PDEs) because of uncalibrated uncertainty estimates.
\cref{fig:probconserv_bayesianno_stefan} supports a similar finding for \bayesiannomethod in the ``hard'' Stefan case: \probconserv is not able to correct the oscillation near the right boundary since the corresponding uncertainty estimates are much lower.
In contrast, \probconserv fixes these oscillations for \method (\cref{fig:probconserv_diverseno_stefan}).
These findings demonstrate the need for uncertainty estimates that are correlated with prediction errors. 
Finally, for large OOD shifts in the ``medium''/``hard'' PDEs, prediction errors remain high for all methods ($\approx\!10^{-2}$) after applying \probconserv, indicating that global conservation constraint alone is not enough to solve these challenging OOD tasks and additional (local) physical constraints may be required.
Overall, we show that with error-correlated UQ from \method, \probconserv can be effective against small to medium OOD shifts.  

\section{Conclusion}
NOs have proven to be a successful class of data-driven methods for efficiently approximating the solution to certain PDE problems on in-domain tasks. 
In this work, we have shown that, despite these promising initial successes, NOs are not robust to OOD shifts in their inputs, in particular to shifts in PDE parameters. 
We show that a computationally-expensive ensemble of NOs provides a strong baseline for good OOD uncertainty estimates; and, motivated by this, we propose a simple scalable alternative, \method, that provides uncertainty estimates that are well-correlated with prediction errors. 
We use these error-correlated uncertainty estimates from \method within the \probconserv framework \citep{hansen2023learning} to develop 
\probconservno. We show that having UQ estimates that are well-correlated with the error are critical for the success of \probconservno in improving the OOD performance. 
Our empirical results demonstrate that \probconservno improves the accuracy of NOs across a wide range of PDE problem settings, in particular in high-error regions of the spatial domain.
The improvements are particularly prominent on problems with shocks and that satisfy conservation laws. 
Future work includes 
extending \probconservno  
to perform updates locally in the areas with the highest error estimates and to further improve and characterize the need for the well-correlated UQ estimates with the error.

\bibliography{references,from_zotero_do_not_edit}  

\begin{thebibliography}{67}
\providecommand{\natexlab}[1]{#1}
\providecommand{\url}[1]{\texttt{#1}}
\expandafter\ifx\csname urlstyle\endcsname\relax
  \providecommand{\doi}[1]{doi: #1}\else
  \providecommand{\doi}{doi: \begingroup \urlstyle{rm}\Url}\fi

\bibitem[Alesiani et~al.(2022)Alesiani, Takamoto, and Niepert]{alesiani2022}
Francesco Alesiani, Makoto Takamoto, and Mathias Niepert.
\newblock Hyper{FNO}: Improving the generalization behavior of {F}ourier
  {N}eural {O}perators.
\newblock In \emph{NeurIPS 2022 Workshop on Machine Learning and Physical
  Sciences}, 2022.

\bibitem[Angelopoulos et~al.(2023)Angelopoulos, Bates,
  et~al.]{angelopoulos2023conformal}
Anastasios~N Angelopoulos, Stephen Bates, et~al.
\newblock Conformal prediction: A gentle introduction.
\newblock \emph{Foundations and Trends{\textregistered} in Machine Learning},
  16\penalty0 (4):\penalty0 494--591, 2023.

\bibitem[Benitez et~al.(2023)Benitez, Furuya, Faucher, Kratsios, Tricoche, and
  de~Hoop]{benitez2023outofdistributional}
J.~Antonio~Lara Benitez, Takashi Furuya, Florian Faucher, Anastasis Kratsios,
  Xavier Tricoche, and Maarten~V. de~Hoop.
\newblock Out-of-distributional risk bounds for neural operators with
  applications to the {Helmholtz} equation.
\newblock \emph{arXiv preprint arXiv:2301.11509}, 2023.

\bibitem[Boullé \& Townsend(2023)Boullé and Townsend]{nicholas2023}
Nicolas Boullé and Alex Townsend.
\newblock A mathematical guide to operator learning.
\newblock \emph{arXiv preprint arXiv:2312.14688}, 2023.

\bibitem[Bourel et~al.(2020)Bourel, Cugliari, Goude, and
  Poggi]{bourel2020boosting}
Mathias Bourel, Jairo Cugliari, Yannig Goude, and Jean-Michel Poggi.
\newblock Boosting diversity in regression ensembles.
\newblock \emph{Statistical Analysis and Data Mining: The ASA Data Science
  Journal}, 2020.

\bibitem[Brunton et~al.(2016)Brunton, Proctor, and
  Kutz]{brunton2016discovering}
Steven~L Brunton, Joshua~L Proctor, and J~Nathan Kutz.
\newblock Discovering governing equations from data by sparse identification of
  nonlinear dynamical systems.
\newblock \emph{Proceedings of the {N}ational {A}cademy of {S}ciences},
  113\penalty0 (15):\penalty0 3932--3937, 2016.

\bibitem[Chatterjee \& Hadi(1988)Chatterjee and Hadi]{ChatterjeeHadi88}
S.~Chatterjee and A.S. Hadi.
\newblock \emph{Sensitivity Analysis in Linear Regression}.
\newblock John Wiley {\&} Sons, New York, 1988.

\bibitem[Conti et~al.(2023)Conti, Gobat, Fresca, Manzoni, and
  Frangi]{conti2023reduced}
Paolo Conti, Giorgio Gobat, Stefania Fresca, Andrea Manzoni, and Attilio
  Frangi.
\newblock Reduced order modeling of parametrized systems through autoencoders
  and sindy approach: continuation of periodic solutions.
\newblock \emph{Computer Methods in Applied Mechanics and Engineering},
  411:\penalty0 116072, 2023.

\bibitem[Dandekar et~al.(2022)Dandekar, Chung, Dixit, Tarek, {Garcia-Valadez},
  Vemula, and Rackauckas]{dandekar2022}
Raj Dandekar, Karen Chung, Vaibhav Dixit, Mohamed Tarek, Aslan
  {Garcia-Valadez}, Krishna~Vishal Vemula, and Chris Rackauckas.
\newblock Bayesian {{Neural Ordinary Differential Equations}}.
\newblock \emph{arXiv preprint arXiv:2012.07244}, 2022.

\bibitem[{de Hoop} et~al.(2022){de Hoop}, Huang, Qian, and
  Stuart]{dehoop2022CostAccuracy}
Maarten~V. {de Hoop}, Daniel~Zhengyu Huang, Elizabeth Qian, and Andrew~M.
  Stuart.
\newblock The {{Cost-Accuracy Trade-Off In Operator Learning With Neural
  Networks}}, August 2022.

\bibitem[Diquigiovanni et~al.(2022)Diquigiovanni, Fontana, and
  Vantini]{diquigiovanni2022conformal}
Jacopo Diquigiovanni, Matteo Fontana, and Simone Vantini.
\newblock Conformal prediction bands for multivariate functional data.
\newblock \emph{Journal of Multivariate Analysis}, 189:\penalty0 104879, 2022.

\bibitem[Edwards(2022)]{cacm_edwards22}
Chris Edwards.
\newblock Neural networks learn to speed up simulations.
\newblock \emph{Communications of the ACM}, 65\penalty0 (5):\penalty0 27--29,
  2022.

\bibitem[Gal \& Ghahramani(2016)Gal and Ghahramani]{gal2016dropout}
Yarin Gal and Zoubin Ghahramani.
\newblock Dropout as a bayesian approximation: Representing model uncertainty
  in deep learning.
\newblock In \emph{International Conference on Machine Learning}, pp.\
  1050--1059. PMLR, 2016.

\bibitem[Gneiting \& Raftery(2007)Gneiting and Raftery]{gneiting2007}
Tilmann Gneiting and Adrian~E. Raftery.
\newblock Strictly proper scoring rules, prediction, and estimation.
\newblock \emph{Journal of the American Statistical Association}, 102\penalty0
  (477):\penalty0 359 -- 378, 2007.

\bibitem[Gneiting et~al.(2007)Gneiting, Balabdaoui, and
  Raftery]{gneiting2007probabilistic}
Tilmann Gneiting, Fadoua Balabdaoui, and Adrian~E Raftery.
\newblock Probabilistic forecasts, calibration and sharpness.
\newblock \emph{Journal of the Royal Statistical Society Series B: Statistical
  Methodology}, 69\penalty0 (2):\penalty0 243--268, 2007.

\bibitem[Graves(2011)]{graves2011practical}
Alex Graves.
\newblock Practical variational inference for neural networks.
\newblock In \emph{Advances in Neural Information Processing Systems},
  volume~24, 2011.

\bibitem[Guo et~al.(2023)Guo, Wu, Zhou, and Zhou]{guo2023ib}
Ling Guo, Hao Wu, Wenwen Zhou, and Tao Zhou.
\newblock Ib-uq: Information bottleneck based uncertainty quantification for
  neural function regression and neural operator learning.
\newblock \emph{arXiv preprint arXiv:2302.03271}, 2023.

\bibitem[Gupta et~al.(2021)Gupta, Xiao, and Bogdan]{gupta2021multiwavelet}
Gaurav Gupta, Xiongye Xiao, and Paul Bogdan.
\newblock Multiwavelet-based operator learning for differential equations.
\newblock In \emph{Advances in Neural Information Processing Systems},
  volume~34, pp.\  24048--24062, 2021.

\bibitem[Hansen et~al.(2023)Hansen, Maddix, Alizadeh, Gupta, and
  Mahoney]{hansen2023learning}
Derek Hansen, Danielle~C. Maddix, Shima Alizadeh, Gaurav Gupta, and Michael~W
  Mahoney.
\newblock Learning physical models that can respect conservation laws.
\newblock In \emph{International Conference on Machine Learning}, volume 202,
  pp.\  12469--12510. PMLR, 2023.

\bibitem[Hodgkinson et~al.(2022)Hodgkinson, {van der Heide}, Roosta, and
  Mahoney]{hodgkinson2022}
Liam Hodgkinson, Chris {van der Heide}, Fred Roosta, and Michael~W. Mahoney.
\newblock Monotonicity and {{Double Descent}} in {{Uncertainty Estimation}}
  with {{Gaussian Processes}}.
\newblock \emph{arXiv preprint arXiv:2210.07612}, 2022.

\bibitem[Hodgkinson et~al.(2023)Hodgkinson, {van der Heide}, Salomone, Roosta,
  and Mahoney]{hodgkinson_IIC_TR}
Liam Hodgkinson, Chris {van der Heide}, Robert Salomone, Fred Roosta, and
  Michael~W. Mahoney.
\newblock The interpolating information criterion for overparameterized models.
\newblock \emph{arXiv preprint arXiv:2307.07785}, 2023.

\bibitem[Hughes(2000)]{hughes2000}
Thomas~J.R. Hughes.
\newblock \emph{The Finite Element Method: Linear Static and Dynamic Finite
  Element Analysis}.
\newblock Dover Publications, 2000.

\bibitem[Kim et~al.(2019)Kim, Mnih, Schwarz, Garnelo, Eslami, Rosenbaum,
  Vinyals, and Teh]{kimAttentiveNeuralProcesses2019}
Hyunjik Kim, Andriy Mnih, Jonathan Schwarz, Marta Garnelo, Ali Eslami, Dan
  Rosenbaum, Oriol Vinyals, and Yee~Whye Teh.
\newblock Attentive {{Neural Processes}}.
\newblock \emph{arXiv preprint arXiv:1901.05761}, 2019.

\bibitem[Kovachki et~al.(2021)Kovachki, Li, Liu, Azizzadenesheli, Bhattacharya,
  Stuart, and Anandkumar]{kovachki2021neural}
Nikola Kovachki, Zongyi Li, Burigede Liu, Kamyar Azizzadenesheli, Kaushik
  Bhattacharya, Andrew Stuart, and Anima Anandkumar.
\newblock Neural operator: Learning maps between function spaces.
\newblock \emph{arXiv preprint arXiv:2108.08481}, 2021.

\bibitem[Krishnapriyan et~al.(2021)Krishnapriyan, Gholami, Zhe, Kirby, and
  Mahoney]{krishnapriyanCharacterizingPossibleFailure2021b}
Aditi~S. Krishnapriyan, Amir Gholami, Shandian Zhe, Robert Kirby, and Michael~W
  Mahoney.
\newblock Characterizing possible failure modes in physics-informed neural
  networks.
\newblock In \emph{Advances in {{Neural Information Processing Systems}}},
  volume~34, pp.\  26548--26560, 2021.

\bibitem[Kuleshov et~al.(2018)Kuleshov, Fenner, and
  Ermon]{kuleshov2018accurate}
Volodymyr Kuleshov, Nathan Fenner, and Stefano Ermon.
\newblock Accurate uncertainties for deep learning using calibrated regression.
\newblock In \emph{International Conference on Machine Learning}, pp.\
  2796--2804. PMLR, 2018.

\bibitem[Lakshminarayanan et~al.(2017)Lakshminarayanan, Pritzel, and
  Blundell]{lakshminarayanan2017Simple}
Balaji Lakshminarayanan, Alexander Pritzel, and Charles Blundell.
\newblock Simple and {{Scalable Predictive Uncertainty Estimation}} using
  {{Deep Ensembles}}.
\newblock \emph{arXiv preprint arXiv:1612.01474}, 2017.

\bibitem[Le~Maître \& Knio(2012)Le~Maître and Knio]{maitre2012}
O.~P. Le~Maître and O.~M. Knio.
\newblock \emph{Spectral Methods for Uncertainty Quantification: With
  Applications to Computational Fluid Dynamics}.
\newblock Springer, 2012.

\bibitem[Lee et~al.(2022)Lee, Yao, and Finn]{lee2022diversify}
Yoonho Lee, Huaxiu Yao, and Chelsea Finn.
\newblock Diversify and disambiguate: Out-of-distribution robustness via
  disagreement.
\newblock In \emph{The Eleventh International Conference on Learning
  Representations}, 2022.

\bibitem[Leiteritz \& Pfl{\"u}ger(2021)Leiteritz and
  Pfl{\"u}ger]{leiteritz2021avoid}
Raphael Leiteritz and Dirk Pfl{\"u}ger.
\newblock How to avoid trivial solutions in physics-informed neural networks.
\newblock \emph{arXiv preprint arXiv:2112.05620}, 2021.

\bibitem[Lessig et~al.(2023)Lessig, Luise, Gong, Langguth, Stadtler, and
  Schultz]{lessig2023}
Christian Lessig, Ilaria Luise, Bing Gong, Michael Langguth, Scarlet Stadtler,
  and Martin Schultz.
\newblock Atmo{R}ep: A stochastic model of atmosphere dynamics using large
  scale representation learning.
\newblock \emph{arXiv preprint arXiv:2308.13280}, 2023.

\bibitem[LeVeque(2002)]{leveque2002}
Randall~J. LeVeque.
\newblock \emph{Finite Volume Methods for Hyperbolic Problems}.
\newblock Cambridge University Press, 2002.

\bibitem[LeVeque(2007)]{leveque2007}
Randall~J. LeVeque.
\newblock \emph{Finite Difference Methods for Ordinary and Partial Differential
  Equations: Steady-State and Time-Dependent Problems}.
\newblock SIAM, 2007.

\bibitem[Li et~al.(2020{\natexlab{a}})Li, Kovachki, Azizzadenesheli, Liu,
  Bhattacharya, Stuart, and Anandkumar]{li2020fourier}
Zongyi Li, Nikola Kovachki, Kamyar Azizzadenesheli, Burigede Liu, Kaushik
  Bhattacharya, Andrew Stuart, and Anima Anandkumar.
\newblock Fourier neural operator for parametric partial differential
  equations.
\newblock In \emph{International Conference on Learning Representations},
  2020{\natexlab{a}}.

\bibitem[Li et~al.(2020{\natexlab{b}})Li, Kovachki, Azizzadenesheli, Liu,
  Bhattacharya, Stuart, and Anandkumar]{li2020graphkernel}
Zongyi Li, Nikola Kovachki, Kamyar Azizzadenesheli, Burigede Liu, Kaushik
  Bhattacharya, Andrew Stuart, and Anima Anandkumar.
\newblock Neural operator: Graph kernel network for partial differential
  equations.
\newblock \emph{arXiv preprint arXiv:2003.03485}, 2020{\natexlab{b}}.

\bibitem[Li et~al.(2021)Li, Zheng, Kovachki, Jin, Chen, Liu, Azizzadenesheli,
  and Anandkumar]{li2021physics}
Zongyi Li, Hongkai Zheng, Nikola Kovachki, David Jin, Haoxuan Chen, Burigede
  Liu, Kamyar Azizzadenesheli, and Anima Anandkumar.
\newblock Physics-informed neural operator for learning partial differential
  equations.
\newblock \emph{arXiv preprint arXiv:2111.03794}, 2021.

\bibitem[Li et~al.(2022{\natexlab{a}})Li, Huang, Liu, and
  Anandkumar]{li2022fourier}
Zongyi Li, Daniel~Zhengyu Huang, Burigede Liu, and Anima Anandkumar.
\newblock Fourier neural operator with learned deformations for pdes on general
  geometries.
\newblock \emph{arXiv preprint arXiv:2207.05209}, 2022{\natexlab{a}}.

\bibitem[Li et~al.(2022{\natexlab{b}})Li, {Liu-Schiaffini}, Kovachki, Liu,
  Azizzadenesheli, Bhattacharya, Stuart, and Anandkumar]{li2022}
Zongyi Li, Miguel {Liu-Schiaffini}, Nikola Kovachki, Burigede Liu, Kamyar
  Azizzadenesheli, Kaushik Bhattacharya, Andrew Stuart, and Anima Anandkumar.
\newblock Learning {{Dissipative Dynamics}} in {{Chaotic Systems}}.
\newblock \emph{arXiv preprint arXiv:2106.06898}, 2022{\natexlab{b}}.

\bibitem[Liu et~al.(2023)Liu, Yu, You, and Tatikola]{liu2023ino}
Ning Liu, Yue Yu, Huaiqian You, and Neeraj Tatikola.
\newblock Ino: Invariant neural operators for learning complex physical systems
  with momentum conservation.
\newblock In \emph{International Conference on Artificial Intelligence and
  Statistics}, pp.\  6822--6838. PMLR, 2023.

\bibitem[Lu et~al.(2019)Lu, Jin, and Karniadakis]{lu2019deeponet}
Lu~Lu, Pengzhan Jin, and George~Em Karniadakis.
\newblock Deeponet: Learning nonlinear operators for identifying differential
  equations based on the universal approximation theorem of operators.
\newblock \emph{arXiv preprint arXiv:1910.03193}, 2019.

\bibitem[Ma et~al.(2024)Ma, Azizzadenesheli, and Anandkumar]{ma2024}
Ziqi Ma, Kamyar Azizzadenesheli, and Anima Anandkumar.
\newblock Calibrated uncertainty quantification for operator learning via
  conformal prediction.
\newblock \emph{arXiv preprint arXiv:2402.01960}, 2024.

\bibitem[Maddix et~al.(2018{\natexlab{a}})Maddix, Sampaio, and
  Gerritsen]{maddix2018_harmonic_avg}
Danielle~C. Maddix, Luiz Sampaio, and Margot Gerritsen.
\newblock Numerical artifacts in the {{Generalized Porous Medium Equation}}:
  {{Why}} harmonic averaging itself is not to blame.
\newblock \emph{Journal of Computational Physics}, 361:\penalty0 280--298,
  2018{\natexlab{a}}.

\bibitem[Maddix et~al.(2018{\natexlab{b}})Maddix, Sampaio, and
  Gerritsen]{maddix2018temp_oscill}
Danielle~C. Maddix, Luiz Sampaio, and Margot Gerritsen.
\newblock Numerical artifacts in the discontinuous {Generalized Porous Medium
  Equation}: How to avoid spurious temporal oscillations.
\newblock \emph{Journal of Computational Physics}, 368:\penalty0 277--298,
  2018{\natexlab{b}}.

\bibitem[Magnani et~al.(2022)Magnani, Kr{\"a}mer, Eschenhagen, Rosasco, and
  Hennig]{magnani2022}
Emilia Magnani, Nicholas Kr{\"a}mer, Runa Eschenhagen, Lorenzo Rosasco, and
  Philipp Hennig.
\newblock Approximate {{Bayesian Neural Operators}}: {{Uncertainty
  Quantification}} for {{Parametric PDEs}}.
\newblock \emph{arXiv preprint arXiv:2208.01565}, 2022.

\bibitem[Moukari et~al.(2019)Moukari, Simon, Picard, and Jurie]{moukari2019n}
Michel Moukari, Lo{\"\i}c Simon, Sylvaine Picard, and Fr{\'e}d{\'e}ric Jurie.
\newblock n-merci: A new metric to evaluate the correlation between predictive
  uncertainty and true error.
\newblock In \emph{2019 IEEE/RSJ International Conference on Intelligent Robots
  and Systems (IROS)}, pp.\  5250--5255. IEEE, 2019.

\bibitem[Négiar et~al.(2023)Négiar, Mahoney, and
  Krishnapriyan]{negiar2022learning}
Geoffrey Négiar, Michael~W. Mahoney, and Aditi~S. Krishnapriyan.
\newblock Learning differentiable solvers for systems with hard constraints.
\newblock In \emph{International Conference on Learning Representations}, 2023.

\bibitem[Pathak et~al.(2022)Pathak, Subramanian, Harrington, Raja,
  Chattopadhyay, Mardani, Kurth, Hall, Li, Azizzadenesheli, Hassanzadeh,
  Kashinath, and Anandkumar]{pathak2022}
Jaideep Pathak, Shashank Subramanian, Peter Harrington, Sanjeev Raja, Ashesh
  Chattopadhyay, Morteza Mardani, Thorsten Kurth, David Hall, Zongyi Li, Kamyar
  Azizzadenesheli, Pedram Hassanzadeh, Karthik Kashinath, and Animashree
  Anandkumar.
\newblock {{FourCastNet}}: {{A Global Data-driven High-resolution Weather
  Model}} using {{Adaptive Fourier Neural Operators}}.
\newblock \emph{arXiv preprint arXiv:2202.11214}, 2022.

\bibitem[Psaros et~al.(2023)Psaros, Meng, Zou, Guo, and
  Karniadakis]{psaros2023Uncertainty}
Apostolos~F. Psaros, Xuhui Meng, Zongren Zou, Ling Guo, and George~Em
  Karniadakis.
\newblock Uncertainty {{Quantification}} in {{Scientific Machine Learning}}:
  {{Methods}}, {{Metrics}}, and {{Comparisons}}.
\newblock \emph{Journal of Computational Physics}, 477:\penalty0 111902, 2023.

\bibitem[Psarosa et~al.(2022)Psarosa, Menga, ~, Guob, and
  Karniadakisa]{apostolos2022}
Apostolos~F Psarosa, Xuhui Menga, Zongren ~, Ling Guob, and George~Em
  Karniadakisa.
\newblock Uncertainty quantification in scientific machine learning: Methods,
  metrics, and comparisons.
\newblock \emph{arXiv preprint arXiv:2201.07766}, 2022.

\bibitem[Raissi et~al.(2019)Raissi, Perdikaris, and
  Karniadakis]{raissi2019physics}
Maziar Raissi, Paris Perdikaris, and George~E Karniadakis.
\newblock Physics-informed neural networks: A deep learning framework for
  solving forward and inverse problems involving nonlinear partial differential
  equations.
\newblock \emph{Journal of Computational Physics}, 378:\penalty0 686--707,
  2019.

\bibitem[Rame \& Cord(2021)Rame and Cord]{rame2021dice}
Alexandre Rame and Matthieu Cord.
\newblock Dice: Diversity in deep ensembles via conditional redundancy
  adversarial estimation.
\newblock \emph{arXiv preprint arXiv:2101.05544}, 2021.

\bibitem[Rezaeiravesh et~al.(2020)Rezaeiravesh, Vinuesa, and
  Schlatter]{Rezaeiravesh2020}
Saleh Rezaeiravesh, Ricardo Vinuesa, and Philipp Schlatter.
\newblock An uncertainty-quantification framework for assessing accuracy,
  sensitivity, and robustness in computational fluid dynamics.
\newblock \emph{arXiv preprint arXiv:2302.03271}, 2020.

\bibitem[Saad et~al.(2023)Saad, Gupta, Alizadeh, and Maddix]{saad2022guiding}
Nadim Saad, Gaurav Gupta, Shima Alizadeh, and Danielle~C. Maddix.
\newblock Guiding continuous operator learning through physics-based boundary
  constraints.
\newblock In \emph{International Conference on Learning Representations}, 2023.

\bibitem[Schwaiger et~al.(2020)Schwaiger, Sinhamahapatra, Gansloser, and
  Roscher]{Schwaiger2020IsUQ}
Adrian Schwaiger, Poulami Sinhamahapatra, Jens Gansloser, and Karsten Roscher.
\newblock Is uncertainty quantification in deep learning sufficient for
  out-of-distribution detection?
\newblock In \emph{AISafety@IJCAI}, 2020.

\bibitem[Sinha et~al.(2020)Sinha, Bharadhwaj, Goyal, Larochelle, Garg, and
  Shkurti]{sinha2020diversity}
Samarth Sinha, Homanga Bharadhwaj, Anirudh Goyal, Hugo Larochelle, Animesh
  Garg, and Florian Shkurti.
\newblock Diversity inducing information bottleneck in model ensembles.
\newblock \emph{arXiv preprint arXiv:2003.04514}, 2020.

\bibitem[Subramanian et~al.(2023)Subramanian, Harrington, Keutzer, Bhimji,
  Morozov, Mahoney, and Gholami]{subramanian2023towards}
Shashank Subramanian, Peter Harrington, Kurt Keutzer, Wahid Bhimji, Dmitriy
  Morozov, Michael Mahoney, and Amir Gholami.
\newblock Towards foundation models for scientific machine learning:
  Characterizing scaling and transfer behavior.
\newblock In \emph{Advances in {{Neural Information Processing Systems}}},
  volume~36, 2023.

\bibitem[Teye et~al.(2018)Teye, Azizpour, and Smith]{teye2018bayesian}
Mattias Teye, Hossein Azizpour, and Kevin Smith.
\newblock Bayesian uncertainty estimation for batch normalized deep networks.
\newblock In \emph{International Conference on Machine Learning}, pp.\
  4907--4916. PMLR, 2018.

\bibitem[Theisen et~al.(2023)Theisen, Kim, Yang, Hodgkinson, and
  Mahoney]{theisen2023}
Ryan Theisen, Hyunsuk Kim, Yaoqing Yang, Liam Hodgkinson, and Michael~W.
  Mahoney.
\newblock When are ensembles really effective?
\newblock \emph{arXiv preprint arXiv:2305.12313}, 2023.

\bibitem[V{\'a}zquez(2007)]{vazquez2007porous}
Juan~Luis V{\'a}zquez.
\newblock \emph{The Porous Medium Equation: Mathematical Theory}.
\newblock Oxford University Press, 2007.

\bibitem[Weber et~al.(2024)Weber, Magnani, Pförtner, and Hennig]{weber2024}
Tobias Weber, Emilia Magnani, Marvin Pförtner, and Philipp Hennig.
\newblock Uncertainty quantificaiton for {F}ourier {N}eural {O}perators.
\newblock In \emph{ICLR 2024 Workshop on AI4DifferentialEquations In Science},
  2024.

\bibitem[Wilson \& Izmailov(2020)Wilson and Izmailov]{wilson2020Bayesian}
Andrew~G Wilson and Pavel Izmailov.
\newblock Bayesian {{Deep Learning}} and a {{Probabilistic Perspective}} of
  {{Generalization}}.
\newblock In \emph{Advances in {{Neural Information Processing Systems}}},
  volume~33, pp.\  4697--4708, 2020.

\bibitem[Wood et~al.(2023)Wood, Mu, Webb, Reeve, Lujan, and
  Brown]{wood2023unified}
Danny Wood, Tingting Mu, Andrew Webb, Henry Reeve, Mikel Lujan, and Gavin
  Brown.
\newblock A unified theory of diversity in ensemble learning.
\newblock \emph{arXiv preprint arXiv:2301.03962}, 2023.

\bibitem[Xiu \& Karniadakis(2002)Xiu and Karniadakis]{xiu2002}
Dongbin Xiu and George Karniadakis.
\newblock The wiener-askey polynomial chaos for stochastic differential
  equations.
\newblock \emph{SIAM Journal on Scientific Computing, 24(2), 619-6442}, 2002.

\bibitem[Yang et~al.(2022)Yang, Kissas, and Perdikaris]{yang2022}
Yibo Yang, Georgios Kissas, and Paris Perdikaris.
\newblock Scalable uncertainty quantification for deep operator networks using
  randomized priors.
\newblock \emph{arXiv preprint arXiv:2203.03048}, 2022.

\bibitem[Yin et~al.(2022)Yin, Kirchmeyer, Franceschi, Rakotomamonjy, and
  Gallinari]{yin2022continuous}
Yuan Yin, Matthieu Kirchmeyer, Jean-Yves Franceschi, Alain Rakotomamonjy, and
  Patrick Gallinari.
\newblock Continuous pde dynamics forecasting with implicit neural
  representations.
\newblock \emph{arXiv preprint arXiv:2209.14855}, 2022.

\bibitem[Zhang et~al.(2020)Zhang, Liu, and Yan]{zhang2020diversified}
Shaofeng Zhang, Meng Liu, and Junchi Yan.
\newblock The diversified ensemble neural network.
\newblock \emph{Advances in Neural Information Processing Systems},
  33:\penalty0 16001--16011, 2020.

\bibitem[Zou et~al.(2023)Zou, Meng, and Karniadakis]{zou2023uncertainty}
Zongren Zou, Xuhui Meng, and George~Em Karniadakis.
\newblock Uncertainty quantification for noisy inputs-outputs in
  physics-informed neural networks and neural operators.
\newblock \emph{arXiv preprint arXiv:2311.11262}, 2023.

\end{thebibliography}
\bibliographystyle{iclr2024_conference}

\newpage
\appendix
\section{Related Work} 
\label{sec:related_work}
\textbf{Numerical methods.} PDEs are ubiquitous throughout science and engineering, where they are used to model the evolution of various physical phenomena. 
These equations are typically solved for different values of PDE physical parameters, e.g., diffusivity in the heat equation, wavespeed in the advection equation, and the Reynolds number in Navier-Stokes equations. 
Solving PDEs typically requires extensive numerical knowledge and computational effort. 
Traditional approaches to solve PDEs (e.g., finite difference~\citep{leveque2007}, finite volume~\citep{leveque2002} and finite element~\citep{hughes2000} methods) can be computationally expensive, as their accuracy is dependent on the level of discretization of the spatial and temporal domains.
Finer meshes are required to achieve high accuracy, resulting in increased computational costs. 
In addition, these approaches require a full re-run from scratch whenever there are changes in PDE parameters, which may not be known a priori. 

\paragraph{SciML works on solving PDEs.}
To alleviate the drawbacks of  numerical methods, recent works in scientific machine learning (SciML) propose to use data-driven approaches to solve PDEs.
These include so-called Physics-informed Neural Networks (PINNs)
~\citep{raissi2019physics}, NOs~\citep{li2020fourier,li2022,lu2019deeponet,gupta2021multiwavelet, yin2022continuous}, and reduced-order models for discovery~\citep{brunton2016discovering,conti2023reduced}. 
By now, it has been shown that PINNs 
have several fundamental challenges associated with its soft constraint approach. In particular, it solves a single instance of the PDE with a fixed set of PDE parameters; is challenging to optimize for PDEs with large parameter values~\citep{krishnapriyanCharacterizingPossibleFailure2021b, cacm_edwards22}; and may return trivial solutions~\citep{leiteritz2021avoid}. 
On the other hand, NOs~\citep{kovachki2021neural,li2020fourier} enjoy appealing properties of discretization invariance and universal approximation, while also achieving low approximation errors on in-domain tasks. 
Being purely data-driven, they are not guaranteed to satisfy all the physical properties of the solution. To address this, existing work has tried to incorporate different physical constraints via regularization~\citep{li2021physics}, within the architecture (e.g., boundary constraints in \citet{saad2022guiding},  invariance in \citet{liu2023ino}, and PDE hard constraints in \citet{negiar2022learning}) or via a projection to enforce conservation laws in \citet{hansen2023learning}. 
Most of these methods do not address the OOD problem that can occur even after enforcing these constraints. 
\citet{subramanian2023towards} show that fine-tuning FNO models on OOD data is typically required to achieve reasonable performance. 
In particular, for significant OOD shifts, few-shot transfer learning requires a large amount of fine-tuning OOD data that may be unavailable for certain applications.
\citet{benitez2023outofdistributional} propose a variant of FNO specifically designed to learn the wavespeed to solution mapping in the Helmholtz equation and show that it performs better OOD. 

\paragraph{UQ for Neural Operators.}
Several Bayesian deep learning methods common for standard neural networks have been shown to work well for NOs on in-domain PDE applications~\citep{psaros2023Uncertainty,zou2023uncertainty}. Commonly used methodologies for approximate Bayesian inference, e.g., the Bayesian Neural Operator~\citep{magnani2022}, DeepEnsembles~\citep{lakshminarayanan2017Simple}, variational inference methods, e.g., Mean-field VI~\citep{graves2011practical,teye2018bayesian} or MC-Dropout~\citep{gal2016dropout} and MCMC approaches, e.g., Hamiltonian Monte Carlo, 
provide, at best, crude approximations of the true posterior distribution in \cref{eq:bma} of this Bayesian model. For instance, DeepEnsembles train the same architecture multiple times to obtain different models that maximize the posterior, i.e., different modes of the posterior. 
In applications to weather forecasting, FourCastNet \citep{pathak2022} generates ensembles by perturbing the initial condition with Gaussian noise. The Bayesian Neural Operator~\citep{magnani2022} uses a last-layer Laplace approximation of the posterior. \citet{weber2024} propose to use a Laplace approximation of the posterior via the last Fourier layer (instead of the last linear layer) to capture the global structure. 
Variational inference approaches approximate the posterior with a density $q(\rmW)$, and sample $\rmW_j \sim q(\rmW)$. 
MCMC methods construct a Markov chain that is asymptotically guaranteed to sample from the true posterior. An alternate approach in \citet{guo2023ib} uses a latent space representation that is assumed to be aware of the confidence of the input data in relation to the region where the training data is located to provide point-wise uncertainty estimates. Recently, \citet{ma2024} have applied distribution-free conformal prediction methods, which have been used for standard classification/regression tasks~\citep{diquigiovanni2022conformal, angelopoulos2023conformal}, to NOs in the function space to provide uncertainty estimates for all points. 
While these methods can provide good in-domain uncertainty estimates,  
most of the scalable UQ methods are not robust to OOD shifts or have not been tested on OOD problems.

\paragraph{Diversity in Ensembles.} There are several works that study the importance of diversity in ensemble-based approaches. \citet{theisen2023} show that disagreement is key for an ensemble to be effective (with respect to accuracy).  \citet{wood2023unified} study the role of diversity in reducing in-domain generalization error. 
Diversity measures directly diversify the outputs from the different models in the ensemble, e.g., via minimizing the mutual information for classification tasks~\citep{lee2022diversify,sinha2020diversity,rame2021dice}, or the $L_2$ distance between the outputs for regression tasks.  
A recent work, DivDis~\citep{lee2022diversify} is the most related to our work in that it uses multiple prediction heads with diverse outputs to improve OOD accuracy in classification tasks. 
In particular, DivDis minimizes the mutual information between the outputs from different prediction heads when given OOD inputs (assumed to be known during training). This differs from our work since in operator learning, diversifying outputs directly does not provide informative uncertainty estimates. 
In applications to atmospheric forecasting, \citet{lessig2023} propose AtmoRep, which is a Transformer-based foundation model that outputs multiple predictions by optimizing a loss function consisting of a MSE loss summed over the prediction heads, a Gaussian statistical loss and a variance regularization term that minimizes diversity across the prediction heads. 
In contrast, we show that low diversity can lead to poor uncertainty estimates on out-of-domain tasks and propose a regularization to maximize diversity in the weights of the prediction heads.

\section{Advantages of Diversity in Ensembling on a Range of PDEs}
In this section, we show that \ensemblenomethod performs well on a wide range of 1-d PDEs including the (degenerate) parabolic GPME family and the linear advection hyperbolic conservation law. We then show the corresponding heatmaps of the weights for each of the FNO models in the ensemble. This diversity is present in the ensemble across these PDEs and various Fourier layers, which motivates our development in enforcing diversity in \method.
\label{app:add_exp}

\subsection{Good Performance of Ensembling across Various PDEs and OOD Shifts}
\label{subsec:ensemble_perf_app}
\begin{figure}[H]
    \centering
    \begin{subfigure}[h]{0.45\textwidth}
    \centering
    \includegraphics[scale=0.37]{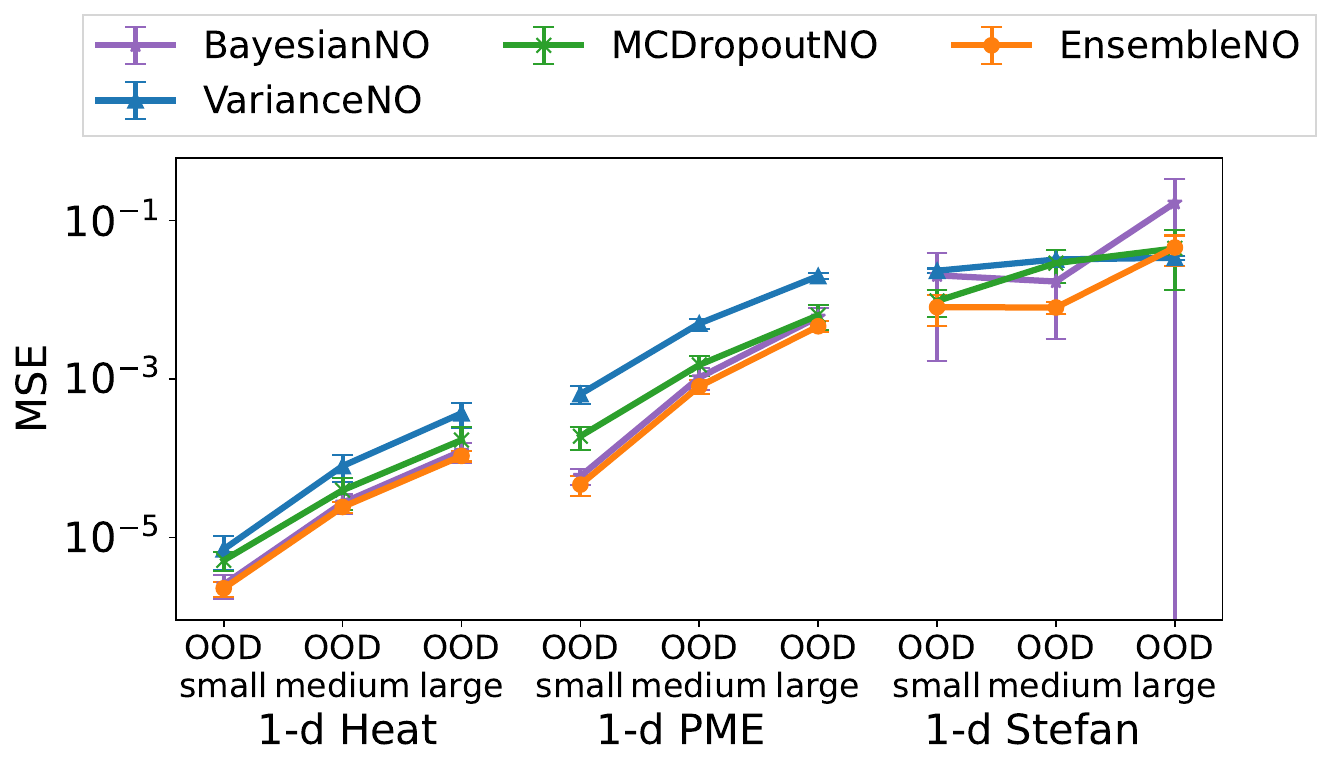}
    \caption{MSE}
    \label{fig:perfvpdeood_mse_nodiverseno}
    \end{subfigure}
    ~~~~~~~~~~
    \begin{subfigure}[h]{0.45\textwidth}
    \centering
    \includegraphics[scale=0.37]{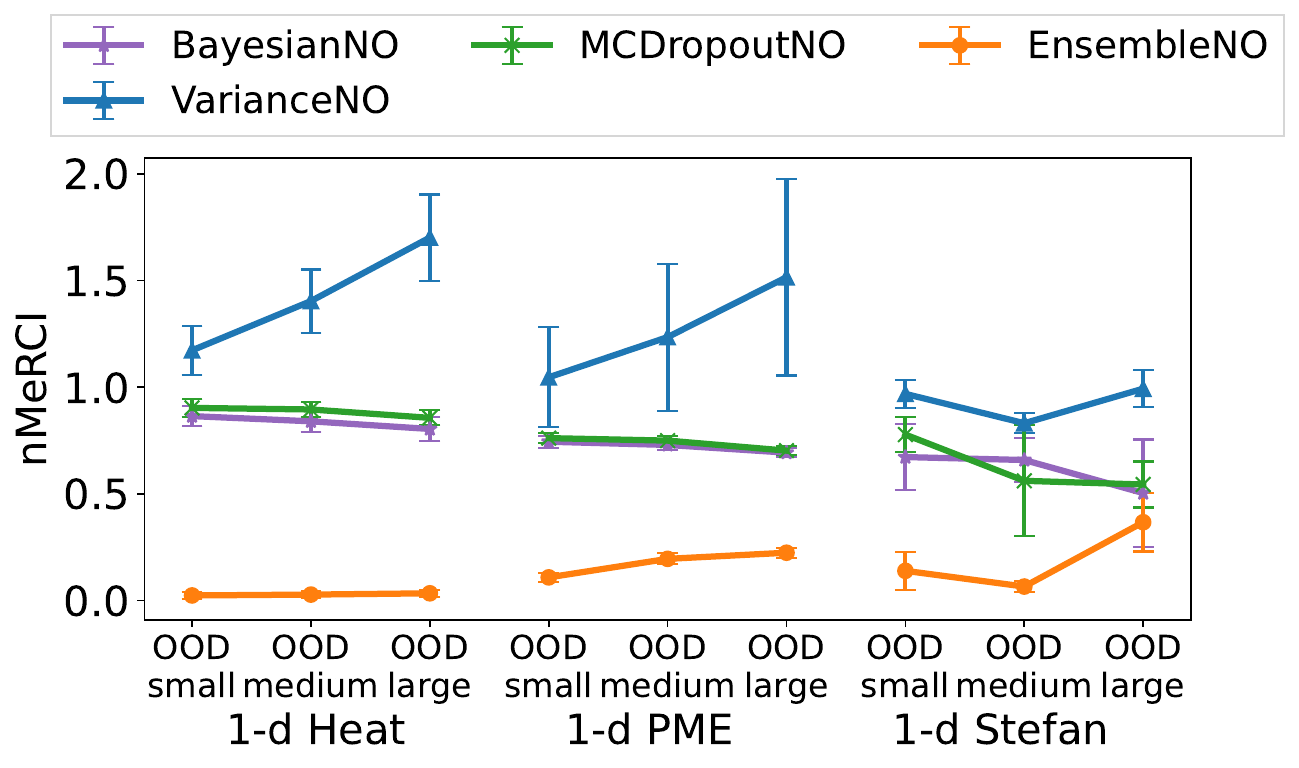}
    \caption{n-MeRCI}
    \label{fig:perfvpdeood_nmerci_nodiverseno}
    \end{subfigure}
    \caption{ 
    MSE $\downarrow$ and n-MeRCI $\downarrow$ metrics for all UQ methods on GPME family of PDEs under small, medium and large OOD shifts. {\bf (a)} MSE increases for all methods with increasing OOD shift and increasing PDE difficulty.
    {\bf (b)} \ensemblenomethod performs significantly better in the n-MeRCI metric compared to other UQ baselines, with generally close to zero values, indicating that its uncertainty estimates are most correlated with prediction errors. 
    }
    \label{fig:perfvpdeood_nodiverseno}
\end{figure}

\cref{fig:perfvpdeood_nodiverseno} illustrates the strong performance of the ensemble compared to various UQ baselines across the GPME benchmarking family of PDEs with increasing difficulty and increasing OOD shifts. \cref{fig:perfvpdeood_mse_nodiverseno} shows that the MSE increases for all methods as the problem difficulty and shift increases. \cref{fig:perfvpdeood_nmerci_nodiverseno} shows that \ensemblenomethod performs significantly better than the baselines with respect to the n-MeRCI metric with close to zero values, indicating that the uncertainty estimates are well-correlated with the prediction error. 

\subsection{Diversity in the Heatmaps of the Ensemble}
\label{app:heatmaps}
\begin{figure}[H]
    \centering
    \begin{subfigure}[h]{0.3\textwidth}
    \centering
    \includegraphics[scale=0.25]{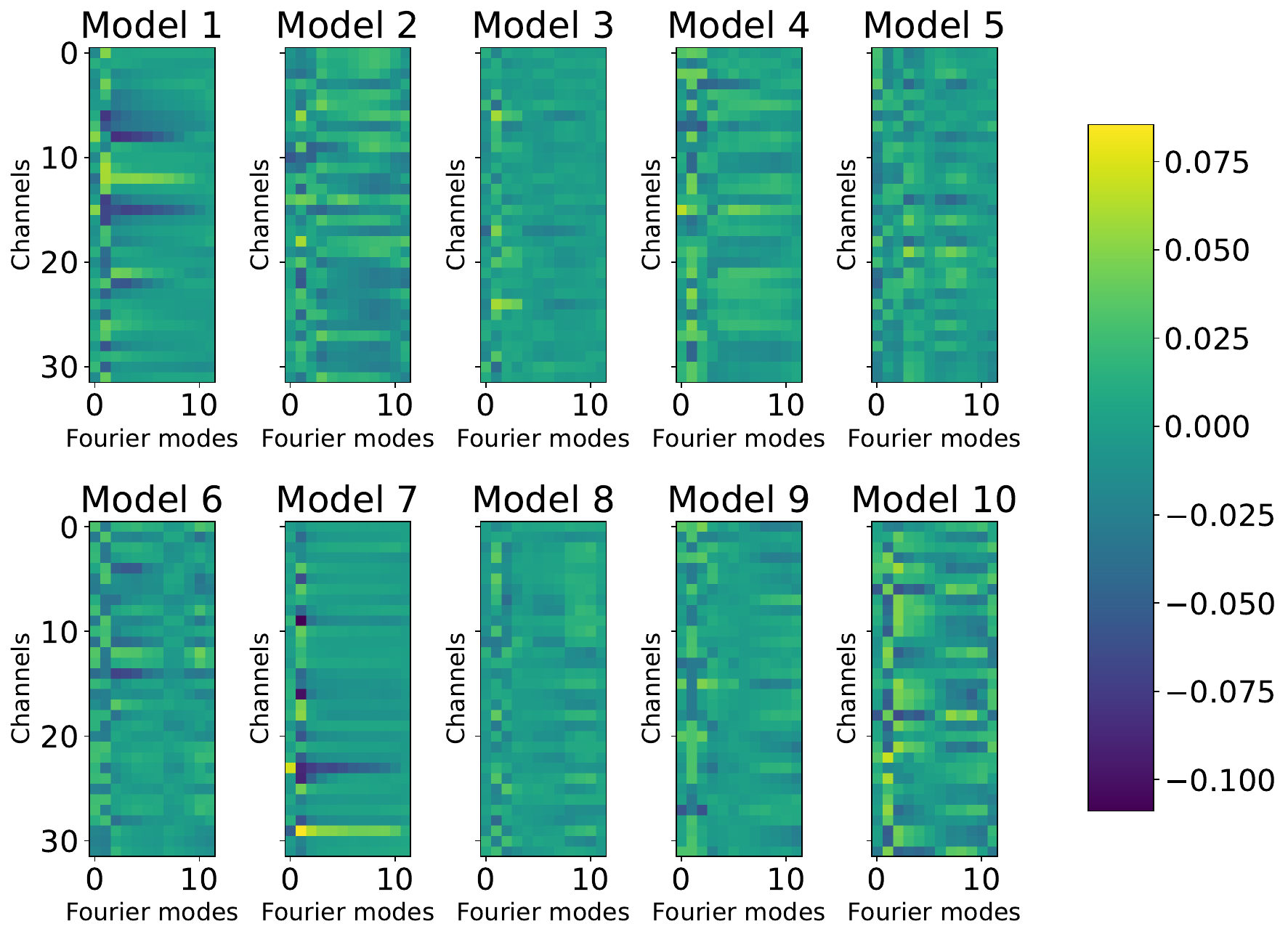}
    \caption{Heatmaps of the first Fourier layer weights from an ensemble of 10 FNO models.}
    \label{fig:ensemble_diversity:heatmaps_heat_layer0}
    \end{subfigure}
    ~~~~~~~~~~~~~~~~~~~~~~~~~~
    \begin{subfigure}[h]{0.3\textwidth}
    \centering
    \includegraphics[scale=0.4]{new_figures/Ensemble_diversity_coefvar_HeatEquation_1D_layer0.pdf}
    \caption{Coefficient of variation (mean/std) of the first Fourier layer weights across the 10 models in the ensemble.}
    \label{fig:ensemble_diversity:coefvar_heat_layer0}
    \end{subfigure}
    
    \caption{First Fourier layer of FNO models trained on 1-d heat equation task with $k^\tr\in[1,5]$.}
    \label{fig:ensemble_diversity_heat_layer0}
\end{figure}

\begin{figure}[H]
    \centering
    \begin{subfigure}[h]{0.3\textwidth}
    \centering
    \includegraphics[scale=0.25]{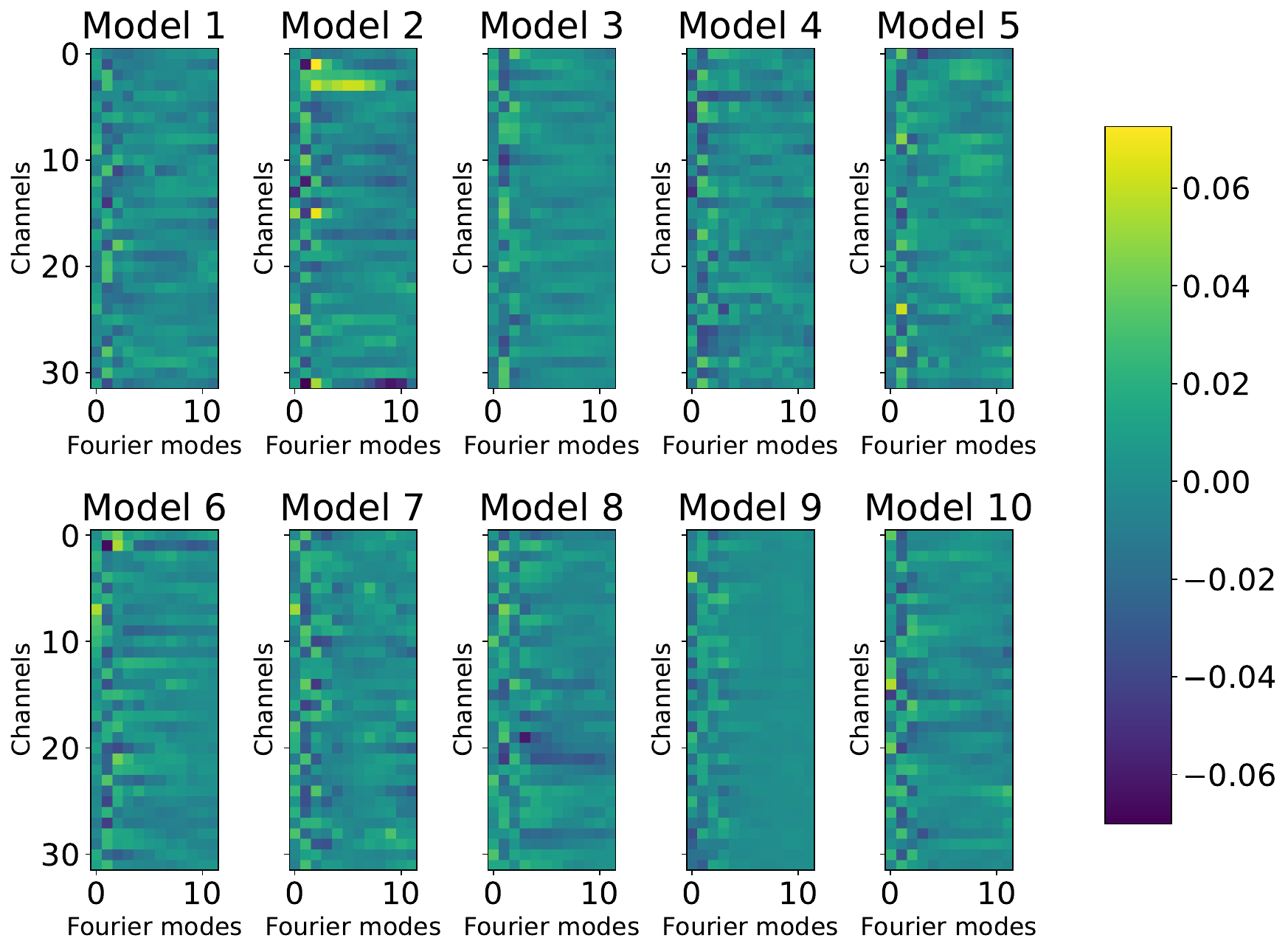}
    \caption{Heatmaps of the last Fourier layer weights from an ensemble of 10 FNO models.}
    \label{fig:ensemble_diversity:heatmaps_heat_last_layer}
    \end{subfigure}
    ~~~~~~~~~~~~~~~~~~~~~~~~~~
    \begin{subfigure}[h]{0.3\textwidth}
    \centering
    \includegraphics[scale=0.4]{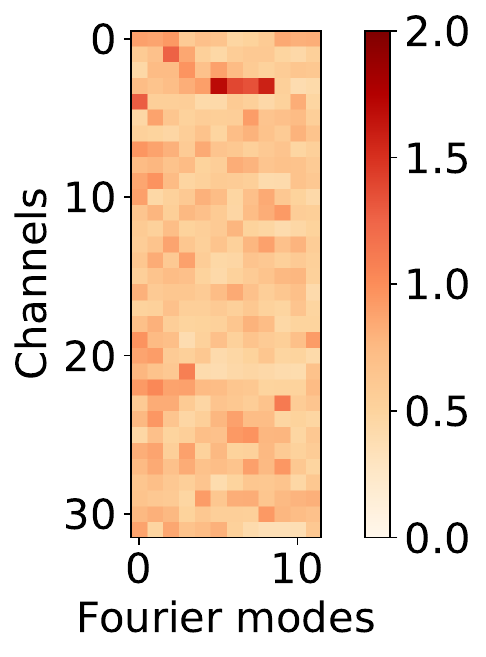}
    \caption{Coefficient of variation (mean/std) of the last Fourier layer weights across the 10 models in the ensemble.}
    \label{fig:ensemble_diversity:coefvar_heat_last_layer}
    \end{subfigure}
    
    \caption{Last Fourier layer of FNO models trained on 1-d heat equation task with $k^\tr\in[1,5]$.}
    \label{fig:ensemble_diversity_heat_layer3}
\end{figure}

\begin{figure}[H]
    \centering
    \begin{subfigure}[h]{0.3\textwidth}
    \centering
    \includegraphics[scale=0.25]{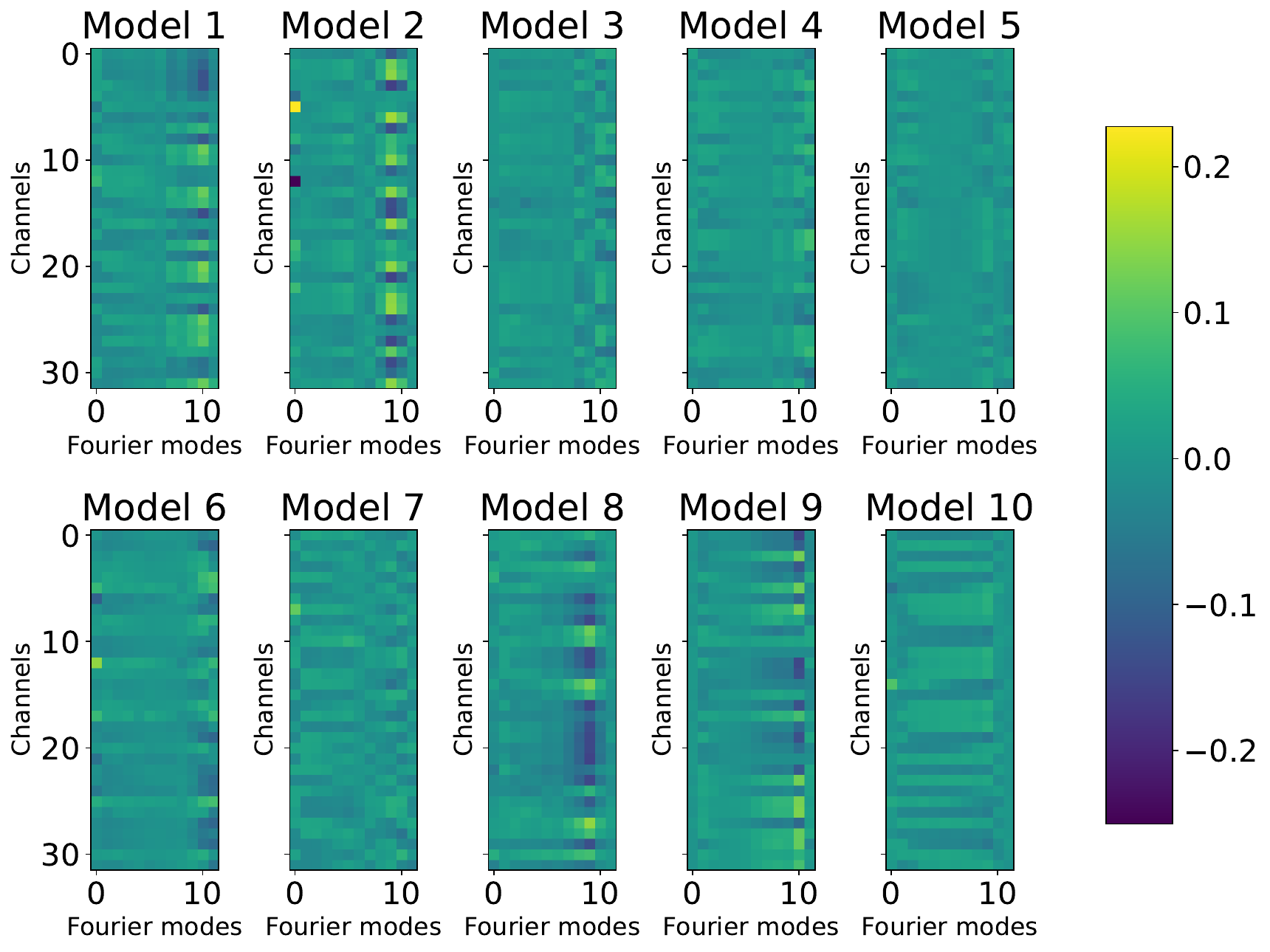}
    \caption{Heatmaps of the first Fourier layer weights from an ensemble of 10 FNO models.}
    \label{fig:ensemble_diversity:heatmaps_pme_layer0}
    \end{subfigure}
    ~~~~~~~~~~~~~~~~~~~~~~~~~~
    \begin{subfigure}[h]{0.3\textwidth}
    \centering
    \includegraphics[scale=0.4]{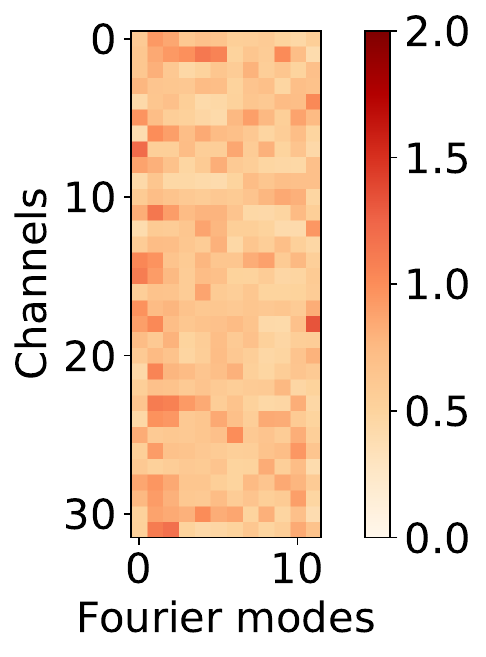}
    \caption{Coefficient of variation (mean/std) of the first Fourier layer weights across the 10 models in the ensemble.}
    \label{fig:ensemble_diversity:coefvar_pme_layer0}
    \end{subfigure}

    \caption{First Fourier layer of FNO models trained on 1-d PME task with $m^\tr\in[2,3]$.}
    \label{fig:ensemble_diversity_pme_layer0}
\end{figure}

\begin{figure}[H]
    \centering
    \begin{subfigure}[h]{0.3\textwidth}
    \centering
    \includegraphics[scale=0.25]{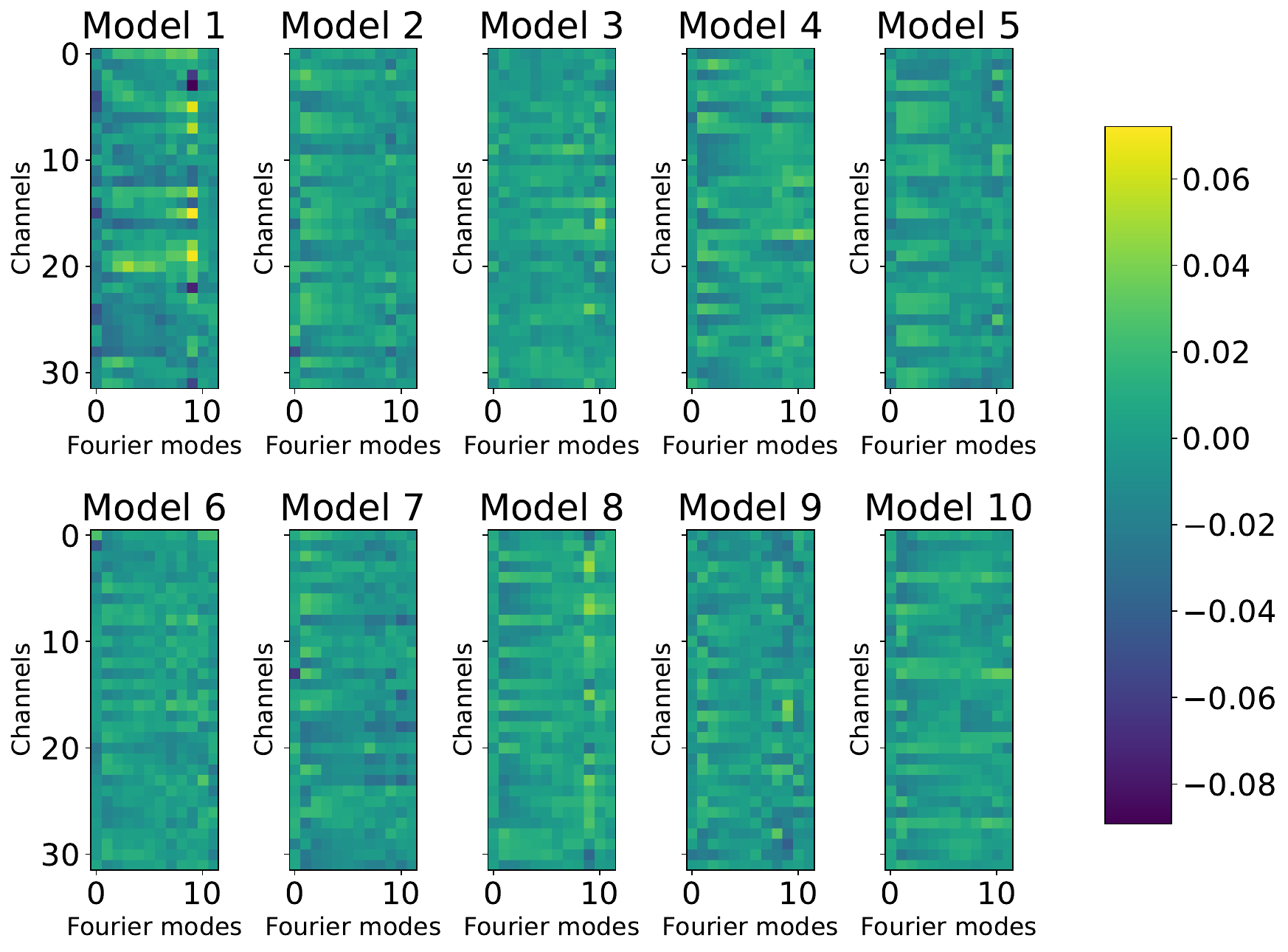}
    \caption{Heatmaps of the last Fourier layer weights from an ensemble of 10 FNO models.}
    \label{fig:ensemble_diversity:heatmaps_pme_last_layer}
    \end{subfigure}
    ~~~~~~~~~~~~~~~~~~~~~~~~~~
    \begin{subfigure}[h]{0.3\textwidth}
    \centering
    \includegraphics[scale=0.4]{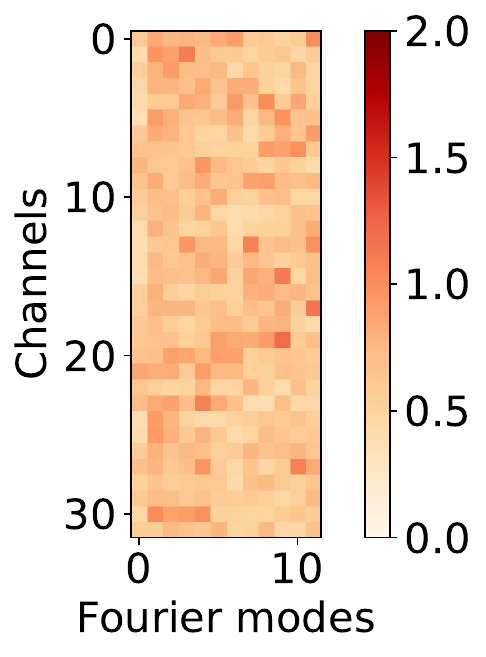}
    \caption{Coefficient of variation (mean/std) of the last Fourier layer weights across the 10 models in the ensemble. }
    \label{fig:ensemble_diversity:coefvar_pme_last_layer}
    \end{subfigure}
    \caption{Last Fourier layer of FNO models trained on 1-d PME task with $m^\tr\in[2,3]$.}
    \label{fig:ensemble_diversity_pme_layer3}
\end{figure}

\begin{figure}[H]
    \centering
    \begin{subfigure}[h]{0.3\textwidth}
    \centering
    \includegraphics[scale=0.25]{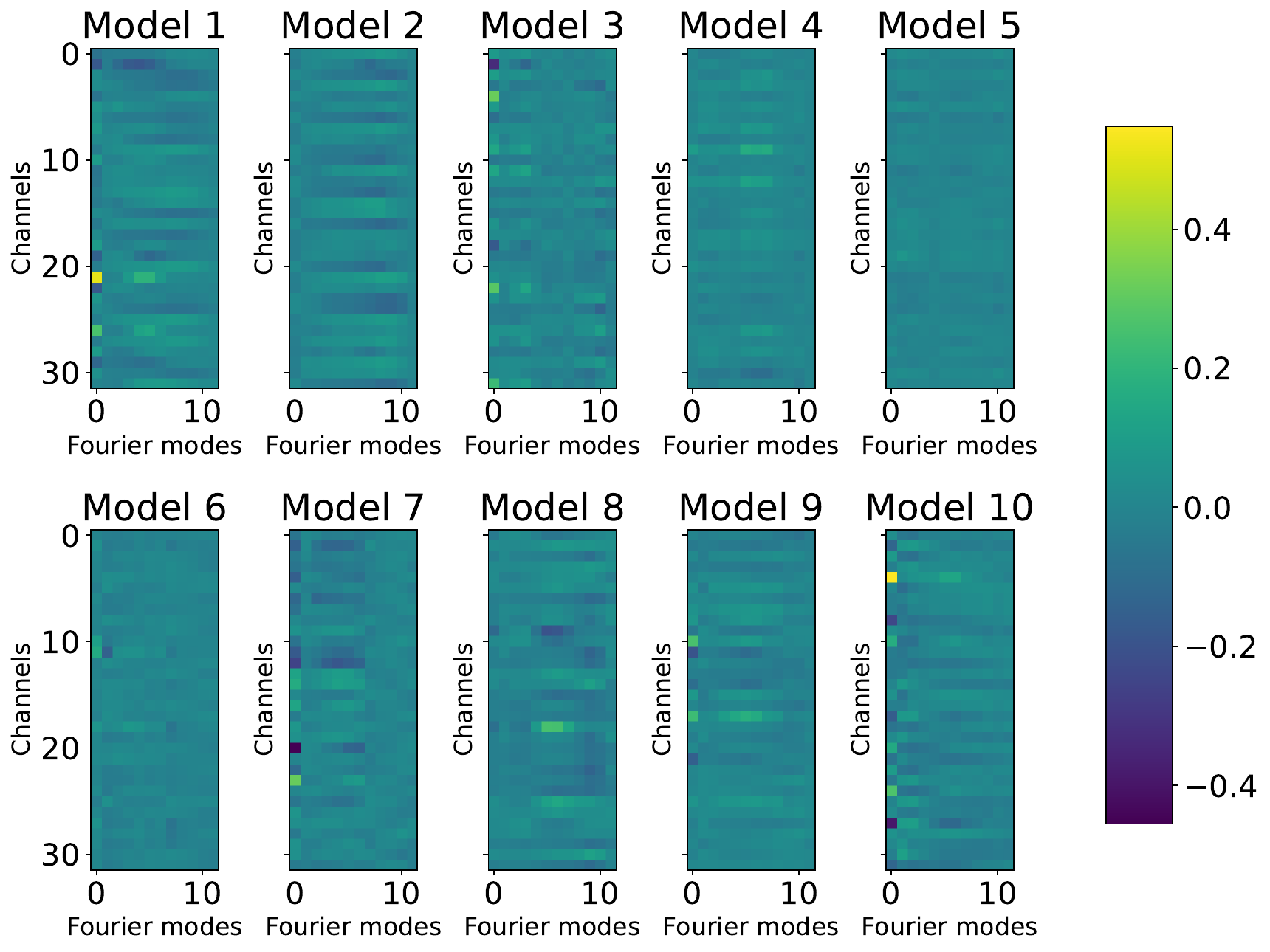}
    \caption{Heatmaps of first Fourier layer weights from an ensemble of 10 FNO models.}
    \label{fig:ensemble_diversity:heatmaps_stefan_layer0}
    \end{subfigure}
    ~~~~~~~~~~~~~~~~~~~~~~~~~~
    \begin{subfigure}[h]{0.3\textwidth}
    \centering
    \includegraphics[scale=0.4]{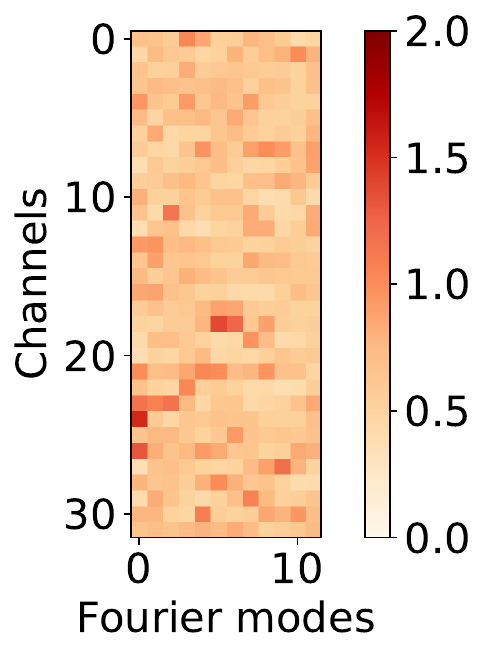}
    \caption{Coefficient of variation (mean/std) of the first Fourier layer weights across the 10 models in the ensemble.}
    \label{fig:ensemble_diversity:coefvar_stefan_layer0}
    \end{subfigure}
    
    \caption{First Fourier layer of FNO models trained on 1-d Stefan task with $u^{*,\tr}\in[0.6,0.65]$.}
    \label{fig:ensemble_diversity_stefan_layer0}
\end{figure}

\begin{figure}[H]
    \centering
    \begin{subfigure}[h]{0.3\textwidth}
    \centering
    \includegraphics[scale=0.25]{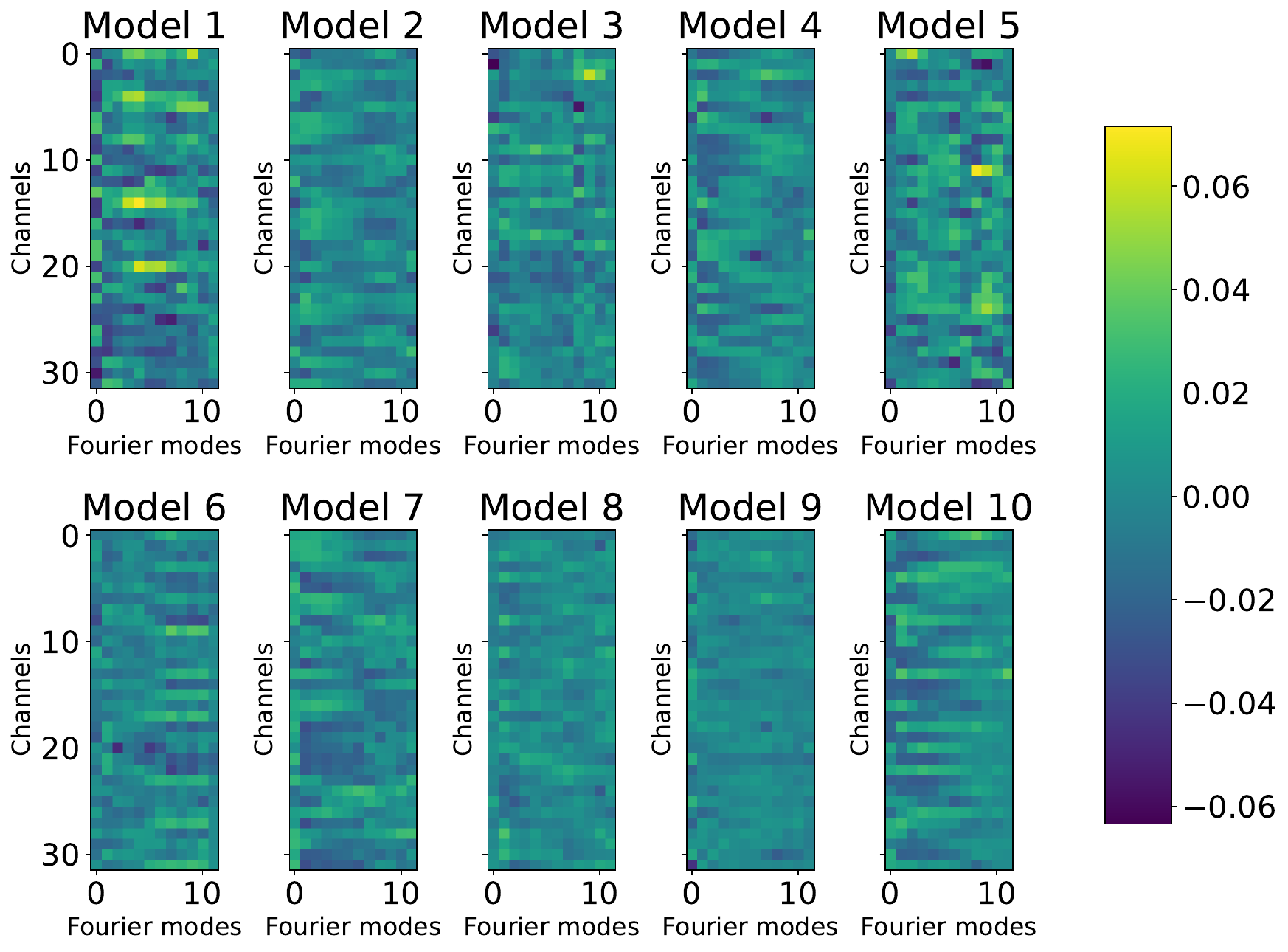}
    \caption{Heatmaps of the last Fourier layer weights from an ensemble of 10 FNO models.}
    \label{fig:ensemble_diversity:heatmaps_stefan_last_layer}
    \end{subfigure}
    ~~~~~~~~~~~~~~~~~~~~~~~~~~
    \begin{subfigure}[h]{0.3\textwidth}
    \centering
    \includegraphics[scale=0.4]{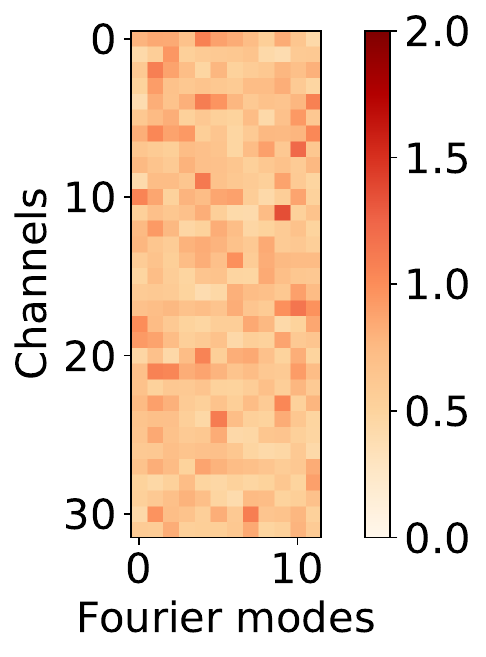}
    \caption{Coefficient of variation (mean/std) of the last Fourier layer weights across the 10 models in the ensemble.}
    \label{fig:ensemble_diversity:coefvar_stefan_last_layer}
    \end{subfigure}
    \caption{Last Fourier layer of FNO models trained on 1-d Stefan task with $u^{*,\tr}\in[0.6,0.65]$.}
\label{fig:ensemble_diversity_stefan_layer3}
\end{figure}

\begin{figure}[H]
    \centering
    \begin{subfigure}[h]{0.3\textwidth}
    \centering
    \includegraphics[scale=0.25]{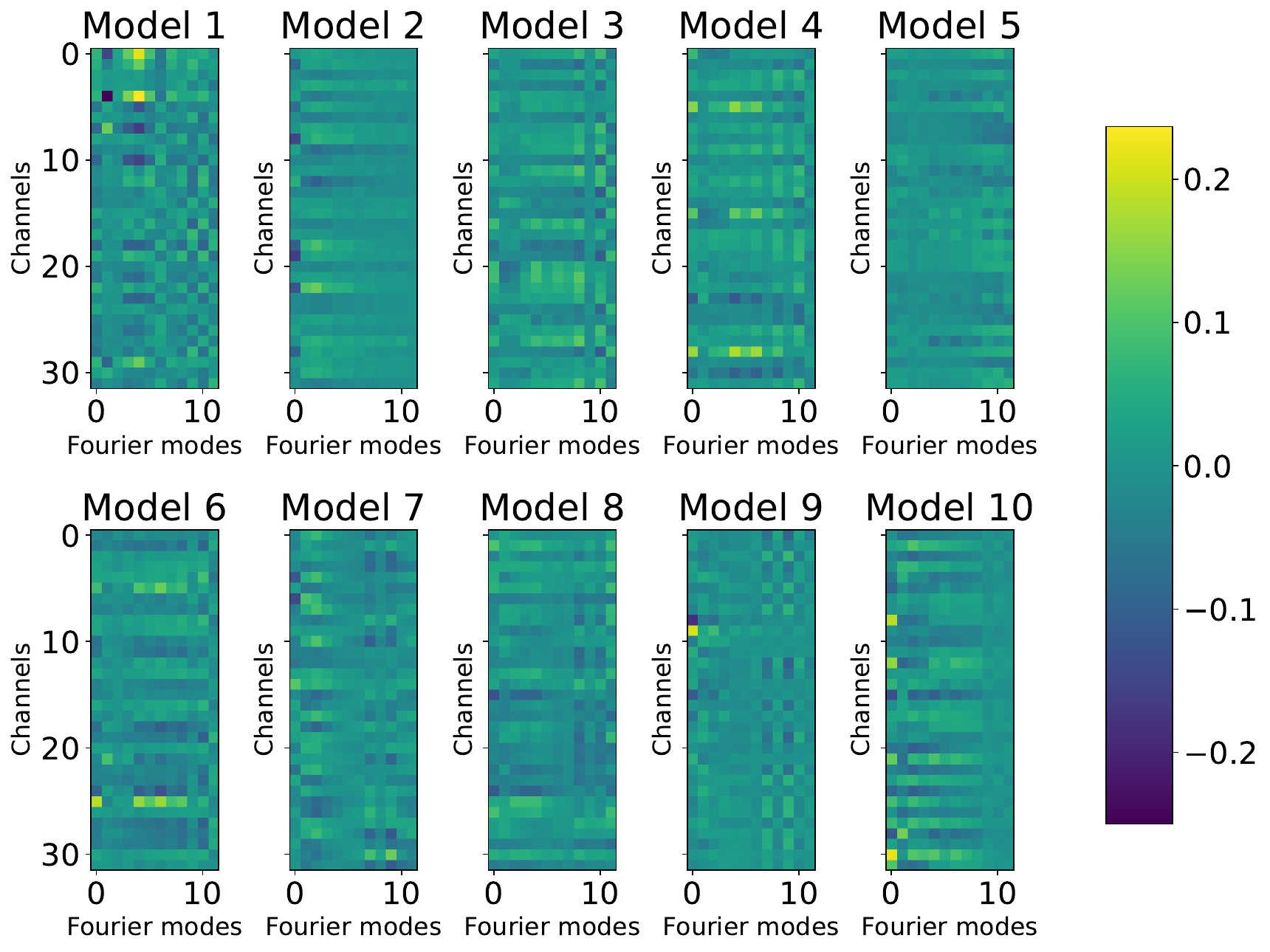}
    \caption{Heatmaps of first Fourier layer weights from an ensemble of 10 FNO models.}
    \label{fig:ensemble_diversity:heatmaps_la_layer0}
    \end{subfigure}
    ~~~~~~~~~~~~~~~~~~~~~~~~~~
    \begin{subfigure}[h]{0.3\textwidth}
    \centering
    \includegraphics[scale=0.4]{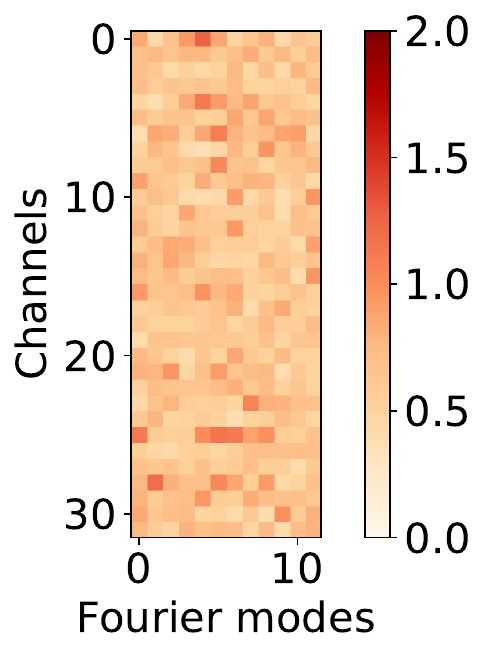}
    \caption{Coefficient of variation (mean/std) of the first Fourier layer weights across the 10 models in the ensemble.}
    \label{fig:ensemble_diversity:coefvar_la_layer0}
    \end{subfigure}
    
    \caption{First Fourier layer of FNO models trained on 1-d linear advection task with $\beta^\tr \in[1,2]$.}
    \label{fig:ensemble_diversity_la_layer0}
\end{figure}

\begin{figure}[H]
    \centering
    \begin{subfigure}[h]{0.3\textwidth}
    \centering
    \includegraphics[scale=0.25]{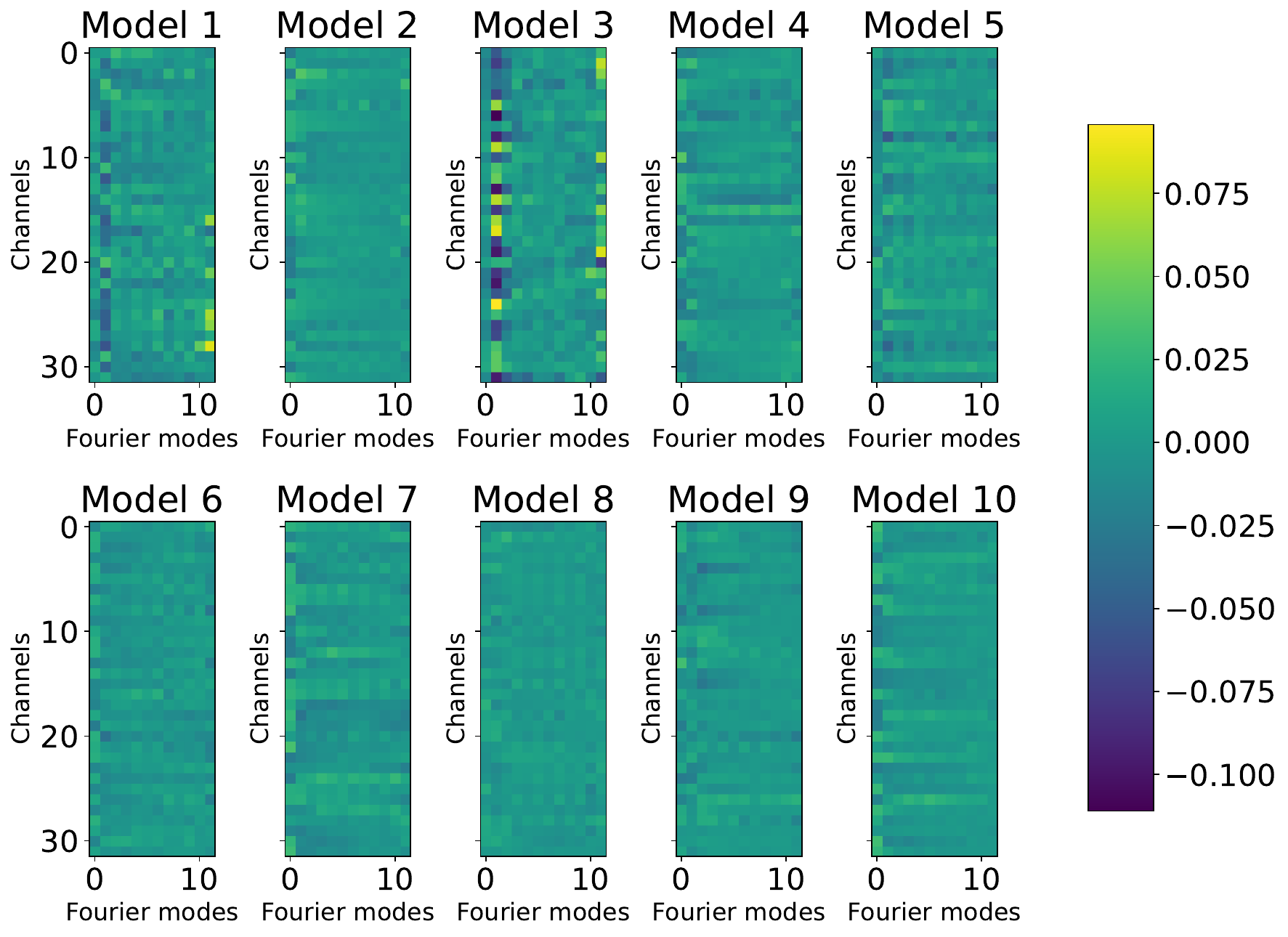}
    \caption{Heatmaps of the last Fourier layer weights from an ensemble of 10 FNO models. }
    \label{fig:ensemble_diversity:heatmaps_la_last}
    \end{subfigure}
    ~~~~~~~~~~~~~~~~~~~~~~~~~~
    \begin{subfigure}[h]{0.3\textwidth}
    \centering
    \includegraphics[scale=0.4]{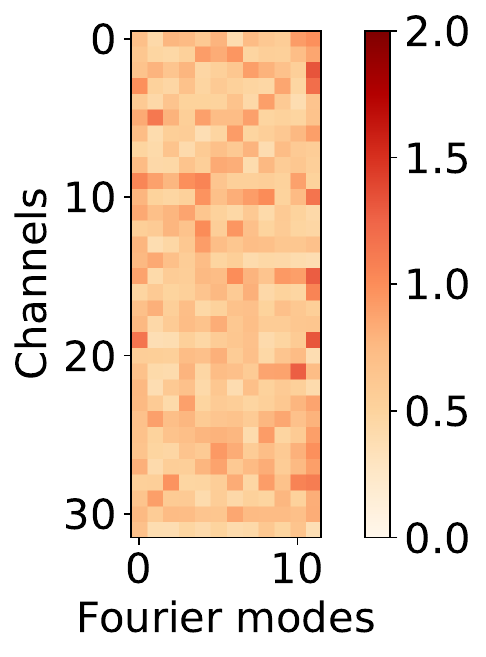}
    \caption{Coefficient of variation (mean/std) of the last Fourier layer weights across the 10 models in the ensemble.}
    \label{fig:ensemble_diversity:coefvar_la_last}
    \end{subfigure}

    \caption{Last Fourier layer of FNO models trained on 1-d linear advection task with $\beta^\tr \in[1,2]$.}
    \label{fig:ensemble_diversity_la_layer3}
\end{figure}

Figures \ref{fig:ensemble_diversity_heat_layer0}-\ref{fig:ensemble_diversity_la_layer3} illustrate the diversity in each FNO model in the ensemble across a wide variety of PDEs with various levels of difficulty and various Fourier layers.  In each figure, (a) shows the heatmaps of the weights in the corresponding Fourier layer across the channels and Fourier modes. This apparent diversity is reinforced in (b), which plots the coefficient of variation, i.e., the (mean/std) across all the models in the ensemble across the channels and Fourier modes.

\section{\method Ablations}
In our proposed \method, we solve the following optimization problem in \cref{eq:optimization} given as 
\begin{equation}
\label{eq:optimization_app}
\hat{\rmW}  = \argmin_{\rmW}  \underbrace{\frac{1}{N M} \sum_{i=1}^N \sum_{m=1}^M{\frac{||\hat{u}_m^{(i)}-  u^{(i)}||^2_{L_2}} {||u^{(i)}||^2_{L_2}}}}_{\text{unconstrained NO loss}} \nonumber  
\underbrace{- \frac{2\lambda_\diverse}{M(M-1)}\sum_{m,k:m< k} ||\rmW_m - \rmW_k||^2_2}_{\text{diversity regularization}}.
\end{equation} In this section, we first study the effect of the hyperparameter $M$, which denotes the number of prediction heads on the accuracy. We then study the effect of various types of diversity regularizations. 

\subsection{Hyperparameter study on the number of prediction heads $M$}
\label{subsec:ablation_num_heads}
\begin{figure}[H]
    \centering
    \begin{subfigure}[h]{0.45\textwidth}
    \centering
    \includegraphics[scale=0.35]{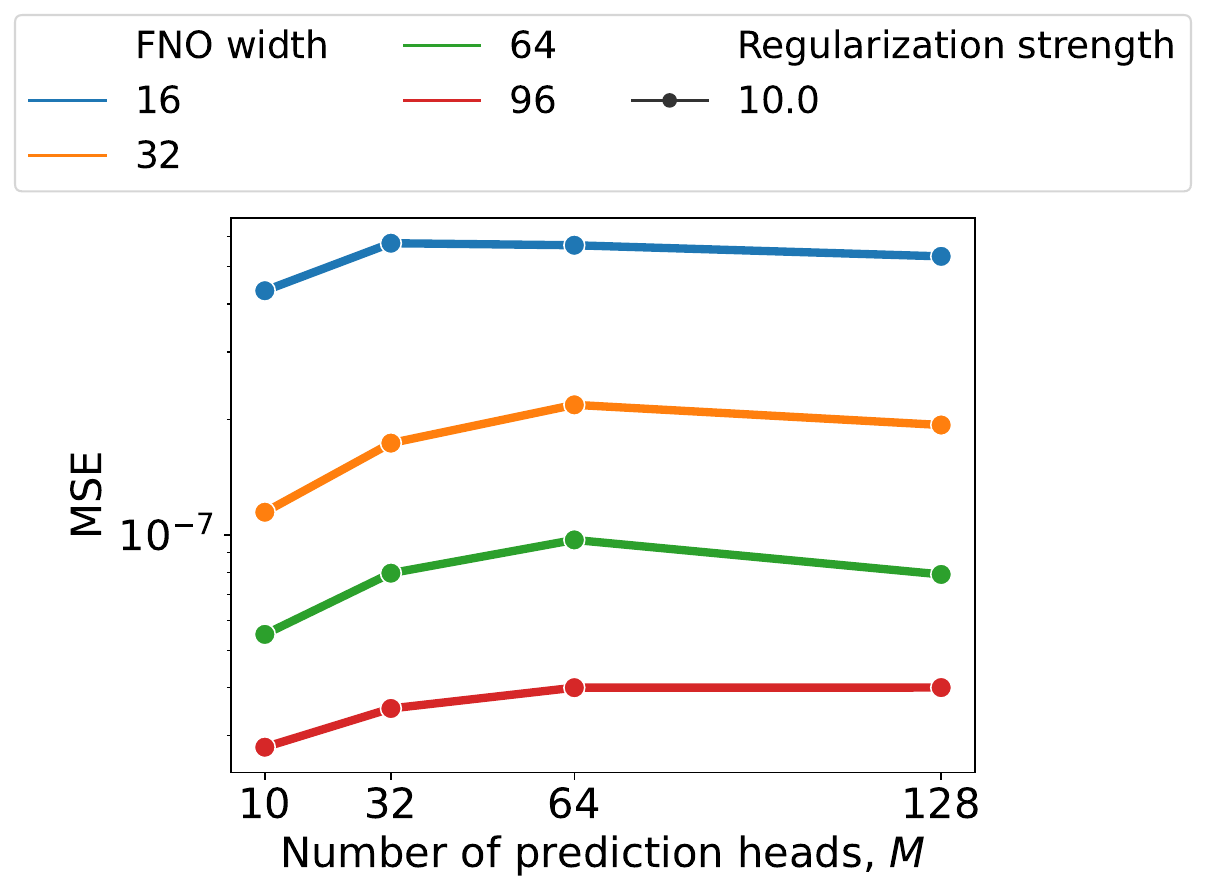}
    \caption{In-domain}
    \end{subfigure}
    ~~
    \begin{subfigure}[h]{0.45\textwidth}
    \centering
    \includegraphics[scale=0.35]{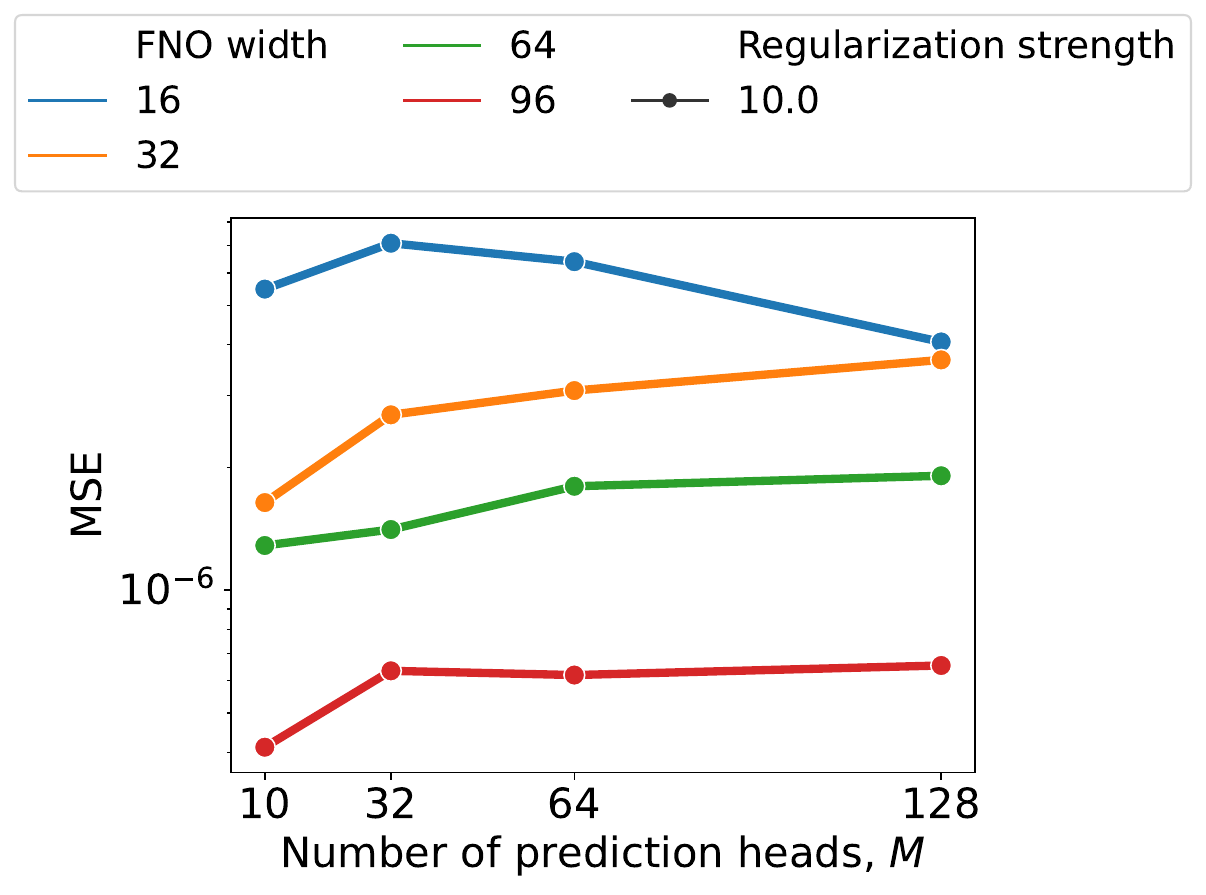}
    \caption{OOD small}
    \end{subfigure}

    \begin{subfigure}[h]{0.45\textwidth}
    \centering
    \includegraphics[scale=0.35]{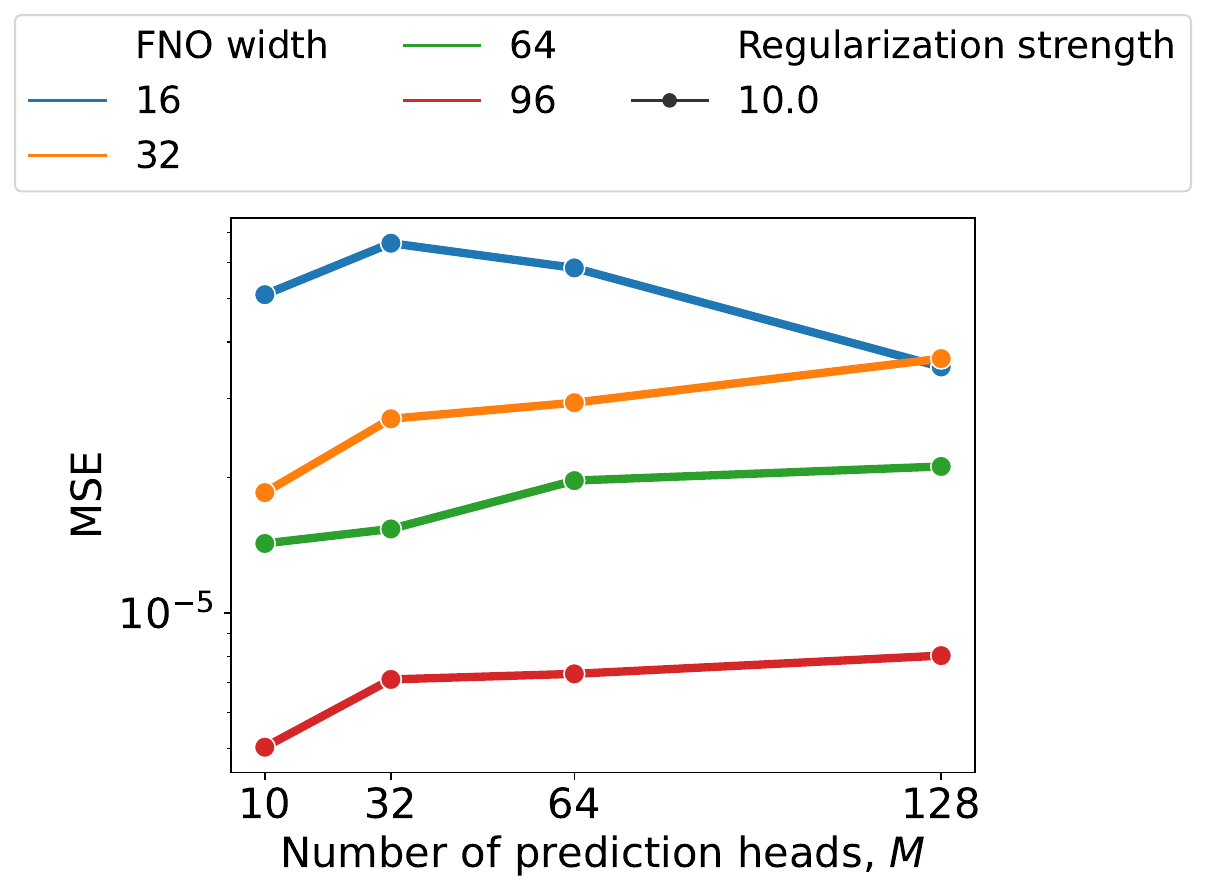}
    \caption{OOD medium}
    \end{subfigure}
    ~~
    \begin{subfigure}[h]{0.45\textwidth}
    \centering
    \includegraphics[scale=0.35]{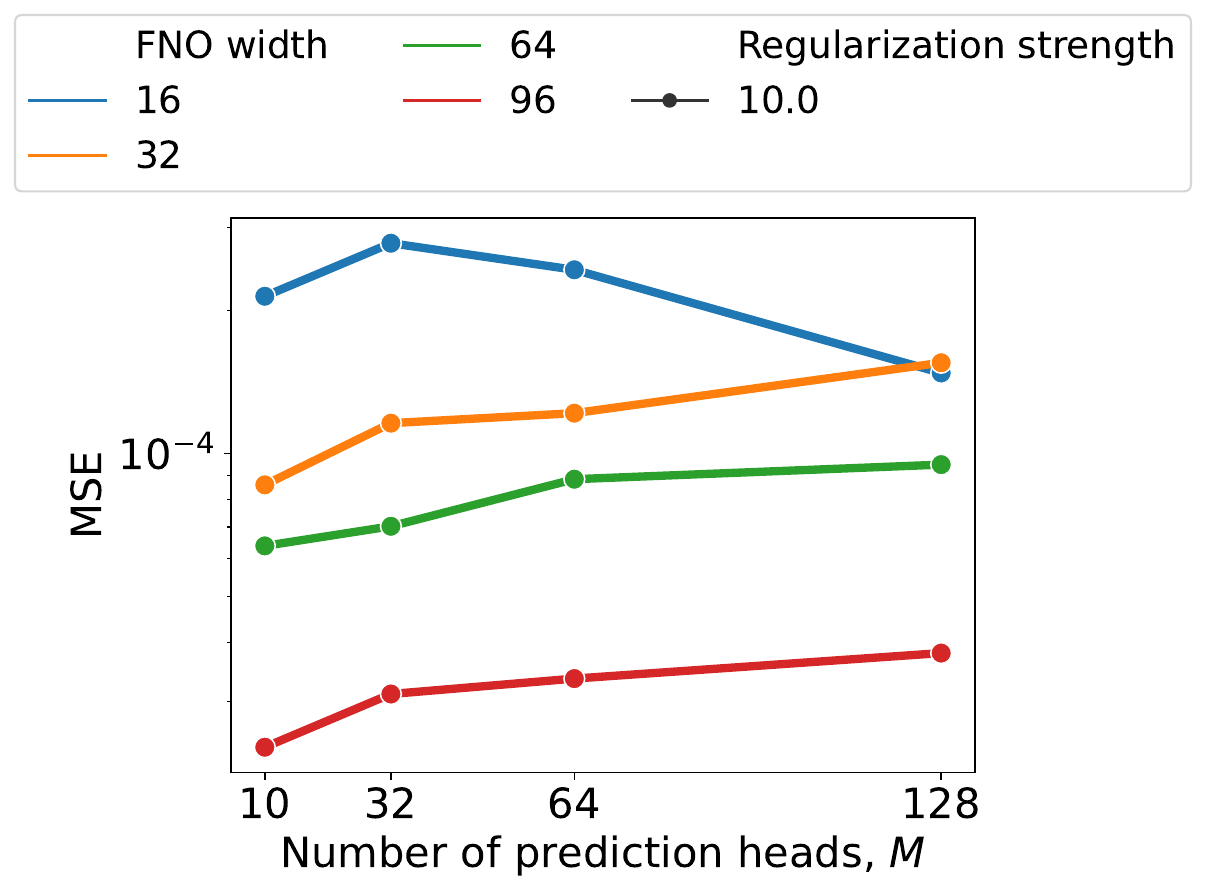}
    \caption{OOD large}
    \end{subfigure}
    \caption{\textbf{Effect of $M$ on MSE.} 
    The MSE metric as a function of the number of prediction heads $M$ for various penultimate layer sizes. The diversity regularization strength $\lambda_{\text{diverse}}$ is fixed at 10. 
    Increasing the number of prediction heads increases MSE mildly likely due to the fact that it is harder to train each of the prediction heads accurately to match the output. 
    The MSE significantly decreases when increasing the FNO channel width, which is expected as the shared parameters of the model increase. 
    } 
    \label{fig:Manalysis_mse}
\end{figure}

\begin{figure}[H]
    \centering
    \begin{subfigure}[h]{\figsizeapp\textwidth}
    \centering
    \includegraphics[scale=\figscaleapp]{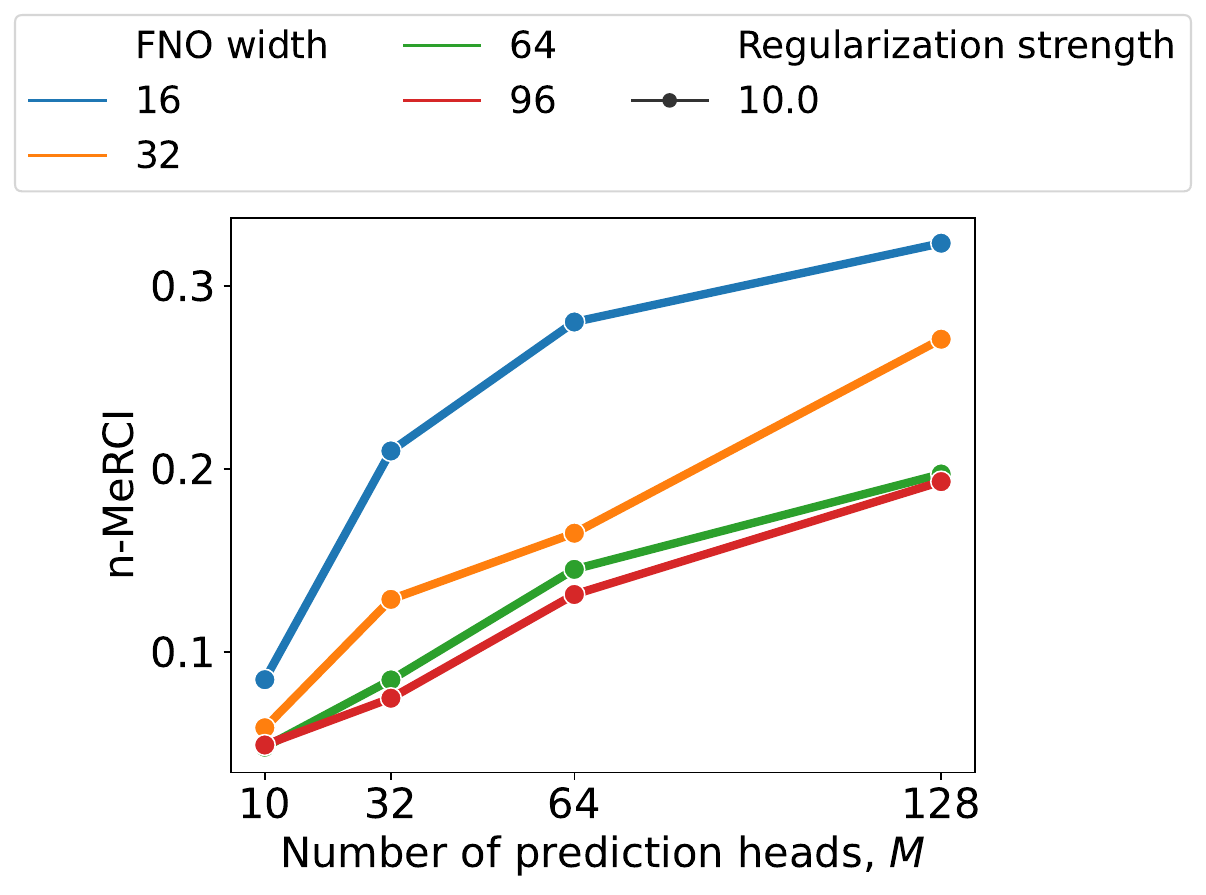}
    \caption{OOD small}
    \end{subfigure}
    ~~ 
    \begin{subfigure}[h]{0.32\textwidth}
    \centering
    \includegraphics[scale=0.28]{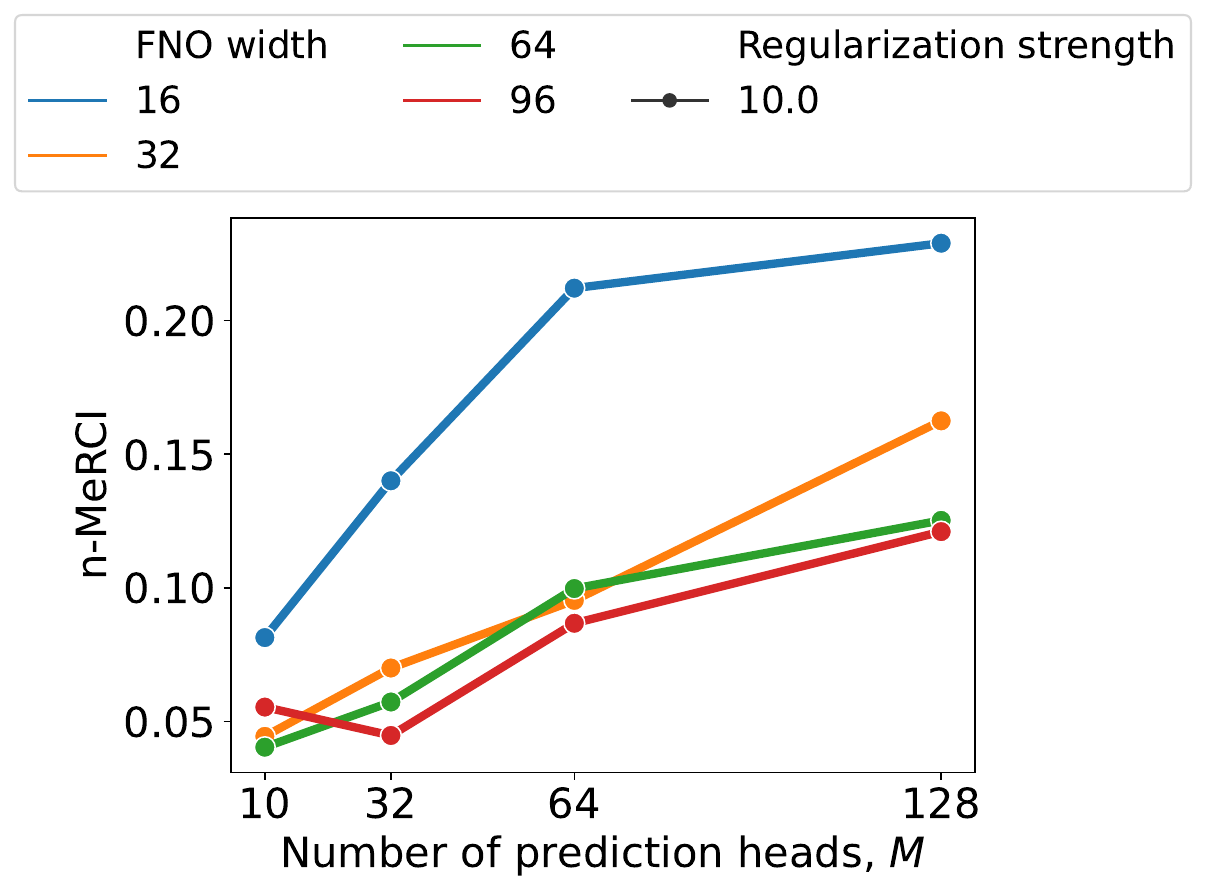}
    \caption{OOD medium}
    \end{subfigure}
    ~~
    \begin{subfigure}[h]{0.32\textwidth}
    \centering
    \includegraphics[scale=0.28]{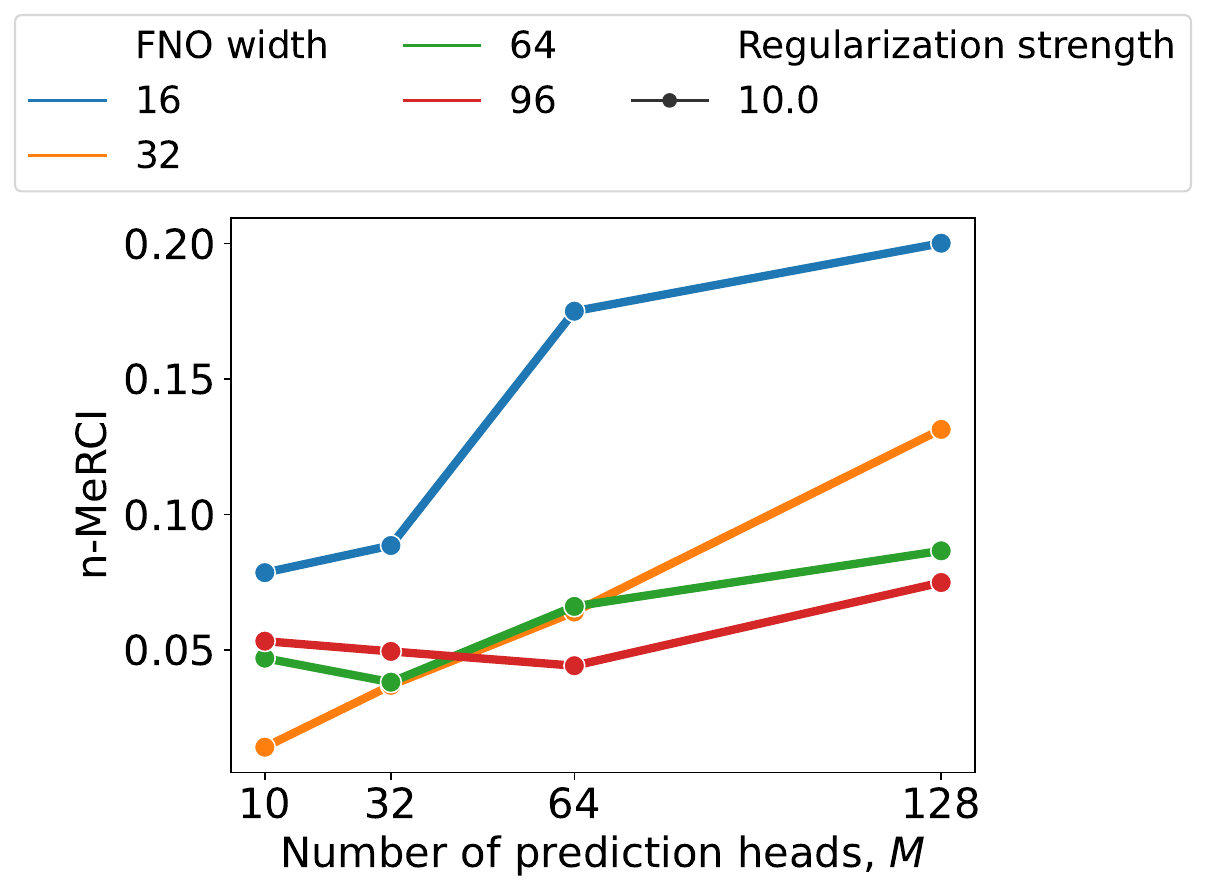}
    \caption{OOD large}
    \end{subfigure}
    \caption{
    \textbf{Effect of $M$ on n-MeRCI.} 
    The n-MeRCI metric as a function of the the number of prediction heads $M$ for various penultimate layer sizes. The diversity regularization strength $\lambda_{\text{diverse}}$ is fixed at 10.
    The UQ estimates vary with the number of prediction heads. A large value of $M$ generally degrades the n-MeRCI metric given a fixed regularization strength because we cannot diversify all possible pairs of prediction heads as it is computationally expensive. 
    } 
    \label{fig:Manalysis_nmerci}
\end{figure}

\begin{figure}[H]
    \centering
    \includegraphics[scale=0.35]{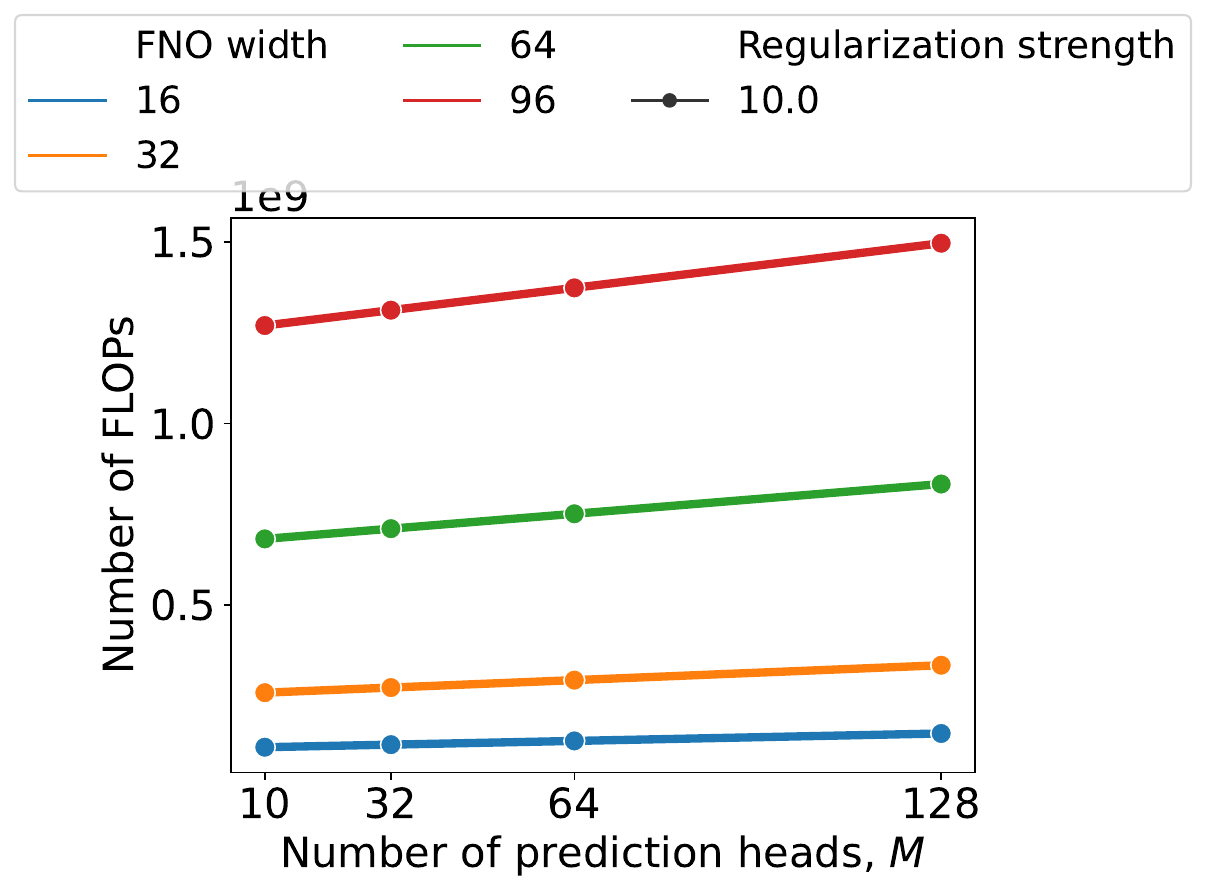}
    \caption{\textbf{Effect of $M$ on FLOPs.} 
    The number of FLOPs  for one forward pass as a function of the number of prediction heads $M$ for various penultimate layer sizes. The diversity regularization strength $\lambda_{\text{diverse}}$ is fixed at 10.
        The number of FLOPs increases mildly with the number of prediction heads.
    } 
    \label{fig:Manalysis_flops}
\end{figure}

In this subsection, we perform a detailed hyperparameter study on the number of prediction heads $M$ across various metrics. We see that our choice of $M=10$ generally gives the lowest MSE (\cref{fig:Manalysis_mse}), n-MeRCI (\cref{fig:Manalysis_nmerci}) and number of FLOPs (\cref{fig:Manalysis_flops}) for various FNO channel widths on in-domain and varying OOD tasks.

\subsection{Diversity Regularization Ablations}
\label{subsec:diversity_reg_app}
\begin{figure}[H]
    \centering
    \begin{subfigure}[t]{\textwidth}
        \centering
        \includegraphics[width=5in, height=0.4in]{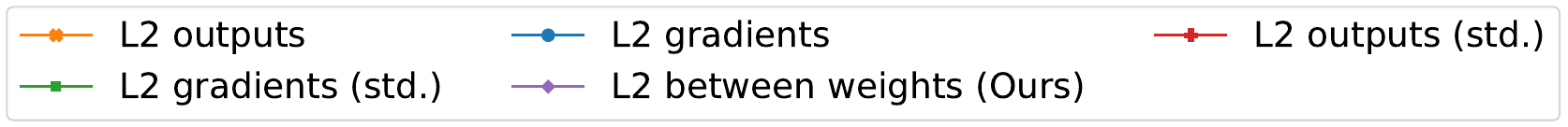}
    \end{subfigure}
    
    \begin{subfigure}[h]{0.30\textwidth}
    \centering
    \includegraphics[scale=0.32]{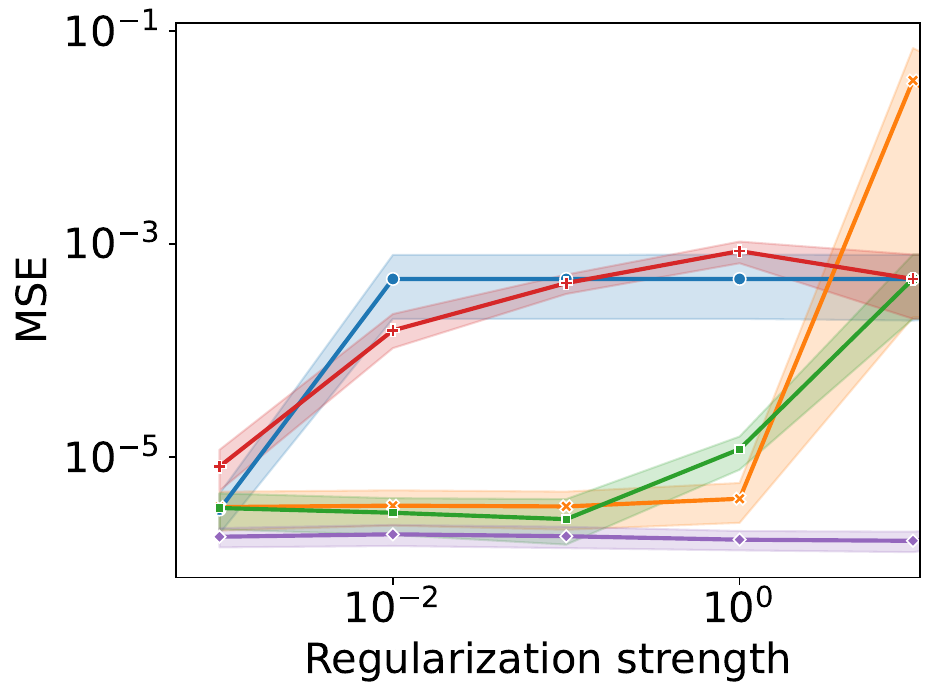}
    \caption{Small OOD shift}
    \end{subfigure}
    ~~
    \begin{subfigure}[h]{0.30\textwidth}
    \centering
    \includegraphics[scale=0.32]{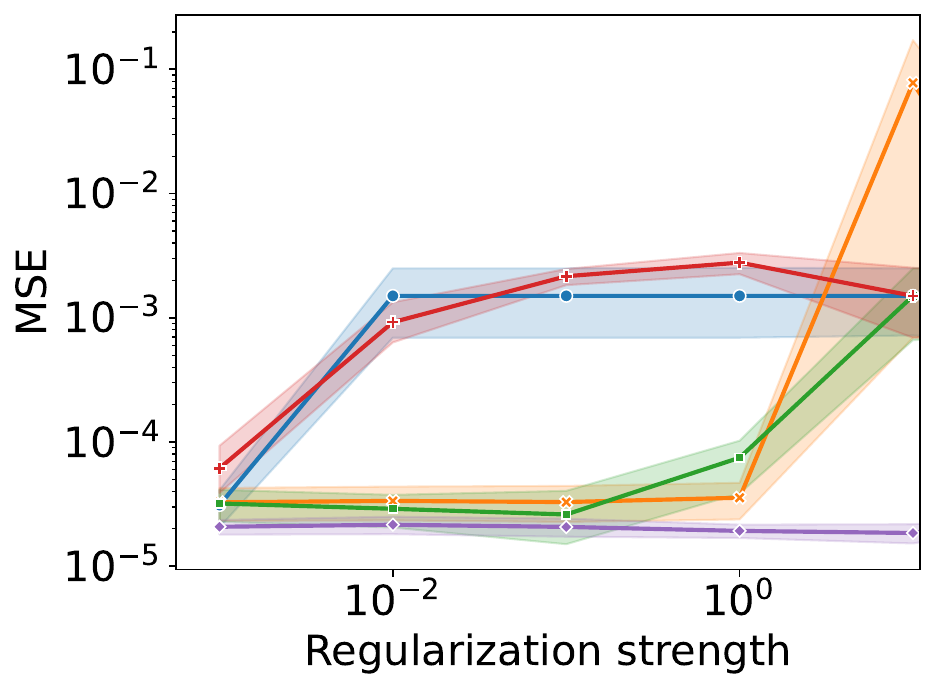}
    \caption{Medium OOD shift}
    \end{subfigure}
    ~~
    \begin{subfigure}[h]{0.30\textwidth}
    \centering
    \includegraphics[scale=0.32]{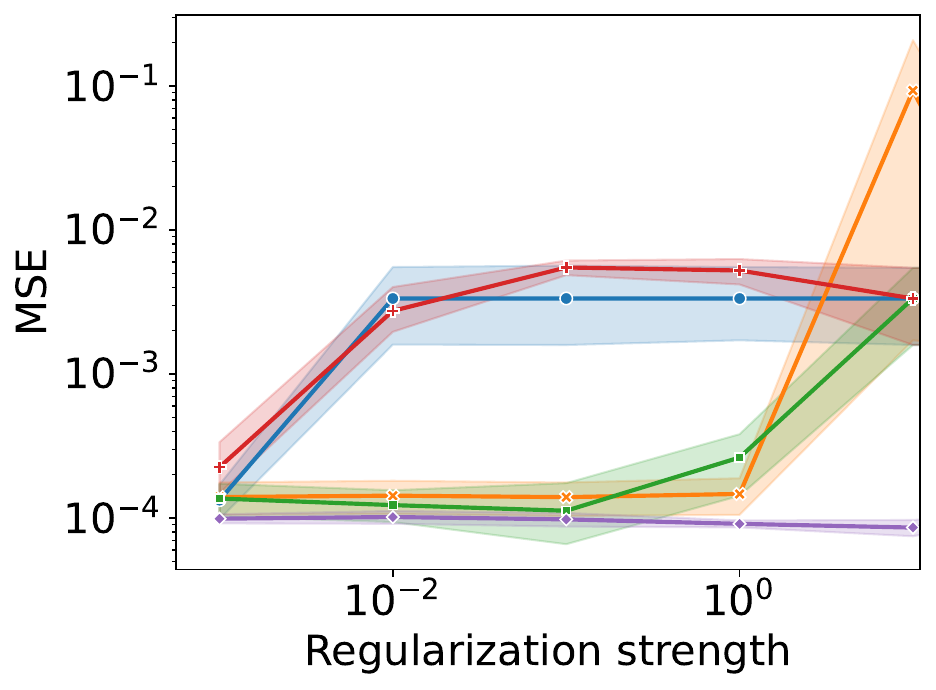}
    \caption{Large OOD shift}
    \end{subfigure}
    
    \caption{{\bf MSE diversity regularization ablation.} Effect of different diversity regularizations  on MSE for small, medium, large OOD shifts in the 1-d heat equation. Our proposed regularization of enforcing diversity over the weights of each head has the best performance across various regularization strengths and OOD shifts.}
    \label{fig:divablation_heat_mse}
\end{figure}

\begin{figure}[H]
    \centering
    \begin{subfigure}[t]{\textwidth}
        \centering
        \includegraphics[width=5in, height=0.4in]{new_figures/divablation_legend.pdf}
    \end{subfigure}
    
    \begin{subfigure}[h]{0.30\textwidth}
    \centering
    \includegraphics[scale=0.32]{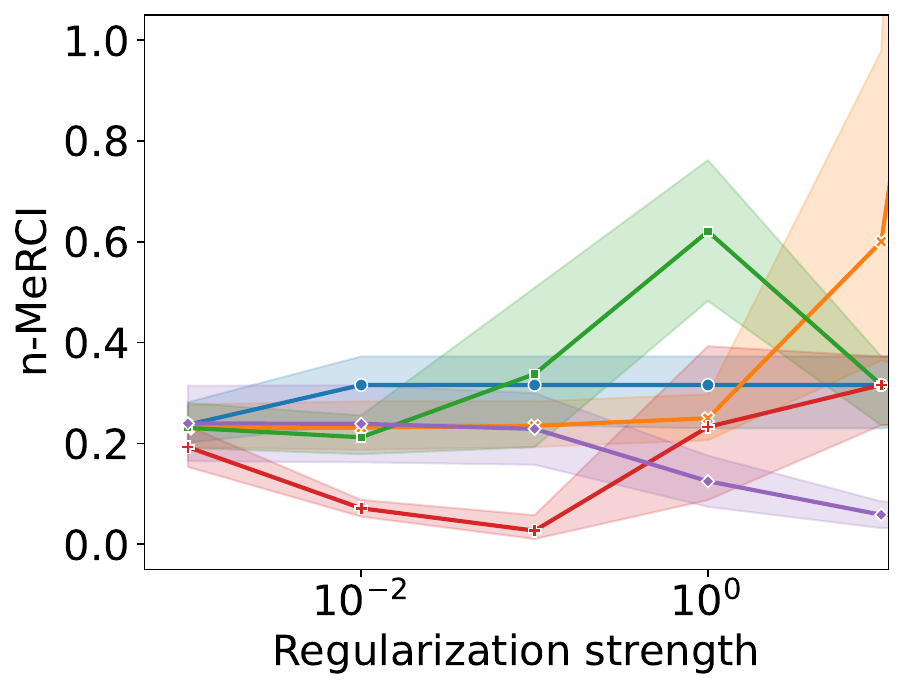}
    \caption{Small OOD shift}
    \end{subfigure}
    ~~
    \begin{subfigure}[h]{0.30\textwidth}
    \centering
    \includegraphics[scale=0.32]{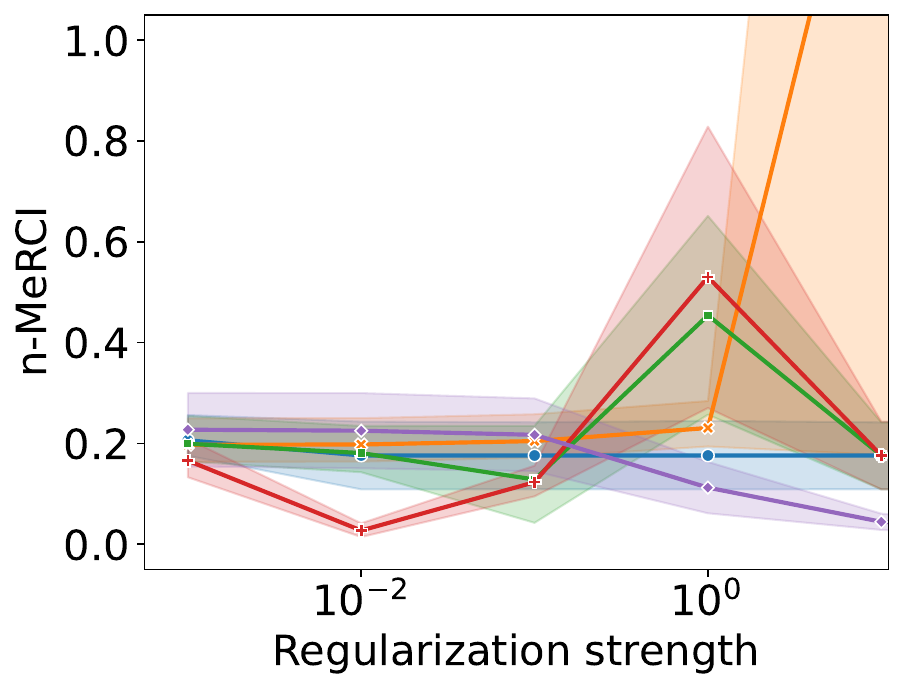}
    \caption{Medium OOD shift}
    \end{subfigure}
    ~~
    \begin{subfigure}[h]{0.30\textwidth}
    \centering
    \includegraphics[scale=0.32]{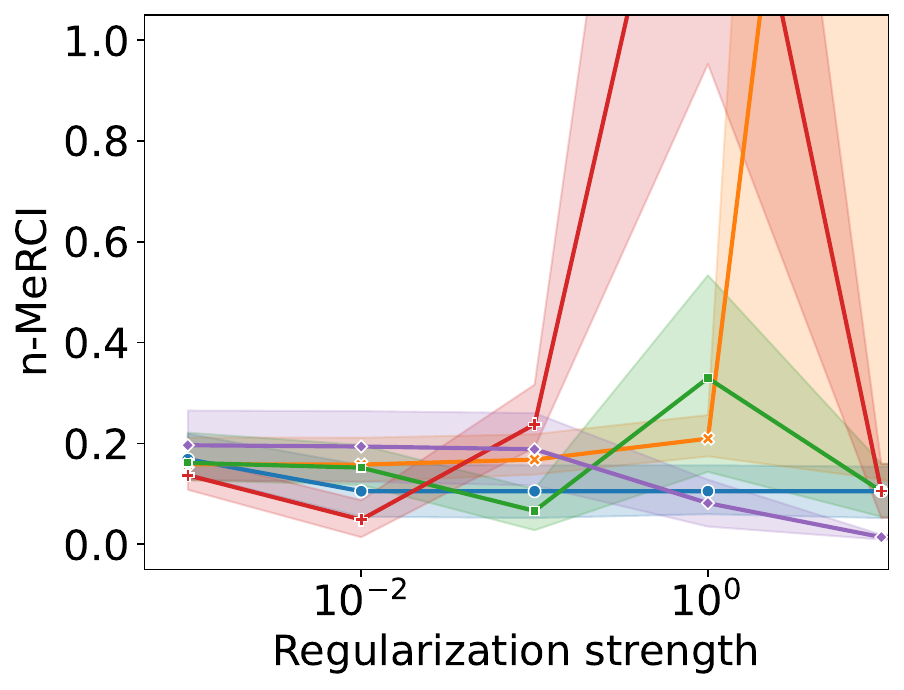}
    \caption{Large OOD shift}
    \end{subfigure}
    
    \caption{{\bf n-MeRCI diversity regularization ablation.} Effect of different diversity regularizations  on n-MeRCI for small, medium, large OOD shifts in the 1-d heat equation. Our proposed regularization of enforcing diversity over the weights of each head has the best performance across various regularization strengths and OOD shifts.}
    \label{fig:divablation_heat_nmerci}
\end{figure}

We test different regularizations in the second term of \cref{eq:optimization_app} to enforce diversity in the last layer heads of \method: maximize the $L_2$ loss between {\bf (a)} the weights of each head (ours), {\bf (b)} the OOD predictions of each head, and {\bf (c)} the gradients with respect to each head. 
For {\bf (b)} and {\bf (c)}, we try two variants: the $L_2$ between the respective quantities after standardizing them to have mean 0 and variance 1, and without standardization. 
Note that {\bf (b)} requires access to a set of OOD inputs without corresponding target outputs. 
Figures \ref{fig:divablation_heat_mse}-\ref{fig:divablation_heat_nmerci} show that the proposed diversity measure results in best MSE and n-MeRCI performance across all OOD shifts.
Regularizations that directly diversify the outputs or gradients are also sensitive to the regularization strength and require a careful trade-off between the prediction loss and regularization penalty. 
Our proposed regularization is more robust to the diversity hyperparameter $\lambda_{\text{diverse}}$, and monotonically improves the performance for reasonable values of this regularization strength. 

\begin{figure}[H]
    \centering
    \begin{subfigure}[h]{\figsizeapp\textwidth}
    \centering
    \includegraphics[scale=\figscaleapp]{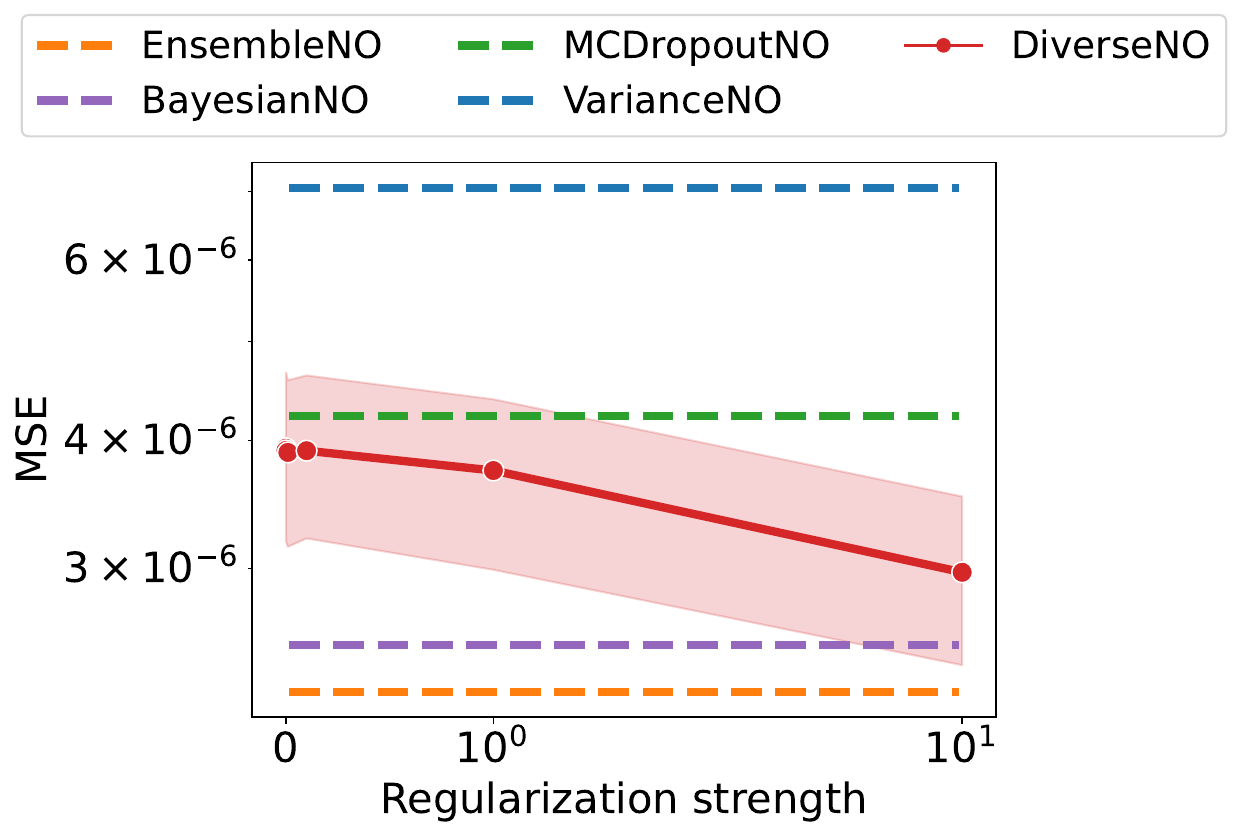}
    \caption{Small OOD shift}
    \end{subfigure}
    ~~
    \begin{subfigure}[h]{\figsizeapp\textwidth}
    \centering
    \includegraphics[scale=\figscaleapp]{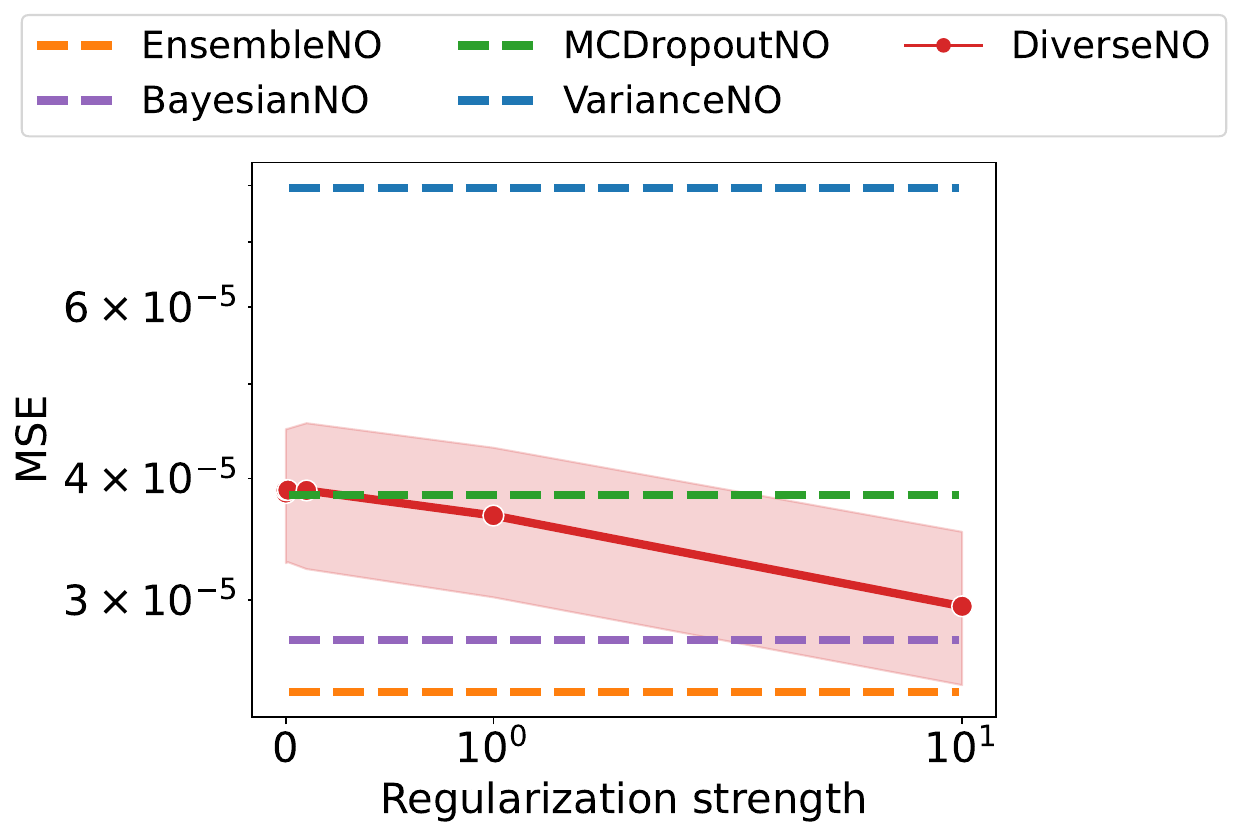}
    \caption{Medium OOD shift}
    \end{subfigure}
    ~~
    \begin{subfigure}[h]{\figsizeapp\textwidth}
    \centering
    \includegraphics[scale=\figscaleapp]{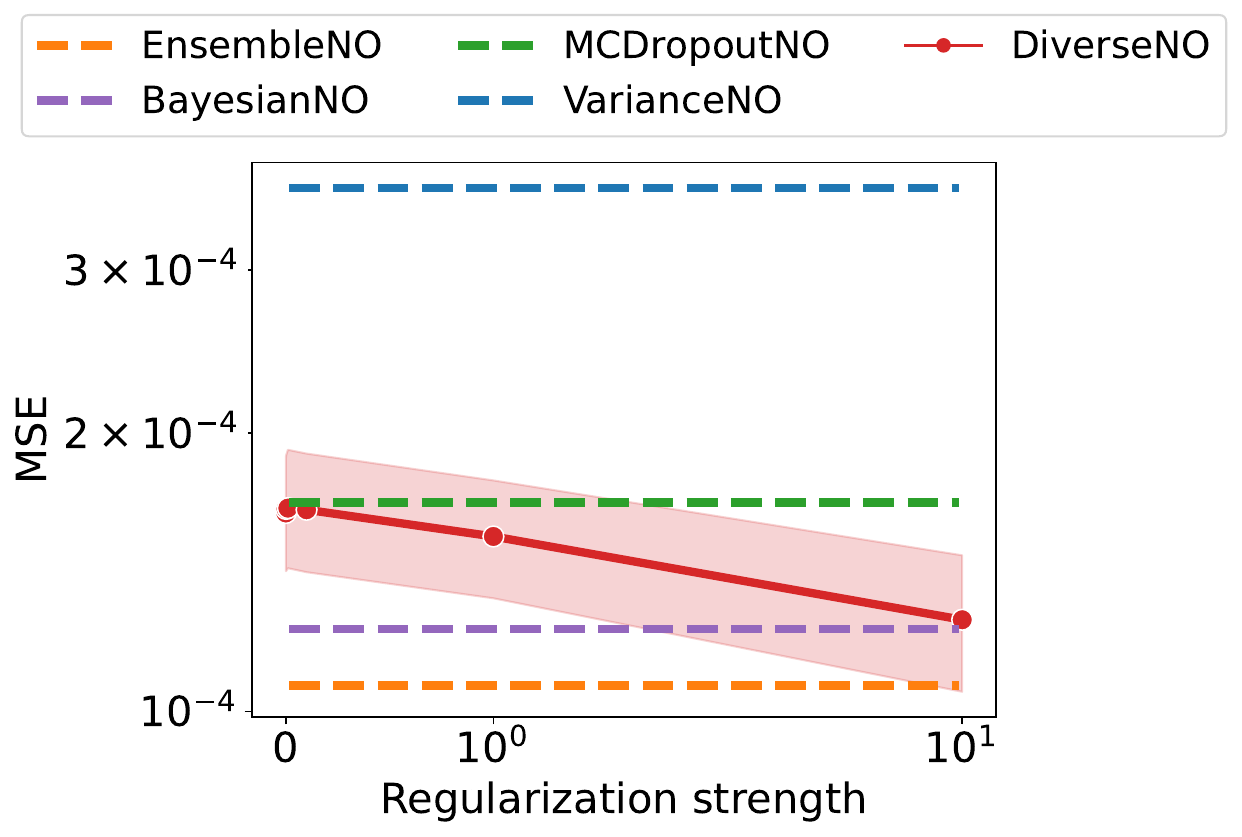}
    \caption{Large OOD shift}
    \end{subfigure}
    
    \caption{{\bf Effect of $\lambda_{\text{diverse}}$ on MSE.}     Comparison of the MSE metric as a function of the diversity regularization strength $\lambda_{\text{diverse}}$ for \method to the MSE of the other baselines. }
    \label{fig:div_regstrength_heat_mse}
\end{figure}

\begin{figure}[H]
    \centering
    \begin{subfigure}[h]{\figsizeapp\textwidth}
    \centering
    \includegraphics[scale=0.28]{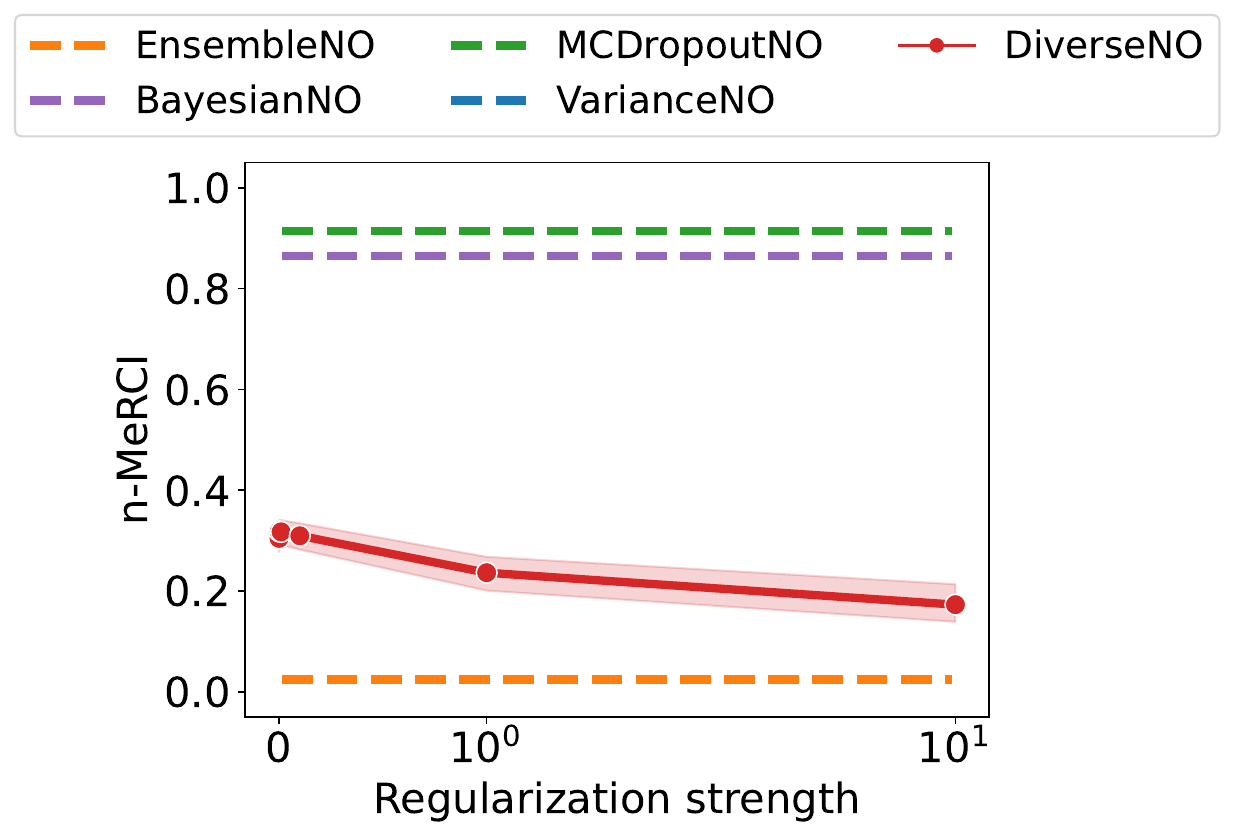}
    \caption{Small OOD shift}
    \end{subfigure}
    ~~
    \begin{subfigure}[h]{\figsizeapp\textwidth}
    \centering
    \includegraphics[scale=0.28]{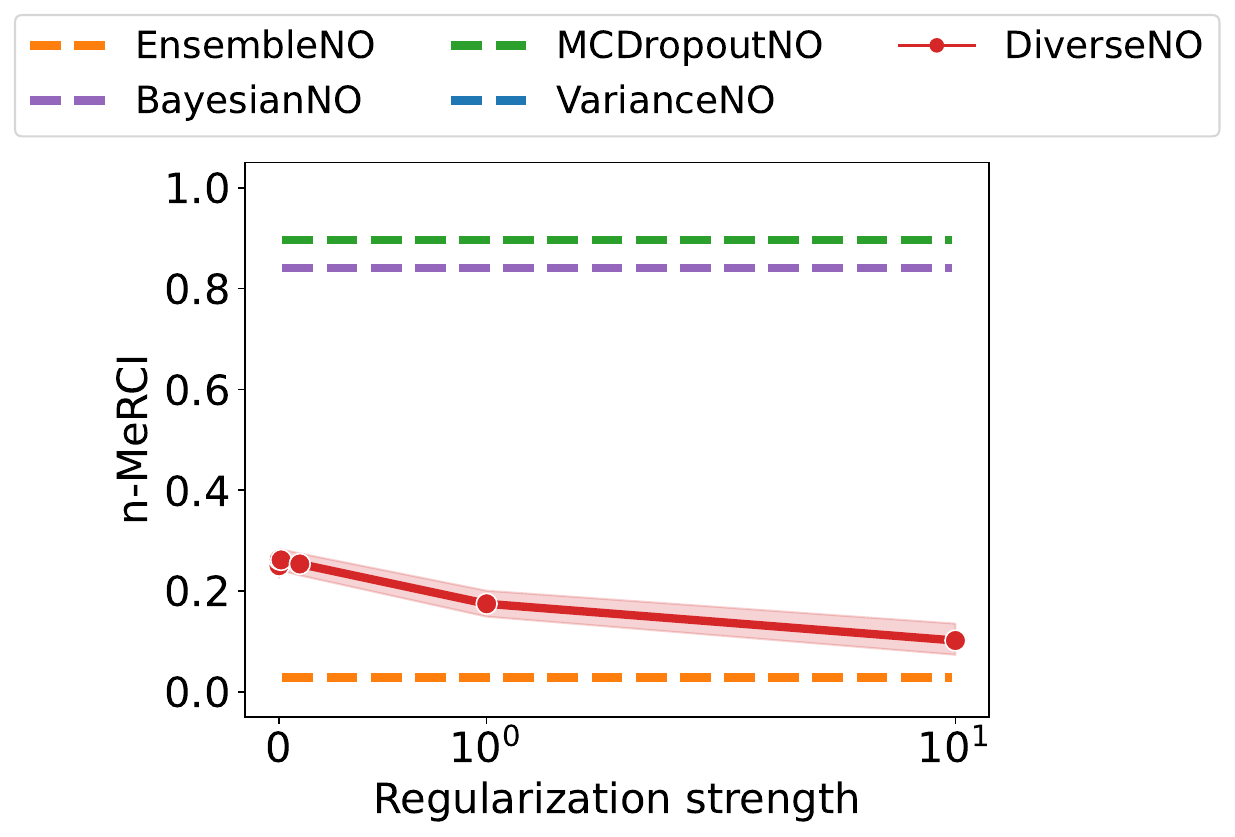}
    \caption{Medium OOD shift}
    \end{subfigure}
    ~~
    \begin{subfigure}[h]{\figsizeapp\textwidth}
    \centering
    \includegraphics[scale=0.28]{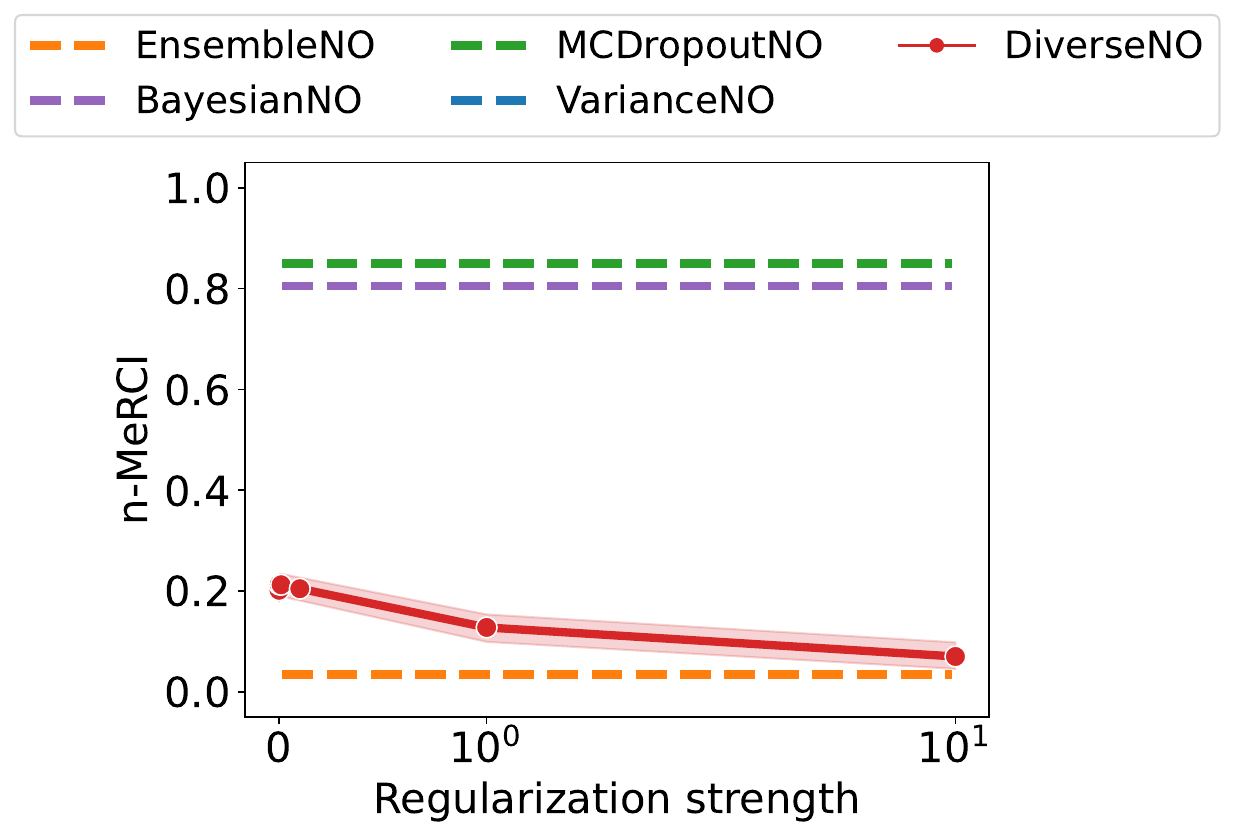}
    \caption{Large OOD shift}
    \end{subfigure}
    
    \caption{{\bf Effect of $\lambda_{\text{diverse}}$ on n-MeRCI.}     Comparison of the n-MeRCI metric as a function of the diversity regularization strength $\lambda_{\text{diverse}}$ for \method to the n-MeRCI of the other baselines. }
    \label{fig:div_regstrength_heat_nemerci}
\end{figure}

The MSE (\cref{fig:div_regstrength_heat_mse}) and especially the n-MeRCI metric (\cref{fig:div_regstrength_heat_nemerci}) monotonically improve (decrease) as a function of $\lambda_{\text{diverse}}$ with larger values of this regularization strength $\approx 10$ being favorable. 
These figures also show that without diversity regularization, i.e., $\lambda_{\text{diverse}} = 0$, there is a large gap in performance between \method and \ensemblenomethod, which is closed with larger $\lambda_{\text{diverse}}$.

\section{PDE Test Problems}
In this section, we provide the details of the various test problems that we study, and the construction of both the training and test OOD datasets. See \cref{tab:summary_pde_tasks} for a summary.
\label{app:test_probs}
\subsection{Generalized Porous Medium Equation (GPME) Family}
The Generalized Porous Medium Equation (GPME) is a family of PDEs parameterized by a (potentially nonlinear) coefficient $k(u)$ ~\cite{maddix2018_harmonic_avg, maddix2018temp_oscill}. 
The GPME models fluid flow through a porous medium, and it has additional applications in heat transfer, groundwater flow, and crystallization, to name a few~\citep{vazquez2007porous}. 
It can be written in the conservative form with flux $F(u) = -k(u)\nabla u$~as:
\begin{align}\label{eq:gpme_app}
u_t - \nabla \cdot (k(u) \nabla u) = 0, \qquad x \in \Omega, t\in \tdomain, 
\end{align}
where $k(u): \xdomain \to \sR$ denotes the (potentially nonlinear) diffusion coefficient. 
We consider three instances of the GPME with increasing levels of difficulty by varying $k(u)$~\citep{hansen2023learning}: 
\textbf{(i)} ``easy'' case with $k(u) = k$, the standard heat equation (linear, constant coefficient parabolic and smooth); 
\textbf{(ii)} ``medium'' case with $k(u) = u^m, m\geq 1$, the Porous Medium Equation (PME) (nonlinear, degenerate parabolic); and 
\textbf{(iii)} ``hard'' case with $k(u) = \vone_{u \geq u^*}, u^* > 0$, the Stefan problem (nonlinear, degenerate parabolic with shock). 

\begin{table}[t]
    \centering
    \caption{Shows the parameter ranges for training and different OOD tasks corresponding to the four tested PDEs. For each PDE, we train the models on a range of parameter inputs (e.g., $k^\tr \in [1, 5]$ for heat equation) and test on increasing OOD shifts (e.g., $k^\te \in [5, 6]$ for a small shift in heat equation). 
    }
    \begin{tabular}{@{}llc@{}}
    \toprule
    PDE & Task & Parameter range \\
    \midrule
    \multirow{3}{*}{\bf Heat Equation} & Train & $k \in [1, 5]$ \\
    \multirow{3}{*}{$u_t - \nabla\cdot(k\nabla u)=0$} & OOD small & $k \in [5, 6]$ \\
    & OOD medium & $k \in [6, 7]$ \\
    & OOD large & $k \in [7, 8]$ \\
    \\
    \multirow{3}{*}{\bf Porous Medium Equation (PME)} & Train & $m \in [2, 3]$ \\
    \multirow{3}{*}{$u_t - \nabla\cdot(u^m\nabla u)=0$} & OOD small & $m \in [1, 2]$ \\
    & OOD medium & $m \in [4, 5]$ \\
    & OOD large & $m \in [5, 6]$ \\
    \\ 
    \multirow{3}{*}{\bf Stefan Equation} & Train & $u^* \in [0.6, 0.65]$ \\
    \multirow{3}{*}{$u_t - \nabla\cdot(\vone_{u\geq u^*}\nabla u)=0$} & OOD small & $u^* \in [0.55, 0.6]$ \\
    & OOD medium & $u^* \in [0.7, 0.75]$ \\
    & OOD large & $u^* \in [0.5, 0.55]$ \\
    \\ 
    \multirow{3}{*}{\bf Linear advection, constant input} & Train & $\beta \in [1,~~2]$ \\
    \multirow{3}{*}{$u_t + \beta u_x=0$} & OOD small & $\beta \in [0.5, 1]$ \\
    & OOD medium & $\beta \in [2.5, 3]$ \\
    & OOD large & $\beta \in [3,3.5]$ \\
    \\
     \multirow{3}{*}{\bf Linear advection, non-constant input} & Train & $a \in [0.45,0.55]$ \\
    \multirow{3}{*}{$u_t + u_x=0$, } & OOD small & $a \in [0.4,0.45]$ \\
   \multirow{3}{*}{$u(x, 0) = \vone_{x \leq a}$} & OOD medium & $a \in [0.6,0.65]$ \\
    & OOD large & $a \in [0.35,0.4]$ \\
    \\
     \multirow{3}{*}{\bf 2-d Darcy Flow} & Train & $k \in [3, 4]$ \\
    \multirow{3}{*}{$-\nabla \cdot (k \nabla u(x)) = 1$} & OOD small & $k \in [4, 4.5]$ \\
    & OOD medium & $k \in [4.5, 5]$ \\
    & OOD large & $k \in [5,6]$ \\
     \bottomrule
    \end{tabular}
    \label{tab:summary_pde_tasks}
\end{table}

\subsubsection{Heat (Diffusion) Equation}
\label{subsec:heat_eqn_app}
Here, we consider the ``easy'' case of GPME in \cref{eq:gpme_app} with a constant diffusion coefficient $k(u):=k$ over a domain $\Omega=[0, 2\pi]$ and $T=1$.
We solve the problem for initial conditions $u(x, 0) = \sin(x), \forall x$, and homogenous Dirichlet boundary conditions $u(0, t) = u(2\pi, t) = 0, \forall t$. 

The training dataset consists of $N=400$ input/output pairs $\{\phi^{(i)}, u^{(i)}\}_{i=1}^N$ where $\phi^{(i)}(x, t):=k^{(i)}, \forall x$, denotes a constant function over the domain representing the value of the diffusivity parameter and $u^{(i)}$ denotes the corresponding solutions at times $t\in[0,1]$. 
The $\phi^{(i)}$ are passed as input to the models without any normalization applied. 
During training, we consider the parameter $k^{(i)} \sim \text{Unif}(1, 5)$, and evaluate the trained NO on small, medium and large OOD ranges for the diffusivity parameter: $k^{\te} \sim  \text{Unif}(5, 6), \text{Unif}(6, 7), \text{Unif}(7, 8)$, respectively.

\subsubsection{Porous Medium Equation (PME)}
\label{subsec:pme_app}
The Porous Medium Equation (PME) with $k(u)=u^m, m\geq 1$ in \cref{eq:gpme_app} represents a ``medium'' case of the GPME.
We solve the problem over the domain $\Omega=[0, 1]$ and $T=1$ for initial conditions $u(x, 0) = 0, \forall x$.

We train the NO to map from a constant function denoting the degree $m$ that identifies $k(u)$ to the corresponding solution for all $t\in[0, 1]$.  
The training dataset consists of $N=400$ input/output pairs $\{\phi^{(i)}, u^{(i)}\}_{i=1}^N$ where $\phi^{(i)}(x, t):=m^{(i)}, \forall x$, is a constant function over the domain representing the degree and $u^{(i)}$ denotes the corresponding solutions at times $t\in[0,1]$.
The $\phi^{(i)}$ are passed as input to the models without any normalization applied. 
During training, we sample the parameter $m^{(i)} \sim \text{Unif}(2, 3)$, and evaluate the trained NO on small, medium and large OOD ranges for the degree: $m^{\te} \sim \text{Unif}(1, 2), \text{Unif}(4, 5), \text{Unif}(5, 6)$, respectively.
With increasing $m$, the solution becomes sharper and more challenging for the learned operator. 

\subsubsection{Stefan Equation}
\label{subsec:stef_app}
The Stefan equation represents a challenging case of the GPME family with a discontinuous and nonlinear diffusivity coefficient $k(u) = \vone_{u \geq u^*}$, where $u^*$ is a parameter denoting the value at the shock position $x^*(t)$, i.e., $u(x^*(t), t)=u^*$. 
We solve the problem over the domain $\Omega=[0, 1]$ and $T=0.1$ for initial conditions $u(x, 0) = 0, \forall x$, and Dirichlet boundary conditions $u(0, t) = 1\:, u(1, t) = 0, \forall t$.

We train the NO to map from a constant function denoting the parameter $u^*$ to the solution for all $t\in[0, 0.1]$.
The training dataset consists of $N=400$ input/output pairs $\{\phi^{(i)}, u^{(i)}\}_{i=1}^N$ where $\phi^{(i)}(x, t)=u{^*}^{(i)}, \forall x$, denotes a constant function over the domain representing the solution value at the shock and $u^{(i)}$ denotes the corresponding solutions at times $t\in[0,0.1]$.
The $\phi^{(i)}$ are passed as input to the models without any normalization applied. 
During training, we sample the parameter $u{^*}^{(i)} \sim \text{Unif}(0.6, 0.65)$, and evaluate the trained NO on small, medium and large OOD ranges: $u{^*}^{\te} \sim \text{Unif}(0.55, 0.6), \text{Unif}(0.7, 0.75), \text{Unif}(0.5, 0.55)$, respectively. 

\subsection{Hyperbolic Linear Advection Equation}
\label{subseq:la_app}
The linear advection equation given by
$$
u_t + \beta u_x = 0, \quad x\in [0, 1], t\in [0, 1], 
$$
describes the motion of a fluid advected by a constant velocity $\beta > 0$.
We consider the following two tasks for the PDE.

\paragraph{Constant parameter to solution mapping.}
We solve the PDE for initial conditions $u(x, 0) = \vone_{x \leq 0.5}, \forall x\in [0, 1]$, and Dirichlet boundary conditions $u(0, t) = 1, u(1, t)=0, \forall t$. The solution is a rightward moving shock (discontinuity) with the speed defined by the parameter $\beta$  \citep{hansen2023learning}. 

We train the NO to map from a constant function denoting the velocity parameter $\beta$ to the solution for all $t\in[0, 0.1]$.
The training dataset consists of $N=400$ input/output pairs $\{\phi^{(i)}, u^{(i)}\}_{i=1}^N$ where $\phi^{(i)}(x, t)=\beta^{(i)}, \forall x$, denotes a constant function over the domain representing the velocity and $u^{(i)}$ denotes the corresponding solutions at times $t\in[0,0.1]$.
During training, we sample the parameter $\beta^{(i)} \sim \text{Unif}(1,2)$, and evaluate the trained NO on small, medium and large OOD ranges for the $\beta$: $\beta^{\te} \sim \text{Unif}(0.5,1),\text{Unif}(2.5,3), \text{Unif}(3,3.5)$, respectively.

\paragraph{Non-constant initial condition to solution mapping.}
We solve the PDE for various initial conditions $u(x, 0) = \vone_{x \leq a}, \forall x\in [0, 1]$, where $a$ denotes the initial shock location.
We use Dirichlet boundary conditions $u(0, t) = 1, u(1, t)=0, \forall t$, and a fixed speed $\beta = 1$. 

We train the NO to map from the non-constant function denoting the initial condition, i.e., $u(x, 0)$ to the solution for all $t\in[0, 0.1]$.
The training dataset consists of $N=400$ input/output pairs $\{\phi^{(i)}, u^{(i)}\}_{i=1}^N$ where $\phi^{(i)}(x, t)=u^{(i)}(x, 0), \forall x$, denotes a constant function over the domain representing the velocity and $u^{(i)}$ denotes the corresponding solutions at times $t\in[0,0.1]$.
During training, we generate initial conditions using shock locations $a^{(i)}\sim \text{Unif}(0.45,0.55)$, and evaluate the trained NO on small, medium and large OOD ranges for the $a$: $a^{\te} \sim \text{Unif}(0.4,0.45),\text{Unif}(0.6,0.65), \text{Unif}(0.35,0.4)$, respectively.

\subsection{2-d Elliptic Darcy Flow}
We consider the steady state solution of the 2-d Darcy flow equation (linear, elliptic):
$$
-\nabla \cdot (k \nabla u(x)) = f(x)\:, \qquad x\in[0,1]^2,
$$
with Dirichlet boundary conditions $u(x)=0$ for all $x$ on the boundary, forcing function $f(x)= 1$, and permeability field defined by the parameter $k$.

We train the NO to map from the constant scalar field over the 2-d domain denoting the permeability $k$ of the surface to the steady-state solution.
The training dataset consists of $N=400$ input/output pairs $\{\phi^{(i)}, u^{(i)}\}_{i=1}^N$ 
where $\phi^{(i)}(x)=k^{(i)}, \forall x$, denotes a constant field over the domain representing the permeability and and $u^{(i)}$ denotes the corresponding solutions representing the unknown pressure.
During training, we sample the permeability parameter $k^{(i)} \sim \text{Unif}(3,4)$, and evaluate the trained NO on small, medium and large OOD ranges for the $k$: $k^{\te} \sim \text{Unif}(4,4.5),\text{Unif}(4.5,5), \text{Unif}(5,6)$, respectively. 

\section{Detailed Experiment Settings}
\label{app:exp_details}
We use the standard optimization procedure for training FNO models \citep{li2020fourier}. In particular, we use the Adam optimizer with a weight decay. We optimize the objective, and learn over batches of a given batch size $B$ (fixed to $B=20$ in our experiments). We use a learning rate scheduler that halves the learning rate after every 50 epochs.

\begin{table}[H]
    \centering
    \caption{Hyperparameters for the base FNO architecture and the UQ methods on top. (\bayesiannomethod does not have additional hyperparameters.)}
    \begin{tabular}{@{}lr@{}}
    \toprule
    Hyperparameter & Values \\
    \midrule
    {\bf Base FNO} \\
     ~~Number of Fourier layers & 4 \\
     ~~Channel width & $\{32, 64\}$ \\
     ~~Number of Fourier modes & 12 \\
     ~~Batch size & 20 \\
     ~~Learning rate & $\{10^{-4}, 10^{-3}, 10^{-2}\}$ \\
    {\bf \bayesiannomethod} & \\
     ~~N/A &  \\
     {\bf \mcdropoutnomethod} & \\
     ~~Dropout probability & $\{0.1, 0.25\}$\\
     ~~Number of dropout masks & 10 \\
     {\bf \ensemblenomethod} & \\
     ~~Number of models & 10 \\
      {\bf \method} & \\
     ~~Number of heads $M$ & 10 \\
     ~~Diversity regularization $\lambda_\diverse$ & $\{10^{-2}, 10^{-1}, 1, 10^{1}, 10^{2}\}$ \\ 
     \bottomrule
    \end{tabular}
    \label{tab:hyperparameters}
\end{table}
\cref{tab:hyperparameters} shows the hyperparameters for the base FNO architecture and the UQ methods used on top of it. 
For all methods except \method, in-domain MSE on validation data is used to select the best hyperparameter configuration. 
For \method, hyperparameter $\lambda_\diverse$ controls the strength of the diversity regularization relative to the prediction loss. 
We select the highest regularization strength $\lambda_\diverse$ that also achieves in-domain validation MSE within 10\% of the best in-domain validation MSE. 
This procedure trades off in-domain prediction errors for higher diversity that is primarily useful for OOD UQ. 

\section{Additional Empirical Results across a Range of PDEs}
In this section, we show the additional empirical results for each PDE on in-domain tasks and with various amounts of OOD shifts ranging from small, medium to large.  

\subsection{Detailed Metric Results and Solution Profiles}
\label{subsec:metrics_app}
In this subsection, we show the detailed metric results and solution profiles for the members of the (degenerate) parabolic GPME and the linear advection hyperbolic conservation law. Tables \ref{tab:results_heat}-\ref{tab:results_la} compare the performances of \method to the UQ baselines on various PDEs under the following metrics~\citep{psaros2023Uncertainty}: mean-squared error (MSE), negative log-likelihood (NLL), normalized Mean Rescaled Confidence Interval (n-MeRCI)~\citep{moukari2019n}, root mean squared calibration error (RMSCE) and continuous ranked probability score (CRPS)~\citep{gneiting2007} across a wide variety of PDEs with varying difficulties. The MSE measures the performance of the mean prediction. The NLL, n-MeRCI, RMSCE and CRPS measure the quality of the uncertainty estimates. The RMSCE measures how well the uncertainty estimates are calibrated and CRPS measures both sharpness and calibration. The n-MeRCI is of particular importance since it measures the correlation of the uncertainty estimates with the prediction error. We see that \ensemblenomethod and our \method have the overall best performance across the various PDEs, especially in the n-MeRCI metric.
\subsubsection{``Easy'' Heat Equation}
\begin{figure}[H]
    \centering
    \hspace{-0.5in}
    \begin{subfigure}[h]{0.30\textwidth}
    \centering
    \includegraphics[scale=0.35]{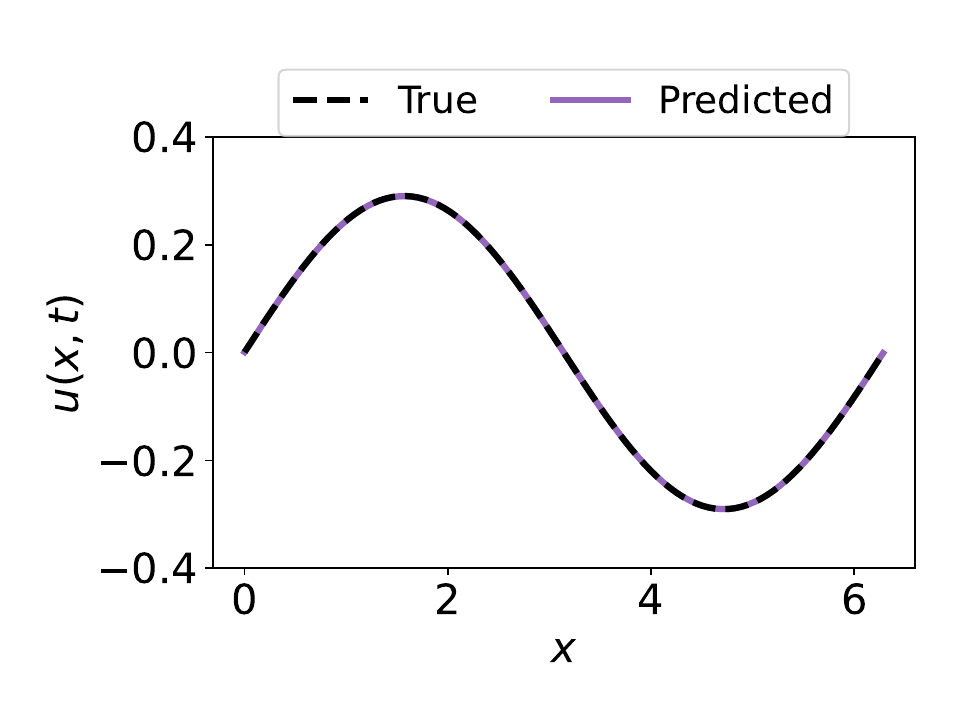}
    \caption{\bayesiannomethod}
    \end{subfigure}
    ~~
    \begin{subfigure}[h]{0.30\textwidth}
    \centering
    \includegraphics[scale=0.35]{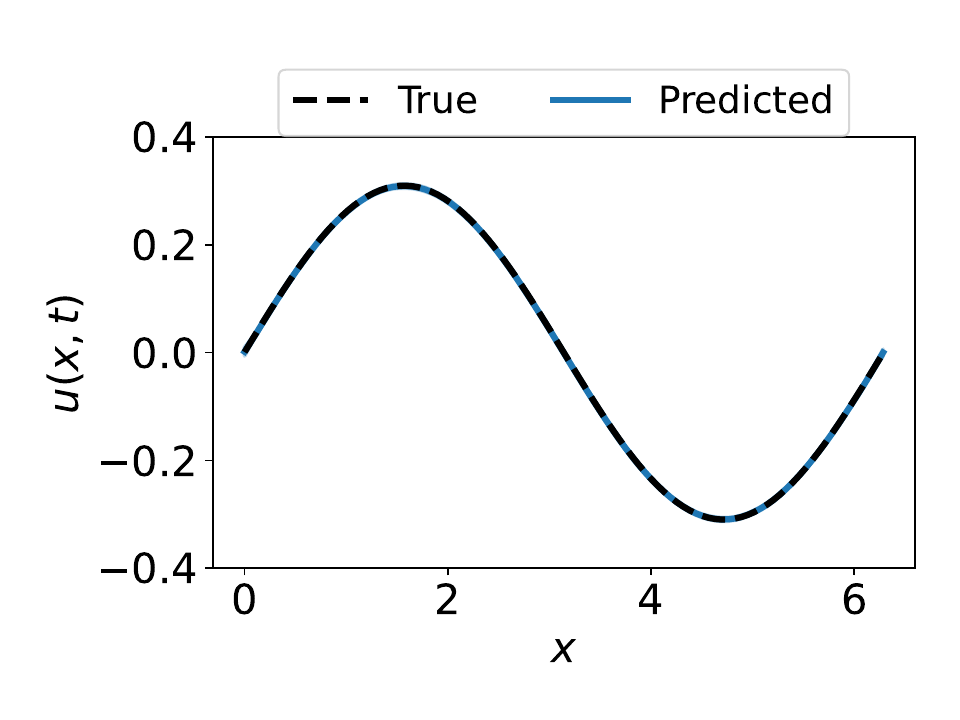}
    \caption{\outputvarmethod}
    \end{subfigure}
    ~~
    \begin{subfigure}[h]{0.30\textwidth}
    \centering
    \includegraphics[scale=0.35]{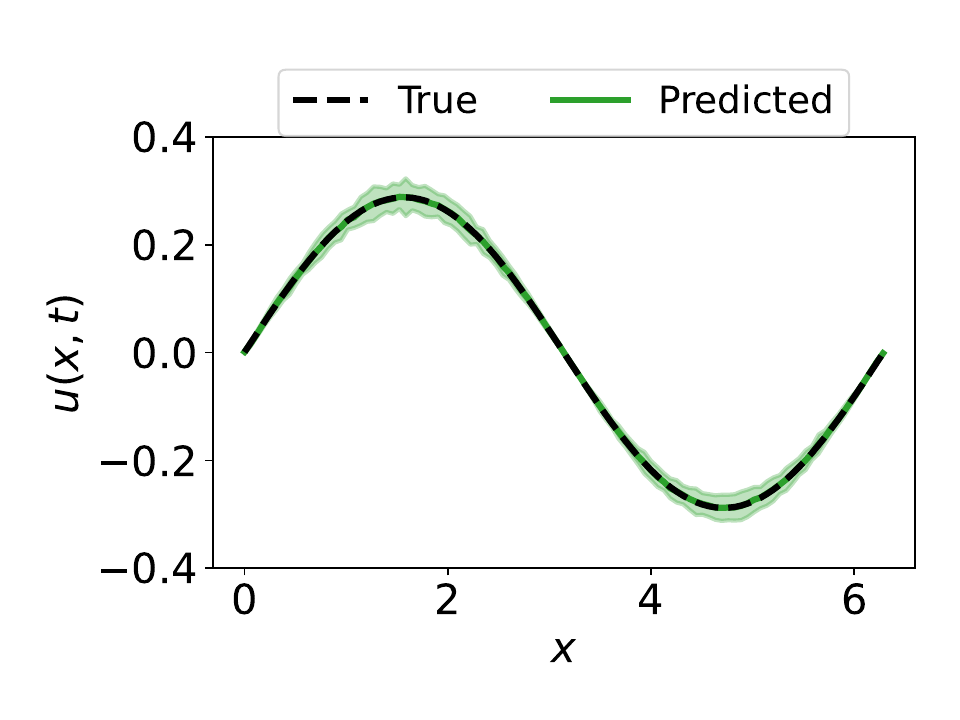}
    \caption{\mcdropoutnomethod}
    \end{subfigure}
    
    \begin{subfigure}[h]{0.30\textwidth}
    \centering
    \includegraphics[scale=0.35]{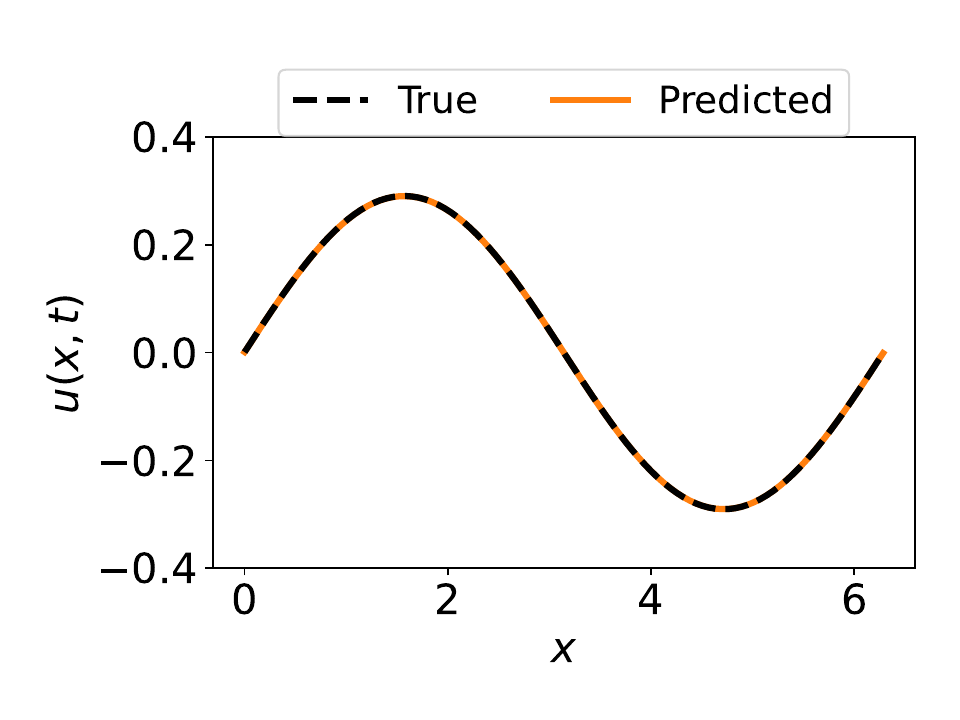}
    \caption{\ensemblenomethod}
    \end{subfigure}
    ~~~~
    \begin{subfigure}[h]{0.30\textwidth}
    \centering
    \includegraphics[scale=0.35]{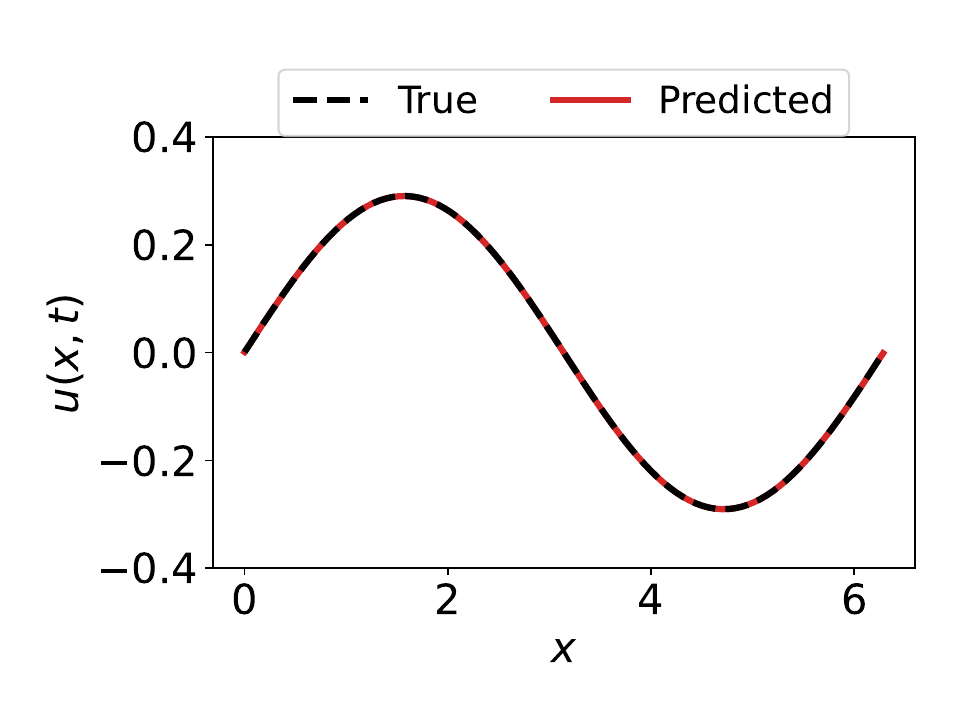}
    \caption{\method}
    \end{subfigure}
    \caption{{\bf 1-d Heat equation, in-domain}, $k^\tr, k^\te\in[1,5]$. 
    Uncertainty estimates from different UQ methods for in-domain values of the input diffusivity coefficient $k$.} 
    \label{fig:solutions_heat_id}
\end{figure}

\begin{figure}[H]
    \centering
    \hspace{-0.5in}
    \begin{subfigure}[h]{0.30\textwidth}
    \centering
    \includegraphics[scale=0.35]{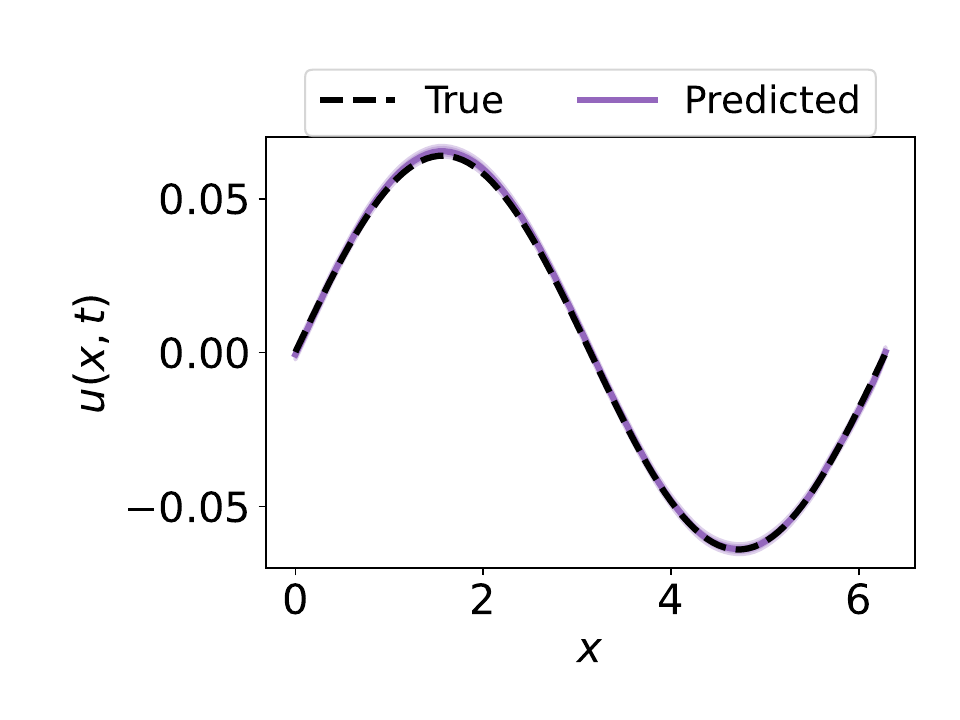}
    \caption{\bayesiannomethod}
    \end{subfigure}
    ~~
    \begin{subfigure}[h]{0.30\textwidth}
    \centering
    \includegraphics[scale=0.35]{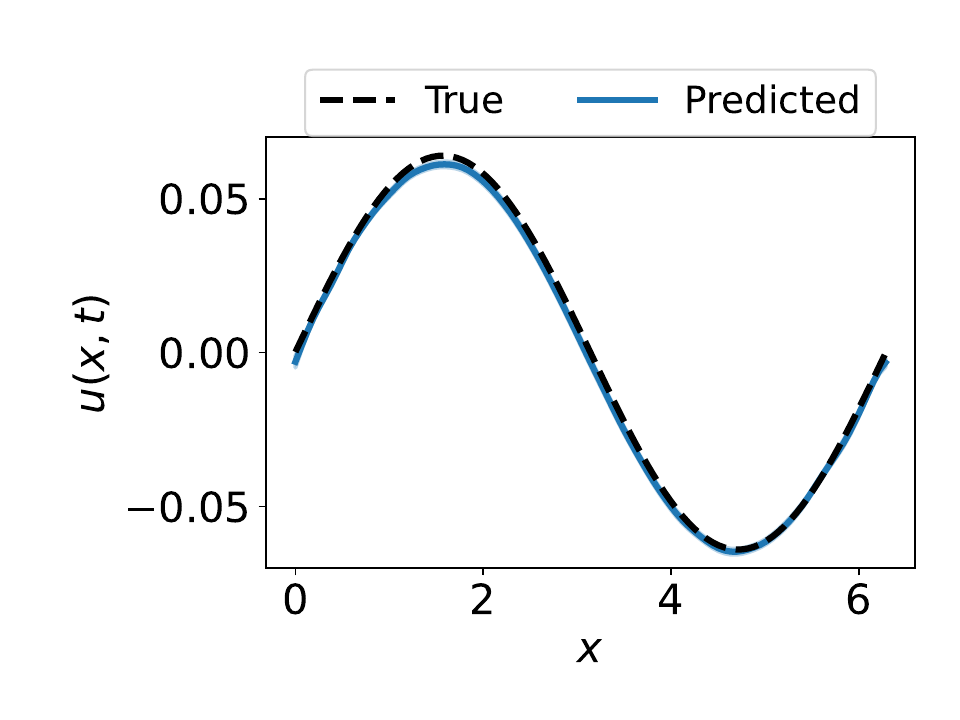}
    \caption{\outputvarmethod}
    \end{subfigure}
    ~~
    \begin{subfigure}[h]{0.30\textwidth}
    \centering
    \includegraphics[scale=0.35]{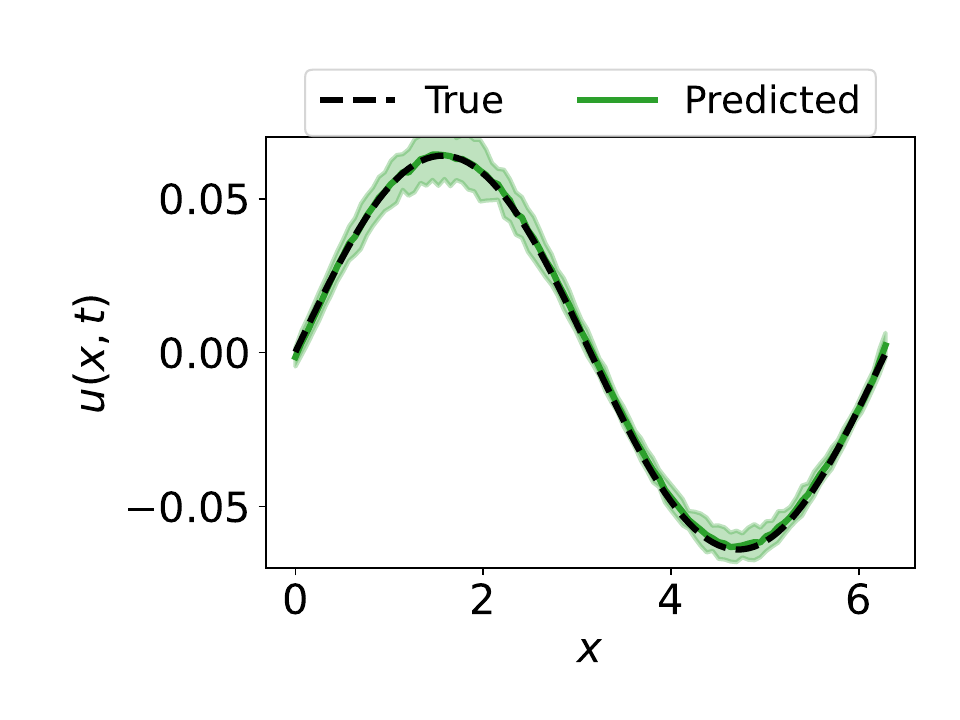}
    \caption{\mcdropoutnomethod}
    \end{subfigure}
    
    \begin{subfigure}[h]{0.30\textwidth}
    \centering
    \includegraphics[scale=0.35]{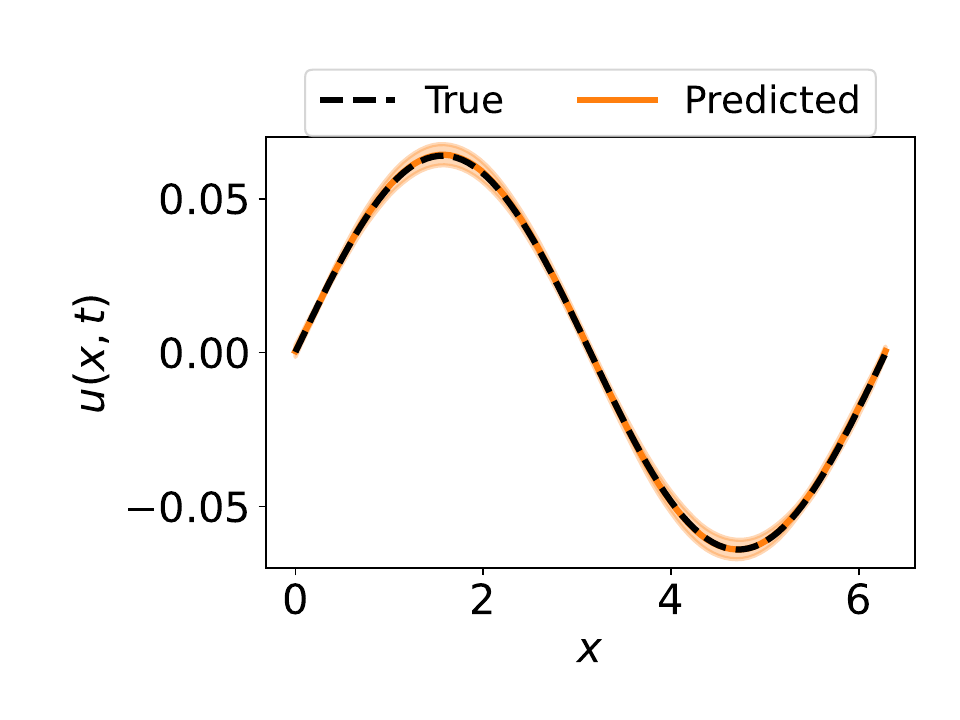}
    \caption{\ensemblenomethod}
    \end{subfigure}
    ~~~~
    \begin{subfigure}[h]{0.30\textwidth}
    \centering
    \includegraphics[scale=0.35]{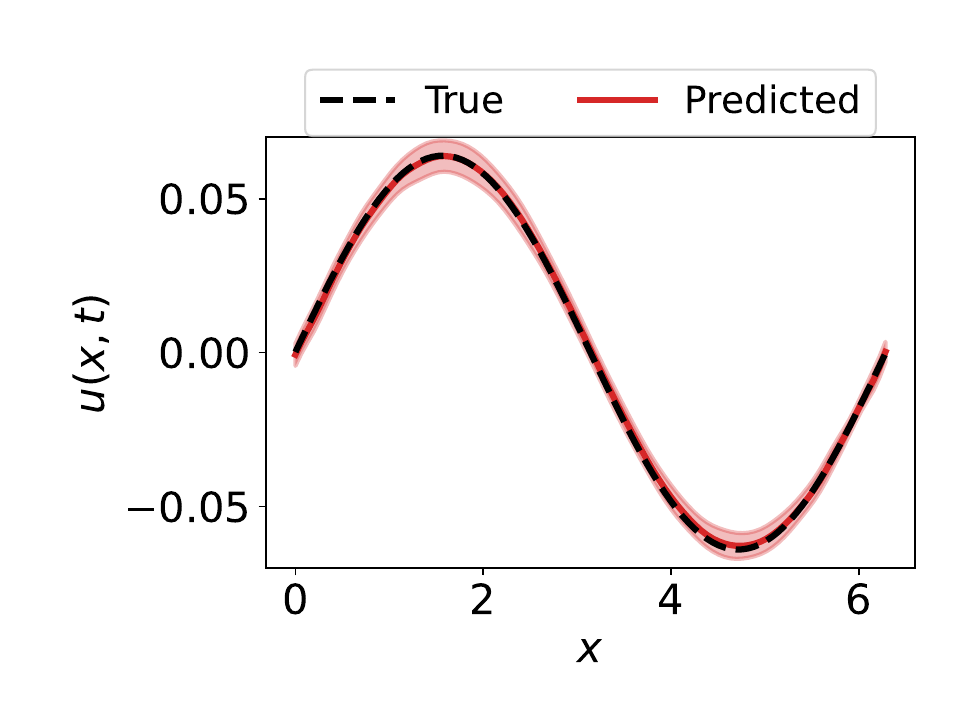}
    \caption{\method}
    \end{subfigure}
    \caption{{\bf 1-d Heat equation, small OOD shift}, $k^\tr\in[1,5], k^\te\in[5,6]$. 
    Uncertainty estimates from different UQ methods under small OOD shifts in the input diffusivity coefficient $k$.} 
    \label{fig:solutions_heat_ood1}
\end{figure}

\begin{figure}[H]
    \centering
    \begin{subfigure}{0.3\textwidth}
    \centering
    \includegraphics[scale=0.35]{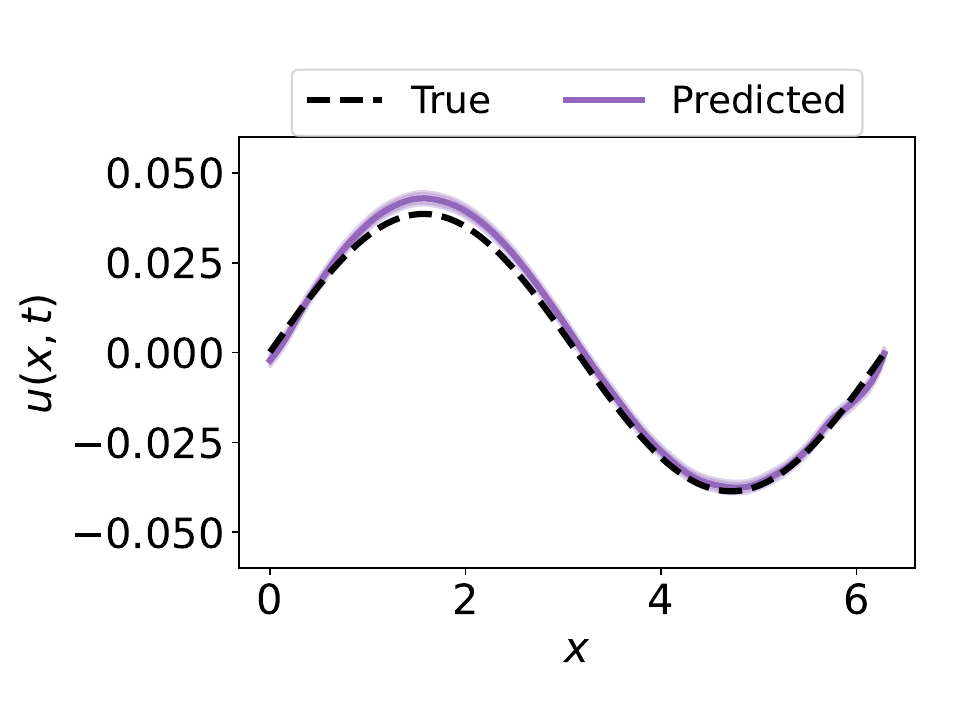}
    \caption{\bayesiannomethod}
    \end{subfigure}
    ~~~~
    \begin{subfigure}{0.3\textwidth}
    \centering
    \includegraphics[scale=0.35]{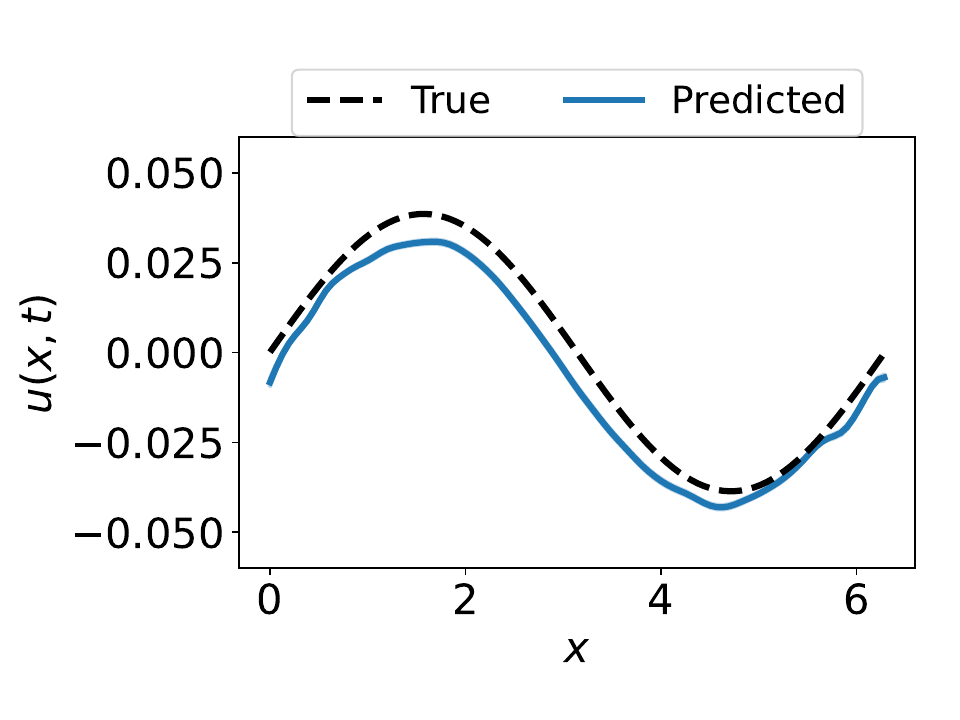}
    \caption{\outputvarmethod}
    \end{subfigure}
    ~~~~
    \begin{subfigure}{0.3\textwidth}
    \centering
    \includegraphics[scale=0.35]{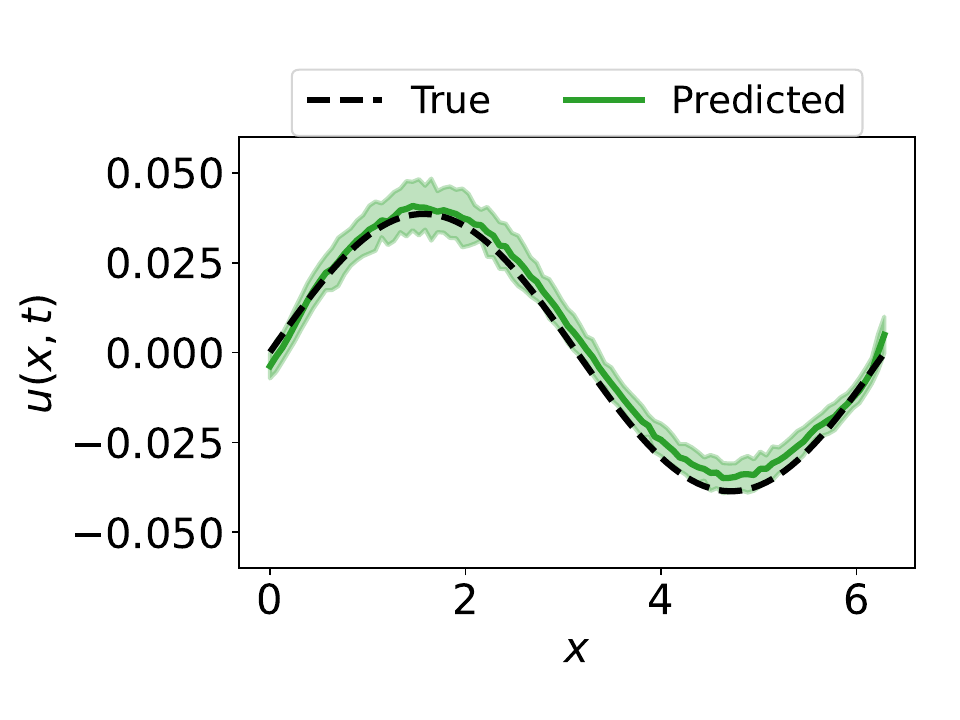}
    \caption{\mcdropoutnomethod}
    \end{subfigure}
    
    \begin{subfigure}{0.3\textwidth}
    \centering
    \includegraphics[scale=0.35]{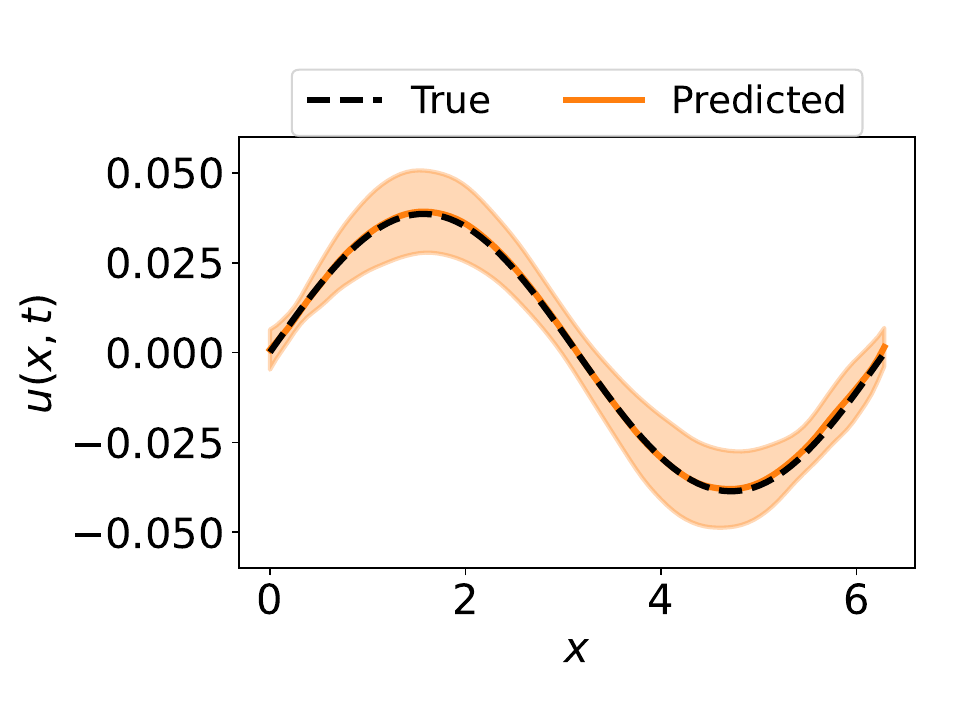}
    \caption{\ensemblenomethod}
    \end{subfigure}
    ~~~~
    \begin{subfigure}{0.3\textwidth}
    \centering
    \includegraphics[scale=0.35]{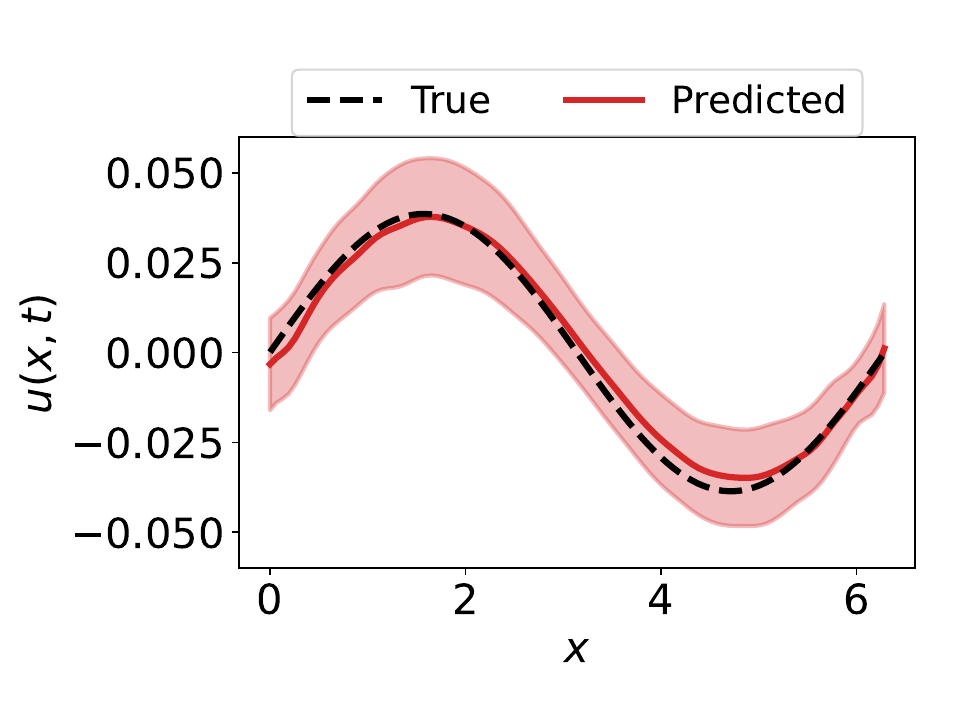}
    \caption{\method}
    \end{subfigure}
    \caption{{\bf 1-d Heat equation, medium OOD shift}, $k^\tr\in[1,5], k^\te\in[6,7]$. 
    Uncertainty estimates from different UQ methods under medium OOD shifts in the  input diffusivity coefficient $k$.} 
    \label{fig:solutions_heat_ood2}
\end{figure}

\begin{figure}[H]
    \centering
    \hspace{-0.5in}
    \begin{subfigure}[h]{0.30\textwidth}
    \centering
    \includegraphics[scale=0.35]{new_figures/HeatEquation_1D_ood3_BayesianFNO2d_new.pdf}
    \caption{\bayesiannomethod}
    \end{subfigure}
    ~~
    \begin{subfigure}[h]{0.30\textwidth}
    \centering
    \includegraphics[scale=0.35]{new_figures/HeatEquation_1D_ood3_OutputVarFNO2d_new.pdf}
    \caption{\outputvarmethod}
    \end{subfigure}
    ~~
    \begin{subfigure}[h]{0.30\textwidth}
    \centering
    \includegraphics[scale=0.35]{new_figures/HeatEquation_1D_ood3_MCDropoutFNO2d_new.pdf}
    \caption{\mcdropoutnomethod}
    \end{subfigure}
    
    \begin{subfigure}[h]{0.30\textwidth}
    \centering
    \includegraphics[scale=0.35]{new_figures/HeatEquation_1D_ood3_EnsembleFNO2d_new.pdf}
    \caption{\ensemblenomethod}
    \end{subfigure}
    ~~~~
    \begin{subfigure}[h]{0.30\textwidth}
    \centering
    \includegraphics[scale=0.35]{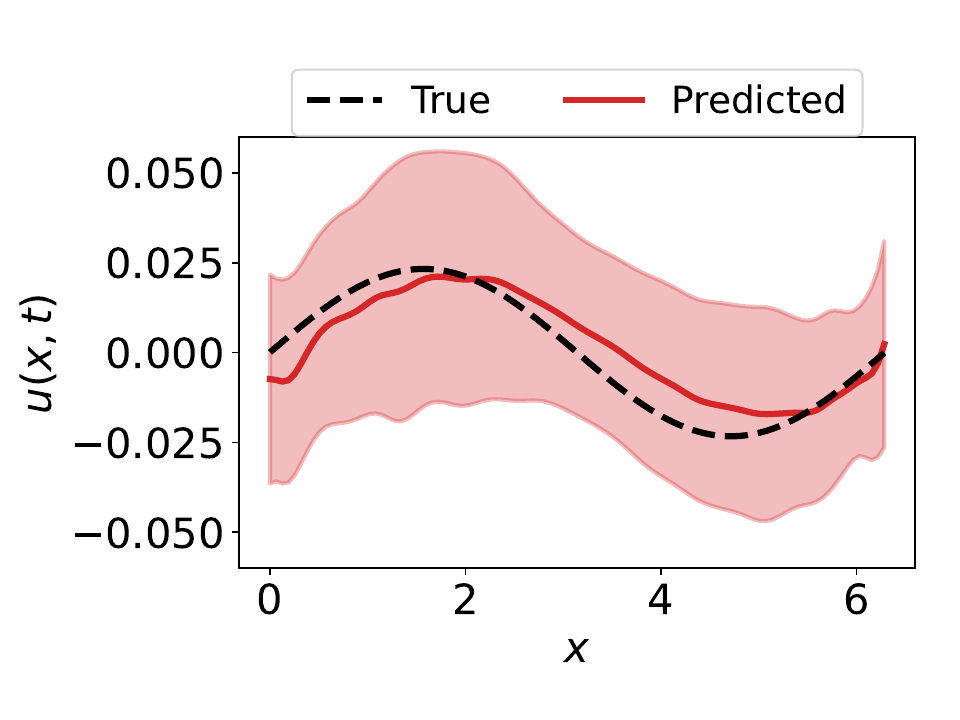}
    \caption{\method}
    \end{subfigure}
   \caption{{\bf 1-d Heat equation, large OOD shift}, $k^\tr\in[1,5], k^\te\in[7,8]$. 
    Uncertainty estimates from different UQ methods under large OOD shifts in the input diffusivity coefficient $k$.}
\label{fig:solutions_heat_ood3_app}
\end{figure}

Figures \ref{fig:solutions_heat_id}-\ref{fig:solutions_heat_ood3_app} show the solution profiles for the ``easy'' smooth and parabolic heat equation with zero Dirichlet boundary conditions for in-domain, small, medium and large OOD shifts, respectively, of the diffusivity parameter $k$. We see that on this ``easy'' case, most methods are very accurate in-domain in \cref{fig:solutions_heat_id} and  perform well in the small shift cases in \cref{fig:solutions_heat_ood1}. Errors in the \bayesiannomethod, \outputvarmethod and \mcdropoutnomethod baselines start to form for medium shifts in \cref{fig:solutions_heat_ood2} and grow in the large shift case in \cref{fig:solutions_heat_ood3_app}. We see both \ensemblenomethod and \method output good uncertainty estimates (3 standard deviations) that contain the true solution within these error bounds. See corresponding metric results in \cref{tab:results_heat}.
\begin{table}[H]
    \centering
    \caption{{\bf 1-d heat equation.} MSE $\downarrow$, NLL $\downarrow$, n-MeRCI $\downarrow$, RMSCE $\downarrow$ and CRPS $\downarrow$ (mean and standard deviation over 5 seeds) metrics for different UQ methods on the 1-d heat equation in-domain and with small, medium and large OOD shifts, where $k^\tr \in [1,5]$. {\bf Bold} indicates values within one standard deviation of the best mean.
    }
    \label{tab:results_heat}
    \resizebox{\textwidth}{!}{
    \begin{tabular}{@{}lccccc@{}}
    \toprule
    & \multicolumn{5}{c}{{\bf In-domain}, $k^\te\in [1, 5]$} \\
     & MSE $\downarrow$ & NLL $\downarrow$ & n-MeRCI $\downarrow$ & RMSCE $\downarrow$ & CRPS $\downarrow$ \\
    \midrule
    \bayesiannomethod & \bf 1.5e-07 (4.9e-08) & -1.3e+04 (2.1e+02) &   0.13 ( 0.07) &   0.19 ( 0.01) & 2.2e-04 (2.8e-05) \\
    \outputvarmethod & 4.1e-07 (2.8e-07) & -1.2e+04 (5.1e+02) &   0.18 ( 0.10) &   0.15 ( 0.02) & 3.1e-04 (9.4e-05) \\
    \mcdropoutnomethod & 3.6e-06 (3.0e-07) & -8.9e+03 (1.8e+02) &   0.16 ( 0.07) &   0.20 ( 0.00) & 1.9e-03 (1.1e-04) \\
    \ensemblenomethod & 1.8e-07 (6.3e-08) & -1.4e+04 (7.5e+02) & \bf   0.05 ( 0.02) & \bf   0.13 ( 0.01) & 1.4e-04 (2.4e-05) \\
    \method & \bf 1.1e-07 (4.3e-08) & \bf -1.5e+04 (7.0e+01) & \bf   0.06 ( 0.02) & \bf   0.13 ( 0.00) & \bf 1.2e-04 (3.5e-06) \\
    \end{tabular}}
    \resizebox{\textwidth}{!}{
    \begin{tabular}{@{}lccccc@{}}
    \midrule
    & \multicolumn{5}{c}{{\bf Small OOD shift}, $k^\te\in [5, 6]$} \\
     & MSE $\downarrow$ & NLL $\downarrow$ & n-MeRCI $\downarrow$ & RMSCE $\downarrow$ & CRPS $\downarrow$ \\
    \midrule
    \bayesiannomethod & 2.5e-06 (8.6e-07) & -8.5e+03 (1.4e+03) &   0.86 ( 0.05) &   0.38 ( 0.02) & 8.9e-04 (1.9e-04) \\
    \outputvarmethod & 7.1e-06 (3.2e-06) & 3.0e+04 (1.2e+04) &   1.17 ( 0.11) &   0.43 ( 0.02) & 1.6e-03 (4.2e-04) \\
    \mcdropoutnomethod & 5.1e-06 (1.4e-06) & -9.4e+03 (2.1e+02) &   0.90 ( 0.04) & \bf   0.25 ( 0.01) & 1.5e-03 (1.0e-04) \\
    \ensemblenomethod & 2.3e-06 (4.9e-07) & \bf -1.1e+04 (7.8e+02) & \bf   0.02 ( 0.02) &   0.37 ( 0.02) & \bf 7.4e-04 (1.2e-04) \\
    \method & \bf 1.7e-06 (4.1e-07) & \bf -1.1e+04 (1.1e+02) &   0.05 ( 0.03) &   0.35 ( 0.01) & \bf 7.3e-04 (5.4e-05) \\
    \end{tabular}
    }
    \resizebox{\textwidth}{!}{
    \begin{tabular}{@{}lccccc@{}}
    \midrule
    & \multicolumn{5}{c}{{\bf Medium OOD shift}, $k^\te\in [6, 7]$} \\
     & MSE $\downarrow$ & NLL $\downarrow$ & n-MeRCI $\downarrow$ & RMSCE $\downarrow$ & CRPS $\downarrow$ \\
    \midrule
    \bayesiannomethod & 2.7e-05 (7.5e-06) & 2.9e+04 (1.0e+04) &   0.84 ( 0.05) &   0.47 ( 0.01) & 3.4e-03 (6.4e-04) \\
    \outputvarmethod & 8.0e-05 (2.9e-05) & 9.0e+05 (2.2e+05) &   1.40 ( 0.15) &   0.49 ( 0.01) & 5.8e-03 (1.4e-03) \\
    \mcdropoutnomethod & 3.9e-05 (1.7e-05) & \bf -7.5e+03 (2.6e+02) &   0.90 ( 0.03) & \bf   0.37 ( 0.01) & 3.4e-03 (6.0e-04) \\
    \ensemblenomethod & 2.4e-05 (3.8e-06) & \bf -8.1e+03 (9.0e+02) &   0.03 ( 0.01) & \bf   0.38 ( 0.02) & \bf 2.5e-03 (3.1e-04) \\
    \method & \bf 1.9e-05 (3.4e-06) & \bf -8.0e+03 (1.0e+02) & \bf   0.02 ( 0.00) & \bf   0.36 ( 0.01) & \bf 2.6e-03 (1.1e-04) \\
    \end{tabular}
    }
    \resizebox{\textwidth}{!}{
    \begin{tabular}{@{}lccccc@{}}
    \midrule
    & \multicolumn{5}{c}{{\bf Large OOD shift}, $k^\te\in [7, 8]$} \\
     & MSE $\downarrow$ & NLL $\downarrow$ & n-MeRCI $\downarrow$ & RMSCE $\downarrow$ & CRPS $\downarrow$ \\
    \midrule
    \bayesiannomethod & 1.2e-04 (3.5e-05) & 1.5e+05 (3.2e+04) &   0.80 ( 0.06) &   0.49 ( 0.01) & 7.5e-03 (1.5e-03) \\
    \outputvarmethod & 3.7e-04 (1.3e-04) & 9.0e+06 (3.0e+06) &   1.70 ( 0.20) &   0.50 ( 0.00) & 1.3e-02 (2.9e-03) \\
    \mcdropoutnomethod & 1.7e-04 (8.0e-05) & -3.0e+03 (1.2e+03) &   0.86 ( 0.04) &   0.44 ( 0.01) & 7.6e-03 (1.7e-03) \\
    \ensemblenomethod & 1.1e-04 (1.6e-05) & \bf -6.6e+03 (9.2e+02) & \bf   0.03 ( 0.02) & \bf   0.37 ( 0.02) & \bf 5.3e-03 (5.9e-04) \\
    \method & \bf 8.8e-05 (1.0e-05) & \bf -6.3e+03 (1.7e+02) & \bf   0.03 ( 0.03) & \bf   0.36 ( 0.02) & \bf 5.8e-03 (8.7e-05) \\
    \bottomrule
    \end{tabular}
    }
\end{table}

\subsubsection{``Medium'' PME}
\begin{figure}[H]
    \centering
    \begin{subfigure}[h]{0.30\textwidth}
    \centering
    \includegraphics[scale=0.35]{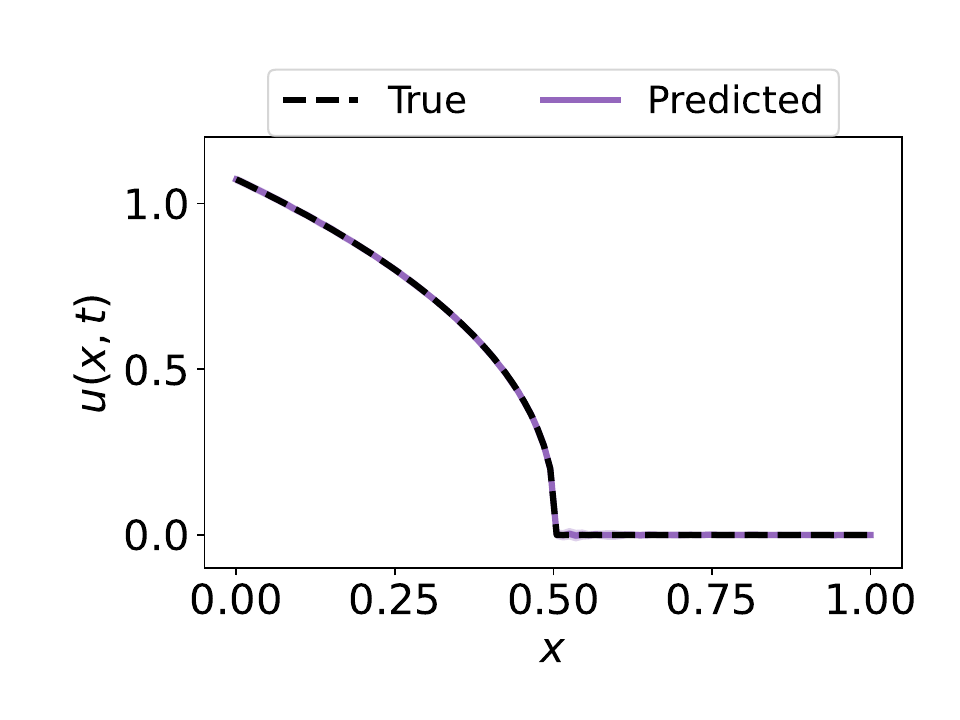}
    \caption{\bayesiannomethod}
    \end{subfigure}
    ~~
    \begin{subfigure}[h]{0.30\textwidth}
    \centering
    \includegraphics[scale=0.35]{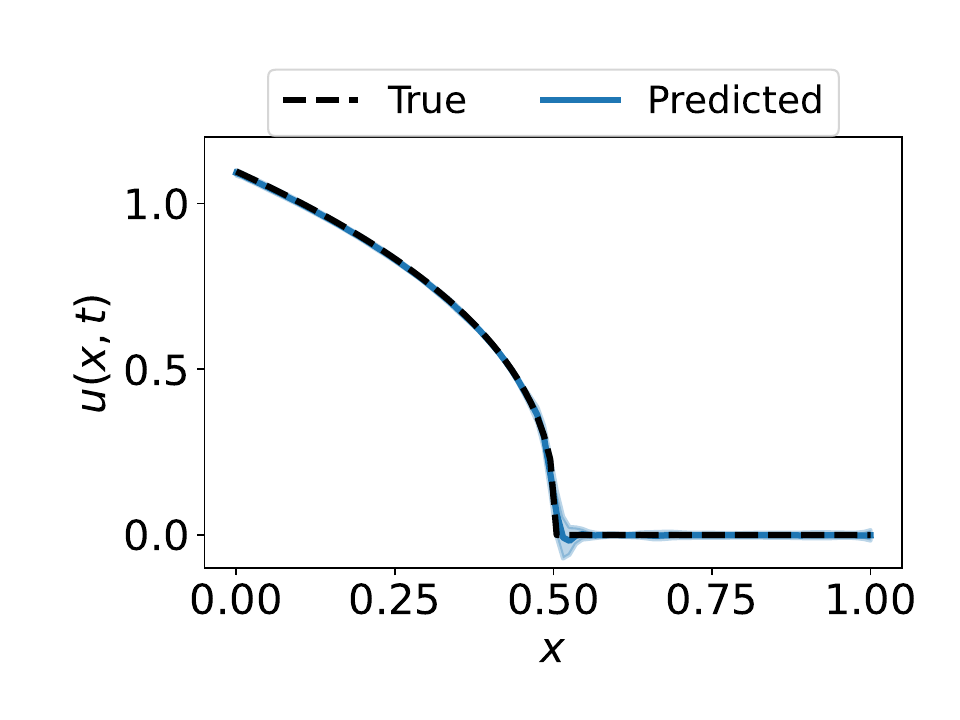}
    \caption{\outputvarmethod}
    \end{subfigure}
    ~~
    \begin{subfigure}[h]{0.30\textwidth}
    \centering
    \includegraphics[scale=0.35]{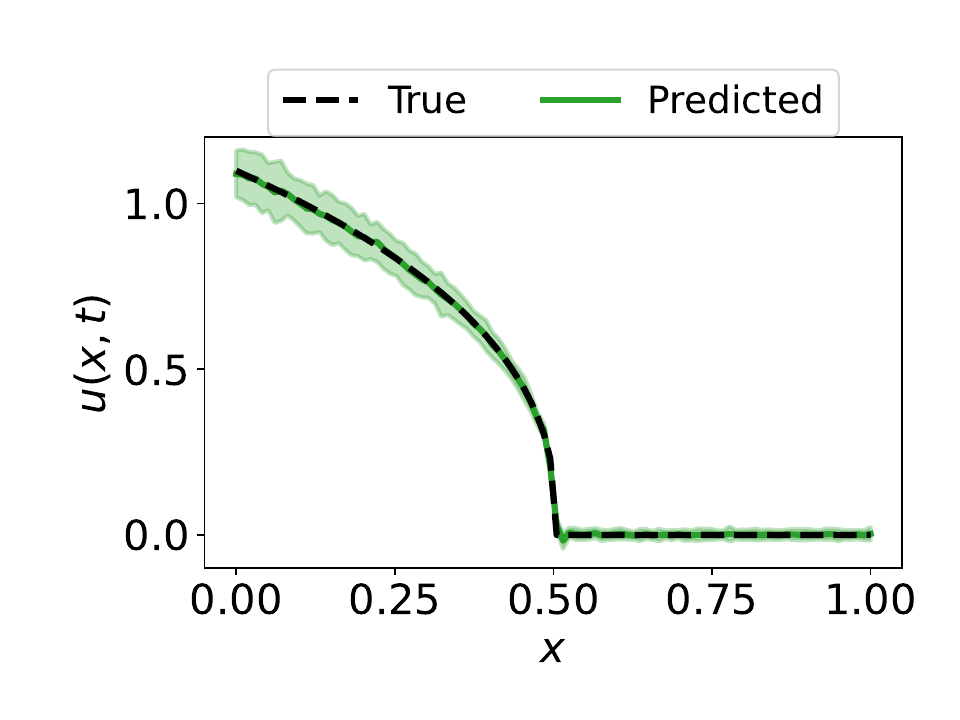}
    \caption{\mcdropoutnomethod}
    \end{subfigure}
    
    \begin{subfigure}[h]{0.30\textwidth}
    \centering
    \includegraphics[scale=0.35]{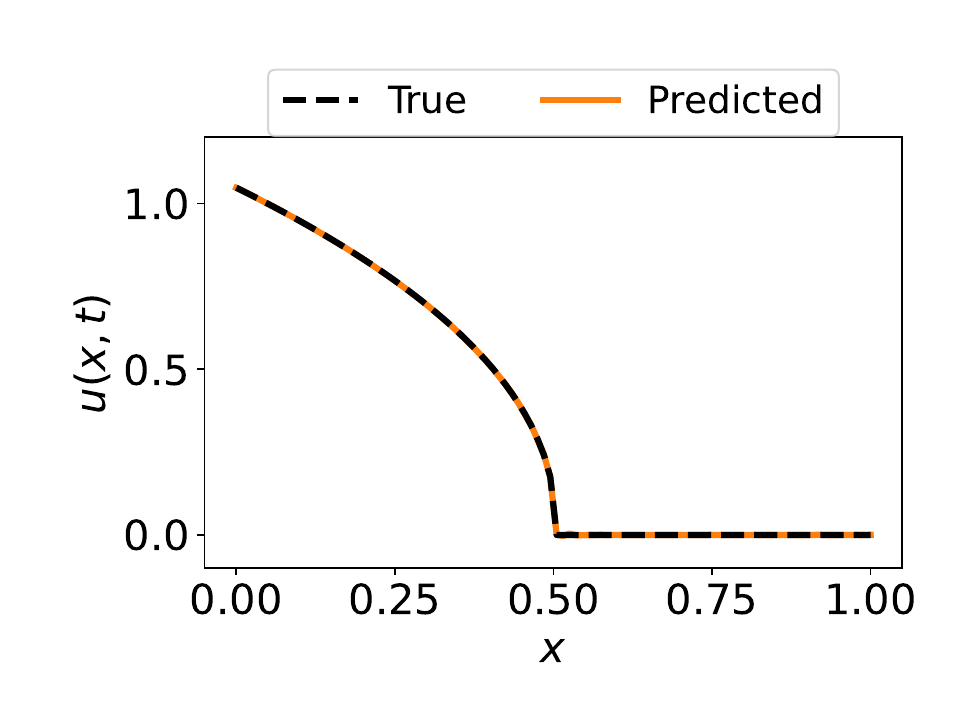}
    \caption{\ensemblenomethod}
    \end{subfigure}
    ~~~~
    \begin{subfigure}[h]{0.30\textwidth}
    \centering
    \includegraphics[scale=0.35]{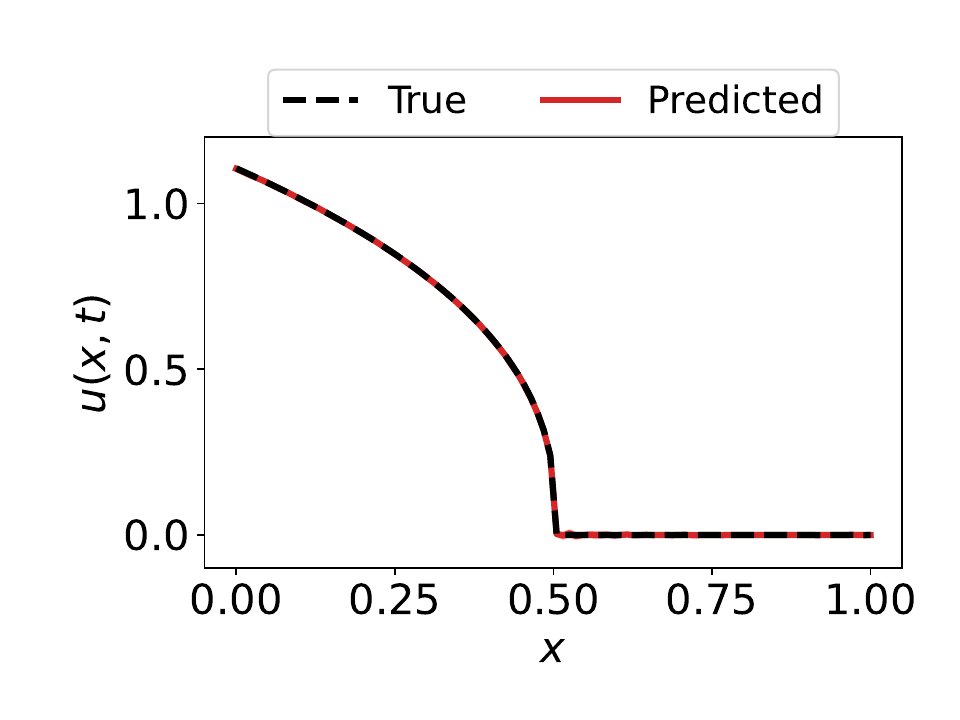}
    \caption{\method}
    \end{subfigure}
    \caption{{\bf 1-d PME, in-domain}, $m^\tr, m^\te \in[2,3]$. 
    Uncertainty estimates from different UQ methods for in-domain values of the power $m$ in the coefficient $k(u) = u^m$.} 
    \label{fig:solutions_pme_id} 
\end{figure}

\begin{figure}[H]
    \centering
    \hspace{-0.5in}
    \begin{subfigure}[h]{0.30\textwidth}
    \centering
    \includegraphics[scale=0.35]{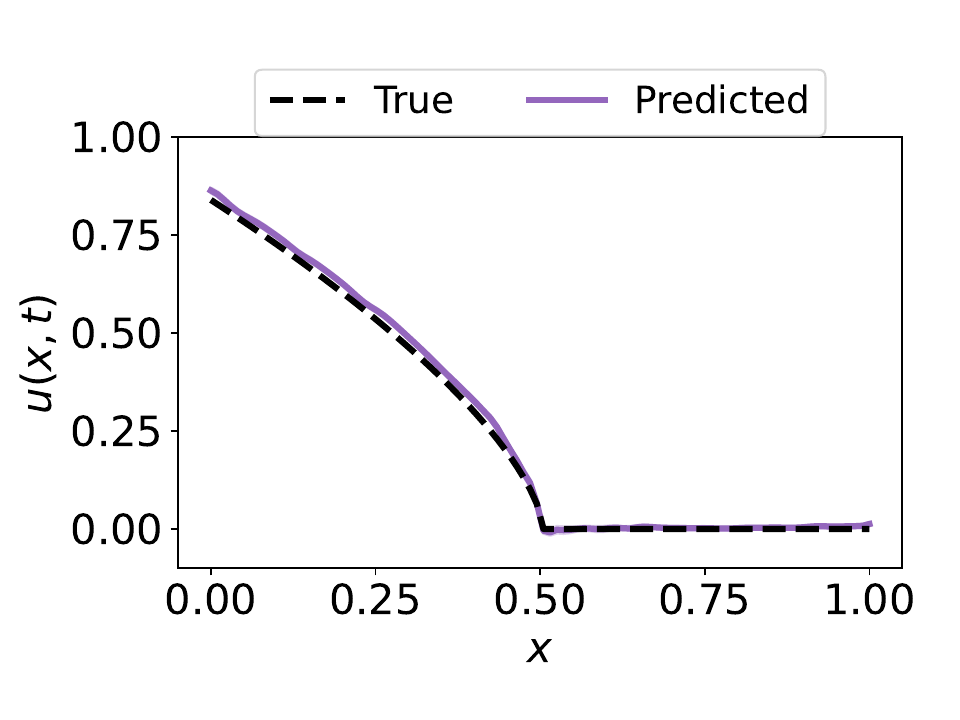}
    \caption{\bayesiannomethod}
    \end{subfigure}
    ~~
    \begin{subfigure}[h]{0.30\textwidth}
    \centering
    \includegraphics[scale=0.35]{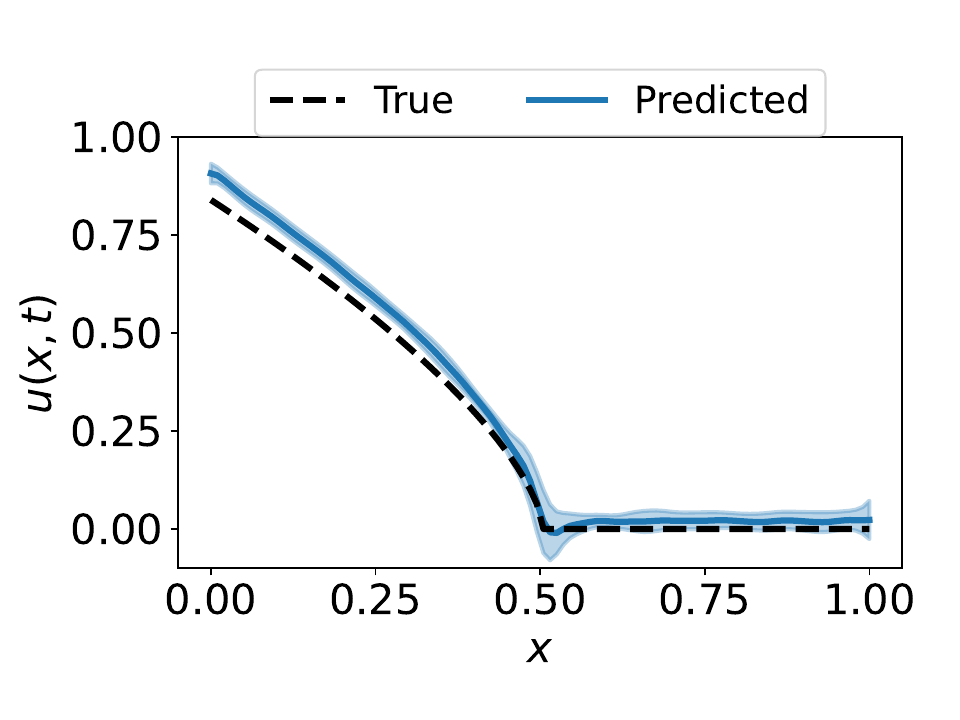}
    \caption{\outputvarmethod}
    \end{subfigure}
    ~~
    \begin{subfigure}[h]{0.30\textwidth}
    \centering
    \includegraphics[scale=0.35]{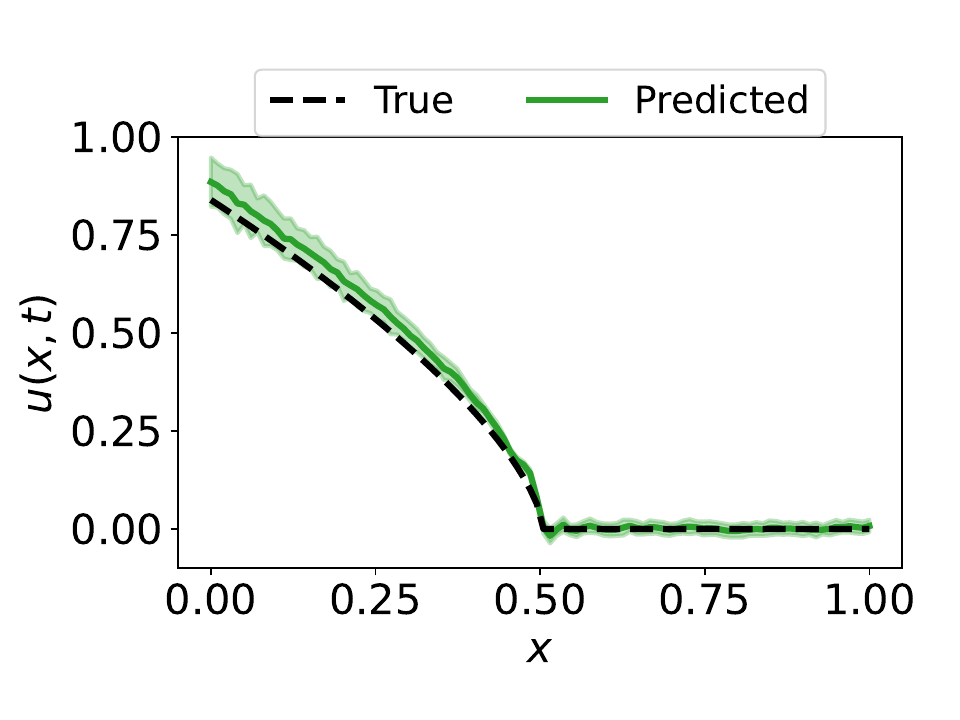}
    \caption{\mcdropoutnomethod}
    \end{subfigure}
    
    \begin{subfigure}[h]{0.30\textwidth}
    \centering
    \includegraphics[scale=0.35]{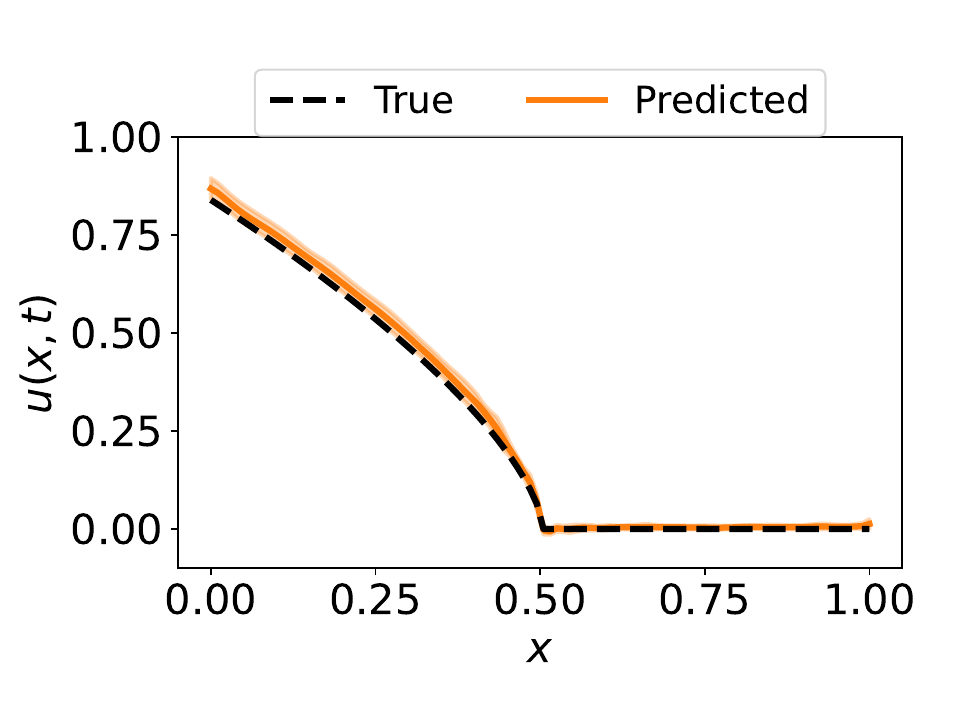}
    \caption{\ensemblenomethod}
    \end{subfigure}
    ~~~~
    \begin{subfigure}[h]{0.30\textwidth}
    \centering
    \includegraphics[scale=0.35]{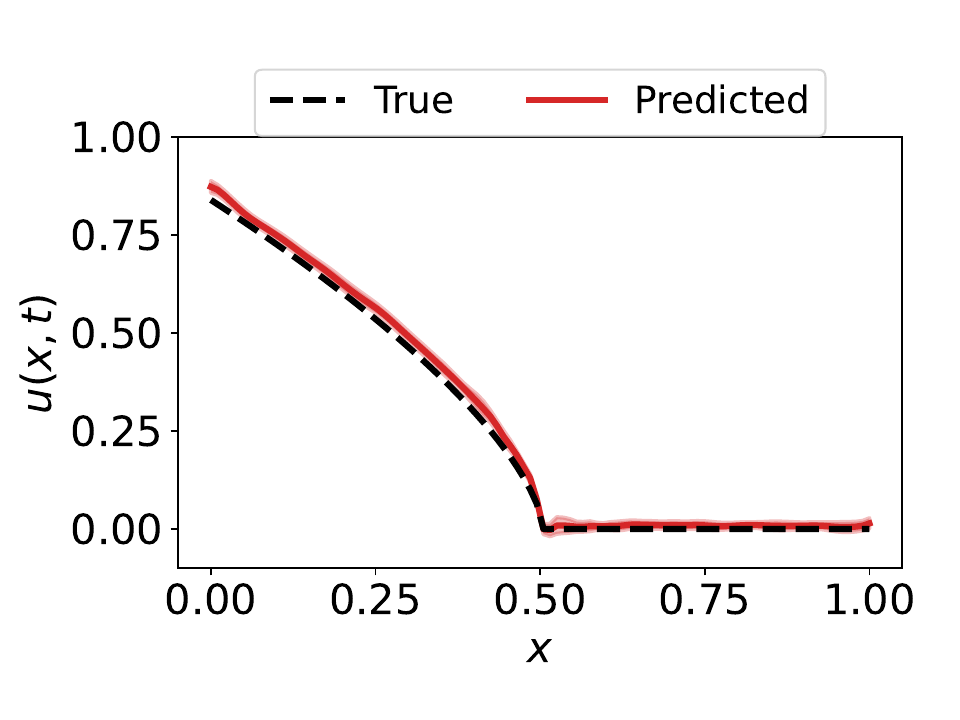}
    \caption{\method}
    \end{subfigure}
    \caption{{\bf 1-d PME, small OOD shift}, $m^\tr\in[2,3], m^\te\in[1,2]$. 
    Uncertainty estimates from different UQ methods under small OOD shifts in the power $m$ in the coefficient $k(u)=u^m$.} 
    \label{fig:solutions_pme_ood2} 
\end{figure}

\begin{figure}[H]
    \centering
    \hspace{-0.5in}
    \begin{subfigure}[h]{0.30\textwidth}
    \centering
    \includegraphics[scale=0.35]{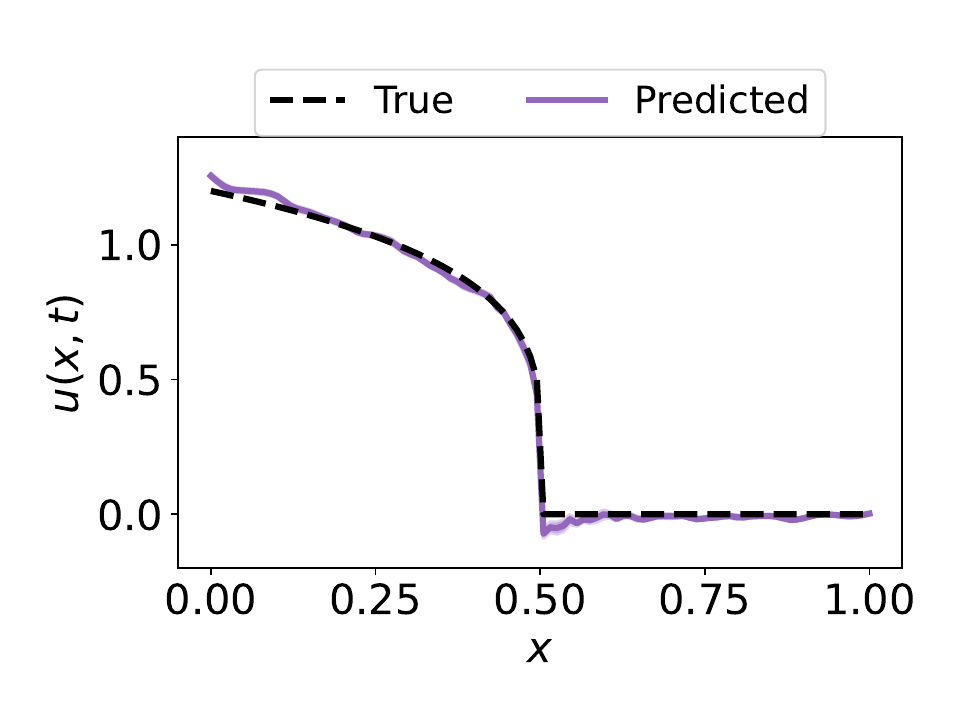}
    \caption{\bayesiannomethod}
    \end{subfigure}
    ~~
    \begin{subfigure}[h]{0.30\textwidth}
    \centering
    \includegraphics[scale=0.35]{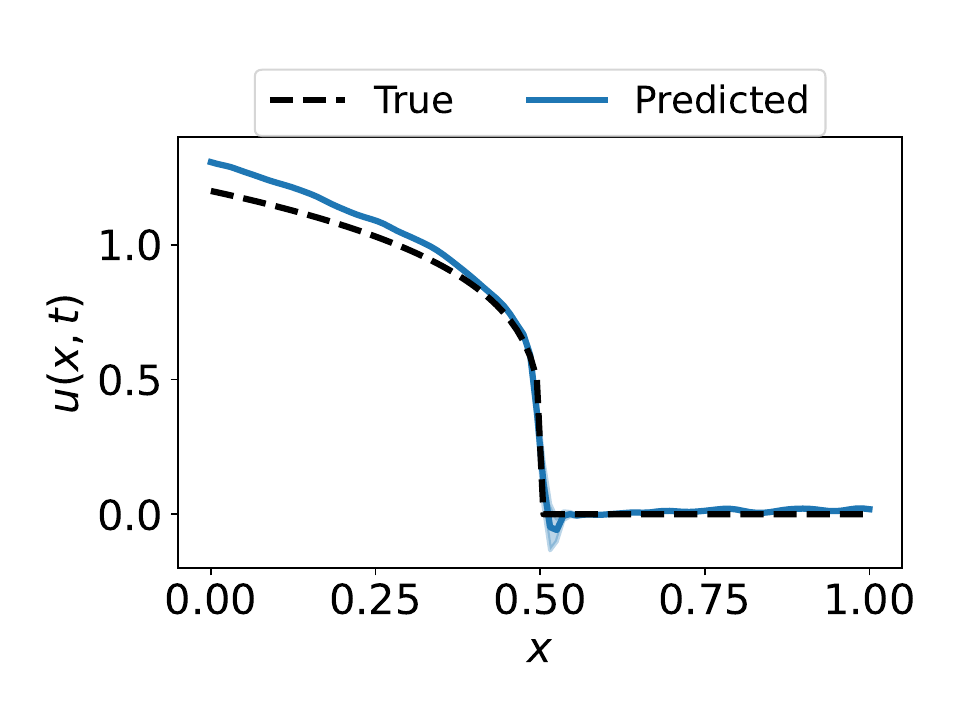}
    \caption{\outputvarmethod}
    \end{subfigure}
    ~~
    \begin{subfigure}[h]{0.30\textwidth}
    \centering
    \includegraphics[scale=0.35]{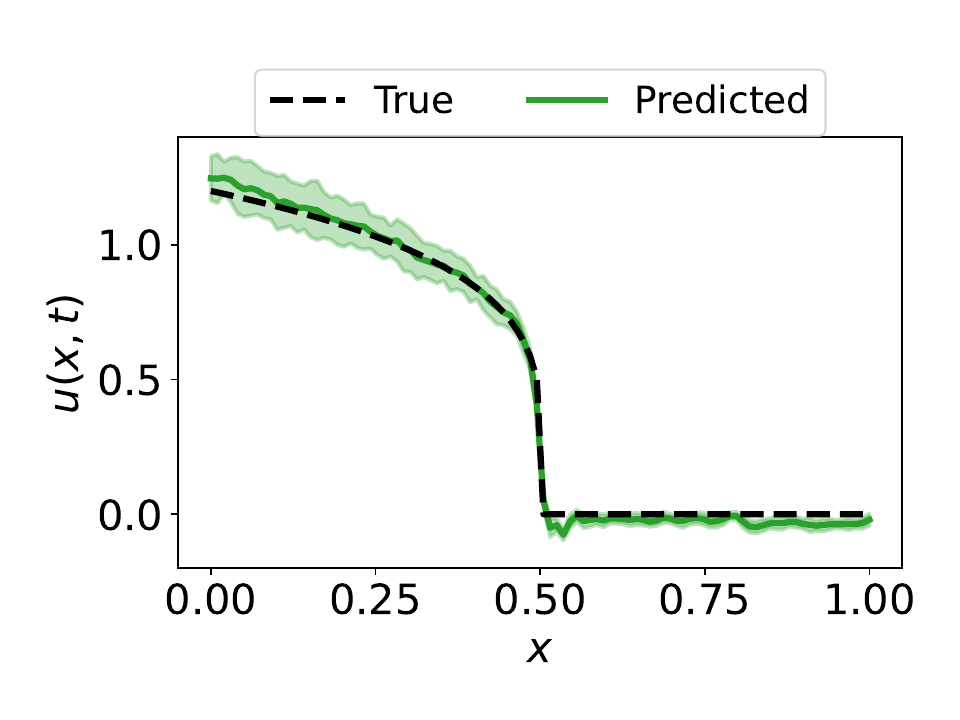}
    \caption{\mcdropoutnomethod}
    \end{subfigure}
    
    \begin{subfigure}[h]{0.30\textwidth}
    \centering
    \includegraphics[scale=0.35]{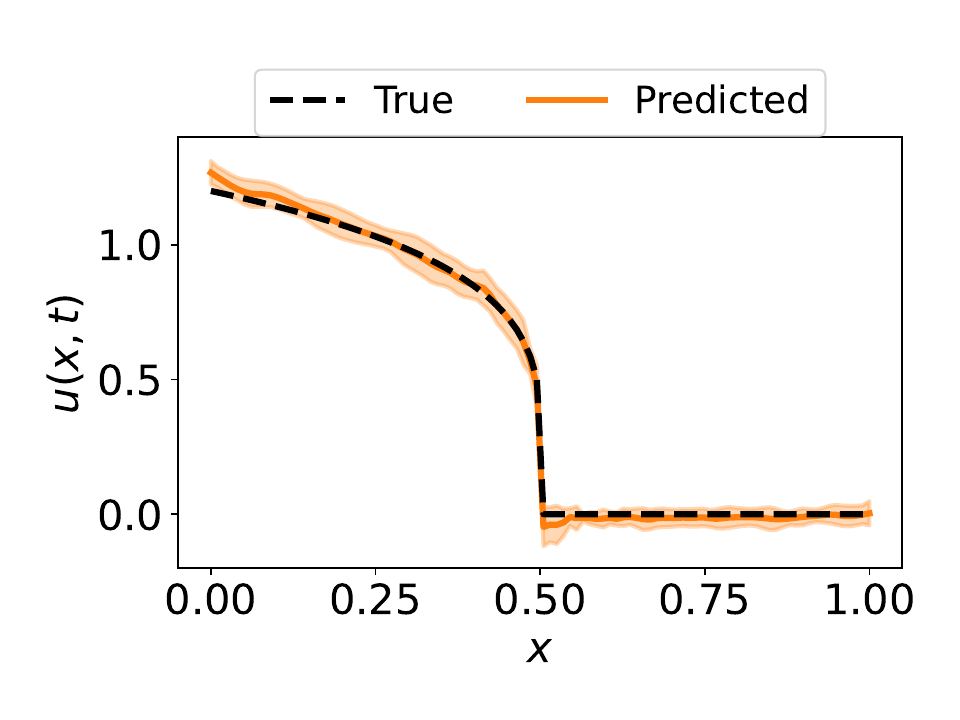}
    \caption{\ensemblenomethod}
    \end{subfigure}
    ~~~~
    \begin{subfigure}[h]{0.30\textwidth}
    \centering
    \includegraphics[scale=0.35]{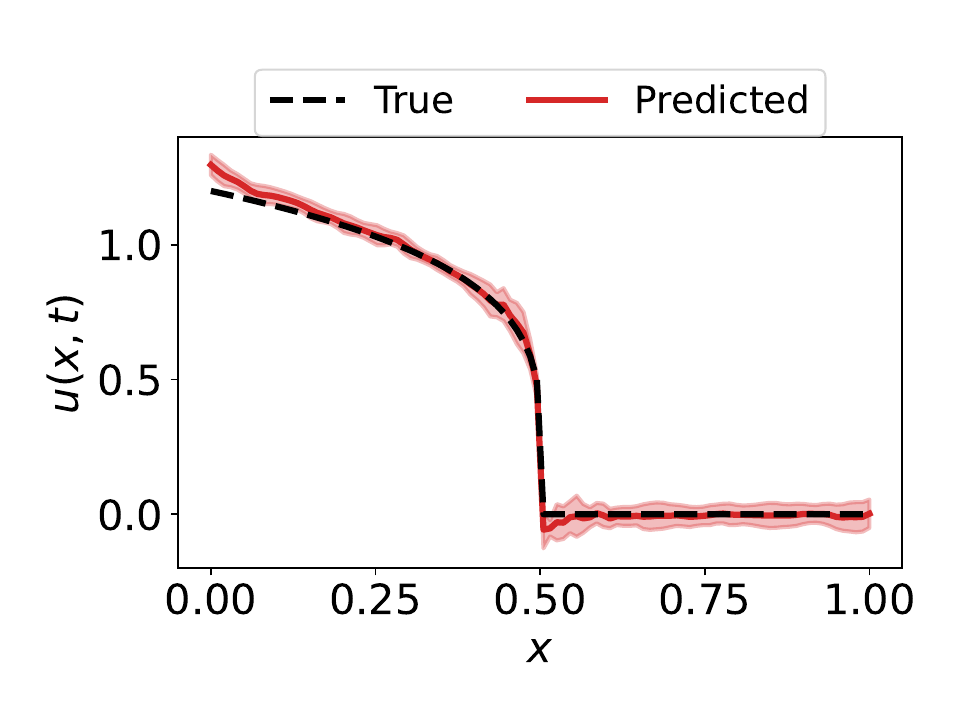}
    \caption{\method}
    \end{subfigure}
    \caption{{\bf 1-d PME, medium OOD shift}, $m^\tr\in[2,3], m^\te\in[4,5]$. 
    Uncertainty estimates from different UQ methods under medium OOD shifts in the power $m$ in the coefficient $k(u)=u^m$.} 
    \label{fig:solutions_pme_ood3} 
\end{figure}

\begin{figure}[H]
    \centering
    \hspace{-0.5in}
    \begin{subfigure}[h]{0.30\textwidth}
    \centering
    \includegraphics[scale=0.35]{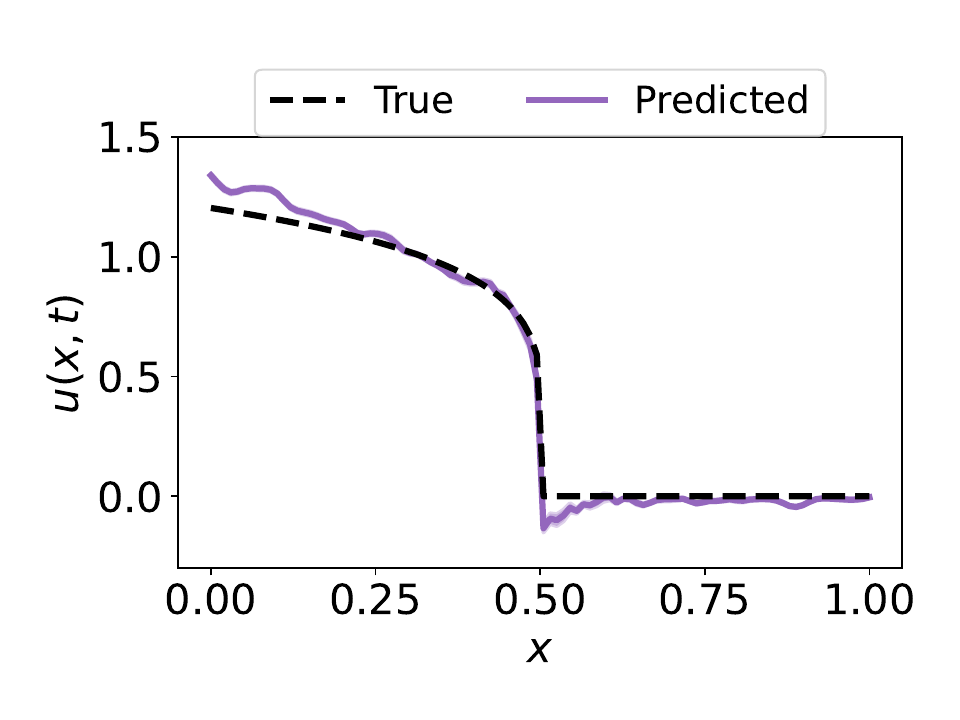}
    \caption{\bayesiannomethod}
    \end{subfigure}
    ~~
    \begin{subfigure}[h]{0.30\textwidth}
    \centering
    \includegraphics[scale=0.35]{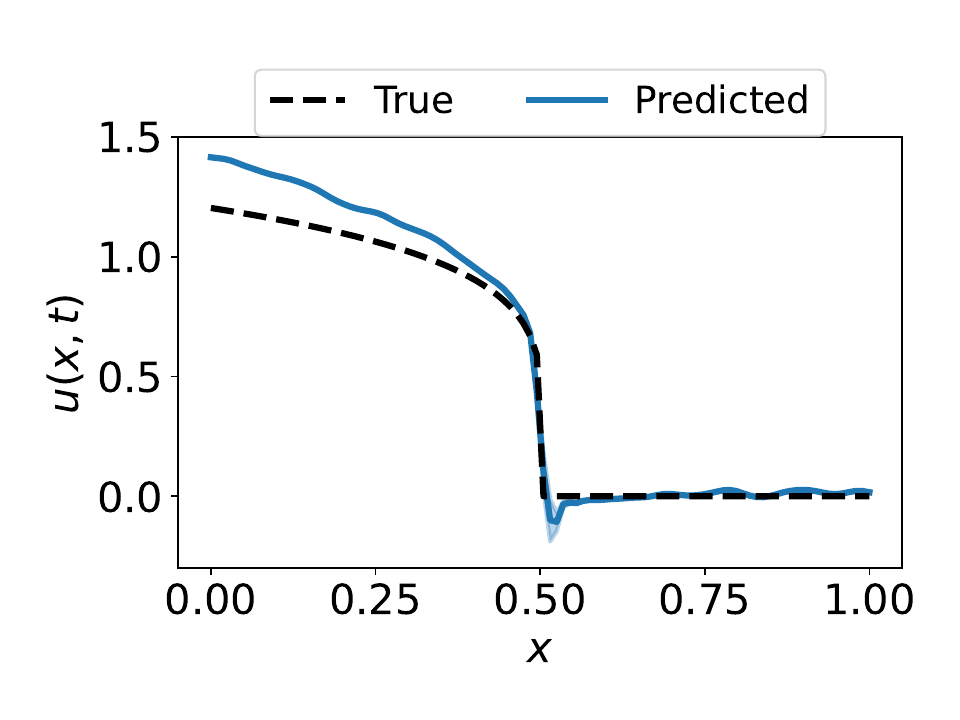}
    \caption{\outputvarmethod}
    \end{subfigure}
    ~~
    \begin{subfigure}[h]{0.30\textwidth}
    \centering
    \includegraphics[scale=0.35]{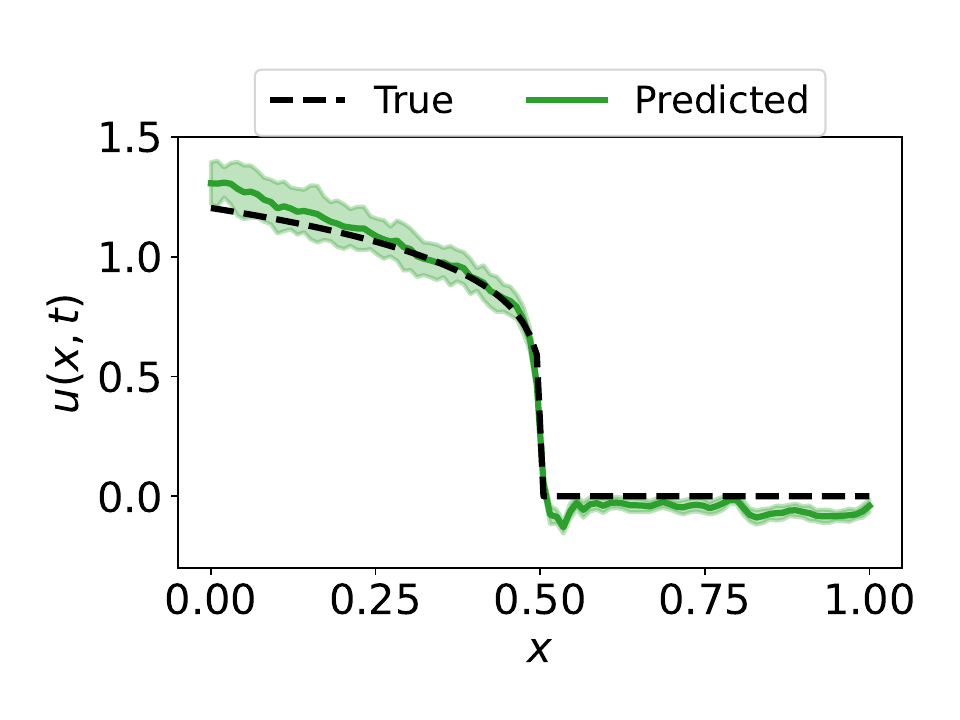}
    \caption{\mcdropoutnomethod}
    \end{subfigure}
    
    \begin{subfigure}[h]{0.30\textwidth}
    \centering
    \includegraphics[scale=0.35]{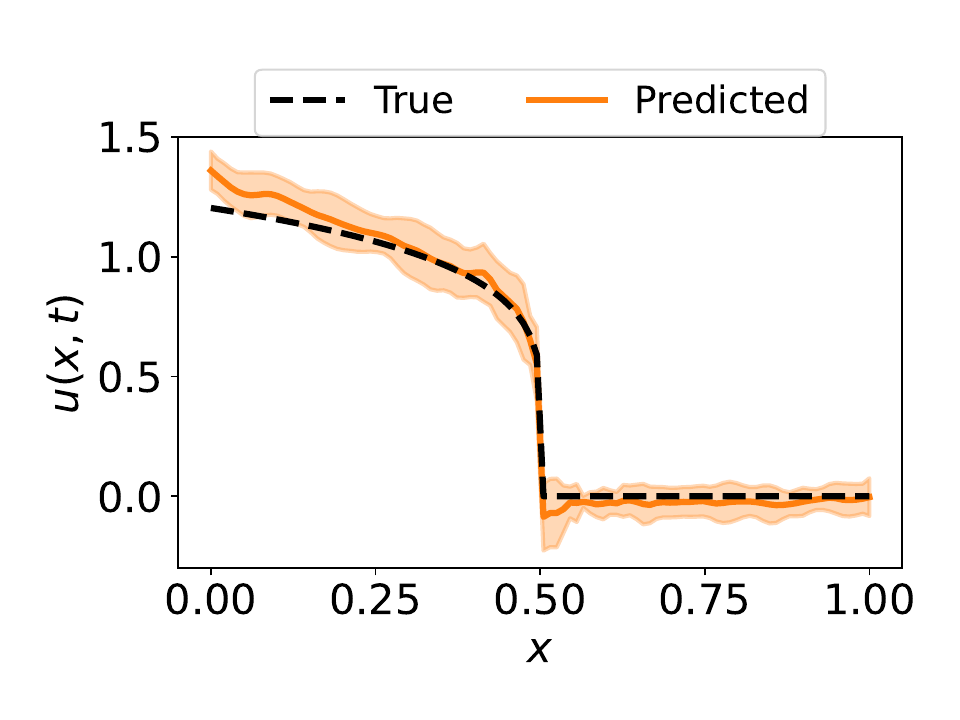}
    \caption{\ensemblenomethod}
    \end{subfigure}
    ~~~~
    \begin{subfigure}[h]{0.30\textwidth}
    \centering
    \includegraphics[scale=0.35]{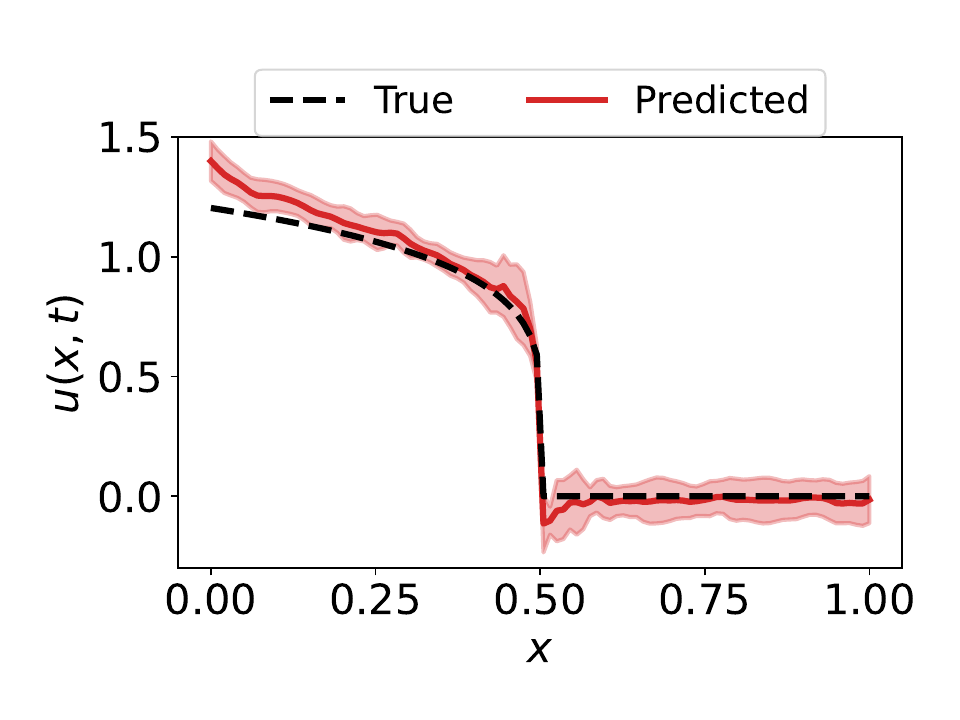}
    \caption{\method}
    \end{subfigure}
    \caption{{\bf 1-d PME, large OOD shift}, $m^\tr\in[2,3], m^\te\in[5,6]$. 
    Uncertainty estimates from different UQ methods under large OOD shifts in the power $m$ in the coefficient $k(u)=u^m$.} 
    \label{fig:solutions_pme_ood4}
\end{figure}

Figures \ref{fig:solutions_pme_id}-\ref{fig:solutions_pme_ood4} show the solution profiles for the ``medium'' degenerate parabolic PME on an in-domain task and for small, medium and large OOD shifts, respectively, of the power $m$ in the monomial coefficient $k(u)=u^m$. The solution for larger values of $m \ge 1$ becomes sharper and more challenging. \cref{fig:solutions_pme_id} shows that all methods perform well on the in-domain task with the exception of \outputvarmethod and \mcdropoutnomethod having small negative oscillations at the sharp corner (degeneracy point), which separates the region with fluid to the left from the region without fluid to the right. \cref{fig:solutions_pme_ood2} shows the solutions on performing inference on an easier task where the values of $m$ are smaller than those trained on and the corresponding solution is smoother. We see in this easier case that all the methods perform reasonably well. There is left boundary error with the growing in time left boundary condition with \outputvarmethod, which grows as the OOD shift increases from medium in \cref{fig:solutions_pme_ood3} to large in \cref{fig:solutions_pme_ood4}. The medium and large shift cases are particularly more challenging since we train on smaller values of $m$ and perform inference on the sharper cases with increased values $m$. As expected as the problems becomes harder and more challenging for larger $m$ (larger shifts), the uncertainty widens for \ensemblenomethod and \method.  See corresponding metric results in \cref{tab:results_pme}.

\begin{table}[H]
    \centering
    \caption{{\bf 1-d PME.} MSE $\downarrow$, NLL $\downarrow$, n-MeRCI $\downarrow$, RMSCE $\downarrow$ and CRPS $\downarrow$ (mean and standard deviation over 5 seeds) metrics for different UQ methods on the 1-d PME in-domain and with small, medium and large OOD shifts, where $m^\tr \in [2,3]$. {\bf Bold} indicates values within one standard deviation of the best mean. 
    }
    \label{tab:results_pme}
    \resizebox{\textwidth}{!}{
    \begin{tabular}{@{}lccccc@{}}
    \toprule
    & \multicolumn{5}{c}{{\bf In-domain}, $m^\te\in [2, 3]$} \\
     & MSE $\downarrow$ & NLL $\downarrow$ & n-MeRCI $\downarrow$ & RMSCE $\downarrow$ & CRPS $\downarrow$ \\
    \midrule
    \bayesiannomethod & 7.3e-07 (8.5e-08) & -1.1e+04 (9.8e+01) &   0.62 ( 0.08) &   0.21 ( 0.00) & 5.6e-04 (3.1e-05) \\
    \outputvarmethod & 8.3e-05 (1.8e-05) & -9.4e+03 (7.6e+02) &   0.55 ( 0.07) & \bf   0.16 ( 0.01) & 2.4e-03 (5.9e-04) \\
    \mcdropoutnomethod & 3.2e-05 (1.2e-05) & -6.7e+03 (7.5e+02) &   0.40 ( 0.05) &   0.22 ( 0.00) & 4.2e-03 (2.4e-04) \\
    \ensemblenomethod & \bf 5.3e-07 (1.8e-07) & \bf -1.3e+04 (4.7e+02) & \bf   0.18 ( 0.10) &   0.18 ( 0.01) & \bf 3.4e-04 (7.4e-05) \\
    \method & 1.8e-06 (2.1e-07) & -1.1e+04 (3.3e+02) & \bf   0.21 ( 0.06) &   0.20 ( 0.01) & 6.0e-04 (5.3e-05) \\
    \end{tabular}
    }
    \resizebox{\textwidth}{!}{
    \begin{tabular}{@{}lccccc@{}}
    \midrule
    & \multicolumn{5}{c}{{\bf Small OOD shift}, $m^\te\in [1, 2]$} \\
     & MSE $\downarrow$ & NLL $\downarrow$ & n-MeRCI $\downarrow$ & RMSCE $\downarrow$ & CRPS $\downarrow$ \\
    \midrule
    \bayesiannomethod & \bf 1.1e-03 (4.0e-04) & 7.7e+05 (2.2e+05) &   1.12 ( 0.07) &   0.44 ( 0.02) & \bf 1.9e-02 (3.5e-03) \\
    \outputvarmethod & 4.0e-03 (2.4e-03) & 2.6e+04 (8.8e+03) &   0.26 ( 0.06) &   0.43 ( 0.03) & 3.6e-02 (1.2e-02) \\
    \mcdropoutnomethod & 2.1e-03 (6.0e-04) & 1.6e+04 (8.6e+03) &   1.18 ( 0.09) & \bf   0.36 ( 0.03) & 2.5e-02 (4.3e-03) \\
    \ensemblenomethod & \bf 1.2e-03 (2.5e-04) & \bf 1.2e+03 (1.8e+03) & \bf   0.14 ( 0.03) &   0.46 ( 0.01) & \bf 1.8e-02 (2.7e-03) \\
    \method & \bf 1.1e-03 (3.7e-04) & 7.8e+03 (6.7e+03) &   0.21 ( 0.04) &   0.42 ( 0.03) & \bf 1.7e-02 (3.8e-03) \\
    \end{tabular}
    }
    \resizebox{\textwidth}{!}{
    \begin{tabular}{@{}lccccc@{}}
    \midrule
    & \multicolumn{5}{c}{{\bf Medium OOD shift}, $m^\te\in [4, 5]$} \\
     & MSE $\downarrow$ & NLL $\downarrow$ & n-MeRCI $\downarrow$ & RMSCE $\downarrow$ & CRPS $\downarrow$ \\
    \midrule
    \bayesiannomethod & 1.0e-03 (3.2e-04) & 1.6e+05 (4.1e+04) &   0.73 ( 0.03) &   0.47 ( 0.01) & 2.1e-02 (3.1e-03) \\
    \outputvarmethod & 5.0e-03 (7.6e-04) & 2.4e+07 (7.7e+06) &   1.23 ( 0.34) &   0.50 ( 0.00) & 5.1e-02 (5.2e-03) \\
    \mcdropoutnomethod & 1.5e-03 (4.2e-04) & 3.7e+03 (2.9e+03) &   0.75 ( 0.02) &   0.42 ( 0.03) & 2.4e-02 (5.2e-03) \\
    \ensemblenomethod & \bf 8.1e-04 (1.6e-04) & \bf -3.1e+03 (1.0e+03) &   0.20 ( 0.03) & \bf   0.38 ( 0.01) & \bf 1.5e-02 (1.5e-03) \\
    \method & 1.1e-03 (3.5e-04) & -1.8e+03 (3.9e+03) & \bf   0.15 ( 0.03) & \bf   0.38 ( 0.02) & 1.8e-02 (3.3e-03) \\
    \end{tabular}
    }
    \resizebox{\textwidth}{!}{
    \begin{tabular}{@{}lccccc@{}}
    \midrule
    & \multicolumn{5}{c}{{\bf Large OOD shift}, $m^\te\in [5, 6]$} \\
     & MSE $\downarrow$ & NLL $\downarrow$ & n-MeRCI $\downarrow$ & RMSCE $\downarrow$ & CRPS $\downarrow$ \\
    \midrule
    \bayesiannomethod & 6.1e-03 (1.9e-03) & 8.0e+05 (2.1e+05) &   0.69 ( 0.02) &   0.49 ( 0.01) & 5.4e-02 (8.7e-03) \\
    \outputvarmethod & 2.0e-02 (1.8e-03) & 6.9e+08 (4.2e+08) &   1.52 ( 0.46) &   0.50 ( 0.00) & 1.0e-01 (5.8e-03) \\
    \mcdropoutnomethod & 6.4e-03 (2.2e-03) & 1.8e+04 (8.3e+03) &   0.70 ( 0.02) &   0.47 ( 0.03) & 5.6e-02 (1.4e-02) \\
    \ensemblenomethod & \bf 4.6e-03 (7.1e-04) & \bf 4.4e+02 (1.9e+03) &   0.22 ( 0.02) & \bf   0.42 ( 0.02) & \bf 4.0e-02 (3.0e-03) \\
    \method & 5.8e-03 (1.7e-03) & \bf 8.2e+02 (4.5e+03) & \bf   0.13 ( 0.05) & \bf   0.41 ( 0.02) & 4.5e-02 (7.5e-03) \\
    \bottomrule
    \end{tabular}
    }
\end{table}

\subsubsection{``Hard'' Stefan Problem}
\begin{figure}[H]
    \centering
    \hspace{-0.5in}
    \begin{subfigure}[h]{0.30\textwidth}
    \centering
    \includegraphics[scale=0.33]{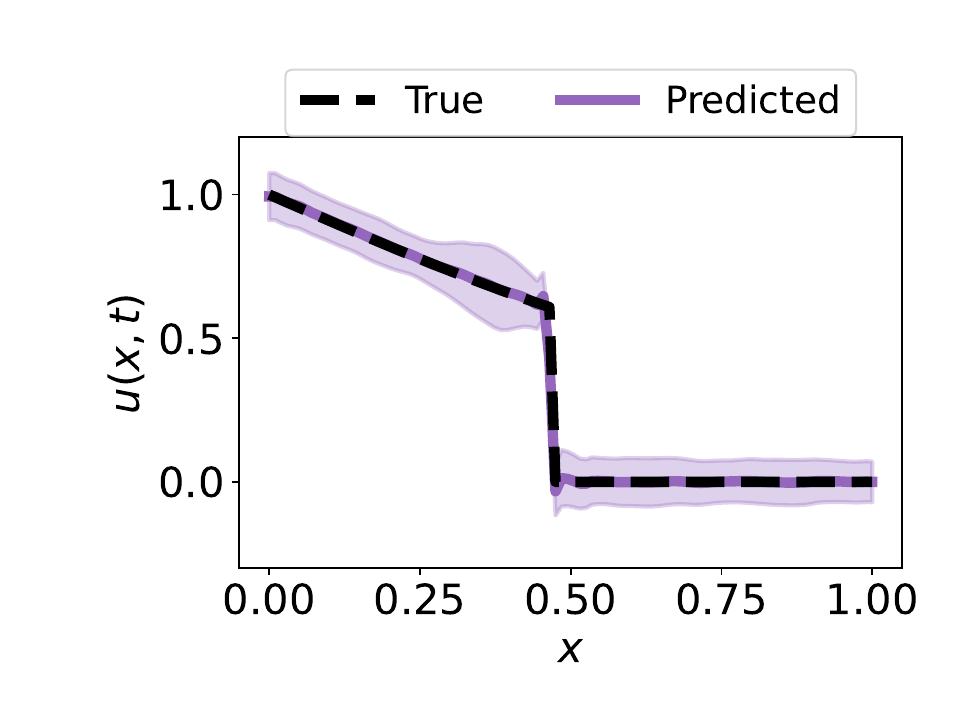}
    \caption{\bayesiannomethod}
    \end{subfigure}
    ~~
    \begin{subfigure}[h]{0.30\textwidth}
    \centering
    \includegraphics[scale=0.33]{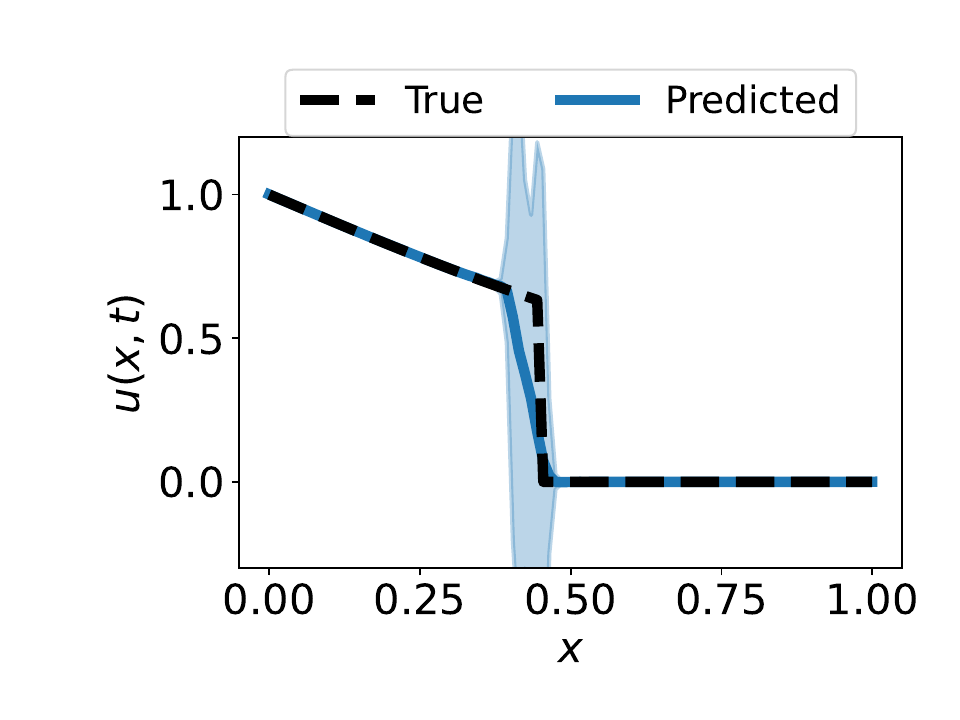}
    \caption{\outputvarmethod}
    \end{subfigure}
    ~~
    \begin{subfigure}[h]{0.30\textwidth}
    \centering
    \includegraphics[scale=0.33]{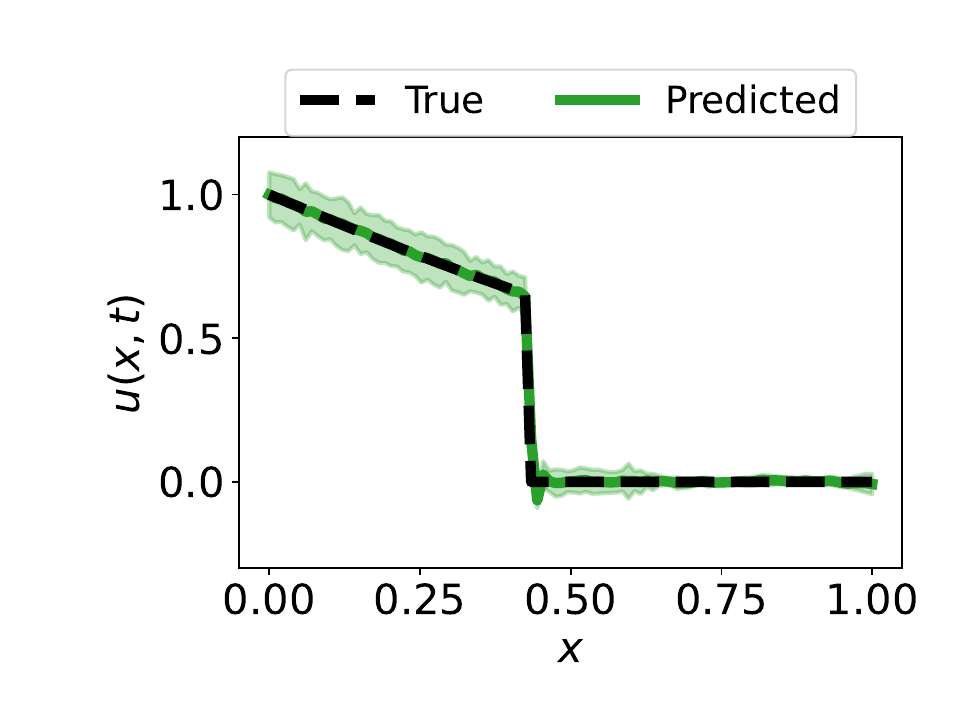}
    \caption{\mcdropoutnomethod}
    \end{subfigure}
    
    \begin{subfigure}[h]{0.30\textwidth}
    \centering
    \includegraphics[scale=0.33]{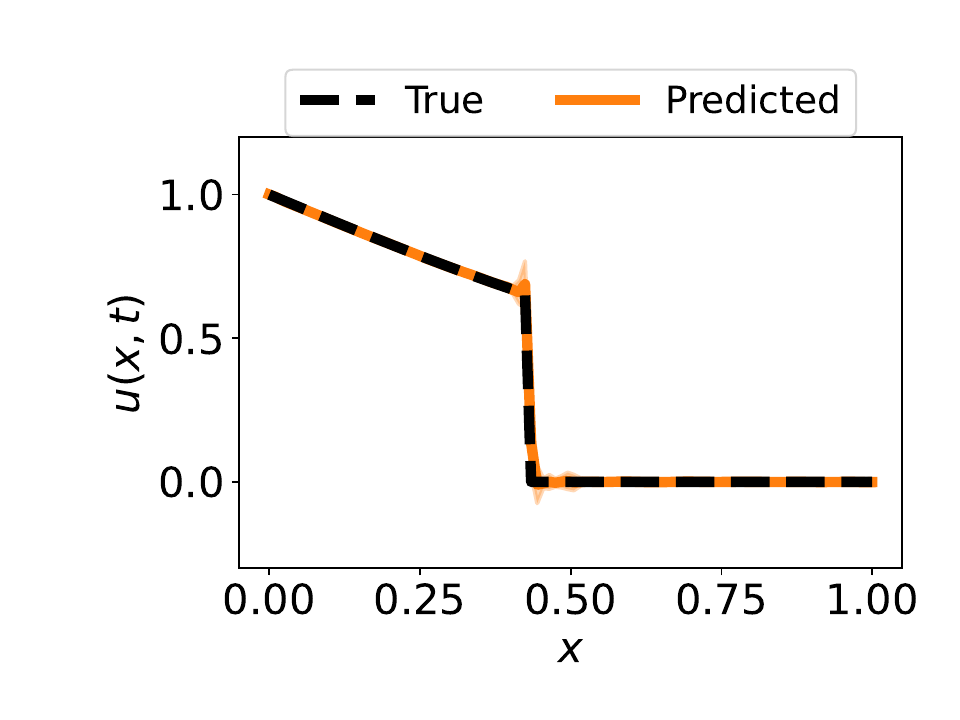}
    \caption{\ensemblenomethod}
    \end{subfigure}
    ~~~~
    \begin{subfigure}[h]{0.30\textwidth}
    \centering
    \includegraphics[scale=0.33]{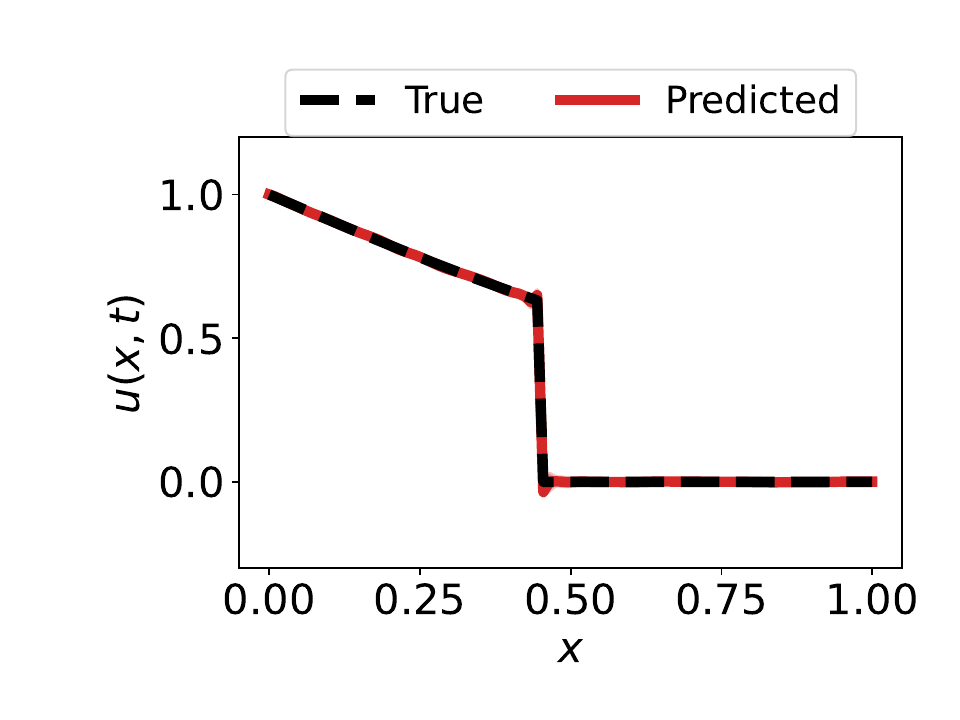}
    \caption{\method}
    \end{subfigure}
    \caption{{\bf 1-d Stefan Equation, in-domain}, $u{^*}^{\tr}, u{^*}^{\te} \in[0.6,0.65]$. 
    Uncertainty estimates from different UQ methods for in-domain values for the solution value at the shock $u(t, x^*(t)) = u^*$ for shock position $x^*(t)$.} 
    \label{fig:solutions_stefan_id}
\end{figure}

\begin{figure}[H]
    \centering
    \hspace{-0.5in}
    \begin{subfigure}[h]{0.30\textwidth}
    \centering
    \includegraphics[scale=0.33]{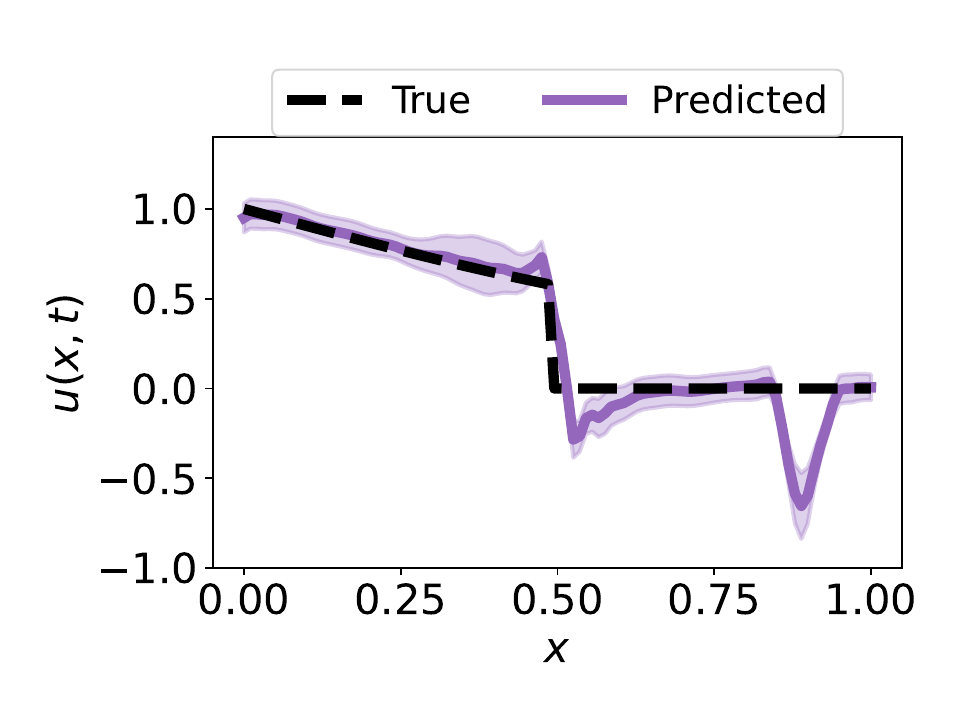}
    \caption{\bayesiannomethod}
    \end{subfigure}
    ~~
    \begin{subfigure}[h]{0.30\textwidth}
    \centering
    \includegraphics[scale=0.33]{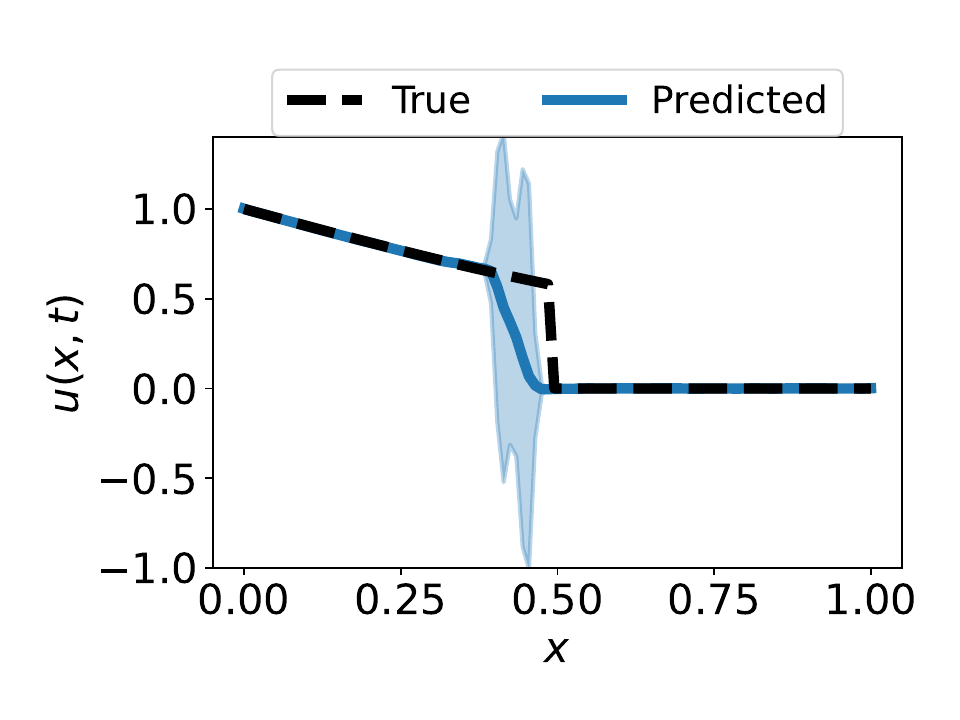}
    \caption{\outputvarmethod}
    \end{subfigure}
    ~~
    \begin{subfigure}[h]{0.30\textwidth}
    \centering
    \includegraphics[scale=0.33]{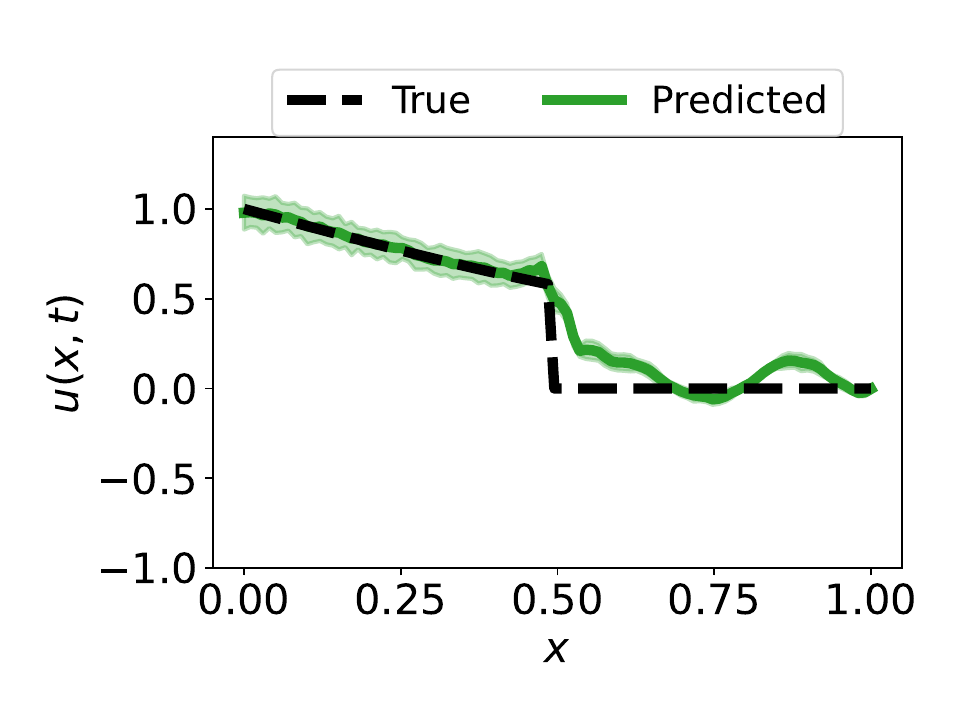}
    \caption{\mcdropoutnomethod}
    \end{subfigure}
    
    \begin{subfigure}[h]{0.30\textwidth}
    \centering
    \includegraphics[scale=0.33]{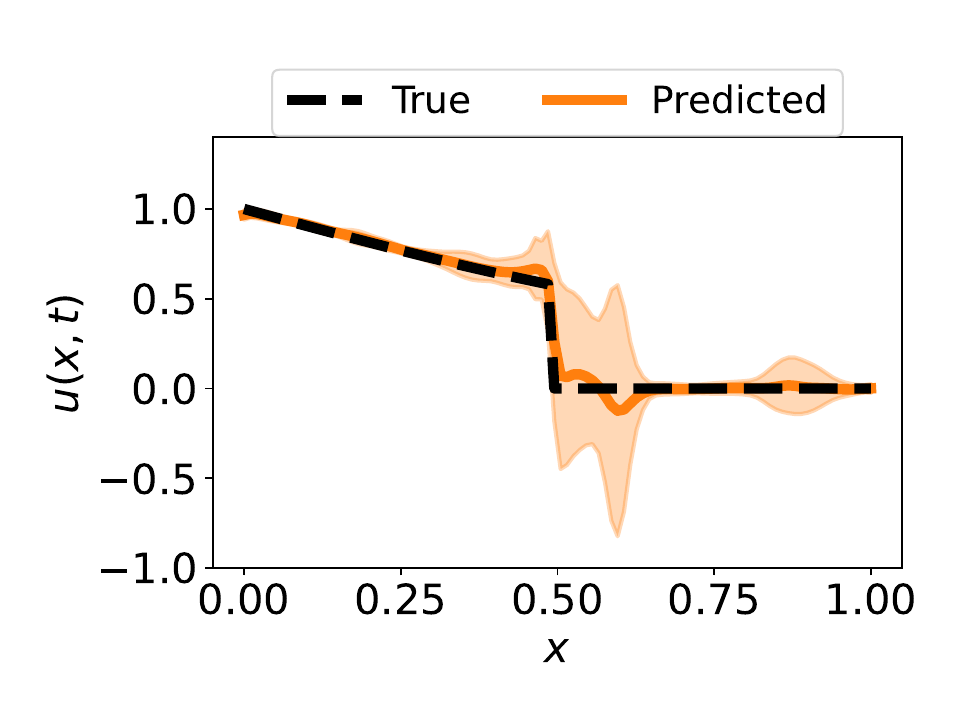}
    \caption{\ensemblenomethod}
    \end{subfigure}
    ~~~~
    \begin{subfigure}[h]{0.30\textwidth}
    \centering
    \includegraphics[scale=0.33]{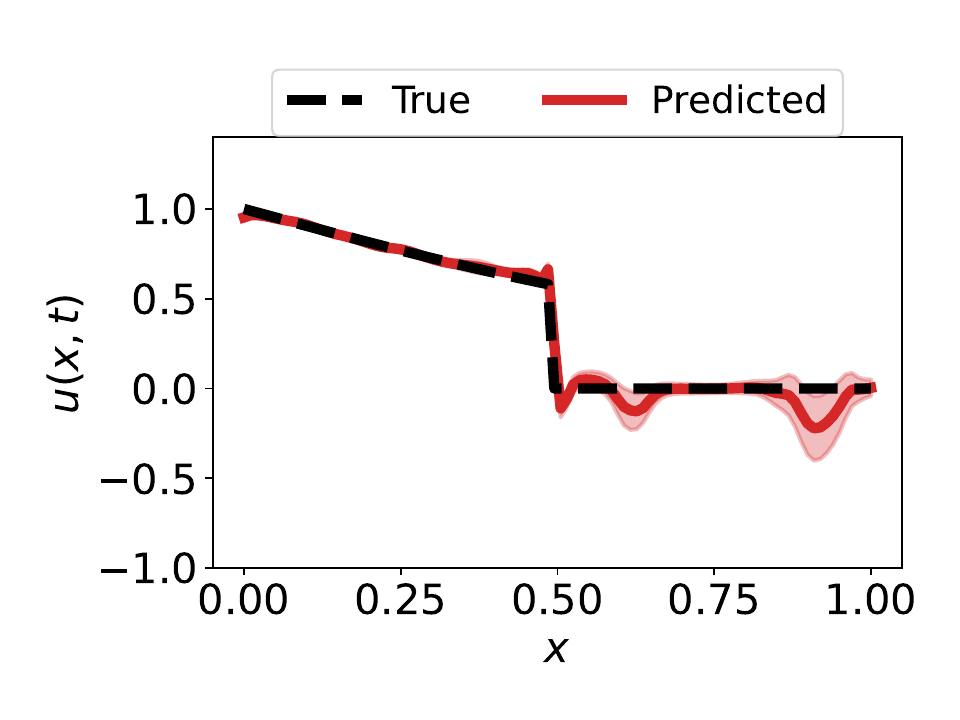}
    \caption{\method}
    \end{subfigure}
    \caption{{\bf 1-d Stefan Equation, small OOD shift}, $u{^*}^{\tr}\in[0.6,0.65], u{^*}^{\te}\in[0.55,0.6]$. 
    Uncertainty estimates from different UQ methods under small OOD shifts in the solution value at the shock $u(t, x^*(t)) = u^*$ for shock position $x^*(t)$.} 
    \label{fig:solutions_stefan_ood_small}
\end{figure}

\begin{figure}[H]
    \centering
    \hspace{-0.5in}
    \begin{subfigure}[t]{0.30\textwidth}
    \centering
    \includegraphics[scale=0.33]{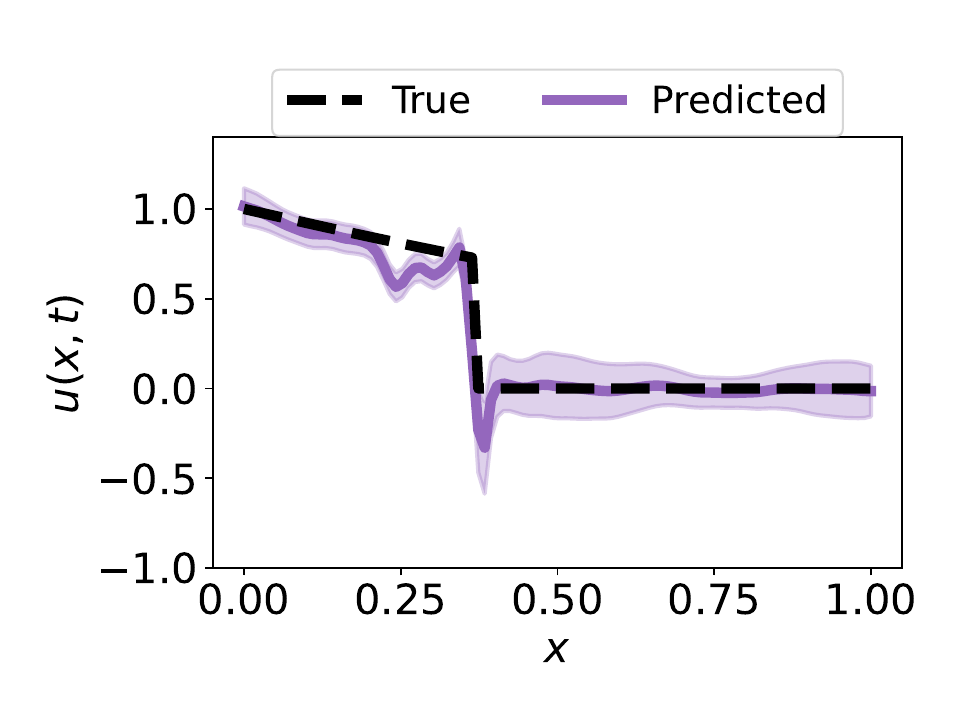}
    \caption{\bayesiannomethod}
    \end{subfigure}
    ~~
    \begin{subfigure}[t]{0.30\textwidth}
    \centering
    \includegraphics[scale=0.33]{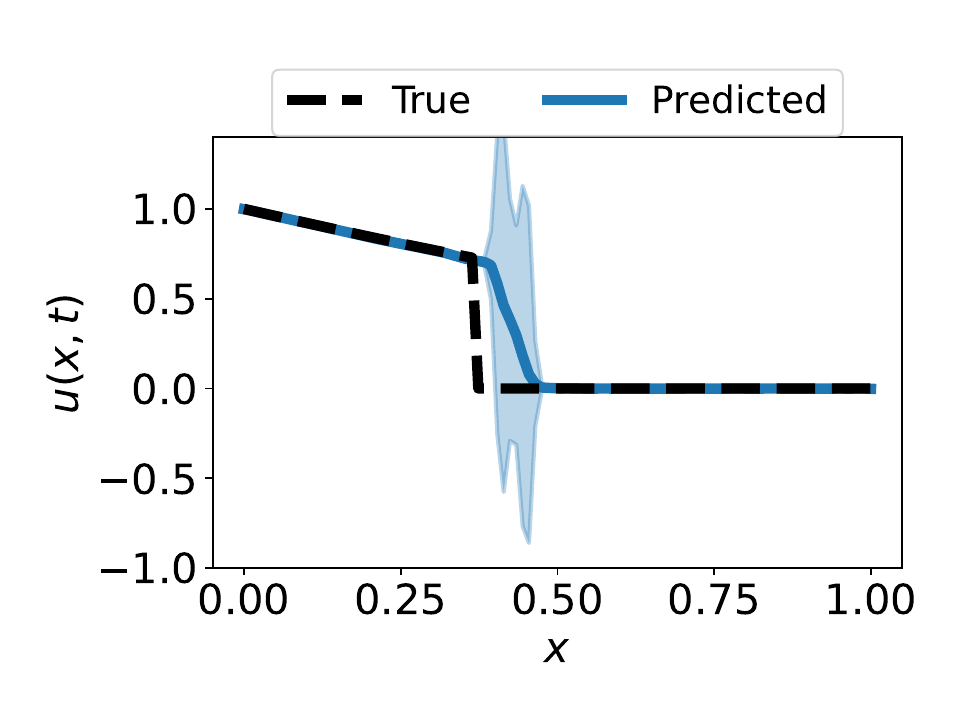}
    \caption{\outputvarmethod}
    \end{subfigure}
    ~~
    \begin{subfigure}[t]{0.30\textwidth}
    \centering
    \includegraphics[scale=0.33]{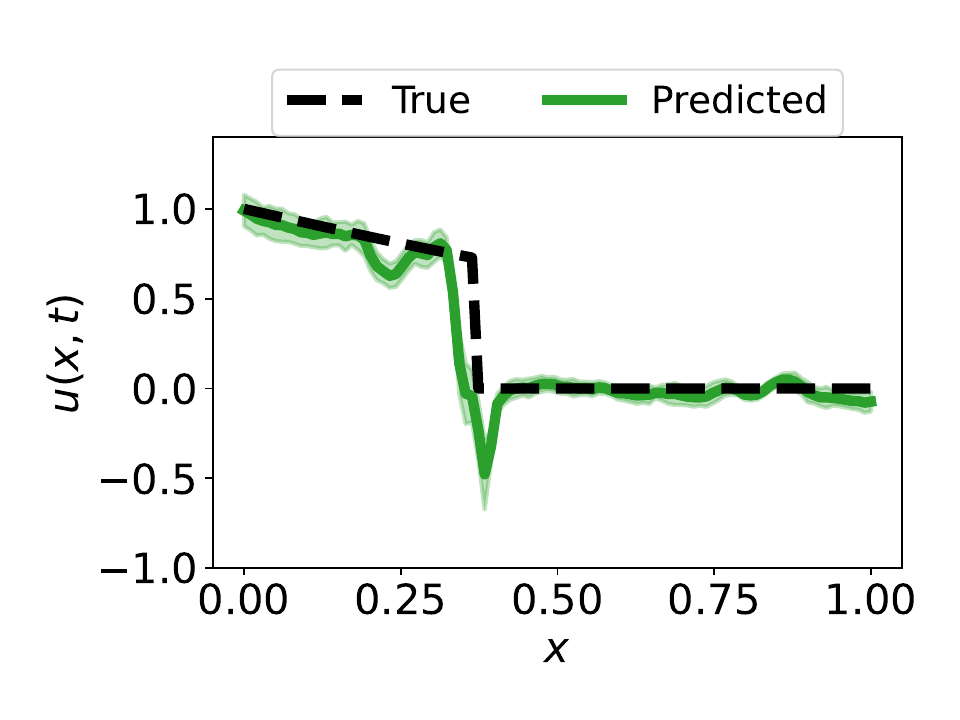}
    \caption{\mcdropoutnomethod}
    \end{subfigure}
    
    \begin{subfigure}[t]{0.30\textwidth}
    \centering
    \includegraphics[scale=0.33]{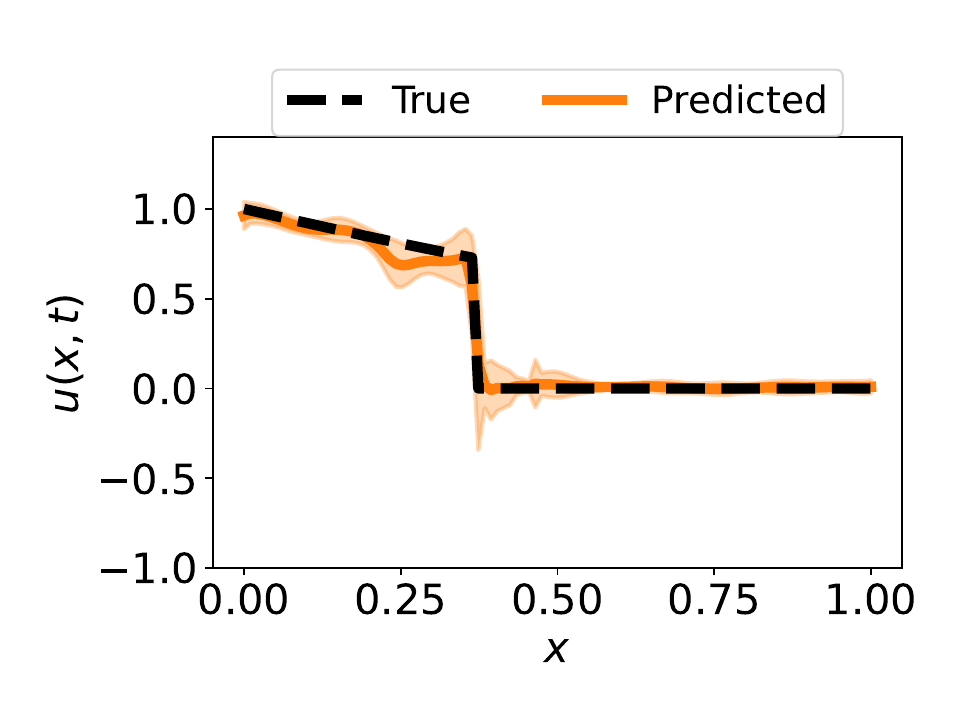}
    \caption{\ensemblenomethod}
    \end{subfigure}
    ~~~~
    \begin{subfigure}[t]{0.30\textwidth}
    \centering
    \includegraphics[scale=0.33]{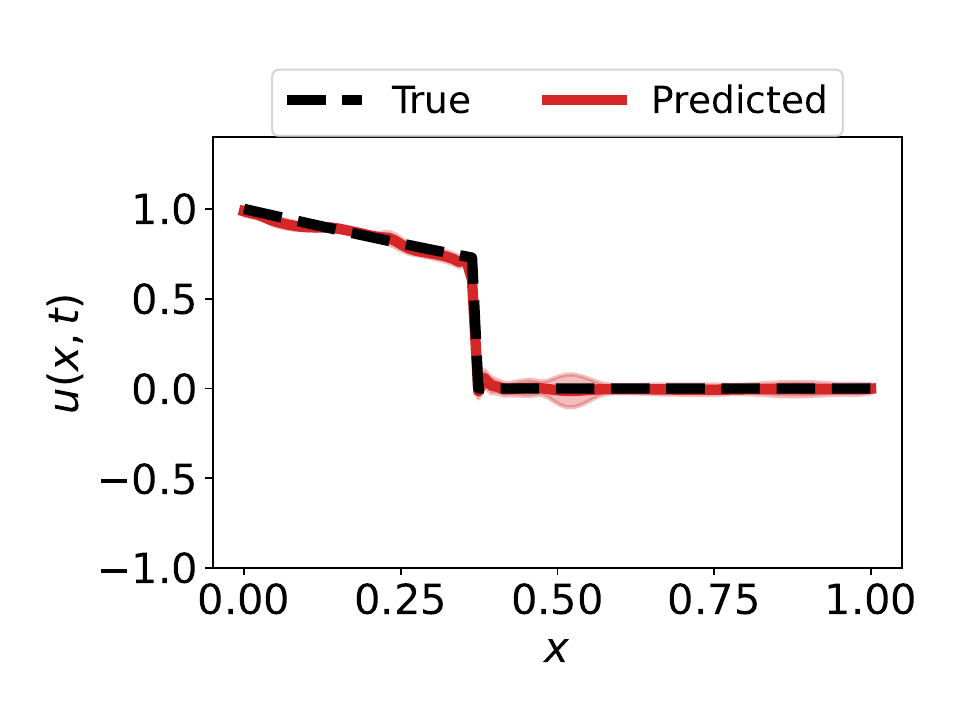}
    \caption{\method}
    \end{subfigure}
     \caption{{\bf 1-d Stefan Equation, medium OOD shift}, $u{^*}^{\tr}\in[0.6,0.65], u{^*}^{\te}\in[0.7,0.75]$. 
    Uncertainty estimates from different UQ methods under medium OOD shifts in the solution value at the shock $u(x^*(t), t) = u^*$ for shock position $x^*(t)$.
    } 
    \label{fig:solutions_stefan_ood_medium}
\end{figure}

\begin{figure}[H]
    \centering
    \hspace{-0.5in}
    \begin{subfigure}[t]{0.30\textwidth}
    \centering
    \includegraphics[scale=0.33]{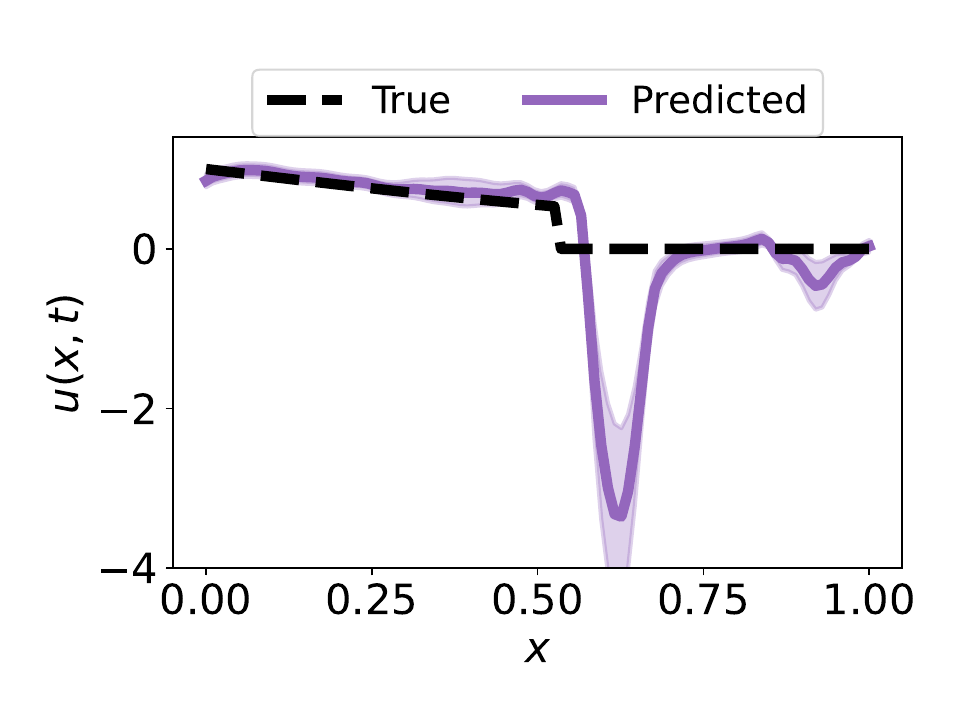}
    \caption{\bayesiannomethod}
    \end{subfigure}
    ~~
    \begin{subfigure}[t]{0.30\textwidth}
    \centering
    \includegraphics[scale=0.33]{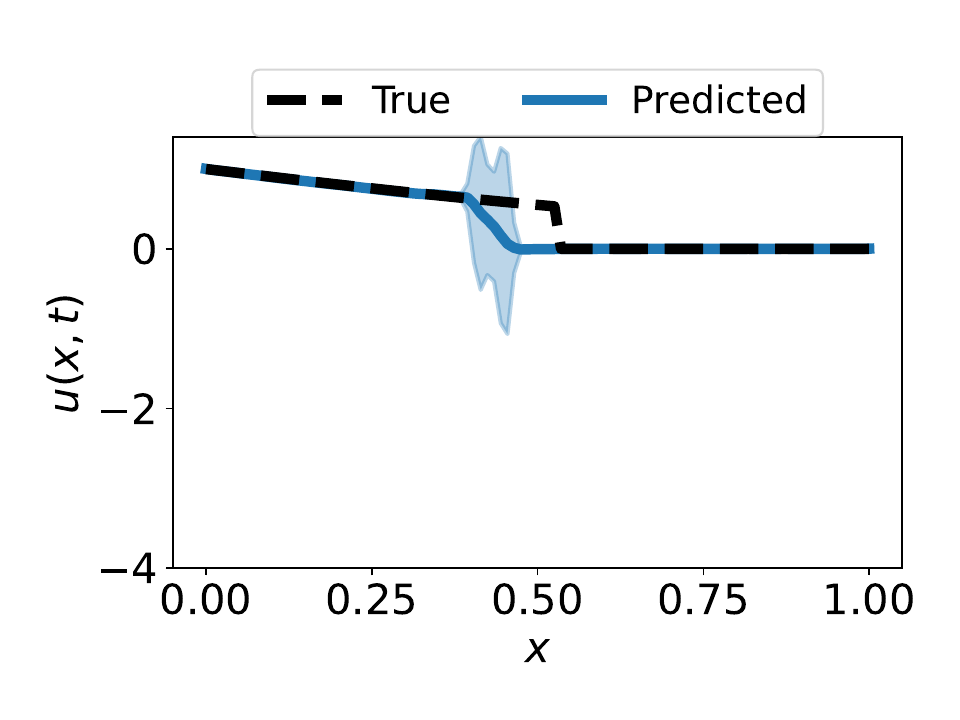}
    \caption{\outputvarmethod}
    \end{subfigure}
    ~~
    \begin{subfigure}[t]{0.30\textwidth}
    \centering
    \includegraphics[scale=0.33]{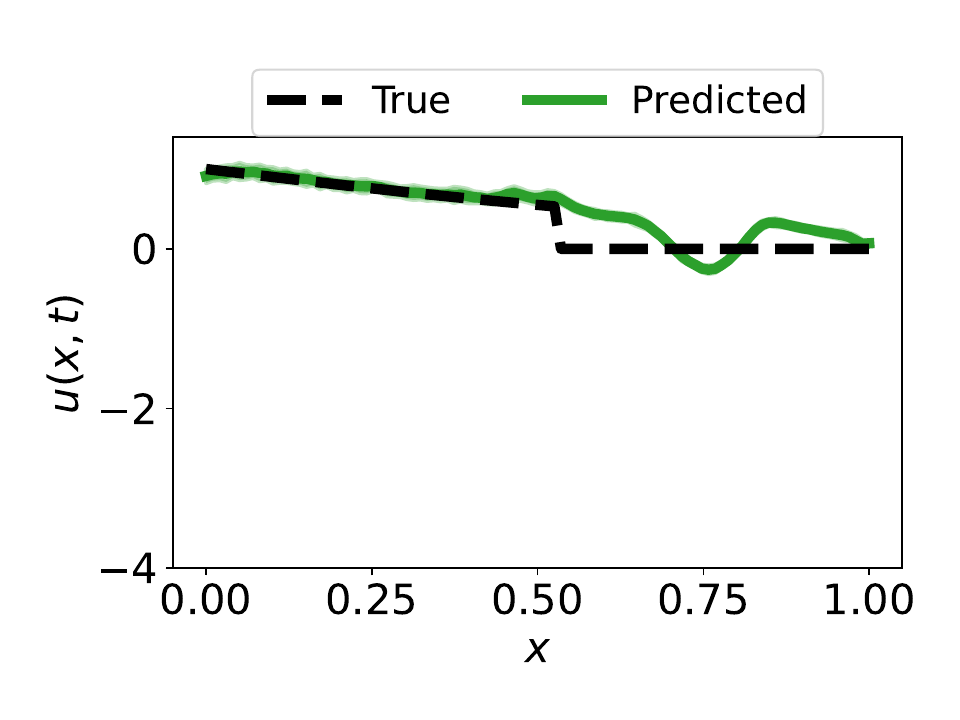}
    \caption{\mcdropoutnomethod}
    \end{subfigure}
    
    \begin{subfigure}[t]{0.30\textwidth}
    \centering
    \includegraphics[scale=0.33]{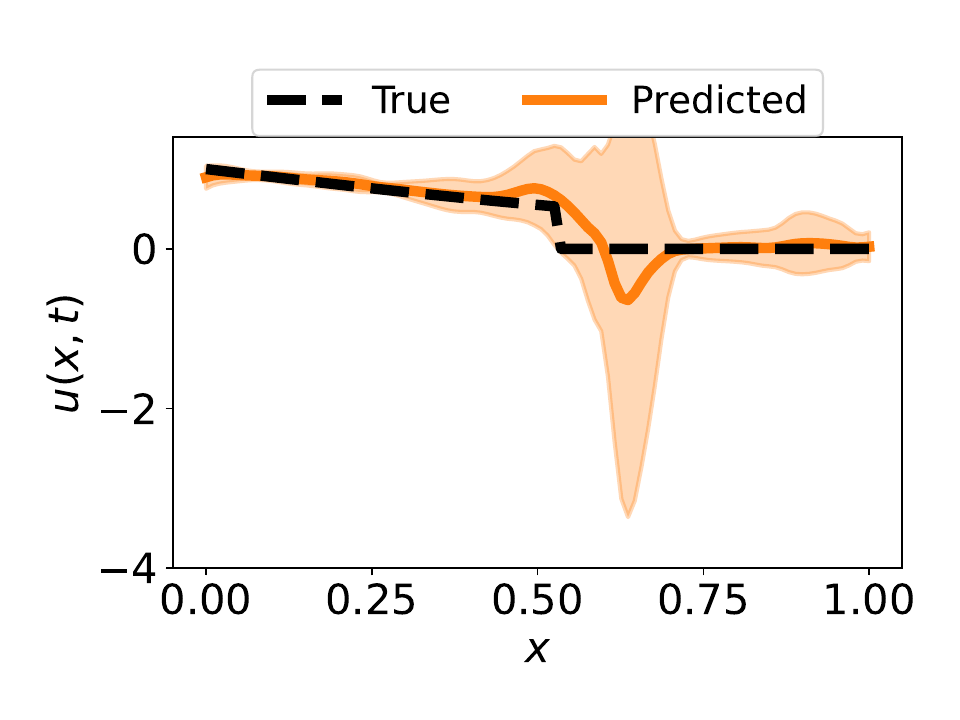}
    \caption{\ensemblenomethod}
    \end{subfigure}
    ~~~~
    \begin{subfigure}[t]{0.30\textwidth}
    \centering
    \includegraphics[scale=0.33]{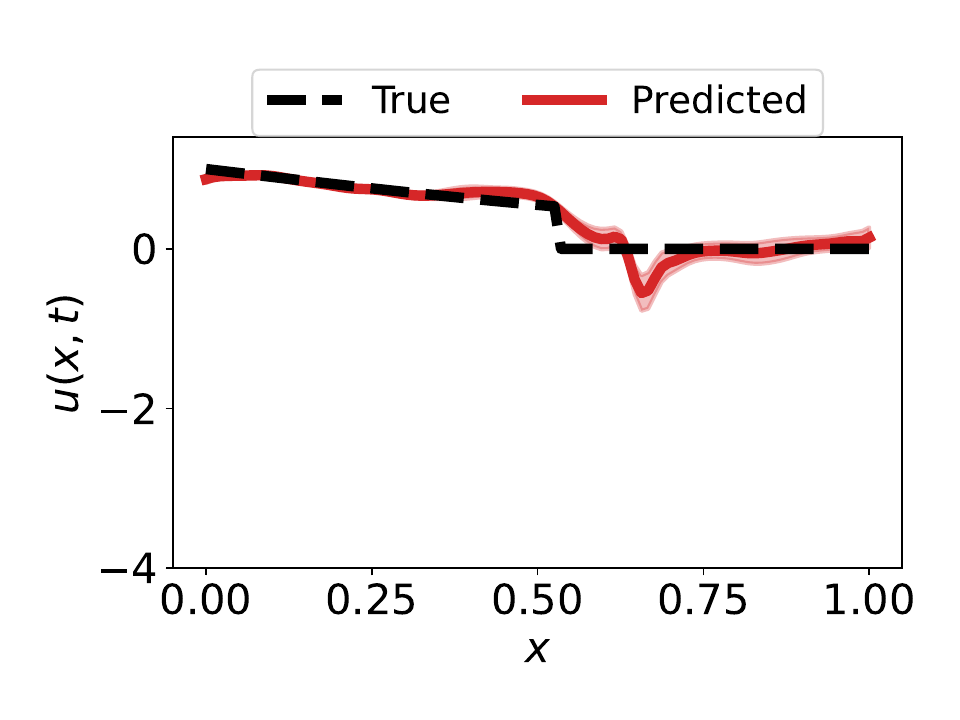}
    \caption{\method}
    \end{subfigure}
       \caption{{\bf 1-d Stefan Equation, large OOD shift}, $u{^*}^{\tr}\in[0.6,0.65], u{^*}^{\te}\in[0.5,0.55]$. 
    Uncertainty estimates from different UQ methods under large OOD shifts in the solution value at the shock $u(x^*(t), t) = u^*$ for shock position $x^*(t)$.}
    \label{fig:solutions_stefan_ood3}
\end{figure}

Figures \ref{fig:solutions_stefan_id}-\ref{fig:solutions_stefan_ood3} show the solution profiles for the ``hard'' degenerate parabolic, discontinuous GPME, i.e., the Stefan equation, on an in-domain task and for small, medium and large OOD shifts, respectively, of the coefficient $u^*$ in $k(u) = \vone_{u \geq u^*}$ in \cref{eq:gpme_app}, where $u^*= u(x^*(t), t)$ denotes the solution value at the shock $x^*(t)$. The parameter $u^*$ also has an effect on the shock position, with smaller values of $u^*$ resulting in a faster shock speed. \cref{fig:solutions_stefan_id} shows that the methods are relatively accurate on the in-domain task with \ensemblenomethod and \method exactly capturing the shock with tight uncertainty bounds. The small Gibbs phenomenon at the shock position that occurs with \ensemblenomethod is damped with \method. \outputvarmethod has large uncertainty around the diffused shock. We see that even for small OOD shifts in \cref{fig:solutions_stefan_ood_small} on this ``hard'' shock problem, several of the baseline methods suffer numerical artifacts of spurious oscillations, being overly diffusive and having an incorrect shock speed, where the predicted shock position either lags or is ahead of the true shock position, which worsens as the OOD shift increases in Figures \ref{fig:solutions_stefan_ood_medium}-\ref{fig:solutions_stefan_ood3}. See corresponding metric results in \cref{tab:results_stefan_app}.

\begin{table}[h]
    \centering
    \caption{{\bf 1-d Stefan equation.} MSE $\downarrow$, NLL $\downarrow$, n-MeRCI $\downarrow$, RMSCE $\downarrow$ and CRPS $\downarrow$ (mean and standard deviation over 5 seeds) metrics for different UQ methods on the 1-d Stefan equation in-domain and with small, medium and large OOD shifts, where $u{^*}^{\tr} \in [0.6, 0.65]$. {\bf Bold} indicates values within one standard deviation of the best mean.
    }
    \label{tab:results_stefan_app}
    \resizebox{\textwidth}{!}{
    \begin{tabular}{@{}lccccc@{}}
    \toprule
    & \multicolumn{5}{c}{{\bf In-domain}, $u{^*}^{\te}\in [0.6, 0.65]$} \\
     & MSE $\downarrow$ & NLL $\downarrow$ & n-MeRCI $\downarrow$ & RMSCE $\downarrow$ & CRPS $\downarrow$ \\
    \midrule
    \bayesiannomethod & \bf 3.9e-04 (9.6e-05) & -4.8e+03 (2.6e+02) & \bf   0.32 ( 0.11) &   0.23 ( 0.00) & 9.3e-03 (1.1e-03) \\
    \outputvarmethod & 8.1e-03 (4.7e-04) & \bf -1.1e+04 (5.2e+02) &   0.75 ( 0.07) &   0.17 ( 0.02) & 1.8e-02 (7.4e-04) \\
    \mcdropoutnomethod & 5.8e-04 (1.6e-04) & -2.5e+03 (6.6e+02) & \bf   0.33 ( 0.13) &   0.17 ( 0.01) & 8.1e-03 (1.1e-03) \\
    \ensemblenomethod & \bf 3.6e-04 (2.9e-05) & -7.4e+03 (8.5e+02) & \bf   0.41 ( 0.23) & \bf   0.12 ( 0.01) & \bf 2.7e-03 (1.2e-04) \\
    \method & \bf 3.7e-04 (5.4e-05) & 8.7e+03 (5.9e+03) & \bf   0.41 ( 0.26) &   0.14 ( 0.01) & 3.3e-03 (4.8e-04) \\
    \end{tabular}
    }
    \resizebox{\textwidth}{!}{
    \begin{tabular}{@{}lccccc@{}}
    \midrule
    & \multicolumn{5}{c}{{\bf Small OOD shift}, $u{^*}^{\te}\in [0.55, 0.6]$} \\
     & MSE $\downarrow$ & NLL $\downarrow$ & n-MeRCI $\downarrow$ & RMSCE $\downarrow$ & CRPS $\downarrow$ \\
    \midrule
    \bayesiannomethod & 2.0e-02 (1.9e-02) & 1.5e+03 (1.3e+03) &   0.67 ( 0.15) & \bf   0.28 ( 0.01) & 4.2e-02 (1.8e-02) \\
    \outputvarmethod & 2.3e-02 (1.6e-03) & 7.0e+06 (3.5e+06) &   0.97 ( 0.07) &   0.40 ( 0.02) & 4.3e-02 (1.7e-03) \\
    \mcdropoutnomethod & \bf 9.6e-03 (3.6e-03) & 2.4e+04 (8.7e+03) &   0.78 ( 0.08) &   0.31 ( 0.01) & 4.1e-02 (4.4e-03) \\
    \ensemblenomethod & \bf 8.1e-03 (3.4e-03) & \bf -5.2e+03 (3.3e+02) & \bf   0.14 ( 0.09) &   0.32 ( 0.01) & \bf 2.5e-02 (3.6e-03) \\
    \method & 1.4e-02 (2.3e-03) & 1.2e+04 (5.5e+03) & \bf   0.14 ( 0.06) &   0.38 ( 0.01) & 3.7e-02 (3.3e-03) \\
    \end{tabular}
    }
    \resizebox{\textwidth}{!}{
    \begin{tabular}{@{}lccccc@{}}
    \midrule
    & \multicolumn{5}{c}{{\bf Medium OOD shift}, $u{^*}^{\te}\in [0.7, 0.75]$} \\
     & MSE $\downarrow$ & NLL $\downarrow$ & n-MeRCI $\downarrow$ & RMSCE $\downarrow$ & CRPS $\downarrow$ \\
    \midrule
    \bayesiannomethod & 1.7e-02 (1.4e-02) & 6.0e+03 (3.6e+03) &   0.66 ( 0.10) & \bf   0.29 ( 0.02) & 4.6e-02 (7.6e-03) \\
    \outputvarmethod & 3.2e-02 (1.3e-03) & 1.9e+07 (8.4e+06) &   0.83 ( 0.05) &   0.40 ( 0.02) & 5.2e-02 (1.4e-03) \\
    \mcdropoutnomethod & 2.9e-02 (1.3e-02) & 2.2e+04 (9.8e+03) &   0.56 ( 0.26) &   0.36 ( 0.03) & 6.4e-02 (1.0e-02) \\
    \ensemblenomethod & \bf 8.0e-03 (1.4e-03) & \bf -4.3e+03 (1.5e+02) & \bf   0.07 ( 0.03) &   0.33 ( 0.02) & \bf 3.3e-02 (3.3e-03) \\
    \method & 1.1e-02 (3.6e-03) & 2.0e+04 (3.8e+03) &   0.14 ( 0.03) &   0.37 ( 0.04) & 4.1e-02 (6.8e-03) \\
    \end{tabular}
    }
    \resizebox{\textwidth}{!}{
    \begin{tabular}{@{}lccccc@{}}
    \midrule
    & \multicolumn{5}{c}{{\bf Large OOD shift}, $u{^*}^{\te}\in [0.5, 0.55]$} \\
     & MSE $\downarrow$ & NLL $\downarrow$ & n-MeRCI $\downarrow$ & RMSCE $\downarrow$ & CRPS $\downarrow$ \\
    \midrule
    \bayesiannomethod & 1.7e-01 (1.7e-01) & 1.9e+04 (5.3e+03) &   0.50 ( 0.25) &   0.40 ( 0.03) & 1.5e-01 (7.2e-02) \\
    \outputvarmethod & \bf 3.4e-02 (1.8e-03) & 2.7e+07 (1.4e+07) &   0.99 ( 0.09) &   0.45 ( 0.01) & \bf 6.6e-02 (1.7e-03) \\
    \mcdropoutnomethod & 4.4e-02 (3.1e-02) & 7.3e+04 (1.2e+04) &   0.54 ( 0.11) &   0.42 ( 0.01) & 1.1e-01 (1.5e-02) \\
    \ensemblenomethod & 4.6e-02 (1.9e-02) & \bf -2.2e+03 (3.9e+02) &   0.37 ( 0.14) & \bf   0.36 ( 0.01) & 8.2e-02 (1.0e-02) \\
    \method & 8.9e-02 (5.1e-02) & 2.1e+04 (6.8e+03) & \bf   0.24 ( 0.11) &   0.43 ( 0.01) & 1.3e-01 (2.9e-02) \\
    \bottomrule
    \end{tabular}
    }
\end{table}

\subsubsection{Hyperbolic Linear Advection Equation}
We show results for the following two tasks: constant parameter input by varying the speed $\beta$ and non-constant initial condition input by varying the initial shock location $a$.

\begin{figure}[H]
    \centering
    \hspace{-0.5in}
    \begin{subfigure}[t]{0.30\textwidth}
    \centering
    \includegraphics[scale=0.35]{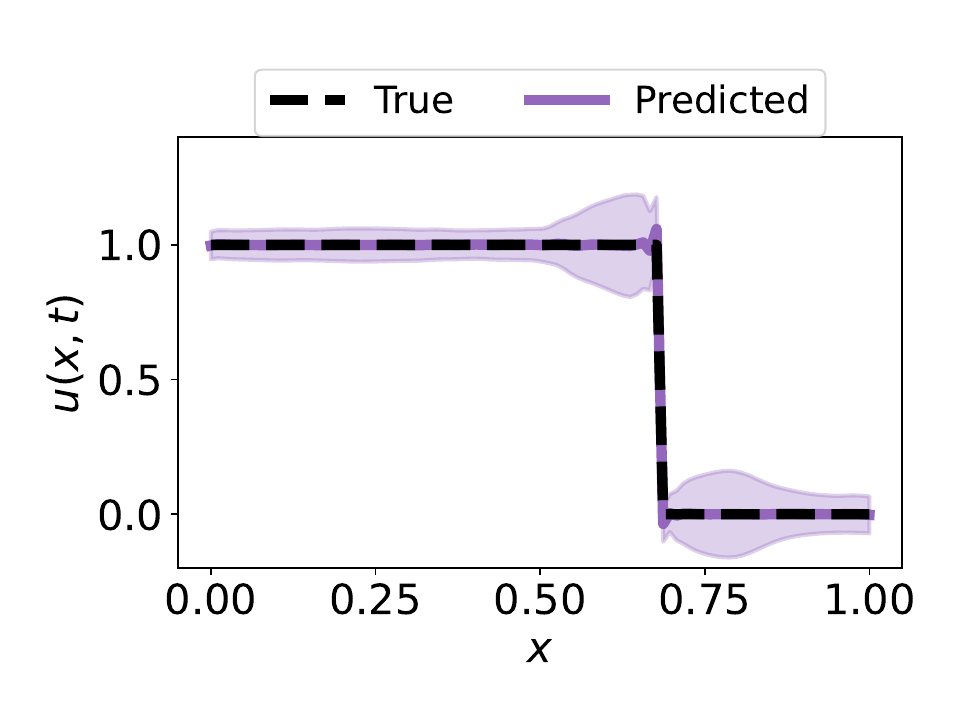}
    \caption{\bayesiannomethod}
    \end{subfigure}
    ~~
    \begin{subfigure}[t]{0.30\textwidth}
    \centering
    \includegraphics[scale=0.35]{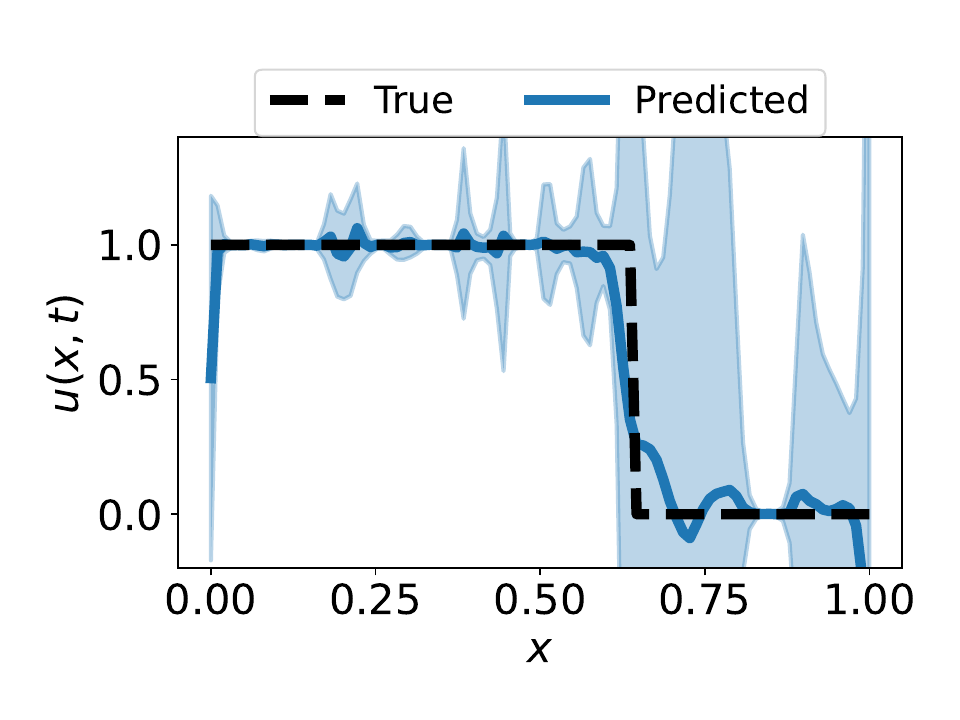}
    \caption{\outputvarmethod}
    \end{subfigure}
    ~~
    \begin{subfigure}[t]{0.30\textwidth}
    \centering
    \includegraphics[scale=0.35]{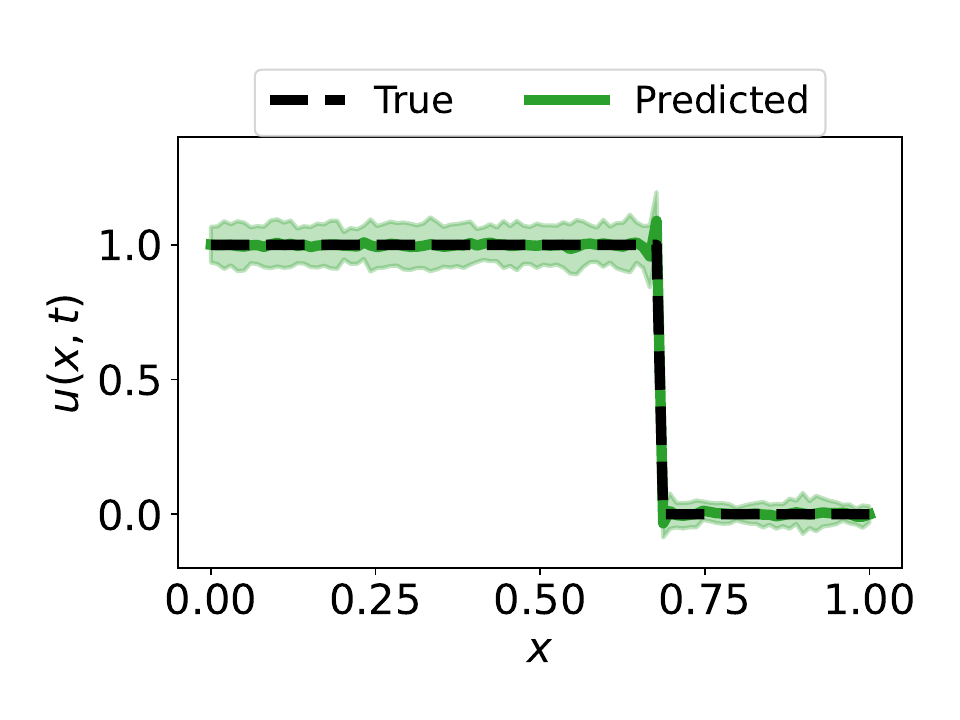}
    \caption{\mcdropoutnomethod}
    \end{subfigure}
    
    \begin{subfigure}[t]{0.30\textwidth}
    \centering
    \includegraphics[scale=0.35]{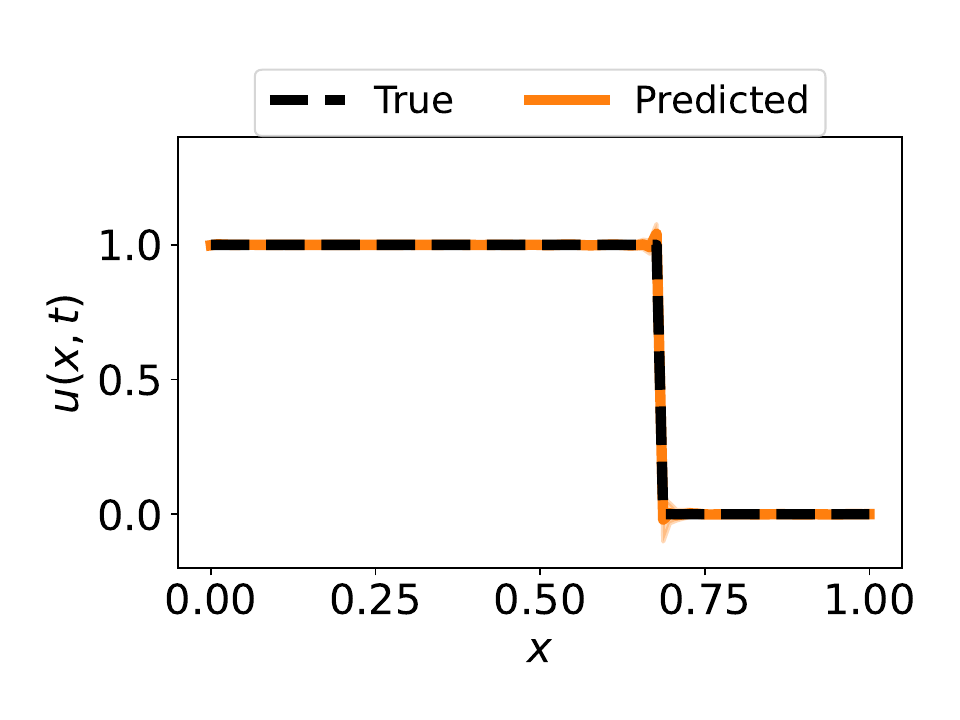}
    \caption{\ensemblenomethod}
    \end{subfigure}
    ~~~~
    \begin{subfigure}[t]{0.30\textwidth}
    \centering
    \includegraphics[scale=0.35]{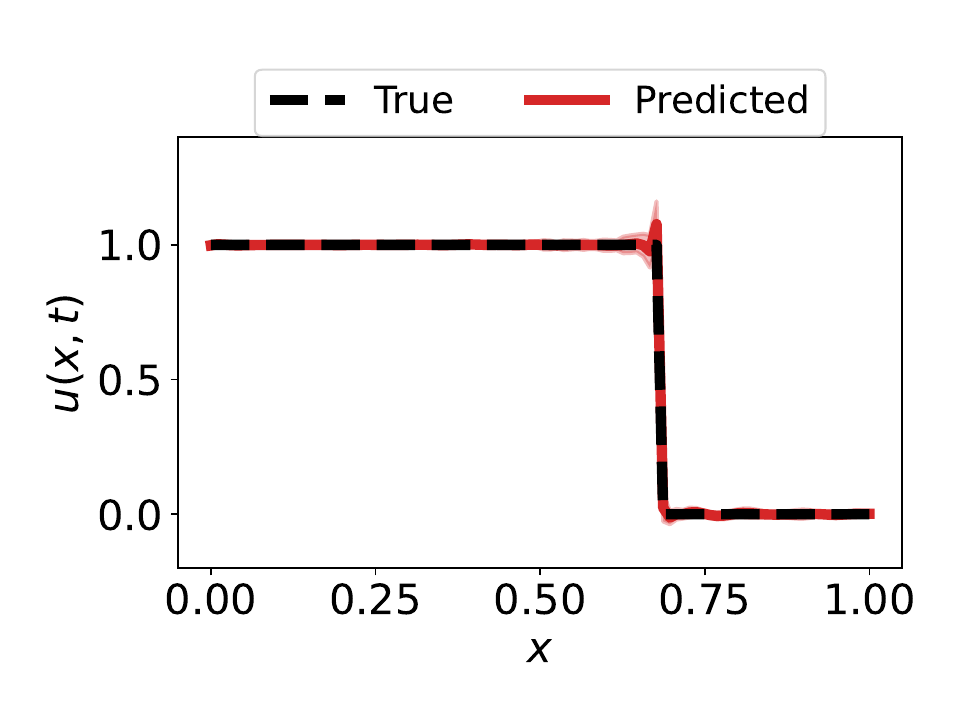}
    \caption{\method}
    \end{subfigure}
    \caption{{\bf 1-d Linear advection, in-domain}, $\beta^\tr, \beta^\te \in [1, 2]$. 
    Uncertainty estimates from different UQ methods for in-domain values of the input velocity $\beta$ coefficient.
    } 
    \label{fig:solutions_la_id}
\end{figure}

\begin{figure}[H]
    \centering
    \hspace{-0.5in}
    \begin{subfigure}[t]{0.30\textwidth}
    \centering
    \includegraphics[scale=0.35]{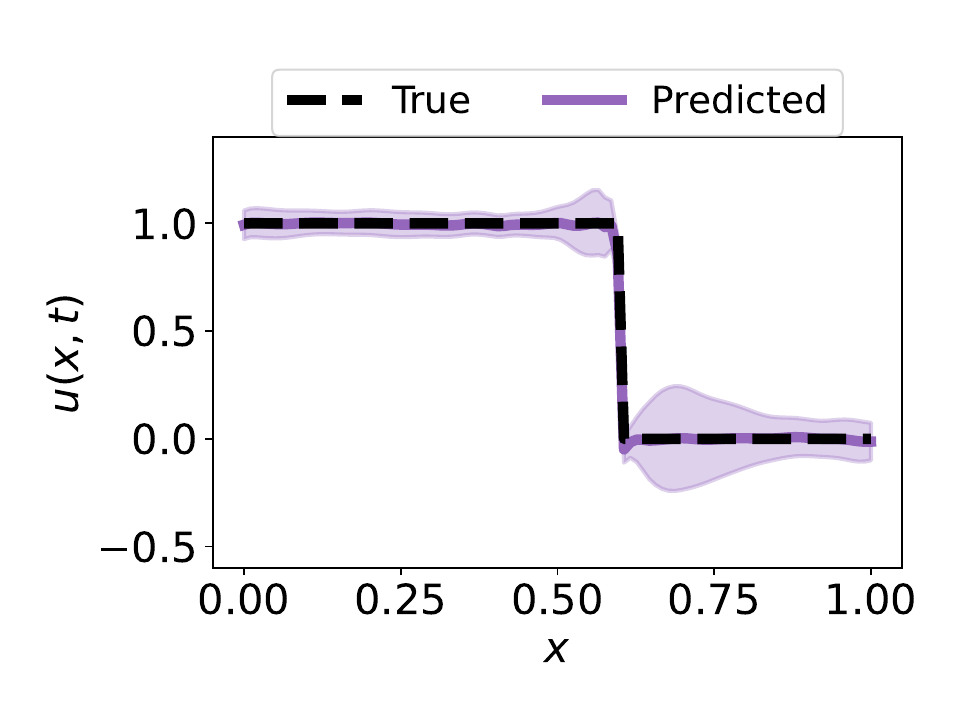}
    \caption{\bayesiannomethod}
    \end{subfigure}
    ~~
    \begin{subfigure}[t]{0.30\textwidth}
    \centering
    \includegraphics[scale=0.35]{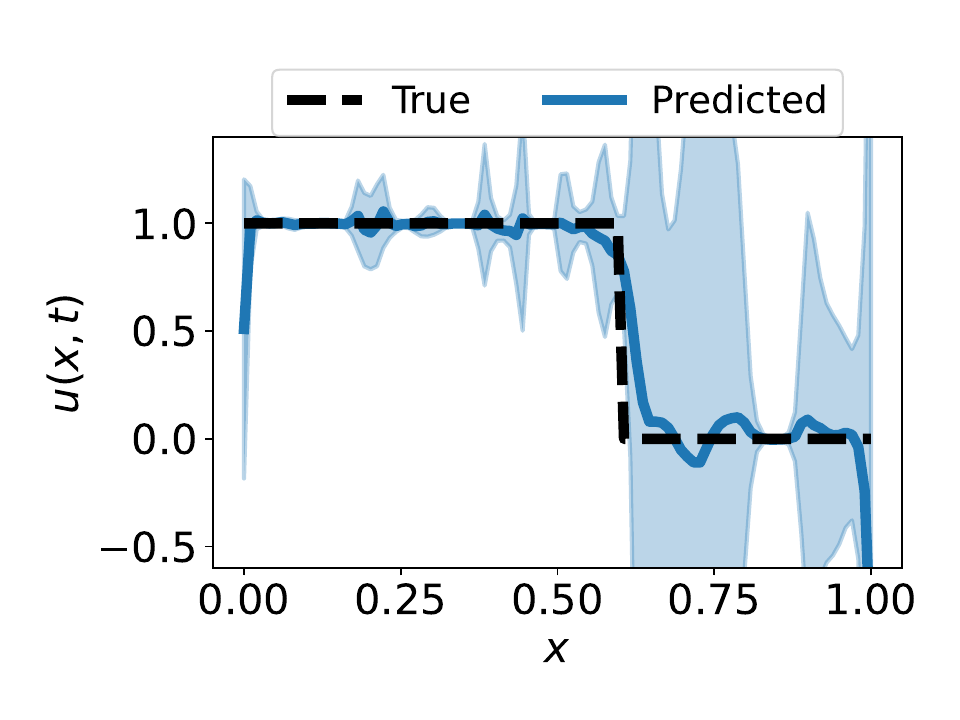}
    \caption{\outputvarmethod}
    \end{subfigure}
    ~~
    \begin{subfigure}[t]{0.30\textwidth}
    \centering
    \includegraphics[scale=0.35]{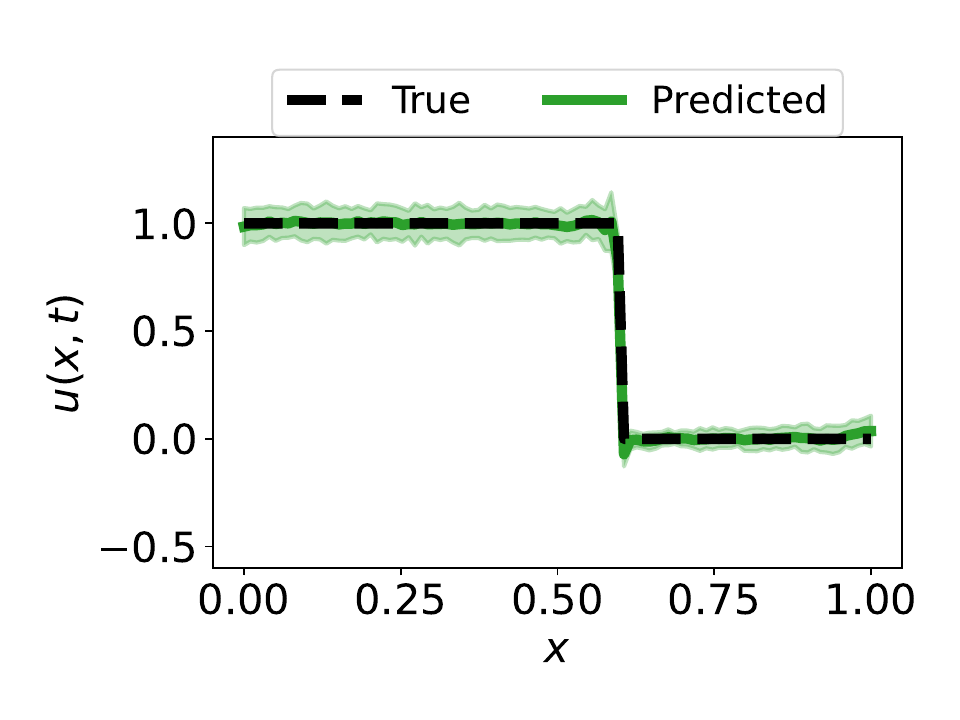}
    \caption{\mcdropoutnomethod}
    \end{subfigure}
    
    \begin{subfigure}[t]{0.30\textwidth}
    \centering
    \includegraphics[scale=0.35]{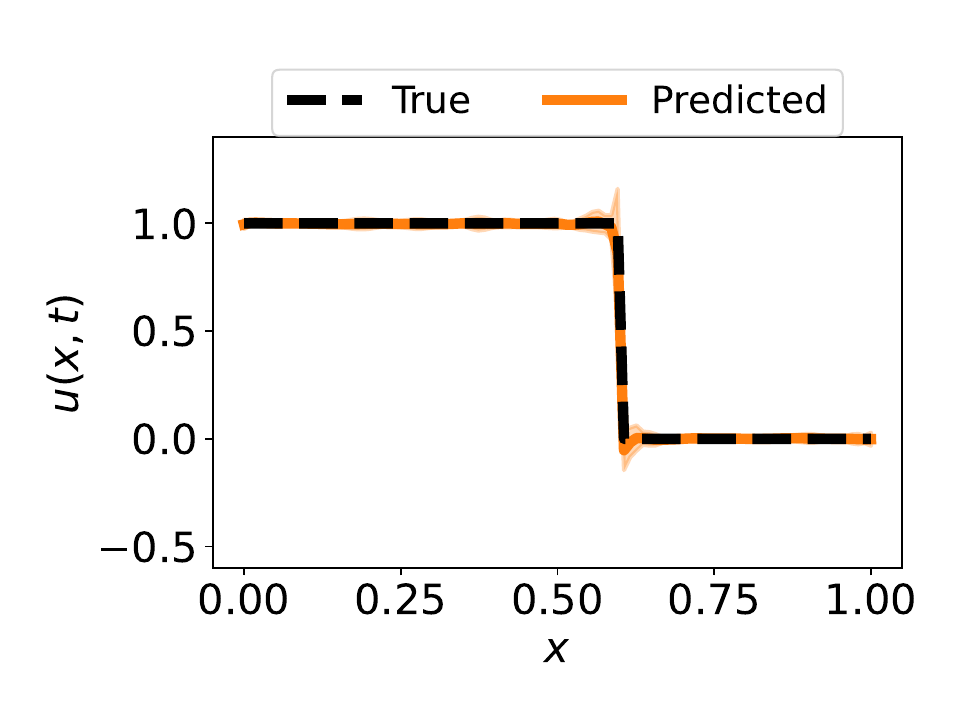}
    \caption{\ensemblenomethod}
    \end{subfigure}
    ~~~~
    \begin{subfigure}[t]{0.30\textwidth}
    \centering
    \includegraphics[scale=0.35]{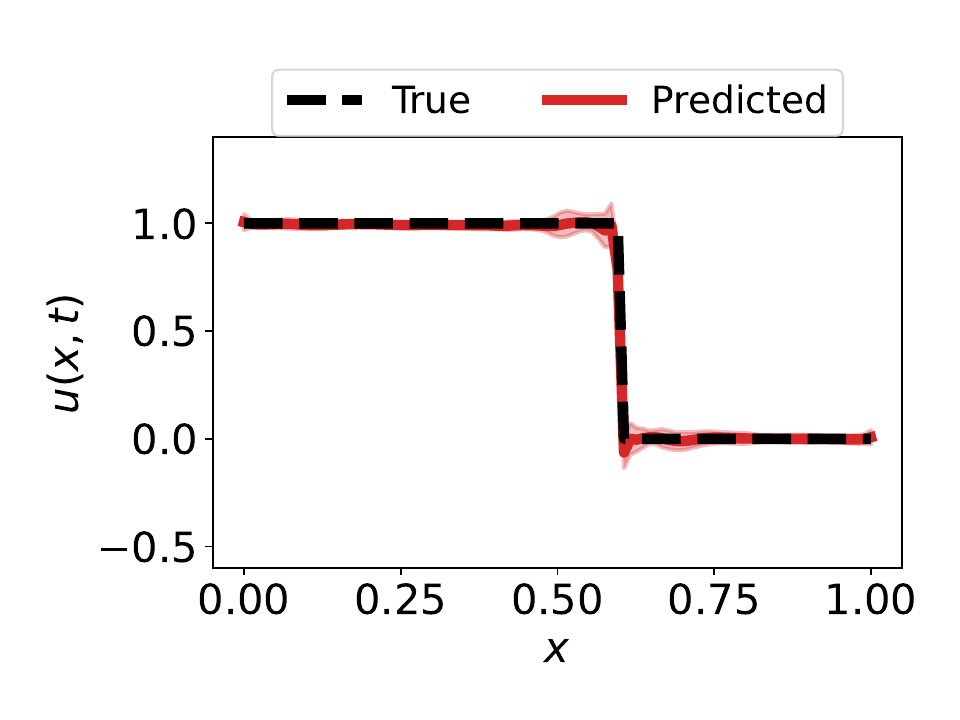}
    \caption{\method}
    \end{subfigure}
    \caption{{\bf 1-d Linear advection, small OOD shift}, $\beta^\tr \in [1, 2], \beta^\te \in [0.5, 1]$. 
    Uncertainty estimates from different UQ methods under small OOD shifts in the input velocity $\beta$ coefficient. 
    } 
    \label{fig:solutions_la_ood1_uqr}
\end{figure}

\begin{figure}[H]
    \centering
    \hspace{-0.5in}
    \begin{subfigure}[t]{0.30\textwidth}
    \centering
    \includegraphics[scale=0.35]{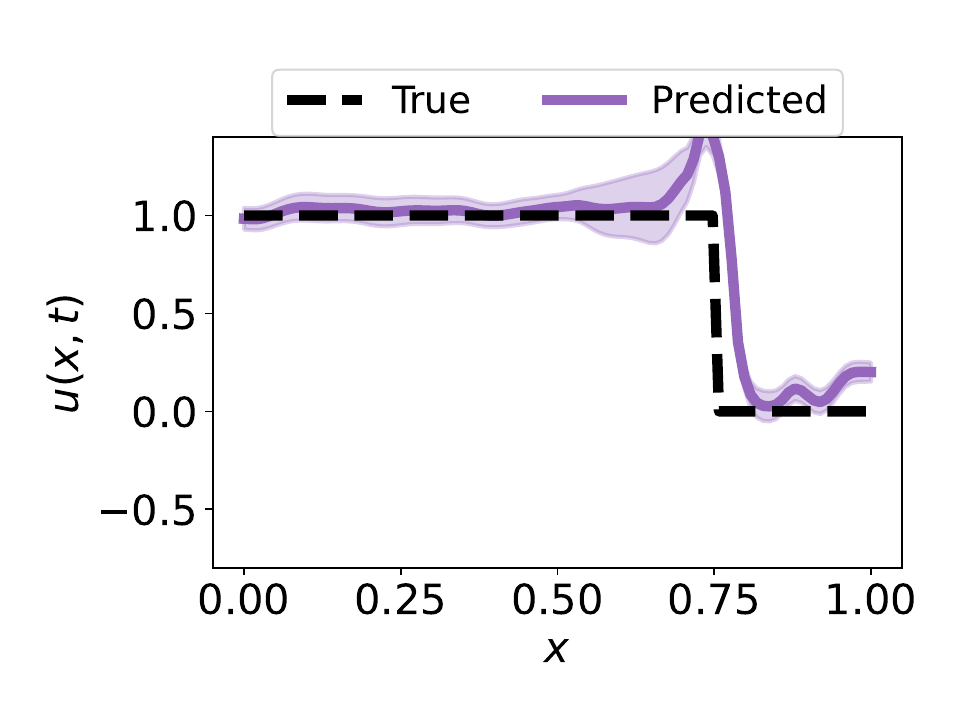}
    \caption{\bayesiannomethod}
    \end{subfigure}
    ~~
    \begin{subfigure}[t]{0.30\textwidth}
    \centering
    \includegraphics[scale=0.35]{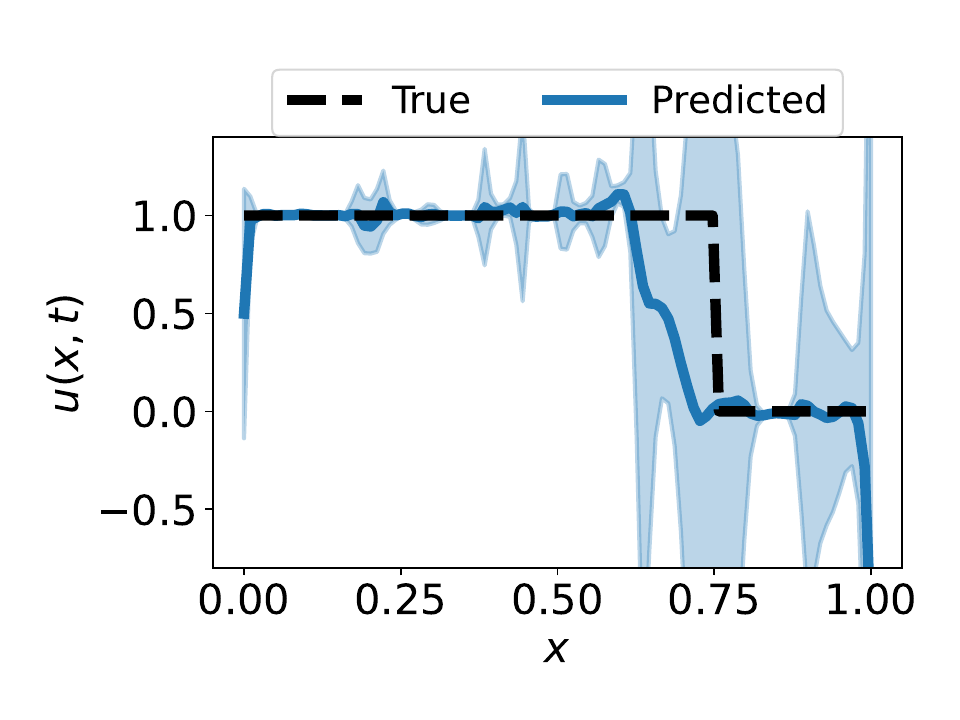}
    \caption{\outputvarmethod}
    \end{subfigure}
    ~~
    \begin{subfigure}[t]{0.30\textwidth}
    \centering
    \includegraphics[scale=0.35]{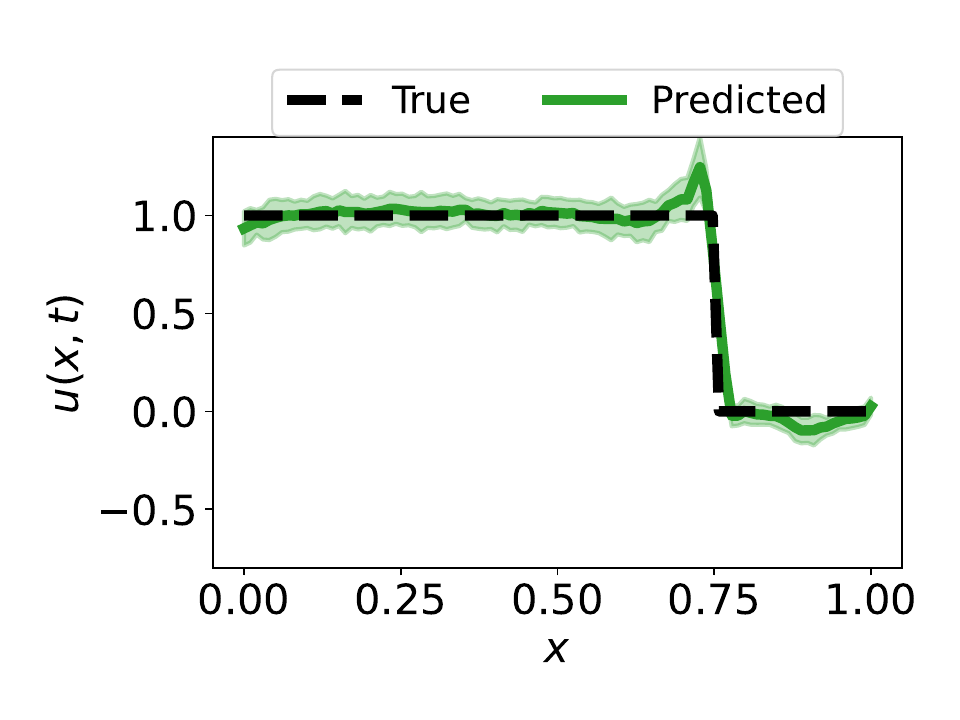}
    \caption{\mcdropoutnomethod}
    \end{subfigure}
    
    \begin{subfigure}[t]{0.30\textwidth}
    \centering
    \includegraphics[scale=0.35]{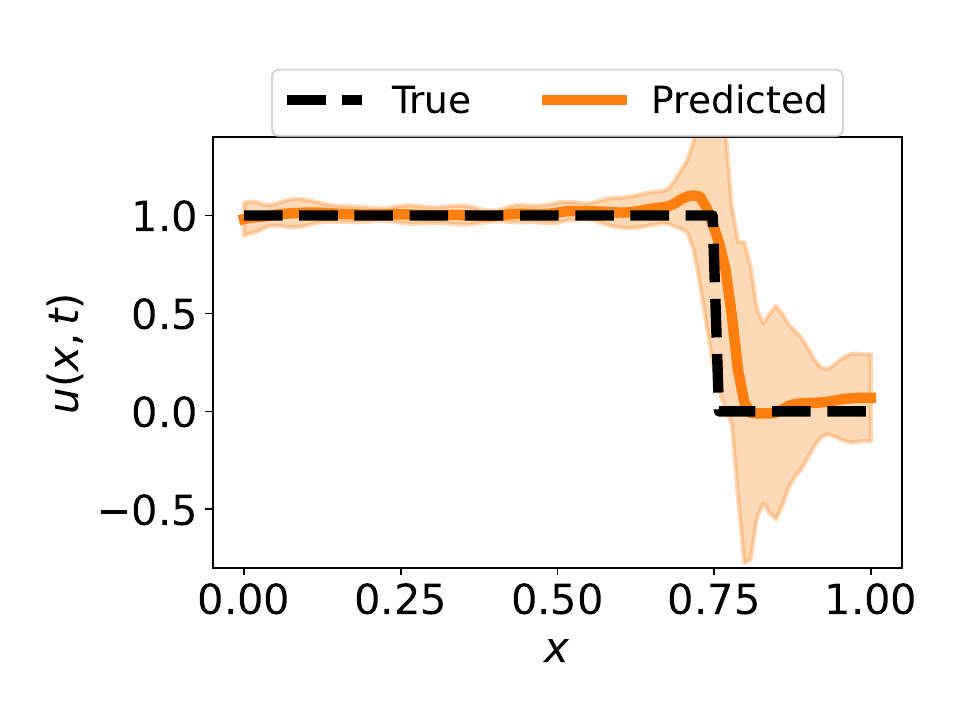}
    \caption{\ensemblenomethod}
    \end{subfigure}
    ~~~~
    \begin{subfigure}[t]{0.30\textwidth}
    \centering
    \includegraphics[scale=0.35]{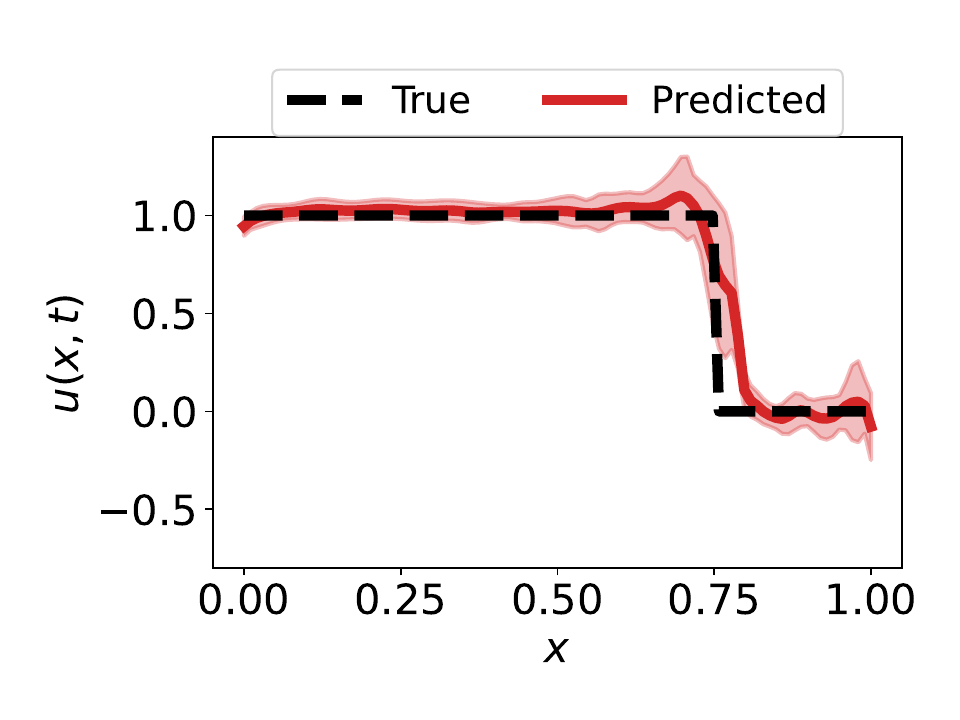}
    \caption{\method}
    \end{subfigure}
    \caption{{\bf 1-d Linear advection, medium OOD shift}, $\beta^\tr \in [1, 2], \beta^\te \in [2.5, 3]$. 
    Uncertainty estimates from different UQ methods under medium OOD shifts in the input velocity $\beta$ coefficient. 
    } 
    \label{fig:solutions_la_ood2_uqr}
\end{figure}

\begin{figure}[H]
    \centering
    \hspace{-0.5in}
    \begin{subfigure}[t]{0.30\textwidth}
    \centering
    \includegraphics[scale=0.35]{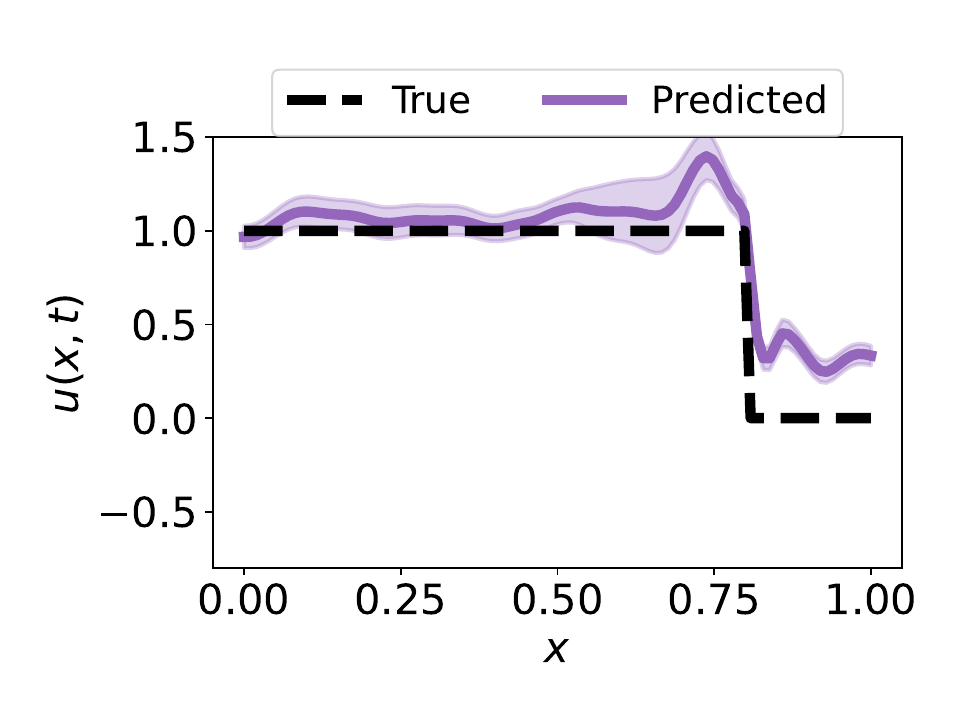}
    \caption{\bayesiannomethod}
    \end{subfigure}
    ~~
    \begin{subfigure}[t]{0.30\textwidth}
    \centering
    \includegraphics[scale=0.35]{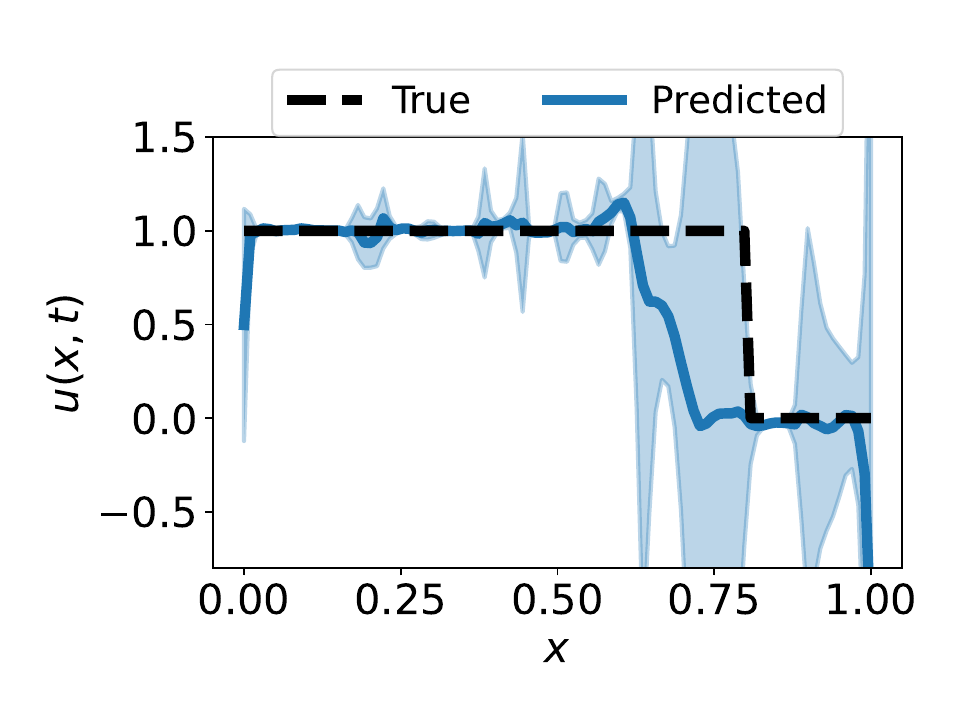}
    \caption{\outputvarmethod}
    \end{subfigure}
    ~~
    \begin{subfigure}[t]{0.30\textwidth}
    \centering
    \includegraphics[scale=0.35]{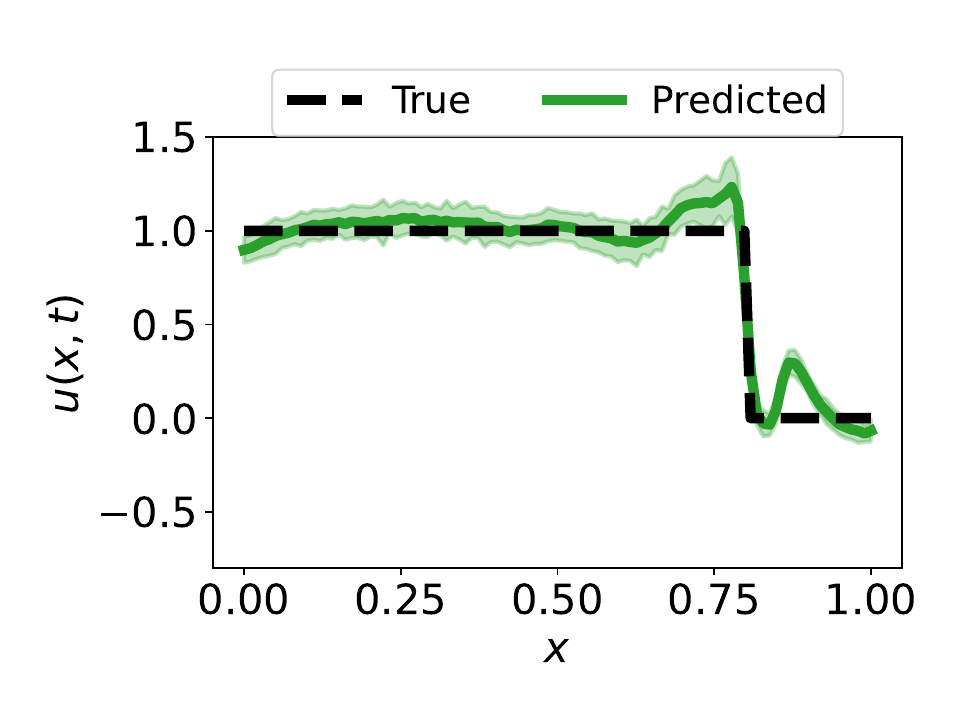}
    \caption{\mcdropoutnomethod}
    \end{subfigure}
    
    \begin{subfigure}[t]{0.30\textwidth}
    \centering
    \includegraphics[scale=0.35]{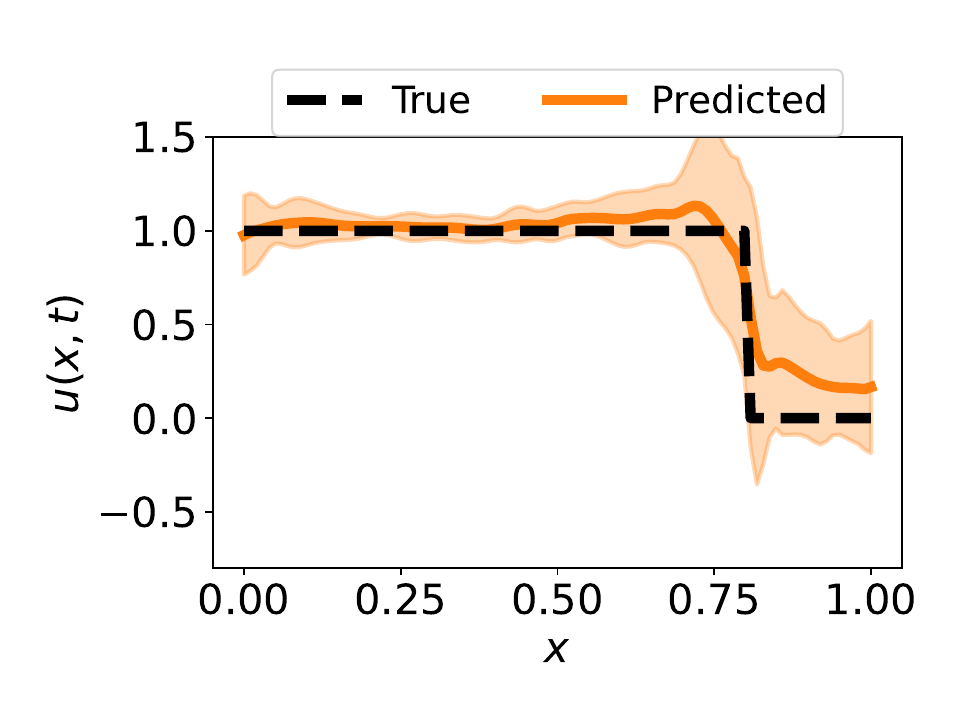}
    \caption{\ensemblenomethod}
    \end{subfigure}
    ~~~~
    \begin{subfigure}[t]{0.30\textwidth}
    \centering
    \includegraphics[scale=0.35]{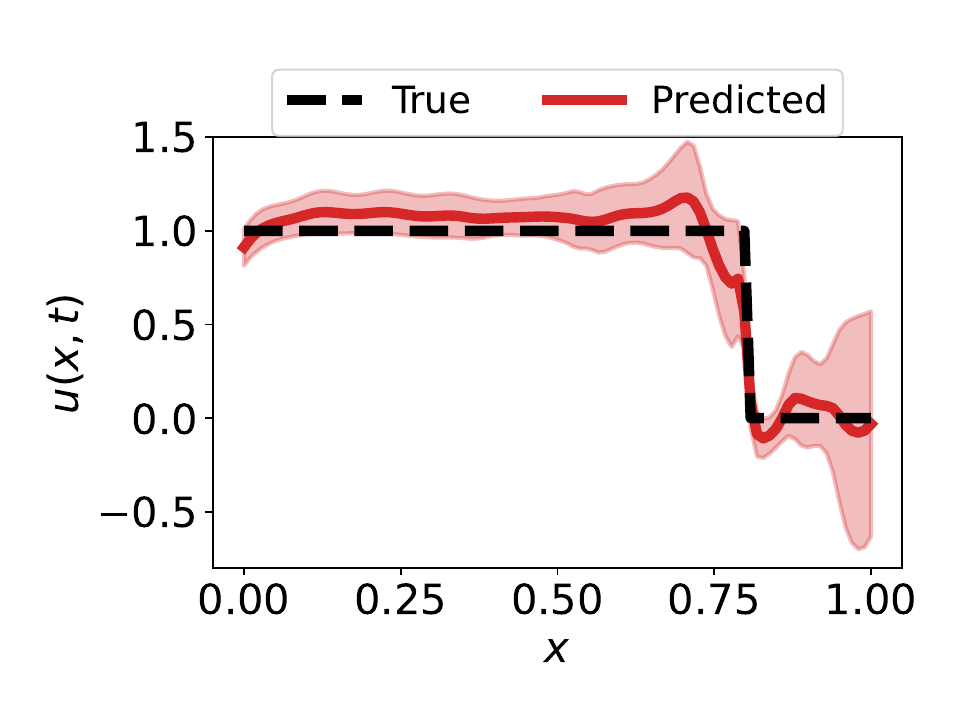}
    \caption{\method}
    \end{subfigure}
    \caption{{\bf 1-d Linear advection, large OOD shift}, $\beta^\tr \in [1, 2], \beta^\te \in [3, 3.5]$. 
    Uncertainty estimates from different UQ methods under large OOD shifts in the input velocity $\beta$ coefficient.} 
    \label{fig:solutions_la_ood3_uqr}
\end{figure}

\paragraph{Constant parameter to solution mapping.} Figures \ref{fig:solutions_la_id}-\ref{fig:solutions_la_ood3_uqr} show the solution profiles for the hyperbolic linear advection equation conditions for an in-domain task, and for small, medium and large OOD shifts, respectively, of the velocity parameter $\beta > 0$. The solution is a rightward moving shock, where $\beta$ controls the shock speed.  We see that in this difficult shock case the baseline methods, e.g., \bayesiannomethod, \outputvarmethod and \mcdropoutnomethod suffer from numerical artifacts of artificial oscillations, being over-diffusive and lagging of the shock position.   Only \ensemblenomethod and our \method accurately capture the shock location and have the largest uncertainty there. See corresponding metric results in \cref{tab:results_la}.

\begin{table}[h]
    \centering
    \caption{{\bf 1-d Linear Advection} MSE $\downarrow$, NLL $\downarrow$, n-MeRCI $\downarrow$, RMSCE $\downarrow$ and CRPS $\downarrow$ (mean and standard deviation over 5 seeds) metrics for different UQ methods on the 1-d linear advection equation in-domain and with small, medium and large OOD shifts, where $\beta^\tr \in [1,2]$. {\bf Bold} indicates values within one standard deviation of the best mean.
    }
    \label{tab:results_la}
    \resizebox{\textwidth}{!}{
    \begin{tabular}{@{}lccccc@{}}
    \toprule
    & \multicolumn{5}{c}{{\bf In-domain}, $\beta^\te \in [1,2]$} \\
     & MSE $\downarrow$ & NLL $\downarrow$ & n-MeRCI $\downarrow$ & RMSCE $\downarrow$ & CRPS $\downarrow$ \\
    \midrule
    \bayesiannomethod & \bf 2.0e-04 (3.6e-05) & -5.7e+03 (1.6e+02) &   0.59 ( 0.08) &   0.23 ( 0.00) & 6.2e-03 (6.2e-04) \\
    \outputvarmethod & 2.2e-02 (3.4e-02) & \bf -1.1e+04 (2.6e+03) &   0.41 ( 0.18) &   0.20 ( 0.03) & 1.8e-02 (1.6e-02) \\
    \mcdropoutnomethod & 2.8e-04 (5.0e-05) & -5.0e+03 (2.3e+02) &   0.67 ( 0.15) &   0.20 ( 0.00) & 7.7e-03 (5.1e-04) \\
    \ensemblenomethod & \bf 2.0e-04 (1.1e-05) & \bf -1.1e+04 (1.8e+02) &   0.40 ( 0.06) & \bf   0.12 ( 0.00) & \bf 1.2e-03 (5.8e-05) \\
    \method & \bf 2.0e-04 (2.6e-05) & -7.2e+03 (9.6e+02) & \bf   0.28 ( 0.08) & \bf   0.12 ( 0.01) & 1.7e-03 (1.1e-04) \\
    \end{tabular}
    }
    \resizebox{\textwidth}{!}{
    \begin{tabular}{@{}lccccc@{}}
    \midrule
    & \multicolumn{5}{c}{{\bf Out-of-domain}, $\beta^\te \in [0.5,1]$} \\
     & MSE $\downarrow$ & NLL $\downarrow$ & n-MeRCI $\downarrow$ & RMSCE $\downarrow$ & CRPS $\downarrow$ \\
    \midrule
    \bayesiannomethod & 6.8e-02 (1.6e-02) & 1.7e+05 (9.3e+04) &   0.77 ( 0.05) & \bf   0.27 ( 0.03) & 1.1e-01 (2.7e-02) \\
    \outputvarmethod & 7.3e-02 (2.8e-02) & 9.4e+07 (6.0e+07) &   0.46 ( 0.20) &   0.40 ( 0.03) & 7.9e-02 (1.2e-02) \\
    \mcdropoutnomethod & \bf 3.0e-02 (9.1e-03) & 5.9e+04 (2.3e+04) &   0.93 ( 0.03) & \bf   0.26 ( 0.03) & 7.3e-02 (1.8e-02) \\
    \ensemblenomethod & \bf 3.0e-02 (3.8e-03) & \bf -3.5e+03 (3.6e+02) & \bf   0.17 ( 0.03) & \bf   0.28 ( 0.02) & \bf 6.0e-02 (5.1e-03) \\
    \method & 4.3e-02 (3.4e-02) & 4.6e+04 (3.2e+04) & \bf   0.16 ( 0.06) &   0.35 ( 0.03) & 7.1e-02 (3.2e-02) \\
    \end{tabular}
    }
    \resizebox{\textwidth}{!}{
    \begin{tabular}{@{}lccccc@{}}
    \midrule
    & \multicolumn{5}{c}{{\bf Out-of-domain}, $\beta^\te \in [2.5,3]$} \\
     & MSE $\downarrow$ & NLL $\downarrow$ & n-MeRCI $\downarrow$ & RMSCE $\downarrow$ & CRPS $\downarrow$ \\
    \midrule
    \bayesiannomethod & 6.4e-03 (1.7e-03) & 4.6e+03 (3.7e+03) &   0.76 ( 0.05) &   0.41 ( 0.02) & 3.7e-02 (4.1e-03) \\
    \outputvarmethod & 4.4e-02 (4.0e-02) & 1.4e+07 (1.7e+07) &   0.42 ( 0.30) &   0.45 ( 0.03) & 3.9e-02 (1.6e-02) \\
    \mcdropoutnomethod & \bf 3.8e-03 (6.8e-04) & 2.9e+03 (2.9e+03) &   0.80 ( 0.05) & \bf   0.30 ( 0.01) & \bf 2.3e-02 (1.6e-03) \\
    \ensemblenomethod & \bf 4.3e-03 (4.2e-04) & \bf -2.3e+03 (7.7e+02) &   0.45 ( 0.04) &   0.44 ( 0.01) & 3.1e-02 (2.7e-03) \\
    \method & 7.4e-03 (2.5e-03) & 1.1e+04 (1.0e+04) & \bf   0.25 ( 0.15) &   0.45 ( 0.03) & 4.1e-02 (2.8e-03) \\
    \end{tabular}
    }
    \resizebox{\textwidth}{!}{
    \begin{tabular}{@{}lccccc@{}}
    \midrule
    & \multicolumn{5}{c}{{\bf Out-of-domain}, $\beta^\te \in [3,3.5]$} \\
     & MSE $\downarrow$ & NLL $\downarrow$ & n-MeRCI $\downarrow$ & RMSCE $\downarrow$ & CRPS $\downarrow$ \\
    \midrule
    \bayesiannomethod & 1.5e-02 (3.4e-03) & 1.5e+04 (6.7e+03) &   0.71 ( 0.06) &   0.46 ( 0.01) & 7.4e-02 (8.5e-03) \\
    \outputvarmethod & 5.3e-02 (4.2e-02) & 1.8e+08 (3.0e+08) & \bf   0.38 ( 0.26) &   0.47 ( 0.02) & 5.1e-02 (1.6e-02) \\
    \mcdropoutnomethod & \bf 7.4e-03 (1.2e-03) & 7.8e+03 (3.6e+03) &   0.75 ( 0.04) & \bf   0.37 ( 0.02) & \bf 3.7e-02 (1.8e-03) \\
    \ensemblenomethod & 1.0e-02 (1.3e-03) & \bf 1.0e+03 (1.0e+03) & \bf   0.42 ( 0.10) &   0.47 ( 0.00) & 6.3e-02 (4.7e-03) \\
    \method & 1.8e-02 (3.7e-03) & 2.1e+04 (1.8e+04) & \bf   0.27 ( 0.17) &   0.47 ( 0.03) & 8.1e-02 (7.0e-03) \\
    \bottomrule
    \end{tabular}
    }
\end{table}

\paragraph{Non-constant initial condition to solution mapping.}

\begin{figure}[H]
    \centering
    \begin{subfigure}[t]{0.3\textwidth}
    \centering
    \includegraphics[scale=0.35]{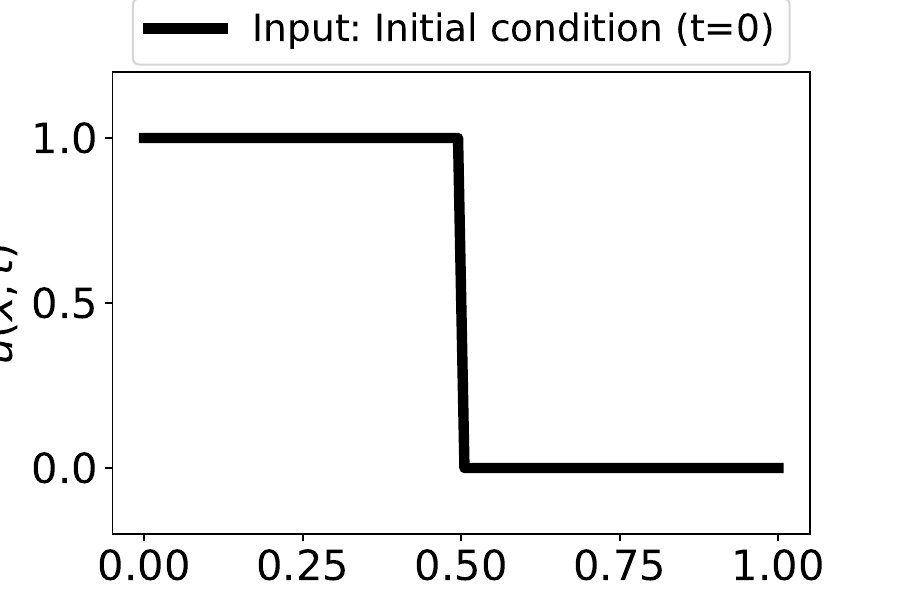}
    \caption{Input}
    \end{subfigure}
    ~~~~
    \begin{subfigure}[t]{0.3\textwidth}
    \centering
    \includegraphics[scale=0.35]{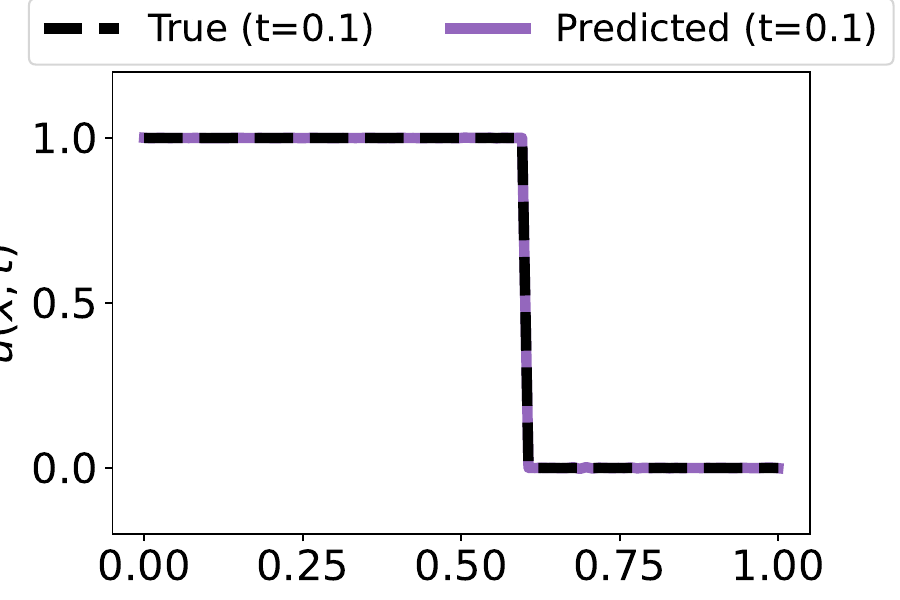}
    \caption{\bayesiannomethod}
    \end{subfigure}
    ~~~~
    \begin{subfigure}[t]{0.3\textwidth}
    \centering
    \includegraphics[scale=0.35]{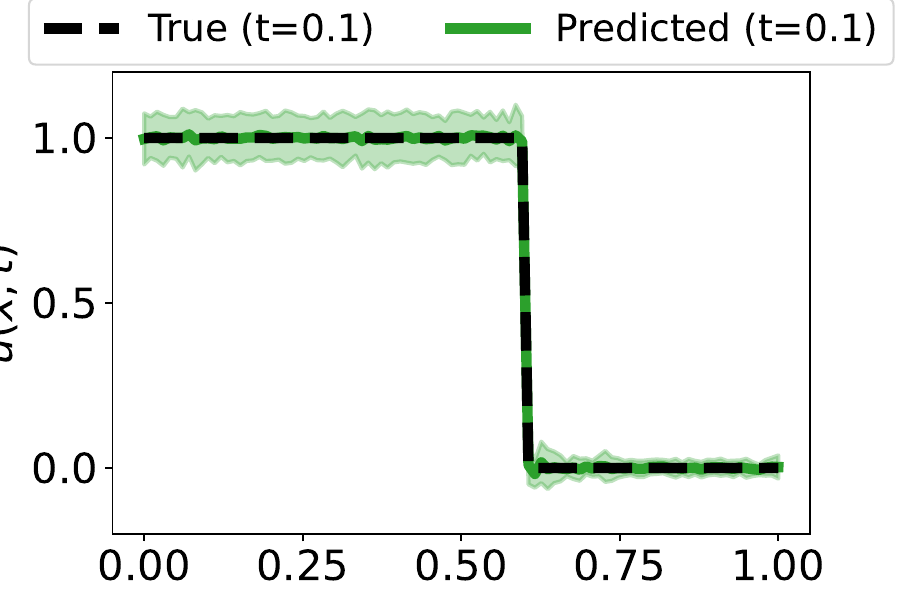}
    \caption{\mcdropoutnomethod}
    \end{subfigure}
    
    \begin{subfigure}[t]{0.3\textwidth}
    \centering
    \includegraphics[scale=0.35]{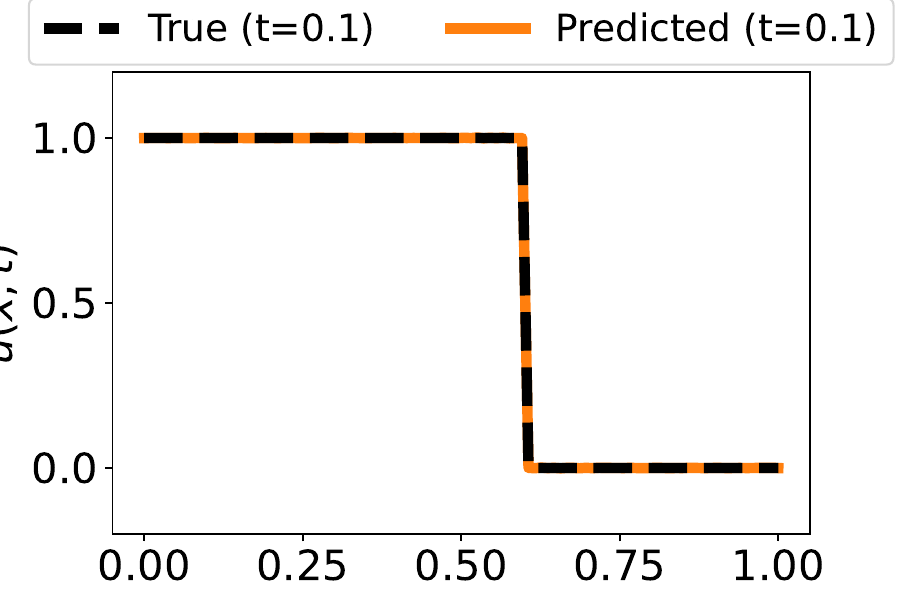}
    \caption{\ensemblenomethod}
    \end{subfigure}
    ~~~~
    \begin{subfigure}[t]{0.3\textwidth}
    \centering
    \includegraphics[scale=0.35]{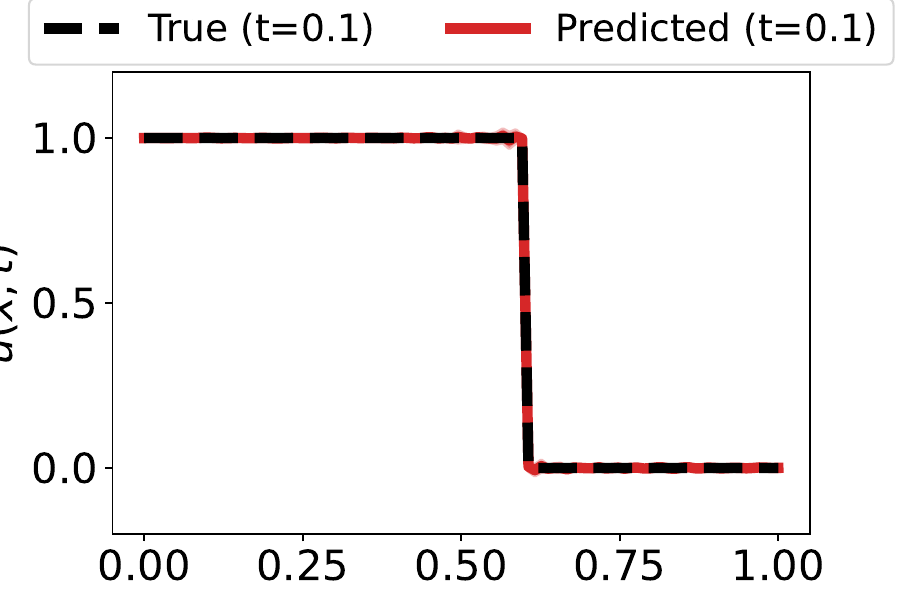}
    \caption{\method}
    \end{subfigure}
    \caption{{\bf 1-d Linear Advection (non-constant input), in-domain.} 
    Uncertainty estimates (3 standard deviations) from various UQ methods in-domain where $a^\tr \in [0.45, 0.55]$. 
    } 
    \label{fig:solutions_la_nonconstant_id}
\end{figure}

Here we test non-constant function input to the NO by varying the initial condition through the initial shock location $a$. We see that every method gives good predictions in-domain (\cref{fig:solutions_la_nonconstant_id}) and has low in-domain MSE (\cref{tab:1dla_nonconstant_mse}). \cref{tab:1dla_nonconstant_mse} also shows that \method performs best or second-best to \ensemblenomethod in MSE, while being computationally cheaper. With respect to the n-MeRCI metric, \cref{tab:1dla_nonconstant_nmerci} shows that \method performs around $1.2\times$ to $1.5\times$ better than \ensemblenomethod and $2.7\times$ to $7.4\times$ better than other baselines on OOD inputs.

\begin{figure}[H]
    \centering
    \begin{subfigure}{0.3\textwidth}
    \centering
    \includegraphics[scale=0.35]{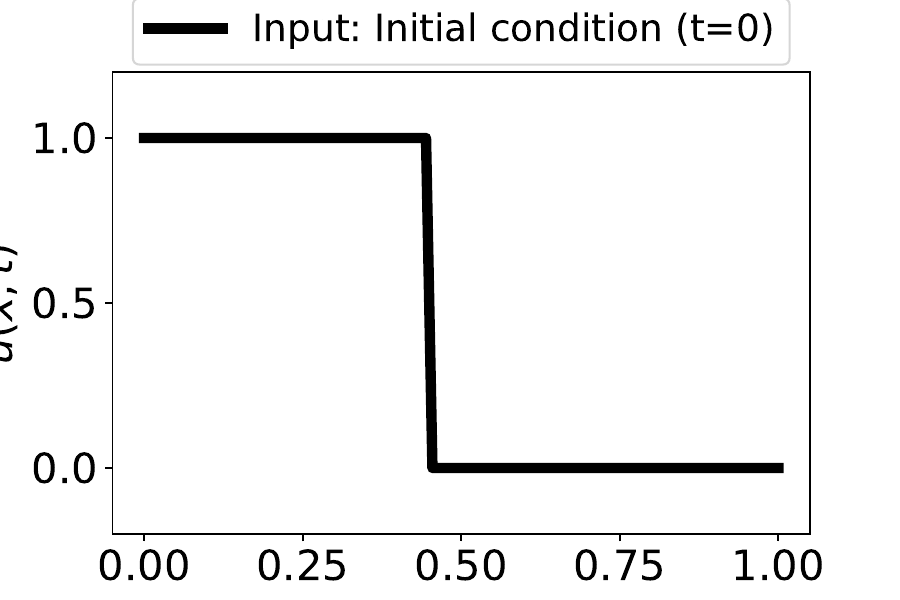}
    \caption{Input}
    \end{subfigure}
    ~~~~
    \begin{subfigure}{0.3\textwidth}
    \centering
    \includegraphics[scale=0.35]{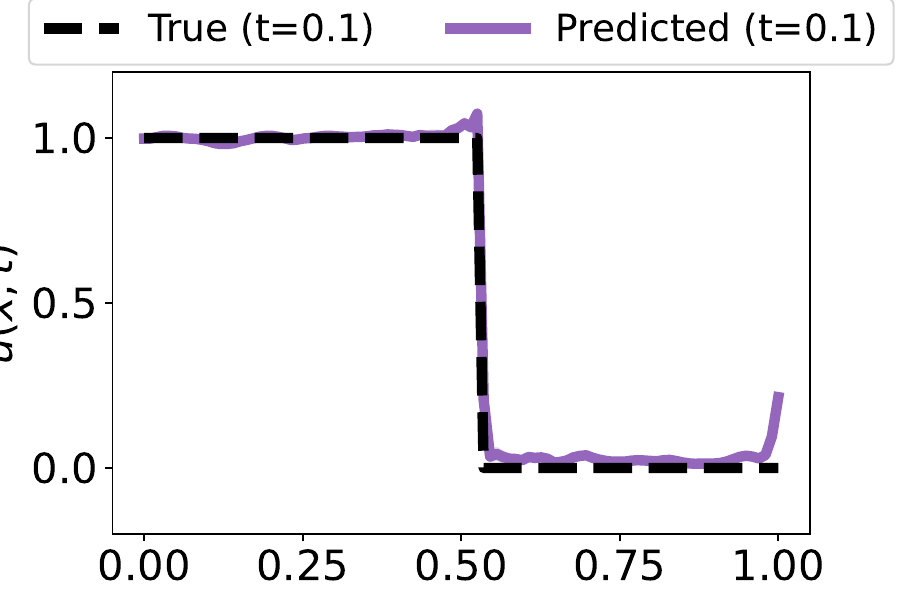}
    \caption{\bayesiannomethod}
    \end{subfigure}
    ~~~~
    \begin{subfigure}{0.3\textwidth}
    \centering
    \includegraphics[scale=0.35]{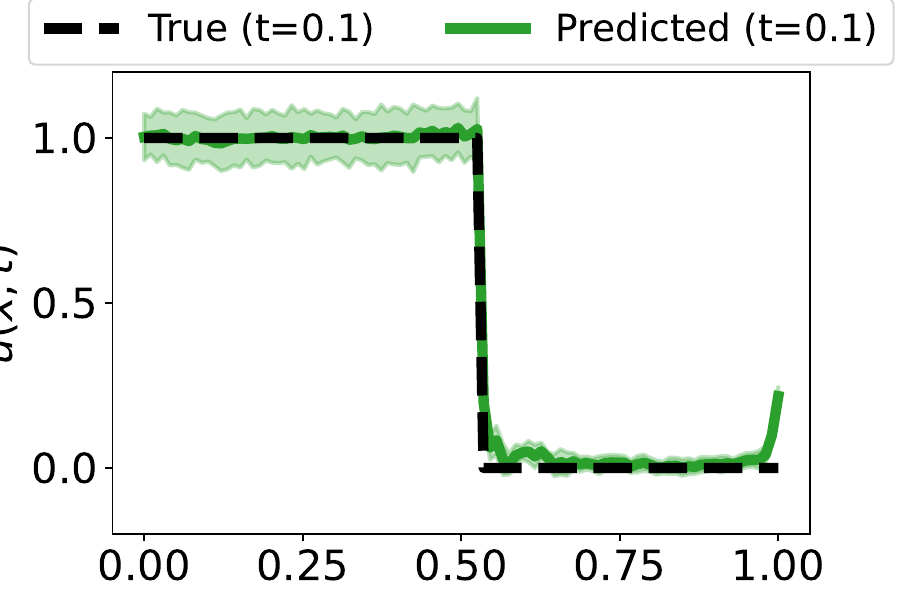}
    \caption{\mcdropoutnomethod}
    \end{subfigure}
    
    \begin{subfigure}{0.3\textwidth}
    \centering
    \includegraphics[scale=0.35]{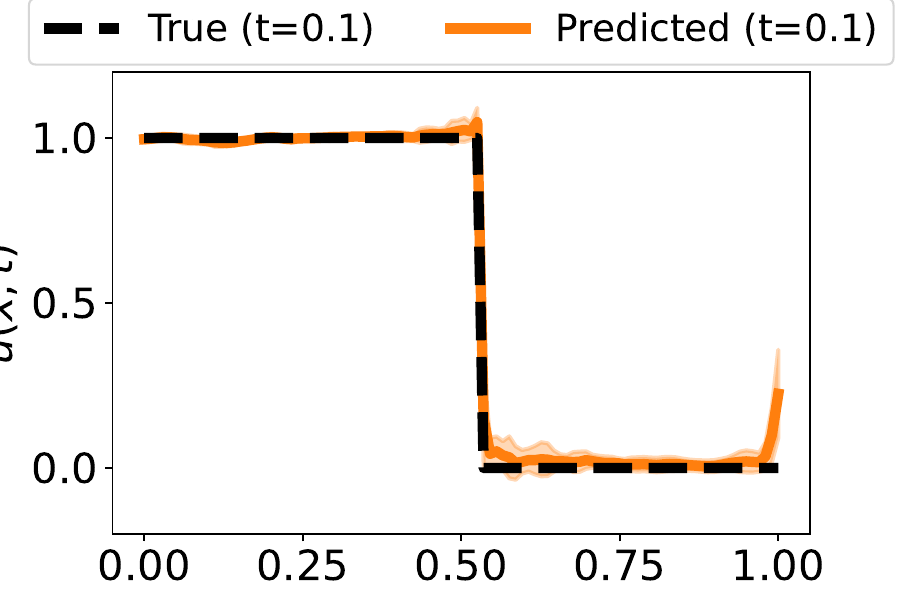}
    \caption{\ensemblenomethod}
    \end{subfigure}
    ~~~~
    \begin{subfigure}{0.3\textwidth}
    \centering
    \includegraphics[scale=0.35]{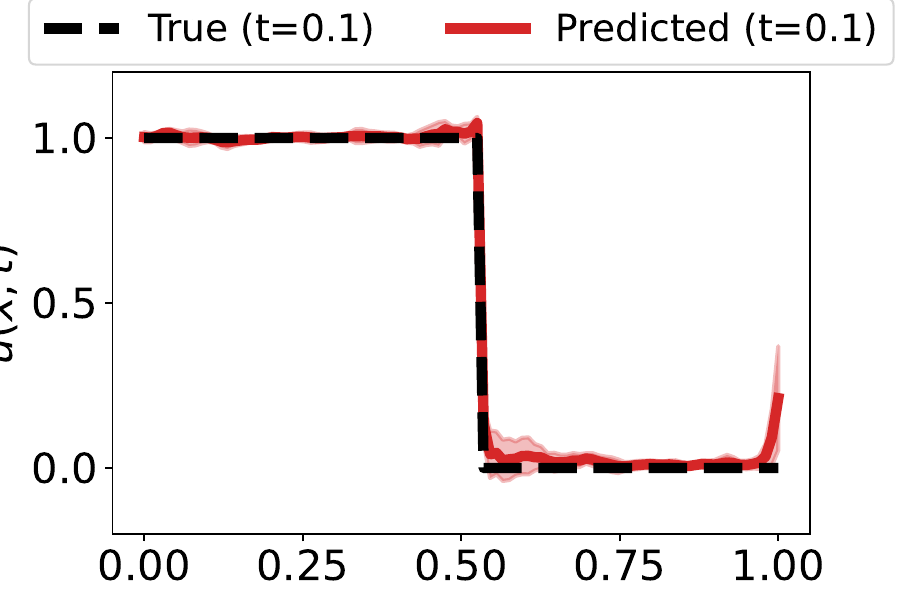}
    \caption{\method}
    \end{subfigure}
    \caption{{\bf 1-d Linear Advection (non-constant input), small OOD shift.} 
    Uncertainty estimates (3 standard deviations) from various UQ methods with small OOD shift with $a^\tr \in [0.45, 0.55]$ and $a^\te\in[0.4,0.45]$. 
    } 
    \label{fig:solutions_la_nonconstant_ood1}
\end{figure}

\begin{figure}[H]
    \centering
    \begin{subfigure}{0.3\textwidth}
    \centering
    \includegraphics[scale=0.35]{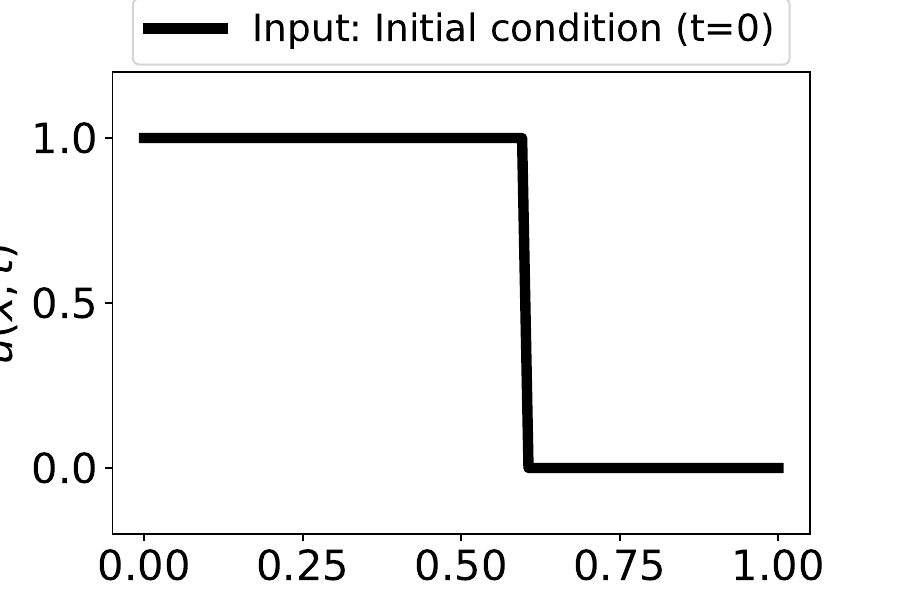}
    \caption{Input}
    \end{subfigure}
    ~~~~
    \begin{subfigure}{0.3\textwidth}
    \centering
    \includegraphics[scale=0.35]{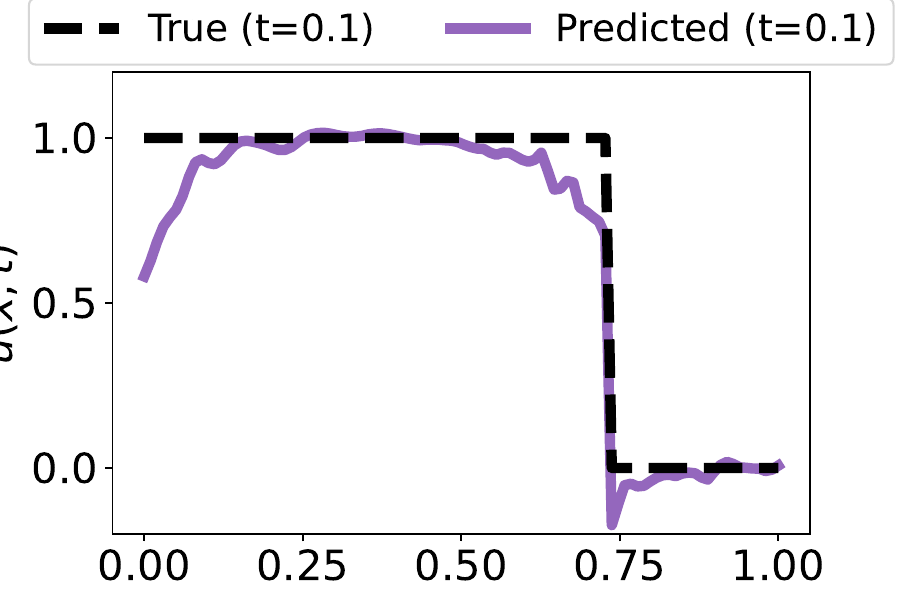}
    \caption{\bayesiannomethod}
    \end{subfigure}
    ~~~~
    \begin{subfigure}{0.3\textwidth}
    \centering
    \includegraphics[scale=0.35]{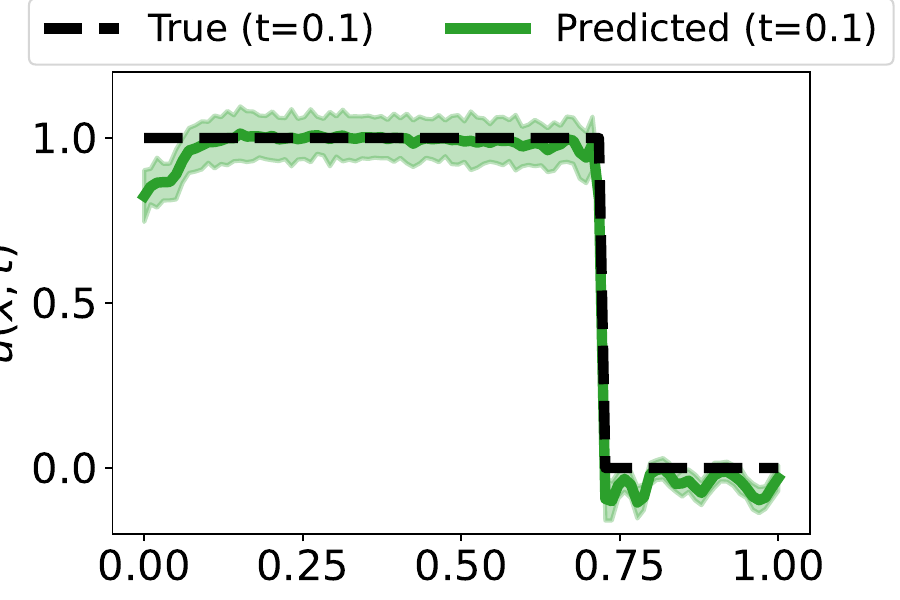}
    \caption{\mcdropoutnomethod}
    \end{subfigure}
    
    \begin{subfigure}{0.3\textwidth}
    \centering
    \includegraphics[scale=0.35]{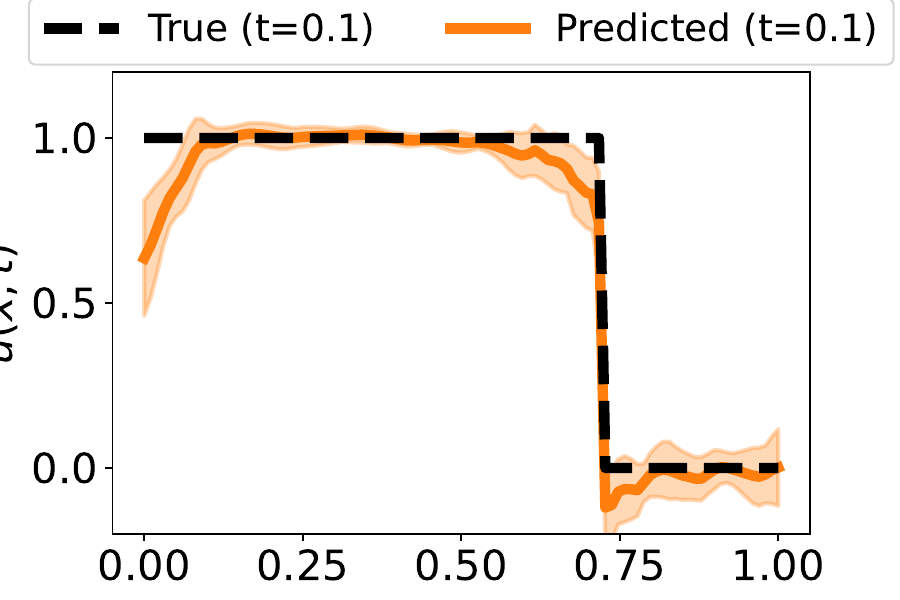}
    \caption{\ensemblenomethod}
    \end{subfigure}
    ~~~~
    \begin{subfigure}{0.3\textwidth}
    \centering
    \includegraphics[scale=0.35]{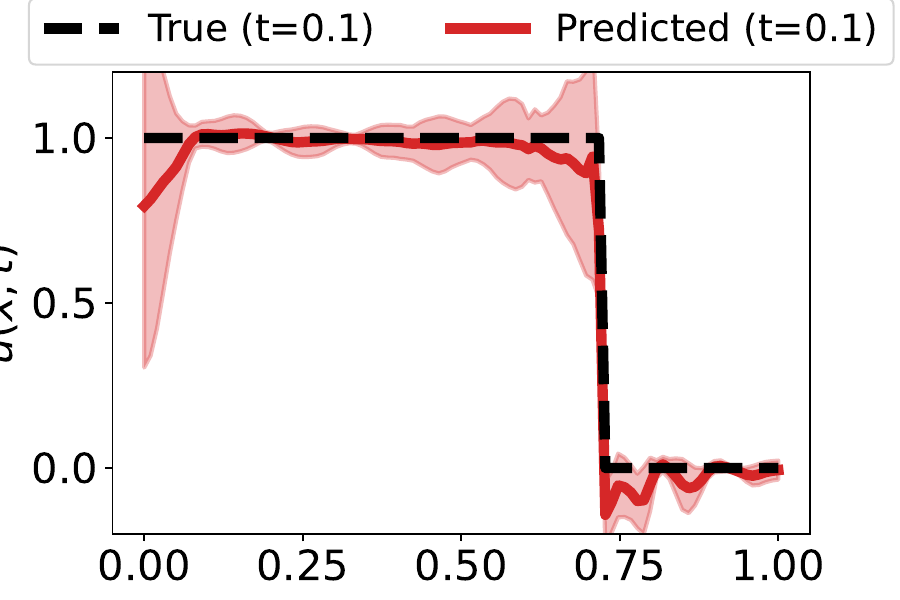}
    \caption{\method}
    \end{subfigure}
    \caption{{\bf 1-d Linear Advection (non-constant input), medium OOD shift.} 
    Uncertainty estimates (3 standard deviations) from various UQ methods with medium OOD shift with $a^\tr \in [0.45, 0.55]$ and $a^\te\in[0.6,0.65]$. 
    } 
    \label{fig:solutions_la_nonconstant_ood3}
\end{figure}

\begin{figure}[H]
    \centering
    \begin{subfigure}{0.3\textwidth}
    \centering
    \includegraphics[scale=0.35]{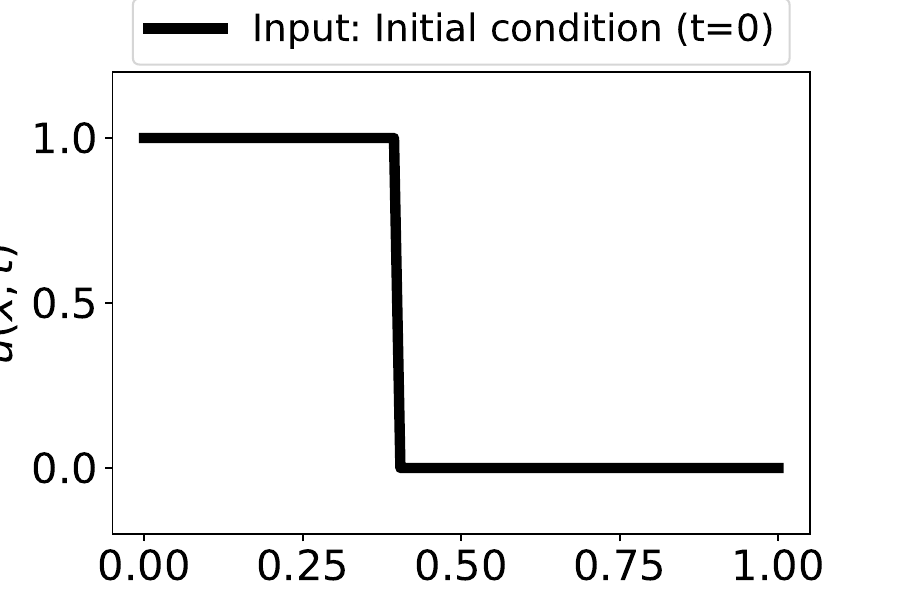}
    \caption{Input}
    \end{subfigure}
    ~~~~
    \begin{subfigure}{0.3\textwidth}
    \centering
    \includegraphics[scale=0.35]{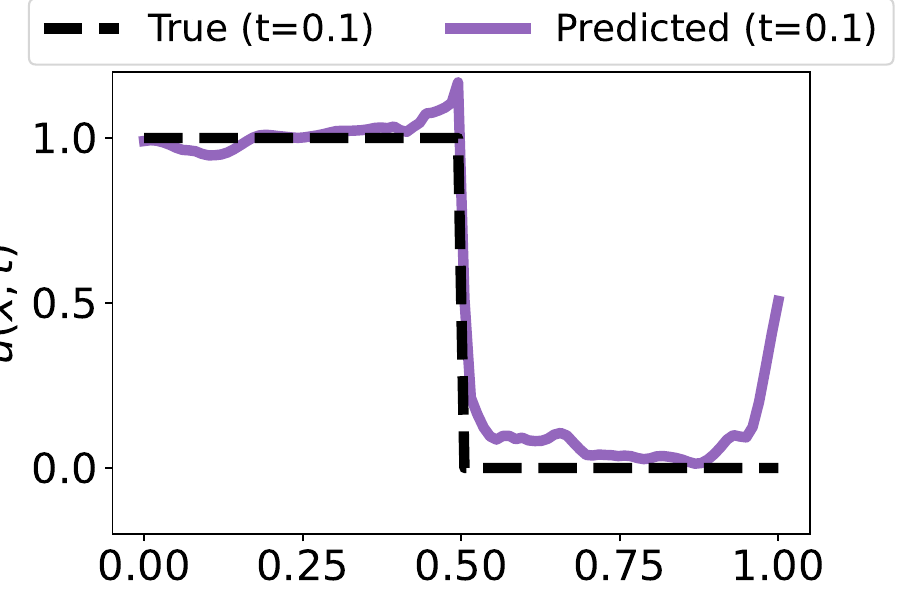}
    \caption{\bayesiannomethod}
    \end{subfigure}
    ~~~~
    \begin{subfigure}{0.3\textwidth}
    \centering
    \includegraphics[scale=0.35]{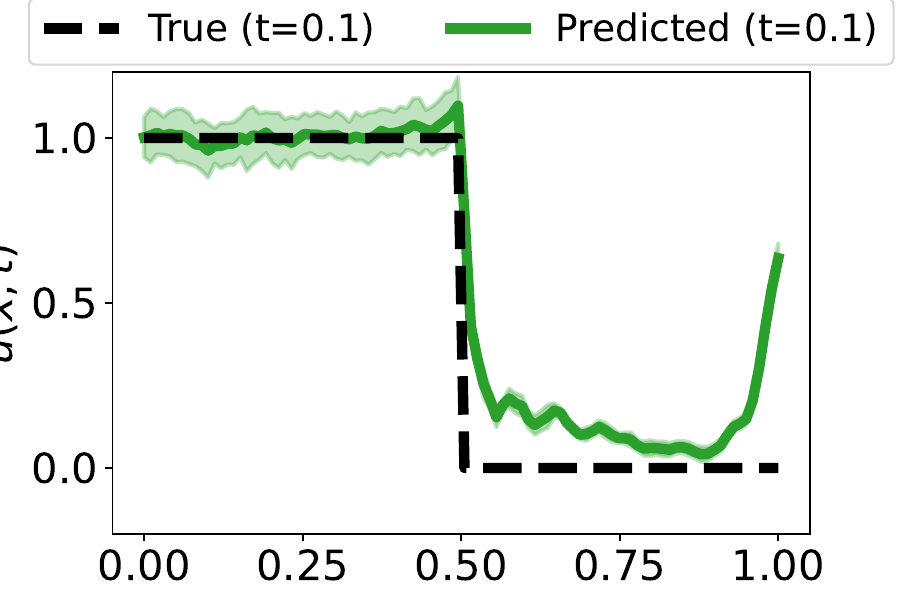}
    \caption{\mcdropoutnomethod}
    \end{subfigure}
    
    \begin{subfigure}{0.3\textwidth}
    \centering
    \includegraphics[scale=0.35]{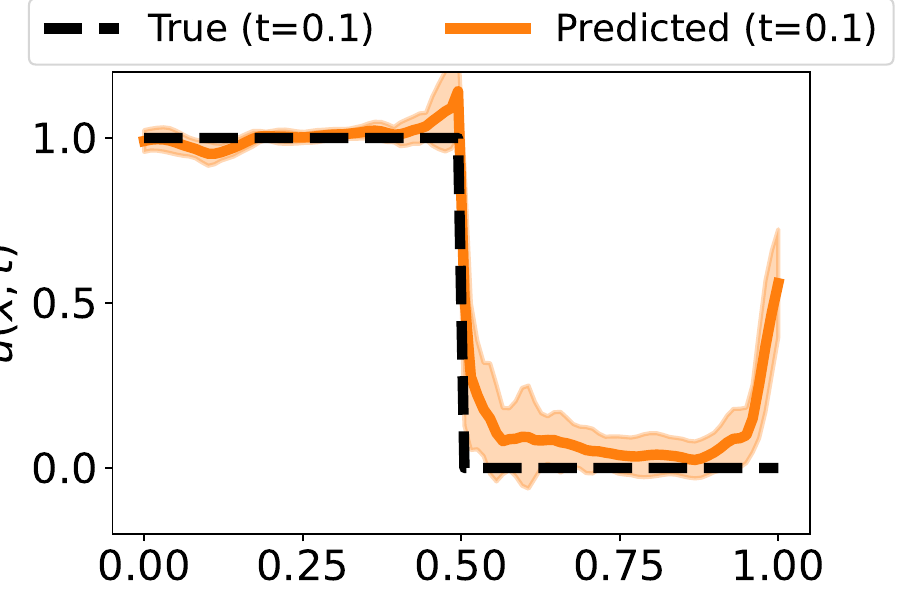}
    \caption{\ensemblenomethod}
    \end{subfigure}
    ~~~~
    \begin{subfigure}{0.3\textwidth}
    \centering
    \includegraphics[scale=0.35]{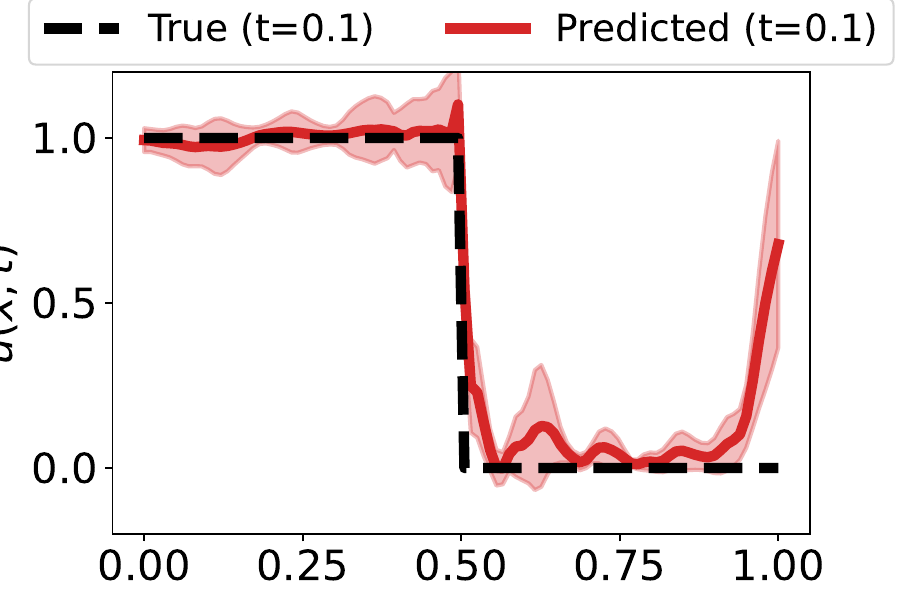}
    \caption{\method}
    \end{subfigure}
    \caption{{\bf 1-d Linear Advection (non-constant input), large OOD shift.} 
    Uncertainty estimates (3 standard deviations) from various UQ methods with large OOD shift with $a^\tr \in [0.45, 0.55]$ and $a^\te\in[0.35,0.4]$. 
    } 
    \label{fig:solutions_la_nonconstant_ood4}
\end{figure}

\cref{fig:solutions_la_nonconstant_ood1,fig:solutions_la_nonconstant_ood3,fig:solutions_la_nonconstant_ood4} show the predictions of the various methods under small, medium and large OOD shifts. 
All methods exhibit high errors near the shock location and at the boundary, particularly for medium (\cref{fig:solutions_la_nonconstant_ood3}) and large OOD shifts (\cref{fig:solutions_la_nonconstant_ood4}). \bayesiannomethod and \mcdropoutnomethod output very low uncertainty despite these high errors. 
In contrast, the UQ from \method is correlated with the error and is highest near the shock and the right boundary.

\begin{table}[H]
    \centering
    \caption{{\bf 1-d Linear Advection (non-constant input)}. MSE $\downarrow$ (mean and standard deviation over 5 seeds) for different UQ methods on the 1-d linear advection equation in-domain and with small, medium and large OOD shifts. 
    Input is a non-constant function $u(x, 0) = 1_{x\leq a}$ with $a^\tr\in[0.45, 0.55]$.
    {\bf Bold} indicates values within one standard deviation of the best mean.
    }
    \label{tab:1dla_nonconstant_mse}
    \begin{tabular}{@{}lcccc@{}}
    \toprule
    Models & In-domain	& OOD-small	& OOD-medium	& OOD-large \\
    & $a^\te\in[0.45, 0.55]$ & $a^\te\in[0.4, 0.45]$ & $a^\te\in[0.6, 0.65]$	& $a^\te\in[0.35, 0.4]$ \\
    \midrule
    \bayesiannomethod & 1.3e-06 (1.2e-06) 	& 3.3e-04 (6.3e-05) & 4.9e-03 (1.4e-03)	& 2.3e-02 (3.2e-03) \\
    \mcdropoutnomethod &  2.7e-05 (5.2e-06) & 	4.2e-04 (6.6e-05)	& 2.9e-03 (5.4e-04)	& 2.9e-02 (4.9e-03) \\
    \ensemblenomethod & \bf 7.5e-07 (6.5e-07)&	\bf 3.1e-04 (5.0e-05)&	4.3e-03 (9.1e-05)&	\bf 2.2e-02 (1.5e-03)\\
    \method & 1.5e-06 (1.1e-06)&	3.4e-04 (9.6e-05)	& \bf 2.7e-03 (7.6e-04)	& 2.8e-02 (5.5e-03) \\
    \bottomrule
    \end{tabular}
\end{table}

\begin{table}[H]
    \centering
    \caption{{\bf 1-d Linear Advection (non-constant input)}. n-MeRCI $\downarrow$ (mean and standard deviation over 5 seeds) for different UQ methods on the 1-d linear advection equation in-domain and with small, medium and large OOD shifts. 
    Input is a non-constant function $u(x, 0) = 1_{x\leq a}$ with $a^\te\in[0.45, 0.55]$.
    {\bf Bold} indicates values within one standard deviation of the best mean. 
    }
    \label{tab:1dla_nonconstant_nmerci}
    \begin{tabular}{@{}lcccc@{}}
    \toprule
        Models & In-domain	& OOD-small	& OOD-medium	& OOD-large \\
        & $a^\te\in[0.45, 0.55]$ & $a^\te\in[0.4, 0.45]$ & $a^\te\in[0.6, 0.65]$	& $a^\te\in[0.35, 0.4]$ \\
    \midrule
    \bayesiannomethod & 0.75 (0.39) &	1.06 (0.01) & 0.94 (0.20) & 	0.87 (0.18) \\
    \mcdropoutnomethod &  0.65 (0.19) &	1.00 (0.01) &	0.89 (0.18) &	0.86 (0.18) \\
    \ensemblenomethod & \bf 0.18 (0.07) & 0.19 (0.03)	& \bf 0.19 (0.05)	& \bf 0.37 (0.11) \\
    \method & \bf 0.22 (0.15) &  \bf 0.16 (0.02) & \bf 0.12 (0.07) &	\bf 0.31 (0.10) \\
    \bottomrule
    \end{tabular}
\end{table}

\subsubsection{Elliptic 2-d Darcy Flow}
Here we provide an elliptic, steady-state 2-d test case, i.e., Darcy Flow, where the solution $u(x)$ denotes the unknown pressure and $k$ the constant permeability field.
\cref{tab:2d_darcy_mse} shows the MSE metric for all methods in-domain and across different OOD shifts.
The in-domain MSE is $\approx 10^{-11}$ for all methods and increases by $10^4$ for the largest OOD shift. 
(\cref{tab:2d_darcy_mse})  also shows that the OOD MSE of \method is improved ($\approx 1.2\times$) upon that of the competing baselines. 
With respect to the meaningful n-MeRCI metric that measures error correlation with uncertainty estimates, \cref{tab:2d_darcy_nmerci} shows that \ensemblenomethod and \method perform $\approx 10\times$ better than the other methods with \method being more computationally efficient. 
Similar trends hold for the CRPS metric in \cref{tab:2d_darcy_crps}, where \method and \ensemblenomethod outperform other baselines by $1.5\times$. 
These 2-d results are consistent with 1-d experiments.

\begin{figure}[H]
    \centering
    \begin{subfigure}[t]{\textwidth}
    \centering
    \includegraphics[scale=0.35]{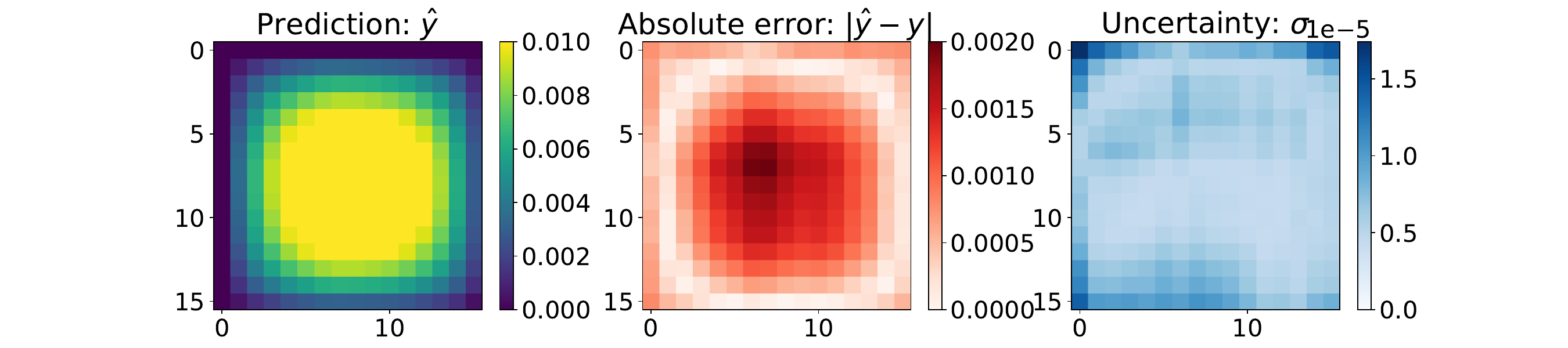}
    \caption{\bayesiannomethod}
    \label{fig:solutions_darcy_ood3_bayesian}
    \end{subfigure}

    \begin{subfigure}[t]{\textwidth}
    \centering
    \includegraphics[scale=0.35]{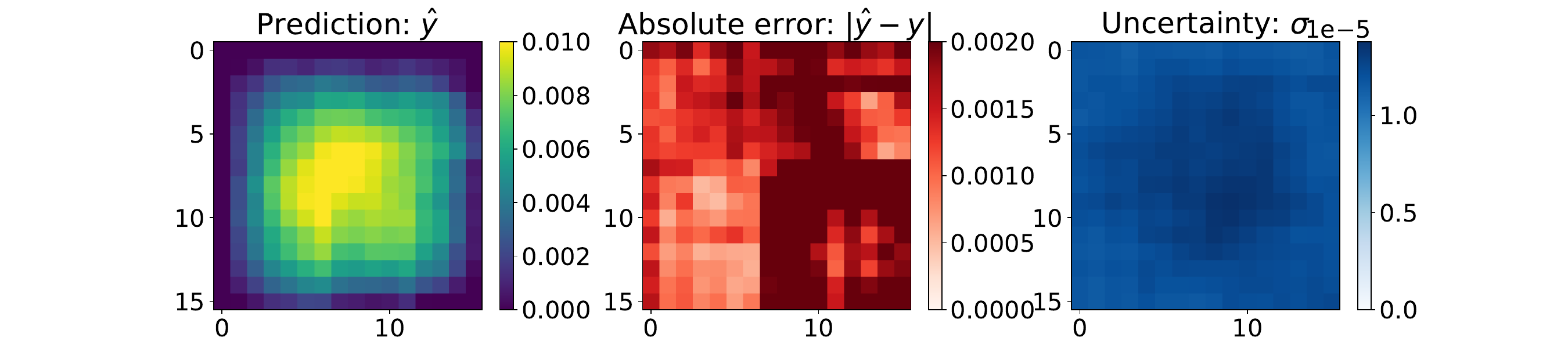}
    \caption{\outputvarmethod}
    \label{fig:solutions_darcy_ood3_variance}
    \end{subfigure}

    \begin{subfigure}[t]{\textwidth}
    \centering
    \includegraphics[scale=0.35]{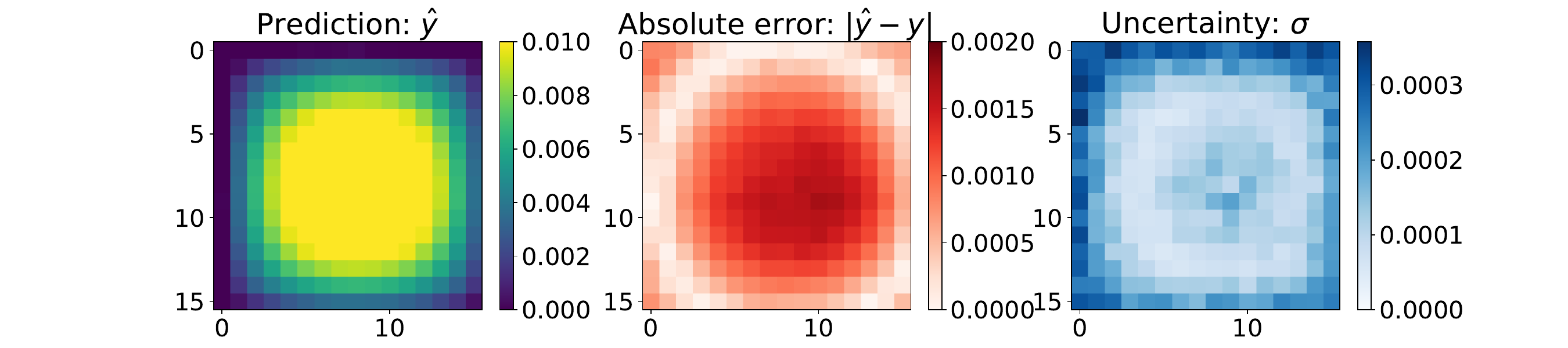}
    \caption{\mcdropoutnomethod}
    \label{fig:solutions_darcy_ood3_mcdropout}
    \end{subfigure}
    
    \begin{subfigure}[t]{\textwidth}
    \centering
    \includegraphics[scale=0.35]{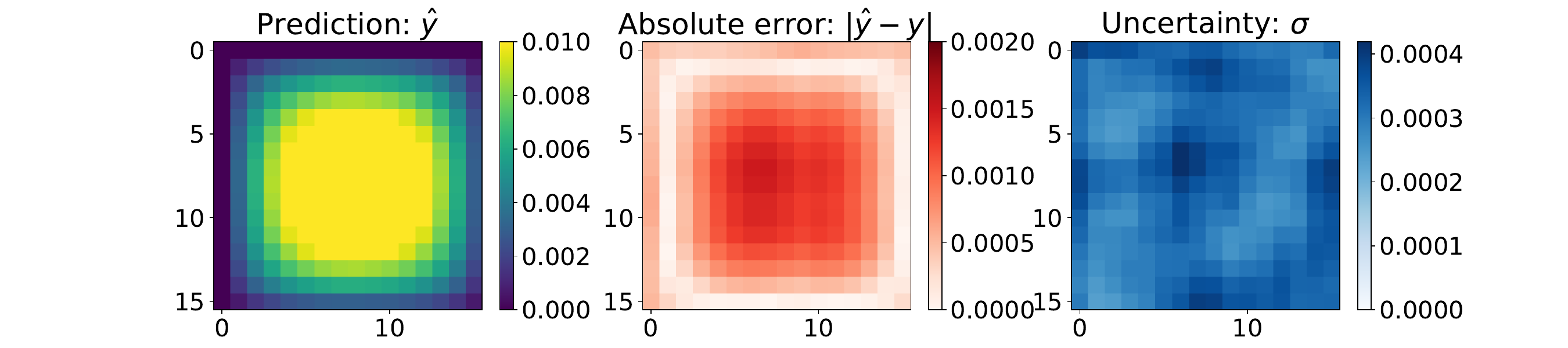}
    \caption{\ensemblenomethod}
    \label{fig:solutions_darcy_ood3_ensemble}
    \end{subfigure}
    
    \begin{subfigure}[t]{\textwidth}
    \centering
    \includegraphics[scale=0.35]{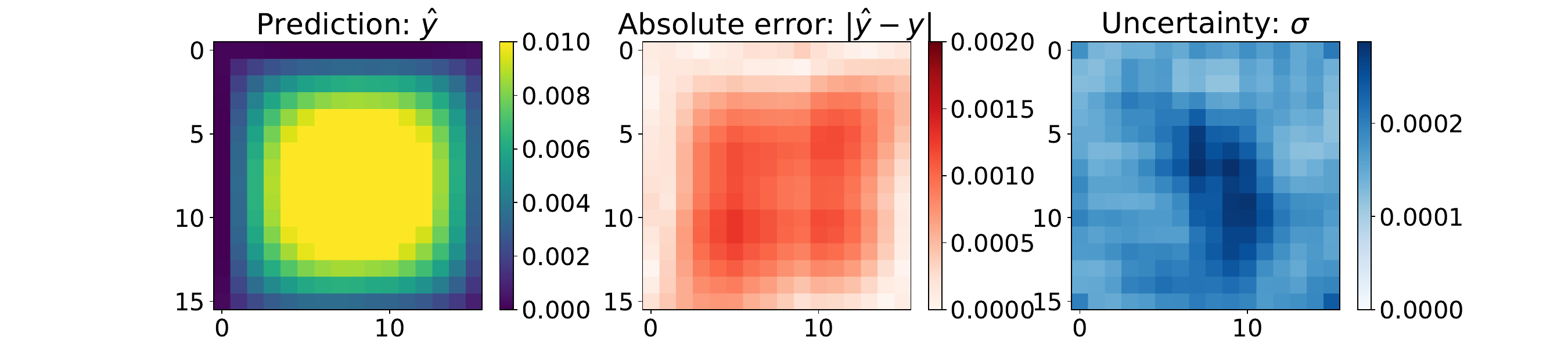}
    \caption{\method}
    \label{fig:solutions_darcy_ood3_diverse}
    \end{subfigure}
    
    \caption{{\bf 2-d Darcy Flow, large OOD shift.} Solution profiles (left column), absolute error (middle column) and uncertainty estimates (right column) from different UQ methods under a large OOD shift ($k^\tr\in[3,4]$ and $k^\te\in[5,6]$).} 
    \label{fig:solutions_darcy_ood3}
\end{figure}

\cref{fig:solutions_darcy_ood3} illustrates the solution profile (left column), absolute error (middle column) and uncertainty plots (right column) under a large OOD shift for our \method model and the baselines. 
The absolute errors from all models except \outputvarmethod  are concentrated around the center of the domain where the pressure values in the solution profile are highest. We see that the errors from \method are slightly lower than that of the baselines. 
\bayesiannomethod (\cref{fig:solutions_darcy_ood3_bayesian}) and \outputvarmethod (\cref{fig:solutions_darcy_ood3_variance}) output uncertainty estimates spread uniformly over the domain. 
Uncertainty estimates from \mcdropoutnomethod (\cref{fig:solutions_darcy_ood3_mcdropout}) are highest near the boundary instead of the center region. 
Uncertainty estimates from \method (\cref{fig:solutions_darcy_ood3_diverse}) are highest around the center, correlating better with the error.

\begin{table}[h]
    \centering
    \caption{{\bf 2-d Darcy Flow}. MSE $\downarrow$ for different UQ methods on the 2-d Darcy Flow equation in-domain and with small, medium and large OOD shifts, where $k^\tr \in [3,4]$.
    {\bf Bold} indicates the best mean. 
    }
    \label{tab:2d_darcy_mse}
    \begin{tabular}{@{}lcccc@{}}
    \toprule
    Models & In-domain	& OOD-small	& OOD-medium	& OOD-large \\
    & $k^\te\in[3, 4]$ & $k^\te\in[4, 4.5]$ & $k^\te\in[4.5, 5]$	& $k^\te\in[5, 6]$ \\
    \midrule
    \bayesiannomethod & 4.3e-11 & 4.6e-09 & 5.6e-08 & 4.2e-07 \\
    \outputvarmethod & \bf 7.7e-12 & \bf 7.4e-10 & \bf 2.9e-08 & 1.0e-06 \\
    \mcdropoutnomethod & 1.2e-09 & 4.7e-09 & 5.4e-08 & 4.4e-07 \\
    \ensemblenomethod & 3.9e-11 & 4.2e-09 & 5.0e-08 & 3.4e-07 \\
    \method & 3.9e-11 & 3.8e-09 & 4.4e-08 & \bf 2.9e-07 \\
    \bottomrule
    \end{tabular}
\end{table}

\begin{table}[h]
    \centering
    \caption{{\bf 2-d Darcy Flow}. n-MeRCI $\downarrow$ for different UQ methods on the 2-d Darcy Flow equation in-domain and with small, medium and large OOD shifts, where $k^\tr \in [3,4]$.
    {\bf Bold} indicates the best mean.
    }
    \label{tab:2d_darcy_nmerci}
    \begin{tabular}{@{}lcccc@{}}
    \toprule
    Models & In-domain	& OOD-small	& OOD-medium	& OOD-large \\
    & $k^\te\in[3, 4]$ & $k^\te\in[4, 4.5]$ & $k^\te\in[4.5, 5]$	& $k^\te\in[5, 6]$ \\
    \midrule
    \bayesiannomethod &  0.47 	& 0.79  & 0.83 	& 0.71  \\
    \outputvarmethod &   0.28 & 	0.78 	& 0.87 	& 0.78  \\
    \mcdropoutnomethod &  0.24  & 	0.86 	& 0.88 	& 0.75  \\
    \ensemblenomethod & 0.11 &	\bf 0.03  &	0.13 & 0.26 \\
    \method & \bf 0.05 &	0.06 	& \bf 0.07 	& \bf 0.24  \\
    \bottomrule
    \end{tabular}
\end{table}

\begin{table}[H]
    \centering
    \caption{{\bf 2-d Darcy Flow}. CRPS $\downarrow$ for different UQ methods on the 2-d Darcy Flow equation in-domain and with small, medium and large OOD shifts, where $k^\tr \in [3,4]$.
    {\bf Bold} indicates the best mean.
    }
    \label{tab:2d_darcy_crps}
    \begin{tabular}{@{}lcccc@{}}
    \toprule
    Models & In-domain	& OOD-small	& OOD-medium	& OOD-large \\
    & $k^\te\in[3, 4]$ & $k^\te\in[4, 4.5]$ & $k^\te\in[4.5, 5]$	& $k^\te\in[5, 6]$ \\
    \midrule
    \bayesiannomethod & 0.034   	& 1.89  & 5.17 	& 10.36  \\
     \outputvarmethod & 0.430   & 	1.59 	& 8.43 	& 30.70  \\
    \mcdropoutnomethod & \bf 0.031   & 	1.70 	& 5.02 	& 10.68  \\
    \ensemblenomethod & 0.033  &	\bf 1.27  &	4.05 & 7.40 \\
    \method & 0.035 & 1.28 	& \bf 4.01 	& \bf 7.19  \\
    \bottomrule
    \end{tabular}
\end{table}

\subsection{Cost performance curves}
\begin{figure}[H]
    \centering
    \begin{subfigure}{0.22\textwidth}
    \centering
    \includegraphics[scale=0.27]{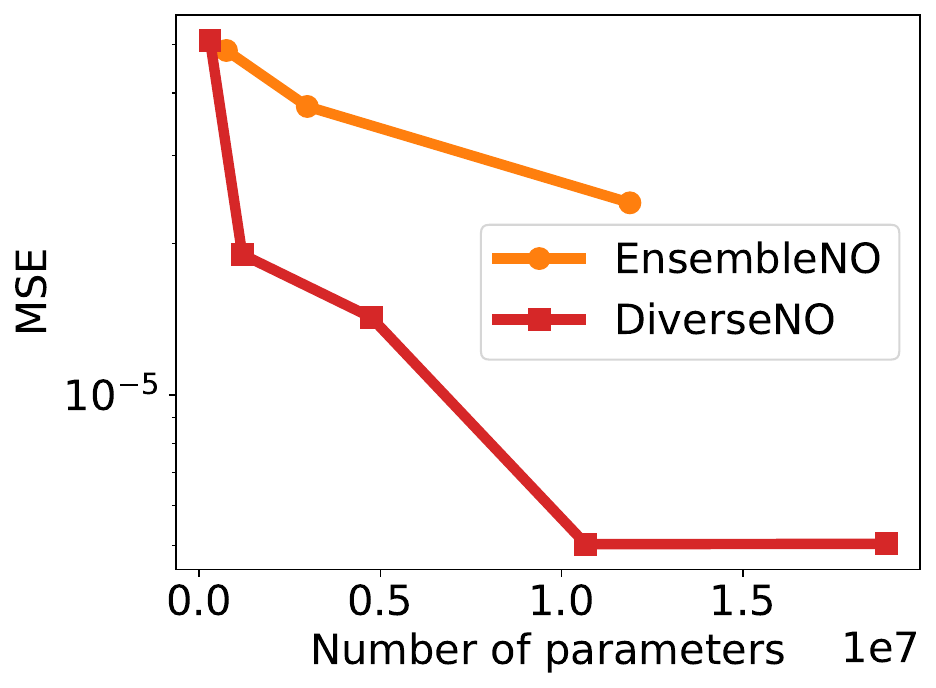}
    \caption{Heat, MSE $\downarrow$}
    \end{subfigure}
    ~~~~
    \begin{subfigure}{0.22\textwidth}
    \centering
    \includegraphics[scale=0.27]{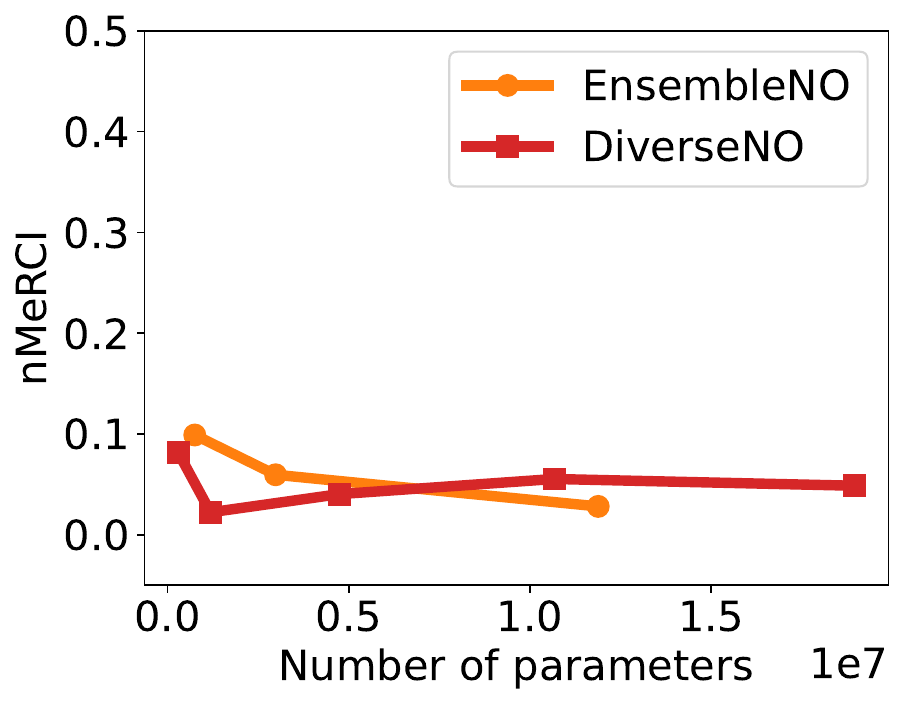}
    \caption{Heat, n-MeRCI $\downarrow$}
    \end{subfigure}
    ~~~~
    \begin{subfigure}{0.22\textwidth}
    \centering
    \includegraphics[scale=0.27]{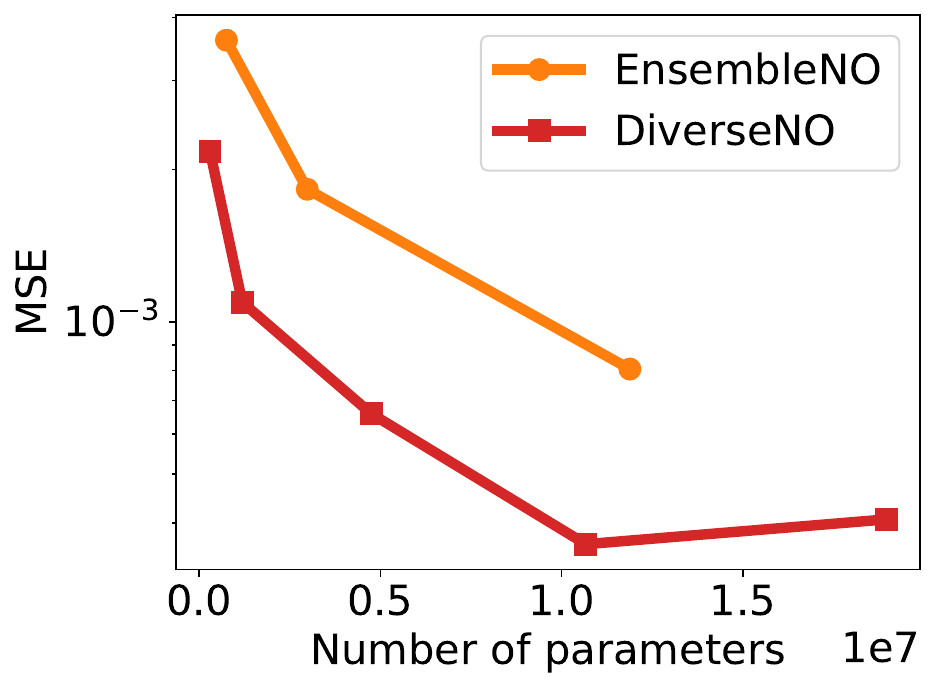}
    \caption{PME, MSE $\downarrow$}
    \end{subfigure}
    ~~~~
    \begin{subfigure}{0.22\textwidth}
    \centering
    \includegraphics[scale=0.27]{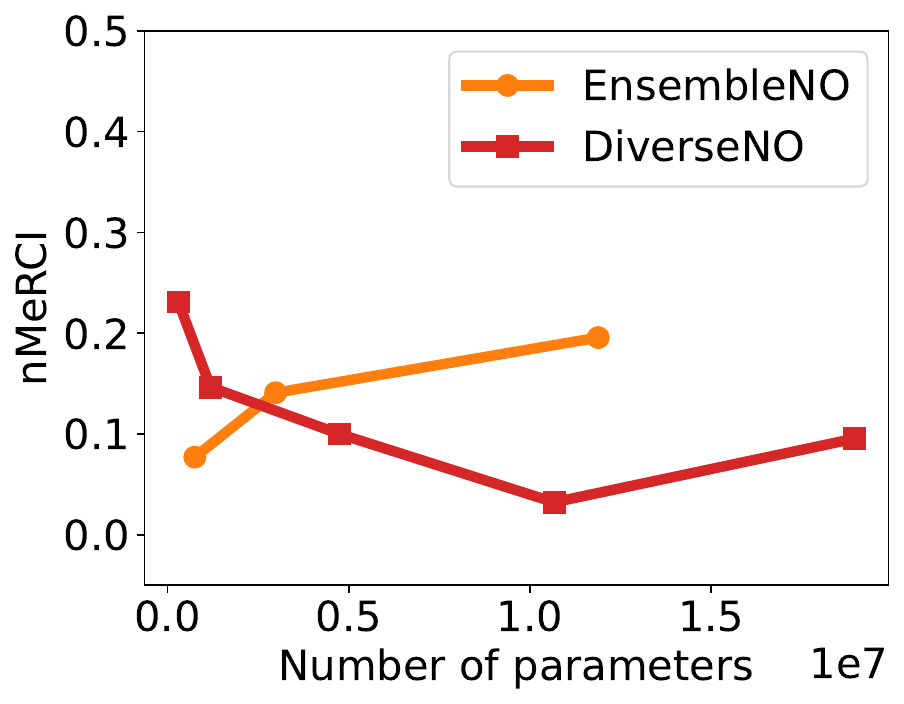}
    \caption{PME, n-MeRCI $\downarrow$}
    \end{subfigure}
    \caption{{\bf Cost-performance tradeoff.} MSE $\downarrow$ and n-MeRCI $\downarrow$ vs number of parameters for \ensemblenomethod and \method with varying number of parameters.
    for {\bf (a-b)} 1-d Heat equation, medium OOD shift, i.e., $k^\tr\in[1,5], k^\te\in[6,7]$ and {\bf (c-d)} 1-d PME, medium OOD shift, i.e., $m^\tr\in[2,3], m^\te\in[4,5]$. 
    } 
    \label{fig:cost_perf_heat_ood3}
\end{figure}

Here we show the the cost performance curves as a function of the number parameters. We see in \cref{fig:cost_perf_heat_ood3} that it has similar trends to the cost performance curves as a  function of the floating point operations (FLOPS) in \cref{fig:costperf_heat_pme}. \method has lower MSE and n-MeRCI values for the same number of parameters as EnsembleNO and hence is more computationally efficient on both the heat equation and PME. 
\label{subsec:cost_perf_app}

\subsection{Effect of \probconserv update}
\label{subsec:probconserv}
\begin{figure}[H]
    \centering
    \hspace{-0.5in}
    \begin{subfigure}[h]{0.30\textwidth}
    \centering
    \includegraphics[scale=0.28]{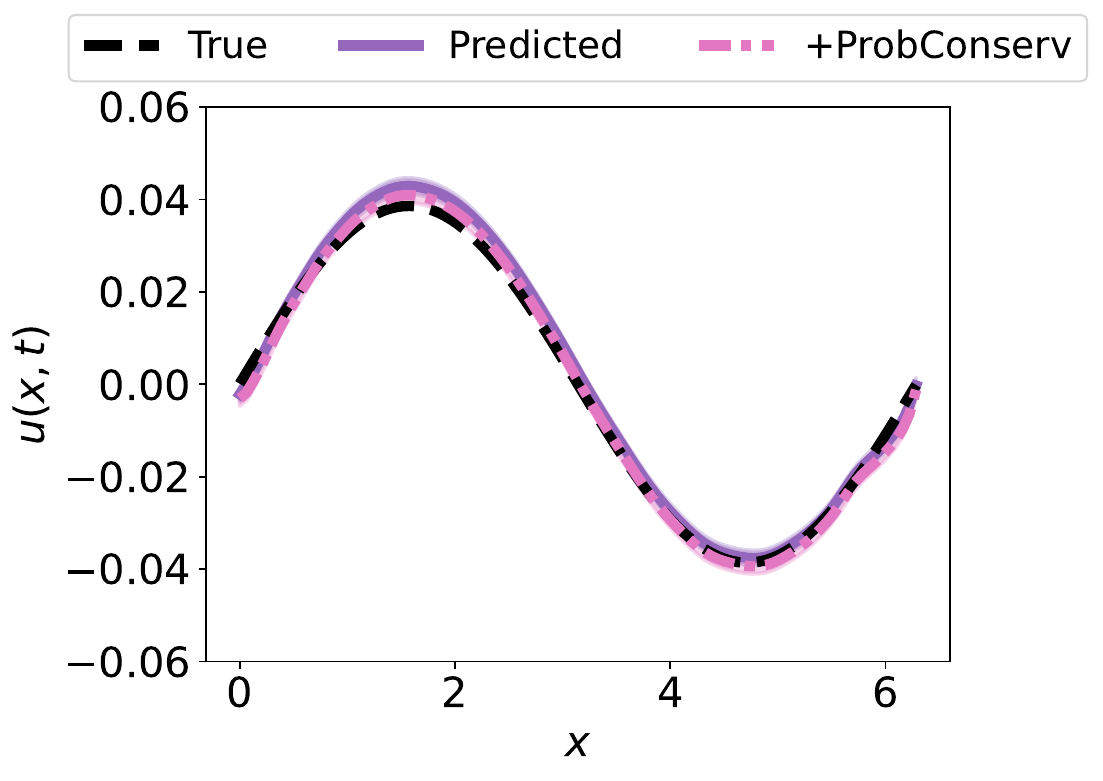}
    \caption{\bayesiannomethod}
    \end{subfigure}
    ~~
    \begin{subfigure}[h]{0.30\textwidth}
    \centering
    \includegraphics[scale=0.28]{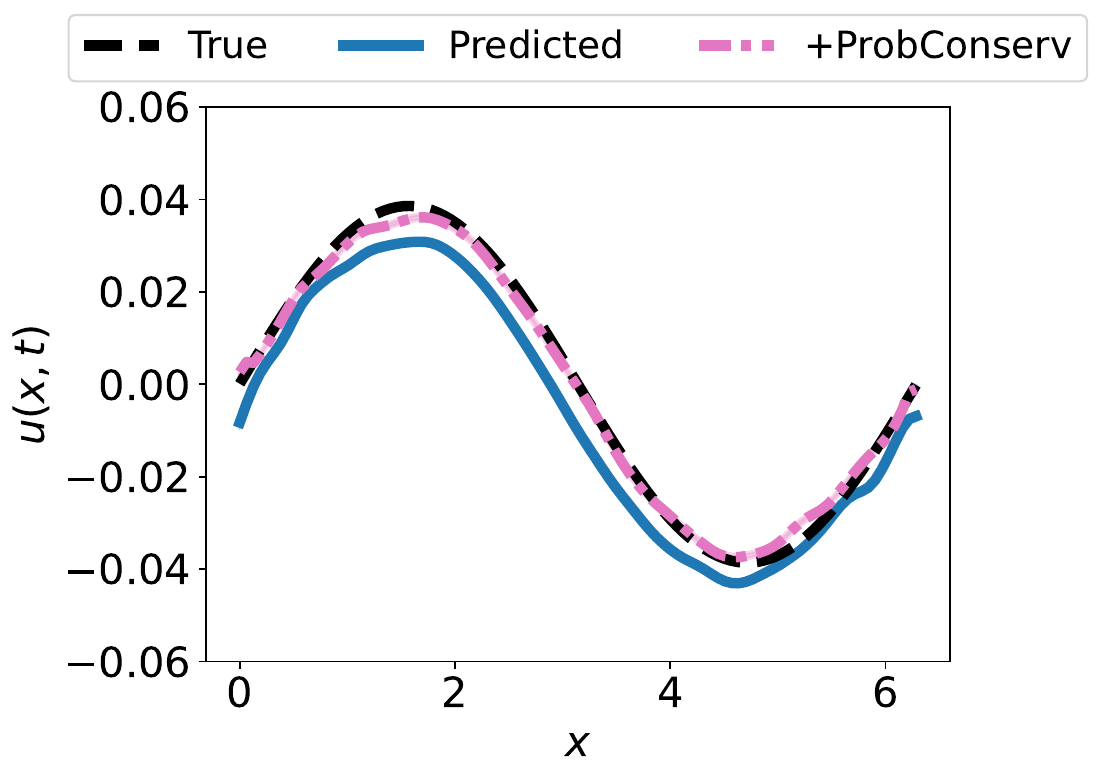}
    \caption{\outputvarmethod}
    \end{subfigure}
    ~~
    \begin{subfigure}[h]{0.30\textwidth}
    \centering
    \includegraphics[scale=0.28]{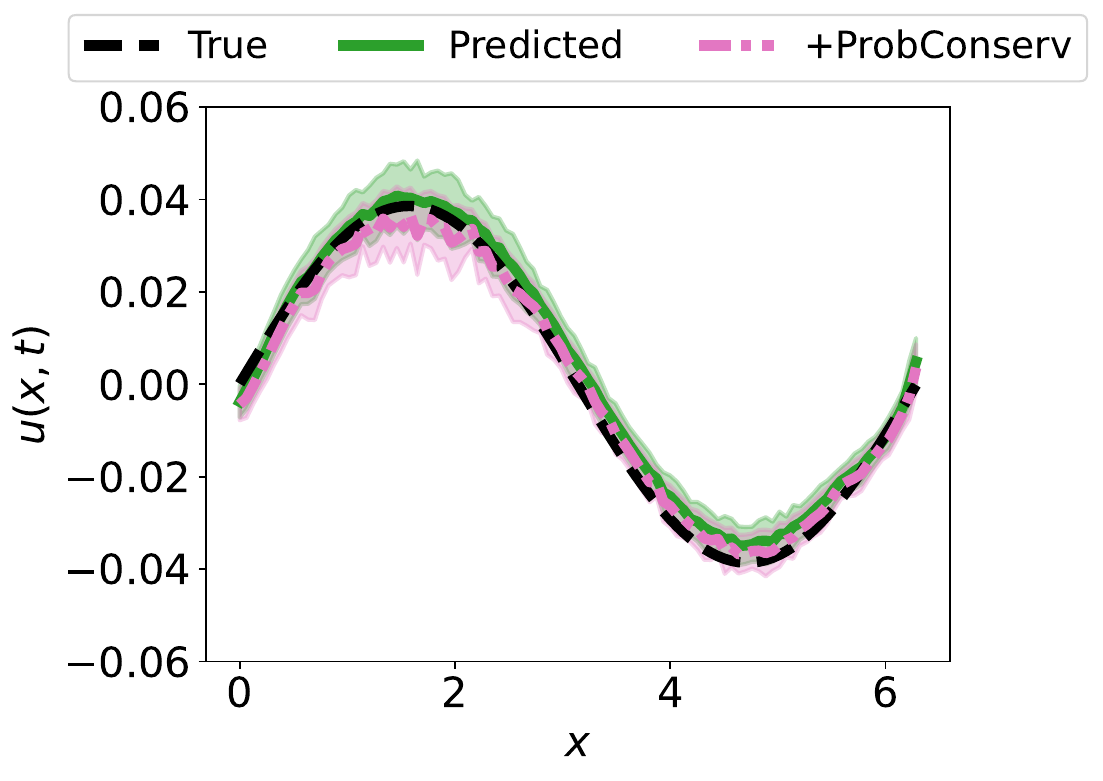}
    \caption{\mcdropoutnomethod}
    \end{subfigure}
    
    \begin{subfigure}[h]{0.30\textwidth}
    \centering
    \includegraphics[scale=0.28]{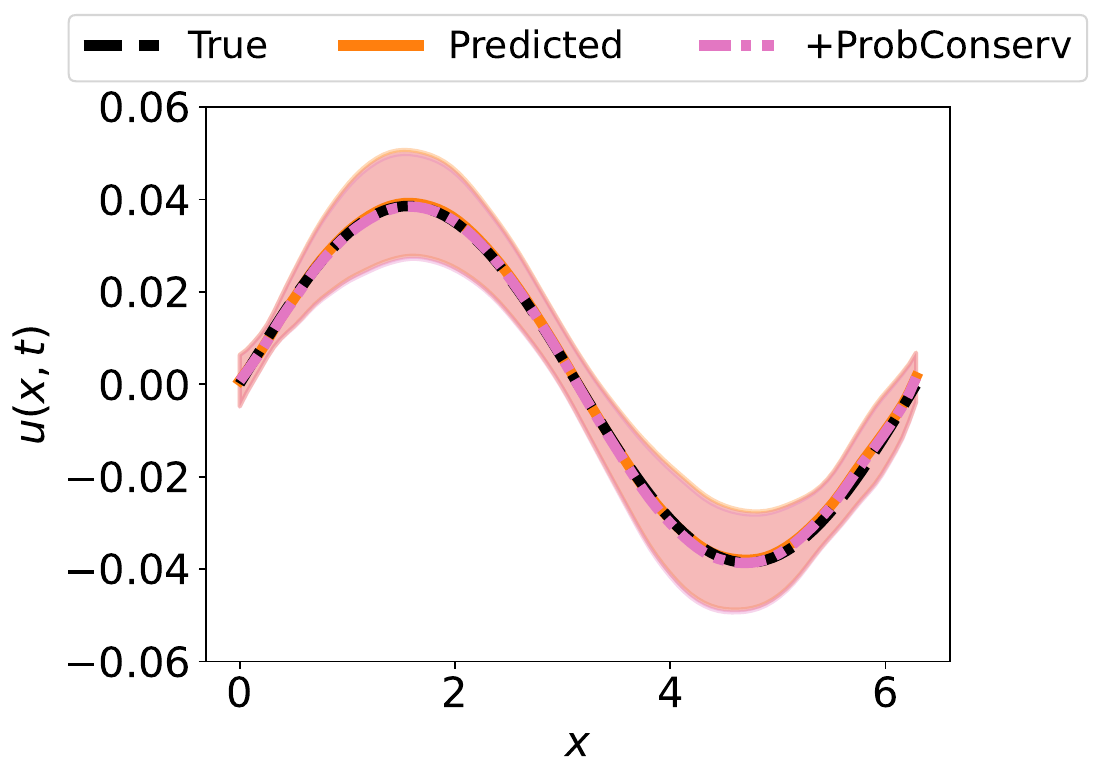}
    \caption{\ensemblenomethod}
    \end{subfigure}
    ~~~~
    \begin{subfigure}[h]{0.30\textwidth}
    \centering
    \includegraphics[scale=0.28]{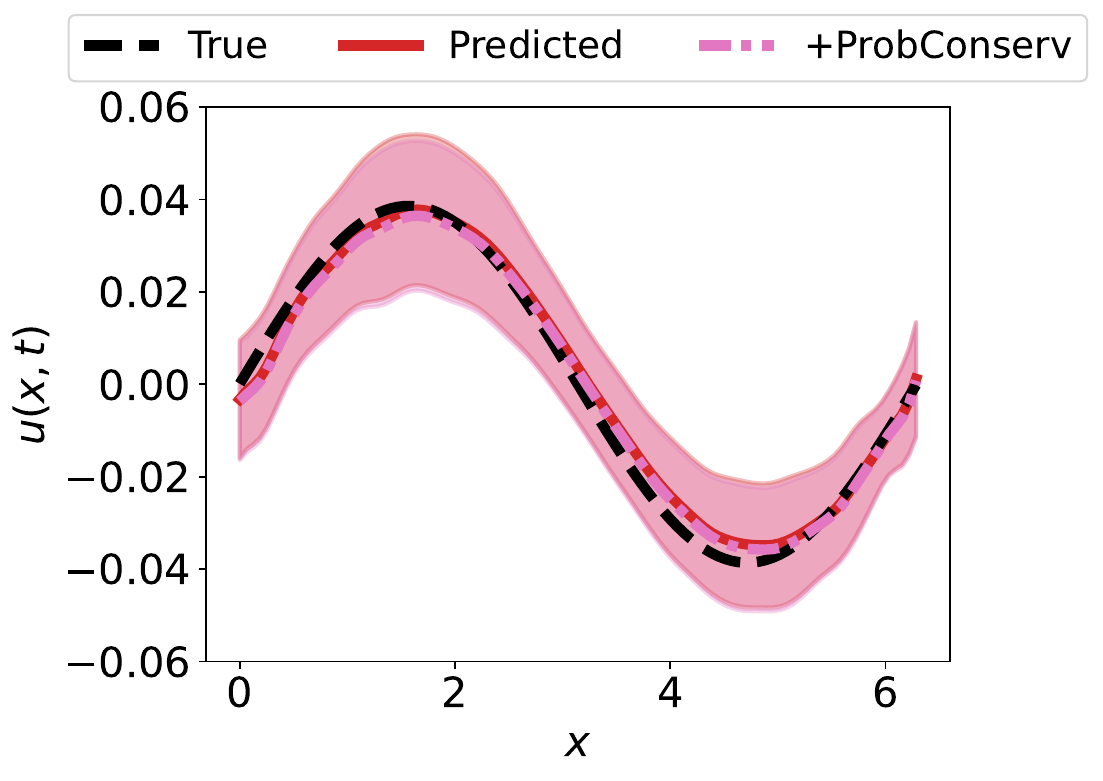}
    \caption{\method}
    \end{subfigure}
    \caption{{\bf 1-d Heat equation, medium OOD shift}, $k^\tr\in[1,5], k^\te\in[6,7]$. 
    Uncertainty estimates from different UQ methods under medium OOD shifts in the input diffusivity coefficient with adding \probconserv onto the uncertainty estimates. \probconserv improves the OOD predictions for all UQ estimates on this ``easy'' task.} 
    \label{fig:probconserv_heat}
\end{figure}

\begin{figure}[H]
    \centering
    \hspace{-0.5in}
    \begin{subfigure}[h]{0.30\textwidth}
    \centering
    \includegraphics[scale=0.28]{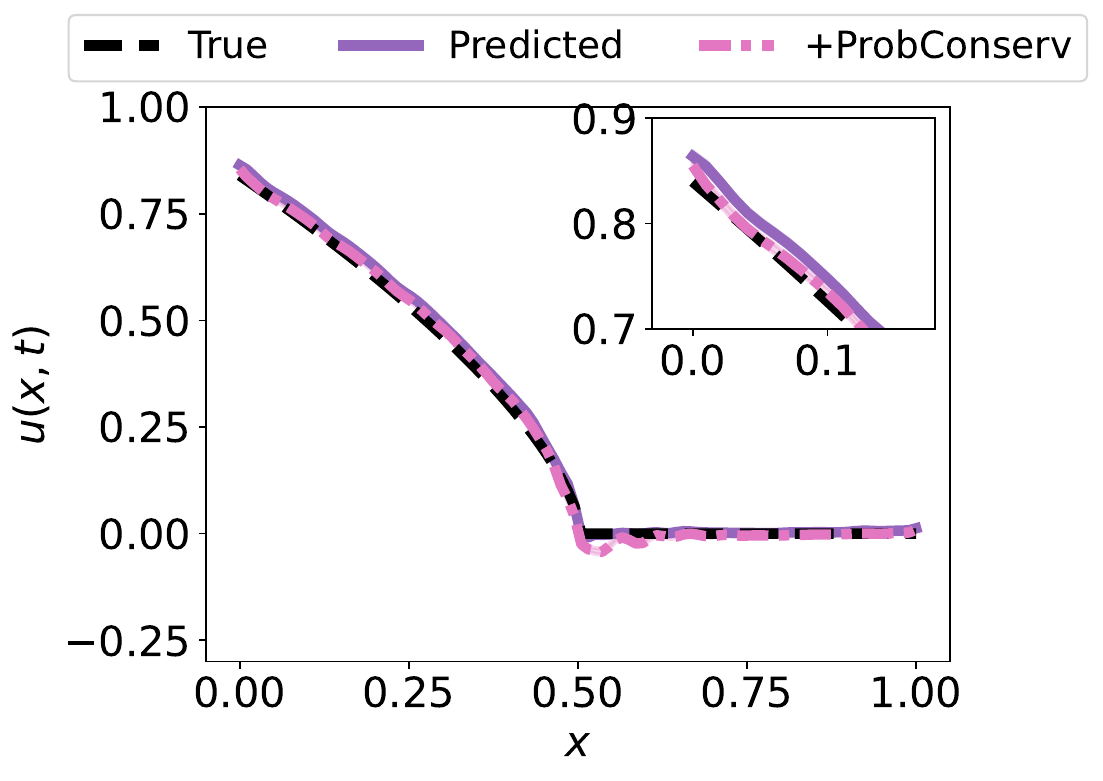}
    \caption{\bayesiannomethod}
    \end{subfigure}
    ~~
    \begin{subfigure}[h]{0.30\textwidth}
    \centering
    \includegraphics[scale=0.28]{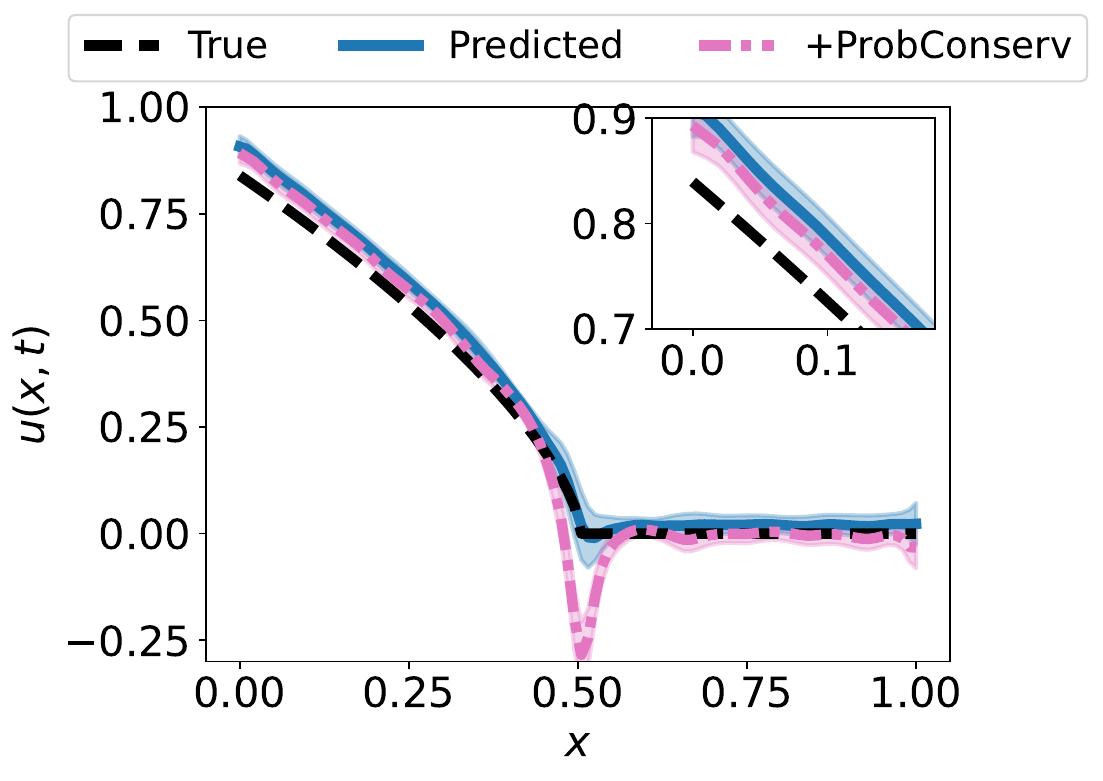}
    \caption{\outputvarmethod}
    \end{subfigure}
    ~~
    \begin{subfigure}[h]{0.30\textwidth}
    \centering
    \includegraphics[scale=0.28]{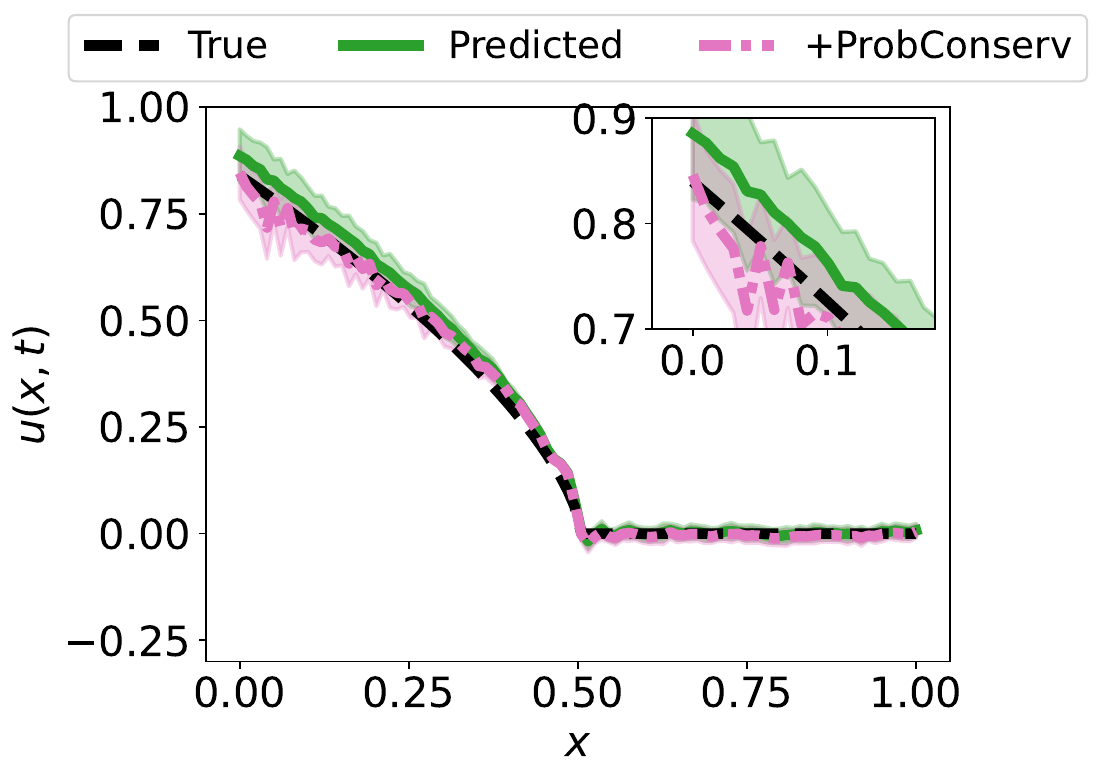}
    \caption{\mcdropoutnomethod}
    \end{subfigure}
    
    \begin{subfigure}[h]{0.30\textwidth}
    \centering
    \includegraphics[scale=0.28]{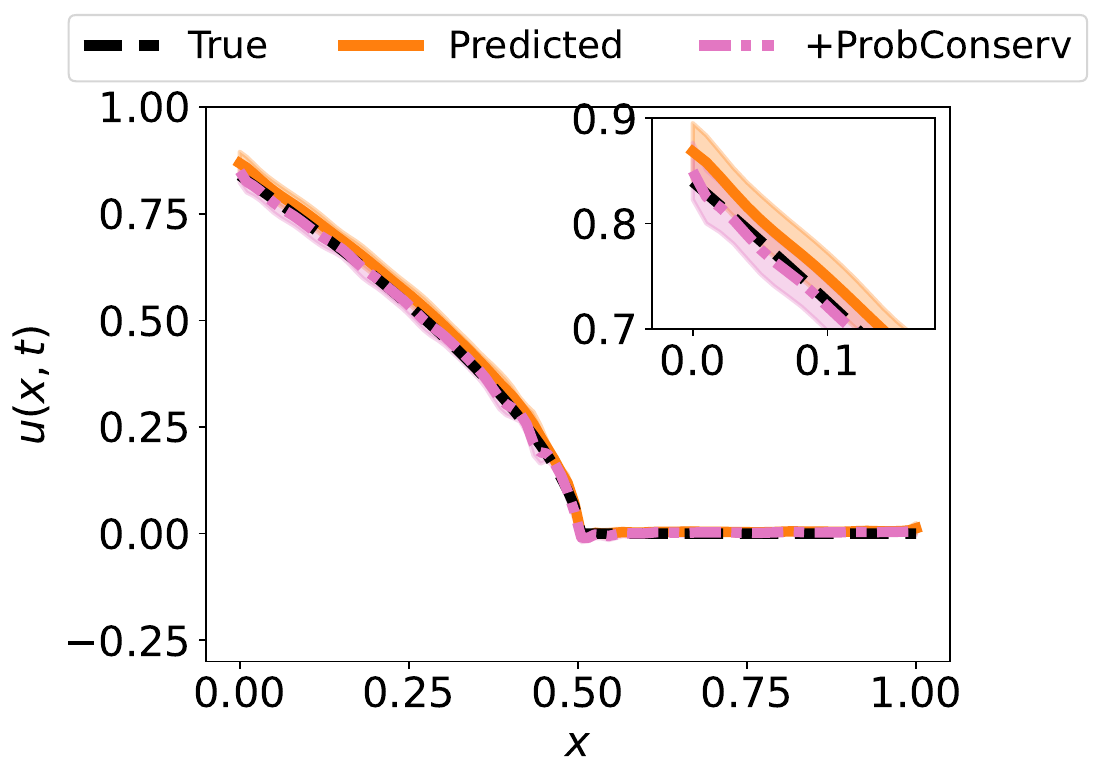}
    \caption{\ensemblenomethod}
    \end{subfigure}
    ~~~~
    \begin{subfigure}[h]{0.30\textwidth}
    \centering
    \includegraphics[scale=0.28]{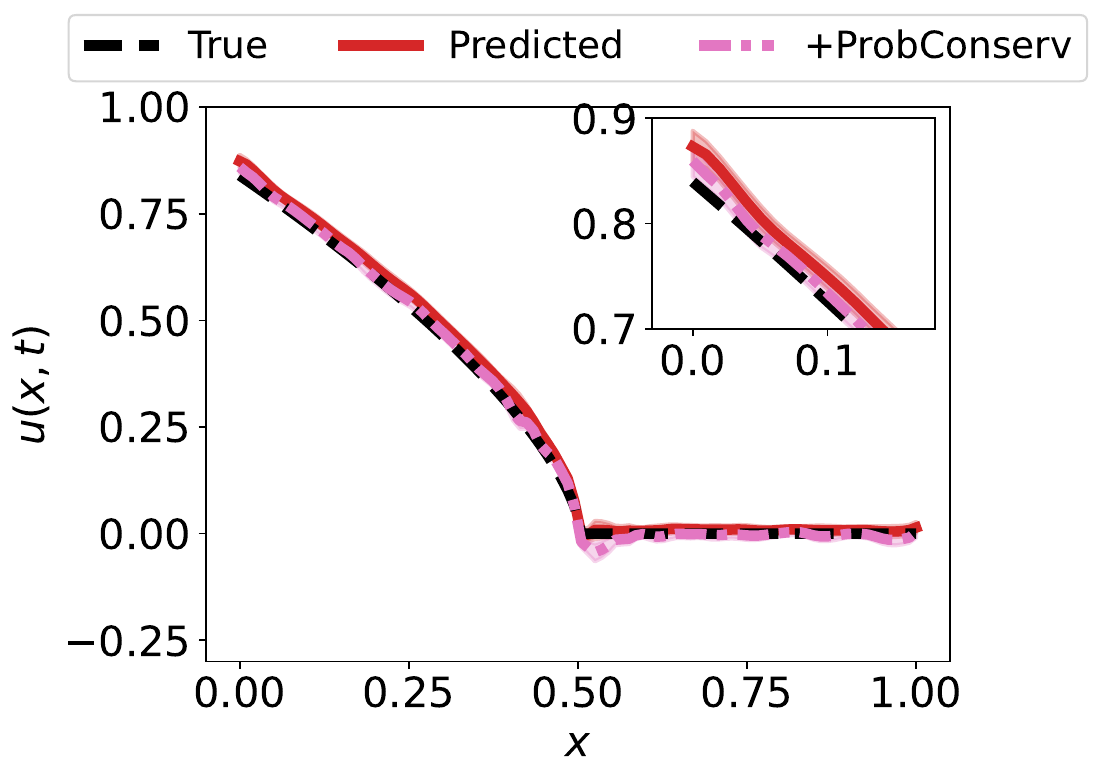}
    \caption{\method}
    \end{subfigure}
    \caption{{\bf 1-d PME equation, small OOD shift}, $m^\tr\in[2,3], m^\te\in[1,2]$.  
    Uncertainty estimates from different UQ methods under small OOD shifts in the power $m$ in the coefficient $k(u)=u^m$ with adding \probconserv onto the uncertainty estimates. \probconserv improves the OOD predictions for all methods except for \outputvarmethod whose UQ estimate is not well-correlated with the error.} 
    \label{fig:probconserv_pme}
\end{figure}

\begin{figure}[H]
    \centering
    \hspace{-0.5in}
    \begin{subfigure}[h]{0.30\textwidth}
    \centering
    \includegraphics[scale=0.28]{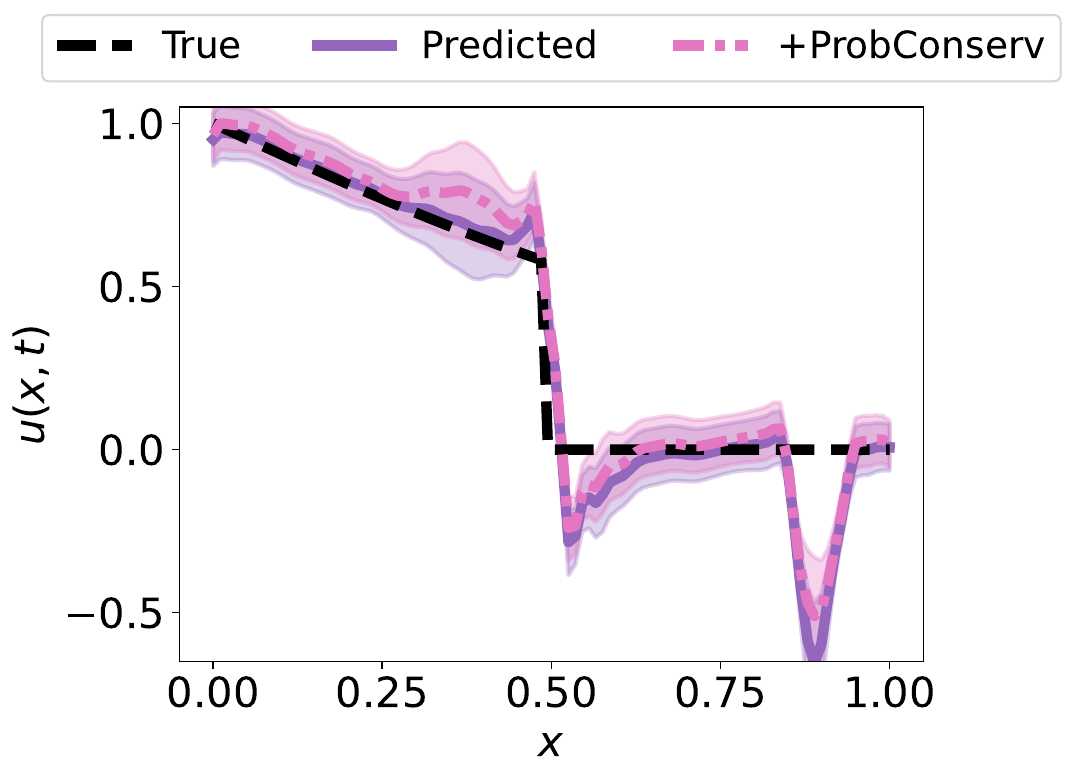}
    \caption{\bayesiannomethod}
    \end{subfigure}
    ~~
    \begin{subfigure}[h]{0.30\textwidth}
    \centering
    \includegraphics[scale=0.28]{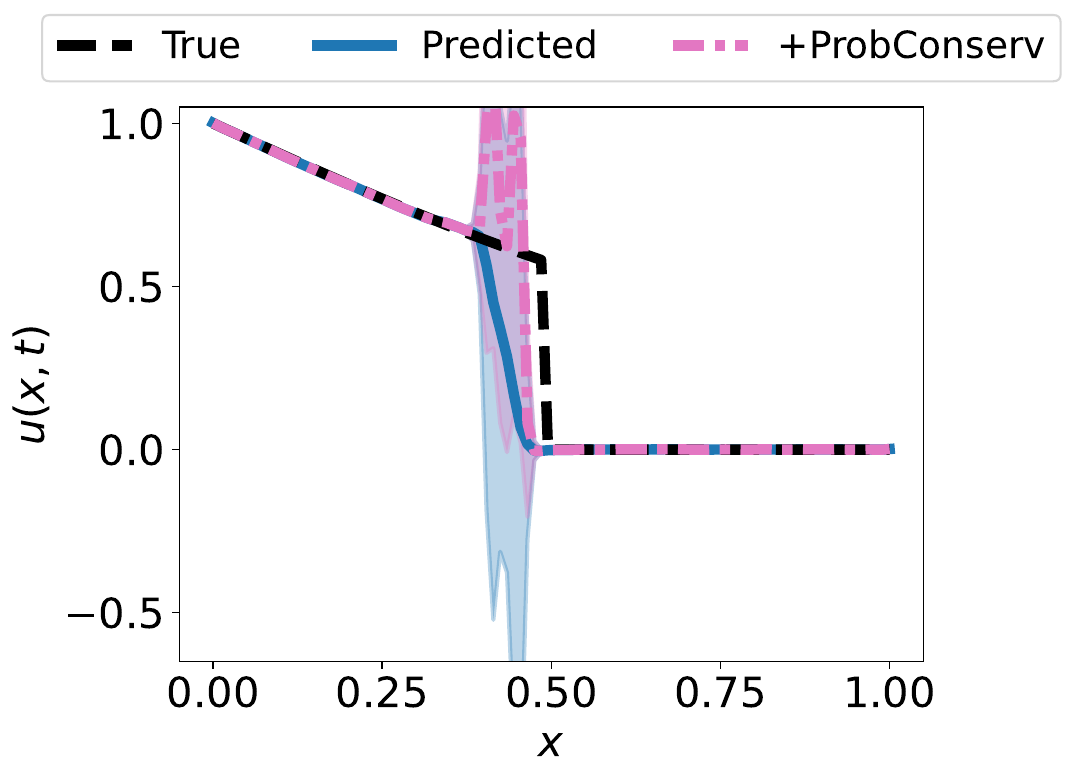}
    \caption{\outputvarmethod}
    \end{subfigure}
    ~~
    \begin{subfigure}[h]{0.30\textwidth}
    \centering
    \includegraphics[scale=0.28]{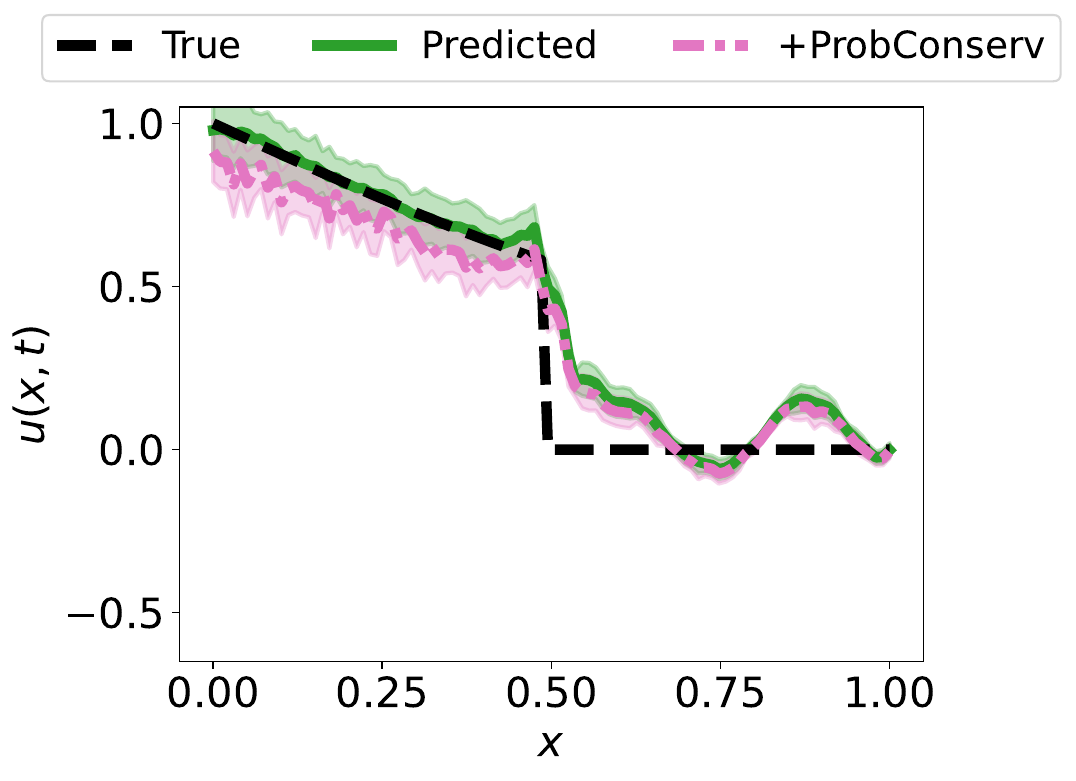}
    \caption{\mcdropoutnomethod}
    \end{subfigure}
    
    \begin{subfigure}[h]{0.30\textwidth}
    \centering
    \includegraphics[scale=0.28]{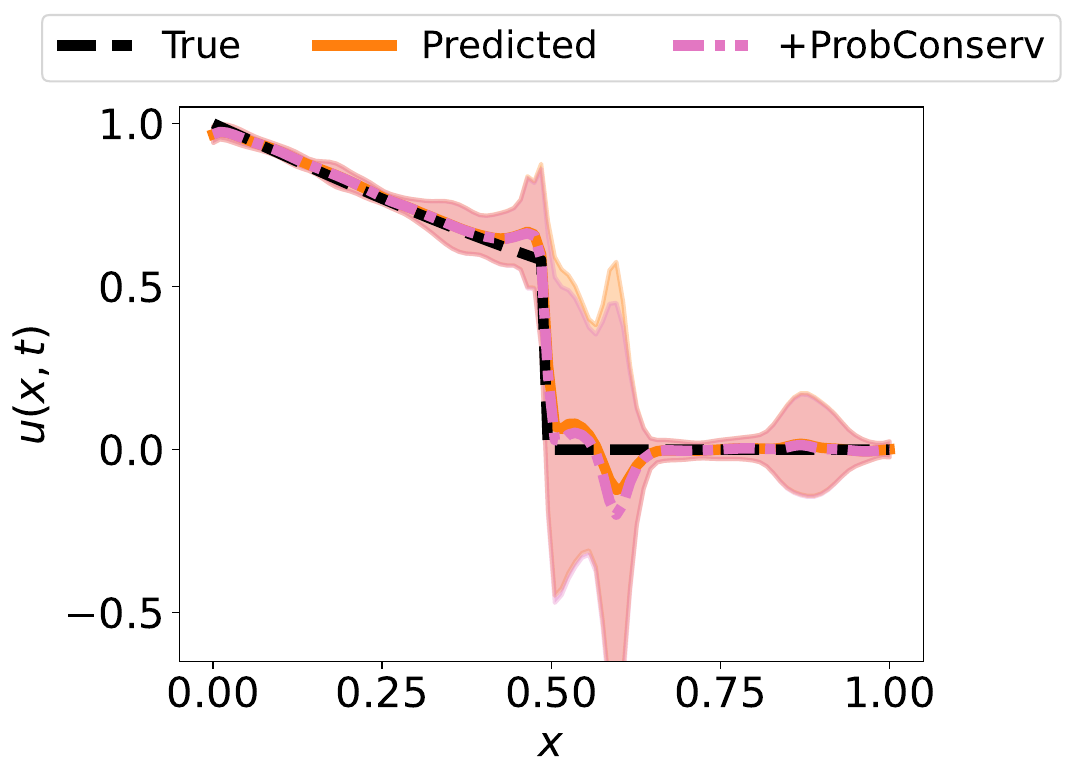}
    \caption{\ensemblenomethod}
    \end{subfigure}
    ~~~~
    \begin{subfigure}[h]{0.30\textwidth}
    \centering
    \includegraphics[scale=0.28]{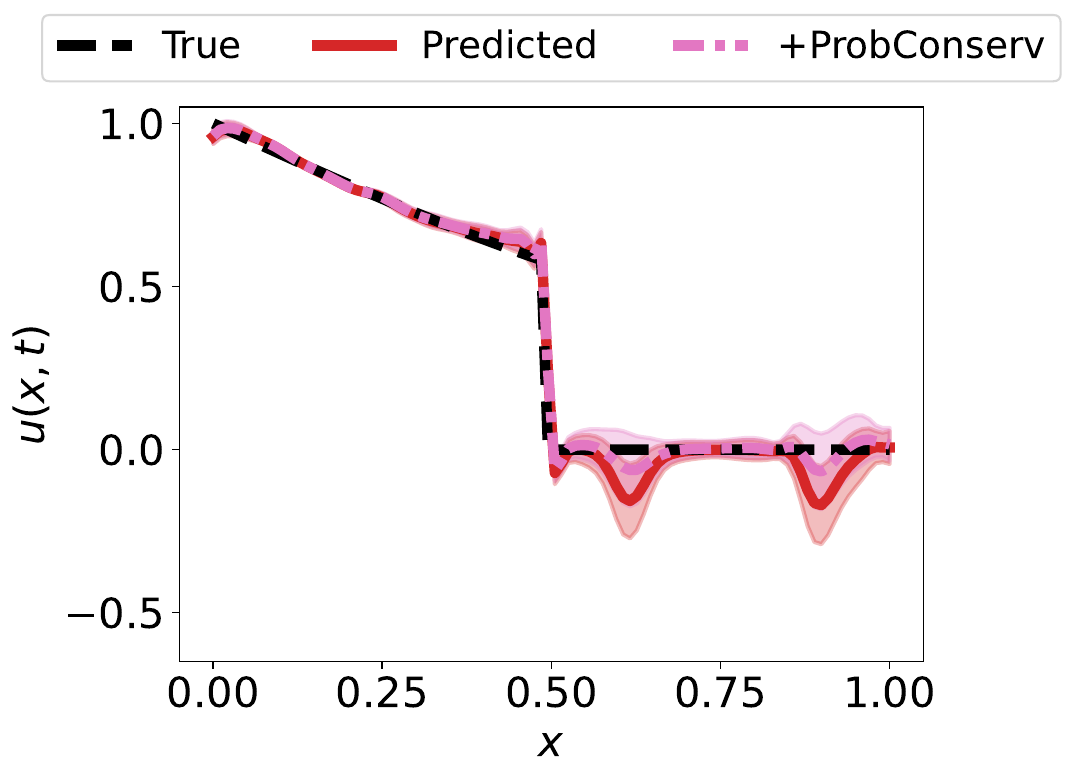}
    \caption{\method}
    \end{subfigure}
    \caption{{\bf 1-d Stefan Equation, small OOD shift}, $u{^*}^{\tr}\in[0.6,0.65], u{^*}^{\te}\in[0.55,0.6]$. 
    Solution profiles after applying \probconserv over different UQ methods under small OOD shifts in the solution value at the shock $u(t, x^*(t)) = u^*$ for shock position $x^*(t)$. \probconserv improves the OOD prediction with \method whose UQ estimate is well-correlated with the error and damps the artificial oscillations.} 
    \label{fig:probconserv_stefan}
\end{figure}

In this section, we report the solution profiles and metrics, which compare the effect of using the various UQ estimates for NOs within the \probconserv framework.  Tables \ref{tab:probconserv_heat_ood}-\ref{tab:probconserv_stefan_ood} show that the conservation error (CE) after applying \probconserv is 0, and that the CE for the unconstrained models can be quite large, especially on the harder tasks.  The results on the ``easy'' heat equation task in \cref{fig:probconserv_heat} and corresponding \cref{tab:probconserv_heat_ood} show that \probconserv improves the MSE for all the UQ methods. The results on the ``harder'' problems in \cref{fig:probconserv_pme} and corresponding \cref{tab:probconserv_pme_ood} for the PME and \cref{fig:probconserv_stefan} and corresponding \cref{tab:probconserv_stefan_ood} for the Stefan equation show the impact of good uncertainty estimates as input to \probconserv. We see that \probconserv improves the solution profiles and MSE for the methods except for \outputvarmethod, whose UQ estimate is not well-correlated with the error.

\begin{table}[H]
    \centering
    \caption{{\bf Effect of \probconserv update, 1-d heat equation.} 
    MSE $\downarrow$, Conservation Error (CE) (should be zero) and n-MeRCI $\downarrow$ (mean and standard deviation over 5 seeds) for different UQ methods with and without \probconserv evaluated on 1-d heat equation with small, medium and large OOD shifts, where $k^\tr\in[1,5]$.
    {\bf Bold} indicates values within one standard deviation of the best mean.
    }
\label{tab:probconserv_heat_ood}

    \resizebox{\textwidth}{!}{%
    \begin{tabular}{@{}lccccccHH@{}}
    \toprule
    &  \multicolumn{6}{c}{{\bf Small OOD shift}, $k^\te \in [5, 6]$} \\
    \midrule
     & \multicolumn{2}{c}{MSE $\downarrow$} & \multicolumn{2}{c}{CE (should be zero)} & \multicolumn{2}{c}{n-MeRCI $\downarrow$} \\ 
     & Standard & + \probconserv & Standard & + \probconserv & Standard & + \probconserv  \\
    \midrule 
    \bayesiannomethod & 2.5e-06 (8.6e-07) & 2.3e-06 (8.5e-07) &   0.01 ( 0.00) &   \bf 0.00 ( 0.00) &   0.86 ( 0.05) &   0.86 ( 0.05) &   0.38 ( 0.02) &   0.37 ( 0.03) \\
    \outputvarmethod & 7.1e-06 (3.2e-06) & 5.5e-06 (1.2e-06) &   0.01 ( 0.01) &   \bf 0.00 ( 0.00) &   1.17 ( 0.11) &   1.17 ( 0.12) &   0.43 ( 0.02) &   0.42 ( 0.02) \\
    \mcdropoutnomethod & 5.1e-06 (1.4e-06) & 4.9e-06 (1.5e-06) &   0.01 ( 0.00) & \bf   0.00 ( 0.00) &   0.90 ( 0.04) &   0.91 ( 0.05) & \bf   0.25 ( 0.01) & \bf   0.24 ( 0.01) \\
    \ensemblenomethod & 2.3e-06 (4.9e-07) & 2.2e-06 (5.2e-07) & \bf   0.00 ( 0.00) & \bf  0.00 ( 0.00) & \bf   0.02 ( 0.02) & \bf   0.02 ( 0.01) &   0.37 ( 0.02) &   0.37 ( 0.03) \\
    \method & \bf 1.7e-06 (4.1e-07) & \bf 1.2e-06 (9.2e-07) &   0.01 ( 0.01) &   \bf 0.00 ( 0.00) &   0.05 ( 0.03) &   0.05 ( 0.03) &   0.35 ( 0.01) &   0.33 ( 0.03) \\
    \end{tabular}
    }
    \resizebox{\textwidth}{!}{%
    \begin{tabular}{@{}lccccccHH@{}}
    \midrule
    &  \multicolumn{6}{c}{{\bf Medium OOD shift}, $k^\te \in [6, 7]$} \\
    \midrule
     & \multicolumn{2}{c}{MSE $\downarrow$} & \multicolumn{2}{c}{CE (should be zero)} & \multicolumn{2}{c}{n-MeRCI $\downarrow$} \\ 
     & Standard & + \probconserv & Standard & + \probconserv & Standard & + \probconserv  \\
    \midrule
    \bayesiannomethod & 2.7e-05 (7.5e-06) & 2.6e-05 (7.7e-06) &   0.02 ( 0.01) &  \bf 0.00 ( 0.00) &   0.84 ( 0.05) &   0.84 ( 0.05) &   0.47 ( 0.01) &   0.47 ( 0.02) \\
    \outputvarmethod & 8.0e-05 (2.9e-05) & 6.7e-05 (1.3e-05) &   0.04 ( 0.03) &   \bf 0.00 ( 0.00) &   1.40 ( 0.15) &   1.41 ( 0.16) &   0.49 ( 0.01) &   0.49 ( 0.01) \\
    \mcdropoutnomethod & 3.9e-05 (1.7e-05) & 3.7e-05 (1.8e-05) &   0.03 ( 0.01) &   \bf 0.00 ( 0.00) &   0.90 ( 0.03) &   0.90 ( 0.04) & \bf   0.37 ( 0.01) & \bf   0.35 ( 0.02) \\
    \ensemblenomethod & 2.4e-05 (3.8e-06) & \bf 2.4e-05 (4.1e-06) & \bf   0.01 ( 0.00) & \bf   0.00 ( 0.00) &   0.03 ( 0.01) & \bf   0.03 ( 0.02) & \bf   0.38 ( 0.02) & \bf   0.37 ( 0.02) \\
    \method & \bf 1.9e-05 (3.4e-06) & \bf 1.3e-05 (1.1e-05) &   0.05 ( 0.03) &   \bf 0.00 ( 0.00) & \bf   0.02 ( 0.00) &   0.08 ( 0.08) & \bf   0.36 ( 0.01) & \bf   0.34 ( 0.05) \\
    \end{tabular}
    }
    \resizebox{\textwidth}{!}{%
    \begin{tabular}{@{}lccccccHH@{}}
    \midrule
    &  \multicolumn{6}{c}{{\bf Large OOD shift}, $k^\te \in [7, 8]$} \\
     & \multicolumn{2}{c}{MSE $\downarrow$} & \multicolumn{2}{c}{CE (should be zero)} & \multicolumn{2}{c}{n-MeRCI $\downarrow$} \\ 
     & Standard & + \probconserv & Standard & + \probconserv & Standard & + \probconserv  \\
    \midrule
    \bayesiannomethod & 1.2e-04 (3.5e-05) & 1.2e-04 (3.6e-05) &   0.04 ( 0.02) & \bf  0.00 ( 0.00) &   0.80 ( 0.06) &   0.80 ( 0.06) &   0.49 ( 0.01) &   0.49 ( 0.01) \\
    \outputvarmethod & 3.7e-04 (1.3e-04) & 3.2e-04 (7.3e-05) &   0.08 ( 0.07) &   0.01 ( 0.01) &   1.70 ( 0.20) &   1.69 ( 0.21) &   0.50 ( 0.00) &   0.50 ( 0.01) \\
    \mcdropoutnomethod & 1.7e-04 (8.0e-05) & 1.6e-04 (8.3e-05) &   0.06 ( 0.02) & \bf  0.00 ( 0.00) &   0.86 ( 0.04) &   0.86 ( 0.04) &   0.44 ( 0.01) &   0.40 ( 0.03) \\
    \ensemblenomethod & 1.1e-04 (1.6e-05) & \bf 1.1e-04 (1.6e-05) & \bf   0.01 ( 0.00) & \bf   0.00 ( 0.00) & \bf   0.03 ( 0.02) & \bf   0.03 ( 0.01) & \bf   0.37 ( 0.02) & \bf   0.37 ( 0.02) \\
    \method & \bf 8.8e-05 (1.0e-05) & \bf 5.8e-05 (5.1e-05) &   0.10 ( 0.06) & \bf  0.00 ( 0.00) & \bf   0.03 ( 0.03) &   0.12 ( 0.11) & \bf   0.36 ( 0.02) & \bf   0.33 ( 0.05) \\
    \bottomrule
    \end{tabular}
    }
\end{table}

\begin{table}[H]
    \centering
    \caption{{\bf Effect of \probconserv update, 1-d PME.} 
    MSE $\downarrow$, Conservation Error (CE) (should be zero) and n-MeRCI $\downarrow$ (mean and standard deviation over 5 seeds) for different UQ methods with and without \probconserv evaluated on 1-d PME with small, medium and large OOD shifts, where $m^\tr\in[2,3]$.
    {\bf Bold} indicates values within one standard deviation of the best mean.
    }
\label{tab:probconserv_pme_ood}

    \resizebox{\textwidth}{!}{%
    \begin{tabular}{@{}lccccccHH@{}}
    \toprule
    &  \multicolumn{6}{c}{{\bf Small OOD shift}, $m^\te \in [1, 2]$} \\
    \midrule
     & \multicolumn{2}{c}{MSE $\downarrow$} & \multicolumn{2}{c}{CE (should be zero)} & \multicolumn{2}{c}{n-MeRCI $\downarrow$} \\ 
     & Standard & + \probconserv & Standard & + \probconserv & Standard & + \probconserv  \\
    \midrule
    \bayesiannomethod & \bf 1.1e-03 (4.0e-04) & 3.8e-04 (1.4e-04) & \bf   0.38 ( 0.07) &   0.02 ( 0.00) &   1.12 ( 0.07) &   1.13 ( 0.07) &   0.44 ( 0.02) &   0.39 ( 0.01) \\
    \outputvarmethod & 4.0e-03 (2.4e-03) & 8.6e-03 (4.3e-03) &   0.75 ( 0.27) & \bf   0.00 ( 0.00) &   0.26 ( 0.06) &   0.22 ( 0.05) &   0.43 ( 0.03) &   0.40 ( 0.01) \\
    \mcdropoutnomethod & 2.1e-03 (6.0e-04) & 8.6e-04 (2.1e-04) &   0.52 ( 0.10) & \bf   0.00 ( 0.00) &   1.18 ( 0.09) &   1.17 ( 0.09) & \bf   0.36 ( 0.03) & \bf   0.26 ( 0.02) \\
    \ensemblenomethod & \bf 1.2e-03 (2.5e-04) & \bf 1.7e-04 (1.1e-04) & \bf   0.40 ( 0.06) &  \bf 0.00 ( 0.00) & \bf   0.14 ( 0.03) & \bf   0.08 ( 0.06) &   0.46 ( 0.01) &   0.32 ( 0.02) \\
    \method & \bf 1.1e-03 (3.7e-04) & 3.9e-04 (6.6e-05) & \bf   0.37 ( 0.09) & \bf  0.00 ( 0.00) &   0.21 ( 0.04) &   0.20 ( 0.04) &   0.42 ( 0.03) &   0.34 ( 0.02) \\
    \end{tabular}
    }
    
    \resizebox{\textwidth}{!}{%
    \begin{tabular}{@{}lccccccHH@{}}
    \midrule
    &  \multicolumn{6}{c}{{\bf Medium OOD shift}, $m^\te \in [4, 5]$} \\
    \midrule
     & \multicolumn{2}{c}{MSE $\downarrow$} & \multicolumn{2}{c}{CE (should be zero)} & \multicolumn{2}{c}{n-MeRCI $\downarrow$} \\ 
     & Standard & + \probconserv & Standard & + \probconserv & Standard & + \probconserv  \\
    \midrule
    \bayesiannomethod & 1.0e-03 (3.2e-04) & 7.8e-04 (1.9e-04) &   0.31 ( 0.08) & \bf  0.00 ( 0.00) &   0.73 ( 0.03) &   0.73 ( 0.03) &   0.47 ( 0.01) &   0.46 ( 0.00) \\
    \outputvarmethod & 5.0e-03 (7.6e-04) & 7.0e-02 (1.0e-02) &   0.90 ( 0.14) &   0.01 ( 0.01) &   1.23 ( 0.34) &   1.25 ( 0.31) &   0.50 ( 0.00) &   0.50 ( 0.00) \\
    \mcdropoutnomethod & 1.5e-03 (4.2e-04) & 1.1e-03 (3.6e-04) &   0.35 ( 0.12) & \bf   0.00 ( 0.00) &   0.75 ( 0.02) &   0.74 ( 0.02) &   0.42 ( 0.03) & \bf   0.36 ( 0.02) \\
    \ensemblenomethod & \bf 8.1e-04 (1.6e-04) & \bf 5.3e-04 (8.1e-05) & \bf   0.27 ( 0.03) & \bf   0.00 ( 0.00) &   0.20 ( 0.03) & \bf   0.14 ( 0.07) & \bf   0.38 ( 0.01) & \bf   0.37 ( 0.01) \\
    \method & 1.1e-03 (3.5e-04) & 9.9e-04 (6.2e-04) &   0.32 ( 0.06) & \bf   0.00 ( 0.00) & \bf   0.15 ( 0.03) & \bf   0.09 ( 0.06) & \bf   0.38 ( 0.02) & \bf   0.37 ( 0.03) \\
    \end{tabular}
    }
    
    \resizebox{\textwidth}{!}{%
    \begin{tabular}{@{}lccccccHH@{}}
    \midrule
    &  \multicolumn{6}{c}{{\bf Large OOD shift}, $m^\te \in [5, 6]$} \\
    \midrule
     & \multicolumn{2}{c}{MSE $\downarrow$} & \multicolumn{2}{c}{CE (should be zero)} & \multicolumn{2}{c}{n-MeRCI $\downarrow$} \\ 
     & Standard & + \probconserv & Standard & + \probconserv & Standard & + \probconserv  \\
    \midrule
    \bayesiannomethod & 6.1e-03 (1.9e-03) & 4.7e-03 (1.3e-03) &   0.83 ( 0.19) &   0.01 ( 0.00) &   0.69 ( 0.02) &   0.69 ( 0.02) &   0.49 ( 0.01) &   0.49 ( 0.01) \\
    \outputvarmethod & 2.0e-02 (1.8e-03) & 3.3e-01 (8.5e-02) &   1.87 ( 0.18) &   0.07 ( 0.06) &   1.52 ( 0.46) &   1.55 ( 0.45) &   0.50 ( 0.00) &   0.50 ( 0.00) \\
    \mcdropoutnomethod & 6.4e-03 (2.2e-03) & 3.5e-03 (1.4e-03) &   0.81 ( 0.24) &  \bf 0.00 ( 0.00) &   0.70 ( 0.02) &   0.66 ( 0.03) &   0.47 ( 0.03) & \bf   0.38 ( 0.02) \\
    \ensemblenomethod & \bf 4.6e-03 (7.1e-04) & \bf 2.5e-03 (5.2e-04) & \bf   0.71 ( 0.05) & \bf  0.00 ( 0.00) &   0.22 ( 0.02) &   0.15 ( 0.06) & \bf   0.42 ( 0.02) &   0.40 ( 0.02) \\
    \method & 5.8e-03 (1.7e-03) & 4.4e-03 (2.6e-03) &   0.82 ( 0.15) & \bf   0.00 ( 0.00) & \bf   0.13 ( 0.05) & \bf   0.06 ( 0.03) & \bf   0.41 ( 0.02) & \bf   0.39 ( 0.03) \\
    \bottomrule
    \end{tabular}
    }
\end{table}

\begin{table}[H]
        \centering
        \caption{{\bf Effect of \probconserv update, 1-d Stefan.} 
        MSE $\downarrow$, Conservation Error (CE) (\textit{should be zero}) and n-MeRCI $\downarrow$ (mean and standard deviation over 5 seeds) for different UQ methods with and without \probconserv evaluated on the 1-d Stefan equation with small, medium and large OOD shifts, where $u{^*}^{\tr}\in[0.6,0.65]$.
        {\bf Bold} indicates values within one standard deviation of the best mean.
        }
    \label{tab:probconserv_stefan_ood}
    
    \resizebox{\textwidth}{!}{%
    \begin{tabular}{@{}lccccccHH@{}}
    \toprule
    &  \multicolumn{6}{c}{{\bf Small OOD shift}, ${u^*}^\te \in [0.55, 0.6]$} \\
    \midrule
     & \multicolumn{2}{c}{MSE $\downarrow$} & \multicolumn{2}{c}{CE (should be zero)} & \multicolumn{2}{c}{n-MeRCI $\downarrow$} \\ 
     & Standard & + \probconserv & Standard & + \probconserv & Standard & + \probconserv  \\
    \midrule
    \bayesiannomethod & 2.0e-02 (1.9e-02) & 1.4e-02 (8.5e-03) &   0.37 ( 0.23) &   \bf 0.00 ( 0.00) &   0.67 ( 0.15) &   0.66 ( 0.17) & \bf   0.28 ( 0.01) & \bf   0.29 ( 0.01) \\
    \outputvarmethod & 2.3e-02 (1.6e-03) & 3.0e-02 (3.6e-03) &   1.08 ( 0.06) & \bf   0.00 ( 0.00) &   0.97 ( 0.07) &   0.95 ( 0.03) &   0.40 ( 0.02) &   0.39 ( 0.02) \\
    \mcdropoutnomethod & \bf 9.6e-03 (3.6e-03) & \bf 9.5e-03 (3.2e-03) &   0.51 ( 0.21) &   \bf 0.00 ( 0.00) &   0.78 ( 0.08) &   0.77 ( 0.08) &   0.31 ( 0.01) &   0.34 ( 0.02) \\
    \ensemblenomethod & \bf 8.1e-03 (3.4e-03) & \bf 9.1e-03 (2.9e-03) & \bf   0.20 ( 0.03) &  \bf 0.00 ( 0.00) & \bf   0.14 ( 0.09) &   0.23 ( 0.06) &   0.32 ( 0.01) &   0.32 ( 0.01) \\
    \method & 1.4e-02 (2.3e-03) & \bf 1.1e-02 (9.2e-04) &   0.27 ( 0.05) &  \bf 0.00 ( 0.00) & \bf   0.14 ( 0.06) & \bf   0.11 ( 0.07) &   0.38 ( 0.01) &   0.37 ( 0.01) \\
    \end{tabular}
    }
    
    \resizebox{\textwidth}{!}{%
    \begin{tabular}{@{}lccccccHH@{}}
    \midrule
    &  \multicolumn{6}{c}{{\bf Medium OOD shift}, ${u^*}^\te \in [0.7, 0.75]$} \\
    \midrule
     & \multicolumn{2}{c}{MSE $\downarrow$} & \multicolumn{2}{c}{CE (should be zero)} & \multicolumn{2}{c}{n-MeRCI $\downarrow$} \\ 
     & Standard & + \probconserv & Standard & + \probconserv & Standard & + \probconserv  \\
    \midrule
    \bayesiannomethod & 1.7e-02 (1.4e-02) & 1.3e-02 (3.8e-03) & \bf   0.55 ( 0.19) &  \bf 0.00 ( 0.00) &   0.66 ( 0.10) &   0.66 ( 0.10) & \bf   0.29 ( 0.02) & \bf   0.29 ( 0.02) \\
    \outputvarmethod & 3.2e-02 (1.3e-03) & 4.1e-02 (5.3e-03) &   1.14 ( 0.05) & \bf   \bf 0.00 ( 0.00) &   0.83 ( 0.05) &   0.84 ( 0.04) &   0.40 ( 0.02) &   0.40 ( 0.02) \\
    \mcdropoutnomethod & 2.9e-02 (1.3e-02) & 1.8e-02 (1.1e-02) &   1.01 ( 0.32) &   \bf 0.00 ( 0.00) &   0.56 ( 0.26) &   0.57 ( 0.25) &   0.36 ( 0.03) &   0.31 ( 0.02) \\
    \ensemblenomethod & \bf 8.0e-03 (1.4e-03) & \bf 7.1e-03 (5.6e-04) & \bf   0.52 ( 0.08) &  \bf 0.00 ( 0.00) & \bf   0.07 ( 0.03) &   0.24 ( 0.11) &   0.33 ( 0.02) &   0.31 ( 0.02) \\
    \method & 1.1e-02 (3.6e-03) & 9.5e-03 (1.8e-03) & \bf   0.59 ( 0.19) &  \bf 0.00 ( 0.00) &   0.14 ( 0.03) & \bf   0.12 ( 0.03) &   0.37 ( 0.04) &   0.38 ( 0.03) \\
    \end{tabular}
    }
    
    \resizebox{\textwidth}{!}{%
    \begin{tabular}{@{}lccccccHH@{}}
    \midrule
    &  \multicolumn{6}{c}{{\bf Large OOD shift}, ${u^*}^\te \in [0.5, 0.55]$} \\
    \midrule
     & \multicolumn{2}{c}{MSE $\downarrow$} & \multicolumn{2}{c}{CE (should be zero)} & \multicolumn{2}{c}{n-MeRCI $\downarrow$} \\ 
     & Standard & + \probconserv & Standard & + \probconserv & Standard & + \probconserv  \\
    \midrule
    \bayesiannomethod & 1.7e-01 (1.7e-01) & 7.7e-02 (4.7e-02) &   1.34 ( 0.94) & \bf  0.00 ( 0.00) &   0.50 ( 0.25) &   0.48 ( 0.27) &   0.40 ( 0.03) &   0.40 ( 0.02) \\
    \outputvarmethod & \bf 3.4e-02 (1.8e-03) & 7.9e-02 (8.8e-03) &   1.72 ( 0.07) &  \bf 0.00 ( 0.00) &   0.99 ( 0.09) &   0.97 ( 0.05) &   0.45 ( 0.01) &   0.46 ( 0.02) \\
    \mcdropoutnomethod & 4.4e-02 (3.1e-02) & \bf 4.0e-02 (2.7e-02) &   1.27 ( 0.41) &  \bf 0.00 ( 0.00) &   0.54 ( 0.11) &   0.44 ( 0.06) &   0.42 ( 0.01) &   0.43 ( 0.01) \\
    \ensemblenomethod & 4.6e-02 (1.9e-02) & \bf 4.7e-02 (1.5e-02) & \bf   0.61 ( 0.13) & \bf   0.00 ( 0.00) &   0.37 ( 0.14) &   0.33 ( 0.13) & \bf   0.36 ( 0.01) & \bf   0.36 ( 0.01) \\
    \method & 8.9e-02 (5.1e-02) & \bf 6.4e-02 (2.6e-02) &   0.90 ( 0.39) &  \bf 0.00 ( 0.00) & \bf   0.24 ( 0.11) & \bf   0.21 ( 0.11) &   0.43 ( 0.01) &   0.42 ( 0.01) \\
    \bottomrule
    \end{tabular}
    }
\end{table}

\end{document}